\documentclass[10pt,twocolumn,letterpaper]{article}

\usepackage{cvpr}              
\usepackage[dvipsnames]{xcolor}

\definecolor{cvprblue}{rgb}{0.21,0.49,0.74}
\usepackage[pagebackref,breaklinks,colorlinks,citecolor=cvprblue]{hyperref}

\usepackage{graphicx}
\usepackage{amsmath}
\usepackage{amssymb}
\usepackage{mathtools}
\usepackage{booktabs}
\usepackage{multirow}
\usepackage{xcolor}
\usepackage{adjustbox}
\usepackage{array}
\usepackage{soul}

\usepackage{nicematrix}
\usepackage{wrapfig}

\newcommand{\tightpara}[1]{\vspace{-3mm}\paragraph{#1}}

\def\paperID{7273} 
\def\confName{CVPR}
\def\confYear{2024}

\begin{document}

\title{PoNQ: a Neural QEM-based Mesh Representation}

\author{
Nissim Maruani\\[-0.5mm]
\small{Inria, Universit\'e C\^ote d'Azur}\\[-1mm]
{\tt\small nissim.maruani@inria.fr}
\and
Maks Ovsjanikov\\[-.5mm]
\small{LIX, École Polytechnique, IP Paris}\\[-1mm]
{\tt\small maks@lix.polytechnique.fr}
\and
Pierre Alliez\\[-.5mm]
\small{Inria, Universit\'e C\^ote d'Azur}\\[-1mm]
{\tt\small pierre.alliez@inria.fr}
\and
Mathieu Desbrun\\[-.5mm]
\small{Inria Saclay - Ecole Polytechnique}\\[-1mm]
{\tt\small mathieu.desbrun@inria.fr}
}

\maketitle
\begin{abstract}
Although polygon meshes have been a standard representation in geometry processing, their irregular and combinatorial nature hinders their suitability for learning-based applications.
In this work, we introduce a novel learnable mesh representation through a set of local 3D sample \emph{Po}ints and their associated \emph{N}ormals and \emph{Q}uadric error metrics (QEM) w.r.t. the underlying shape, which we denote PoNQ.
A global mesh is directly derived from PoNQ by efficiently leveraging the knowledge of the local quadric errors. Besides marking the first use of QEM within a neural shape representation, our contribution guarantees both topological and geometrical properties by ensuring that a PoNQ mesh does not self-intersect and is always the boundary of a volume. Notably, our representation does not rely on a regular grid, is supervised directly by the target \textit{surface} alone, and also handles open surfaces with boundaries and/or sharp features. 
We demonstrate the efficacy of PoNQ through a learning-based mesh prediction from SDF grids and show that our method surpasses recent state-of-the-art techniques in terms of both surface and edge-based metrics.
\end{abstract}\vspace*{-8mm}

\section{Introduction}
\label{sec:intro}

\begin{figure}[t!]
\centering
 \begin{subfigure}{.15\textwidth}
  \centering
  \includegraphics[width=\linewidth]{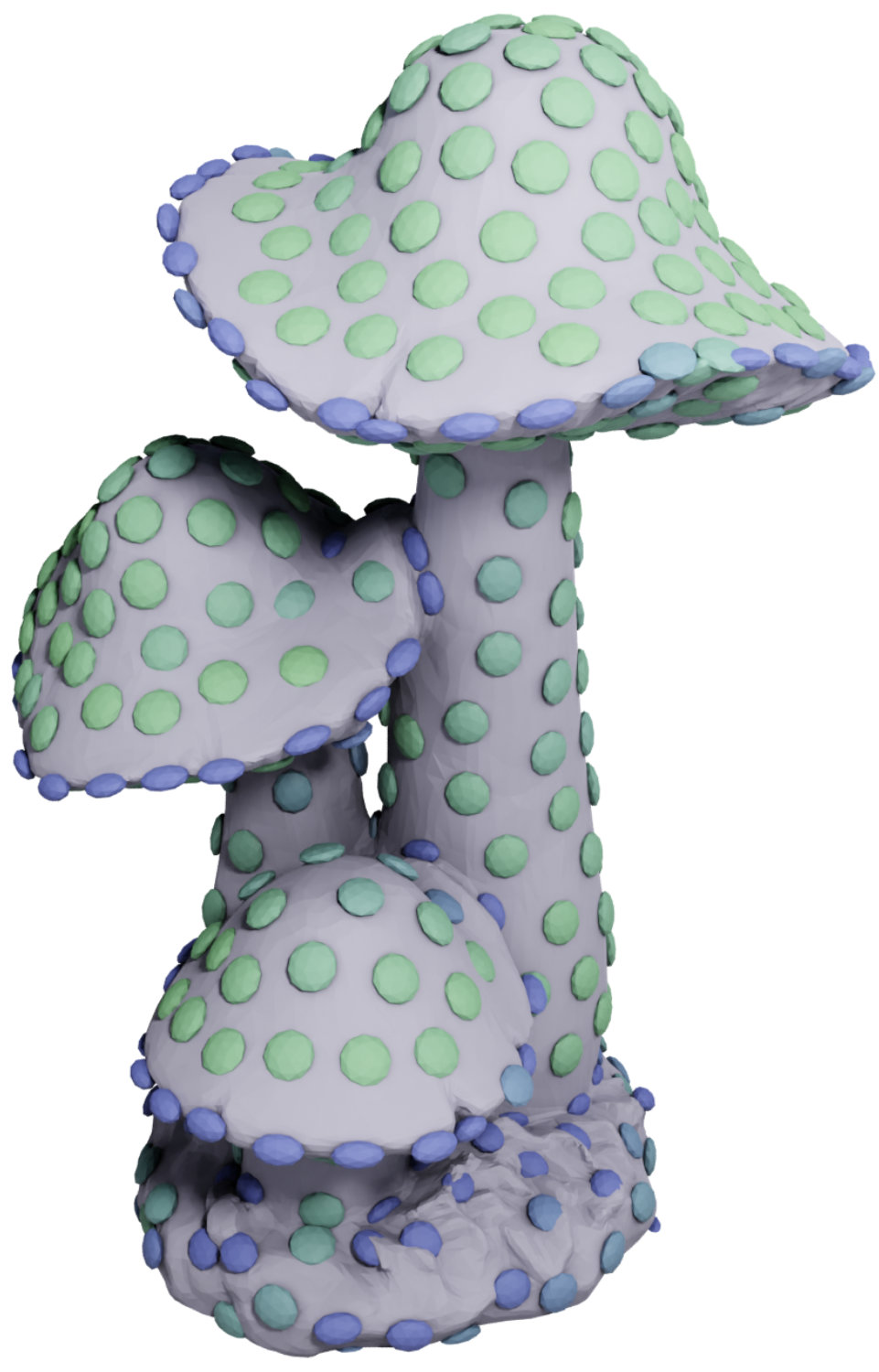} 
\end{subfigure}
\begin{subfigure}{.15\textwidth}
  \centering    
  \includegraphics[width=\linewidth]{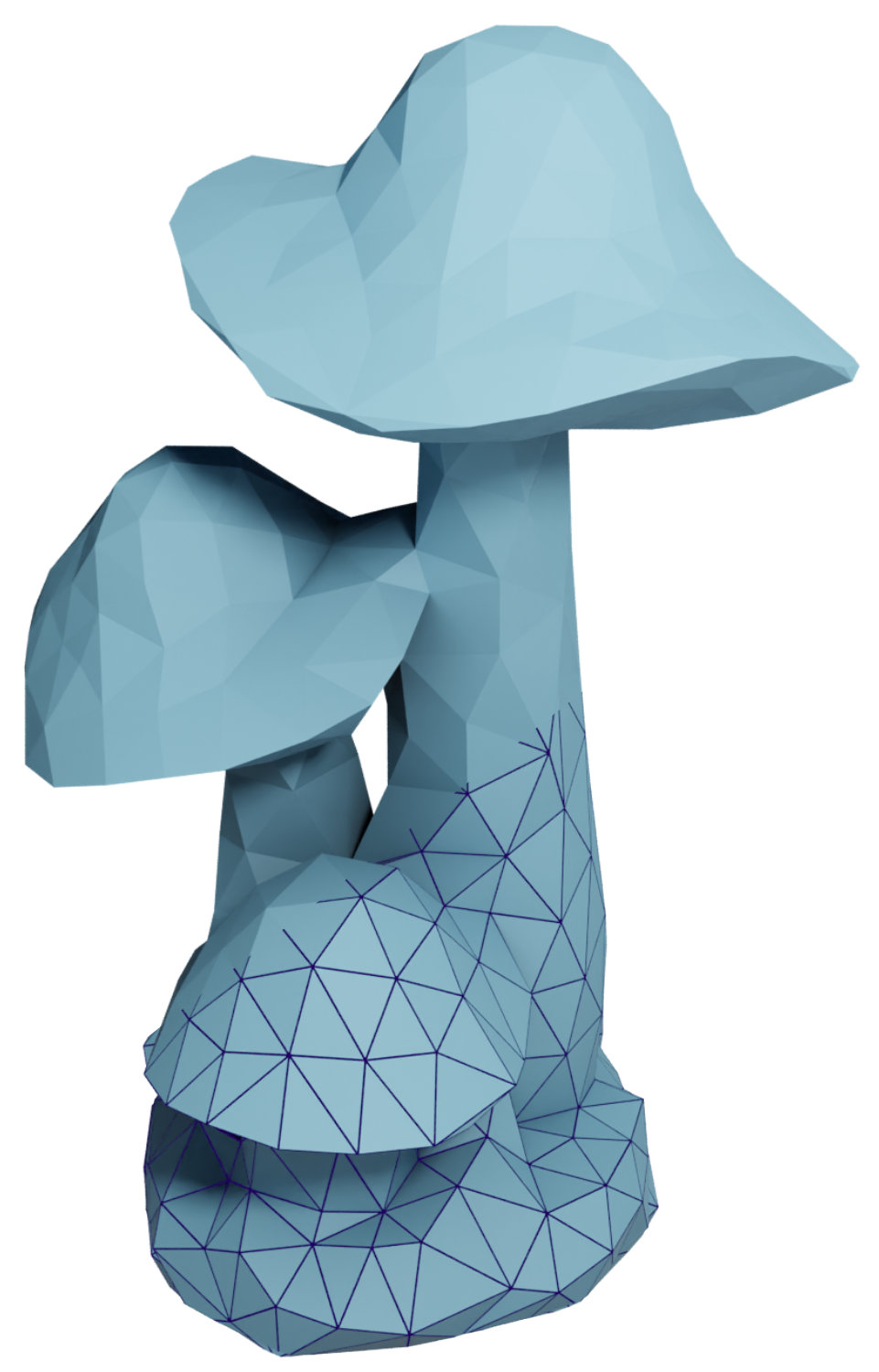}
\end{subfigure}
\begin{subfigure}{.15\textwidth}
  \centering    
  \includegraphics[width=\linewidth]{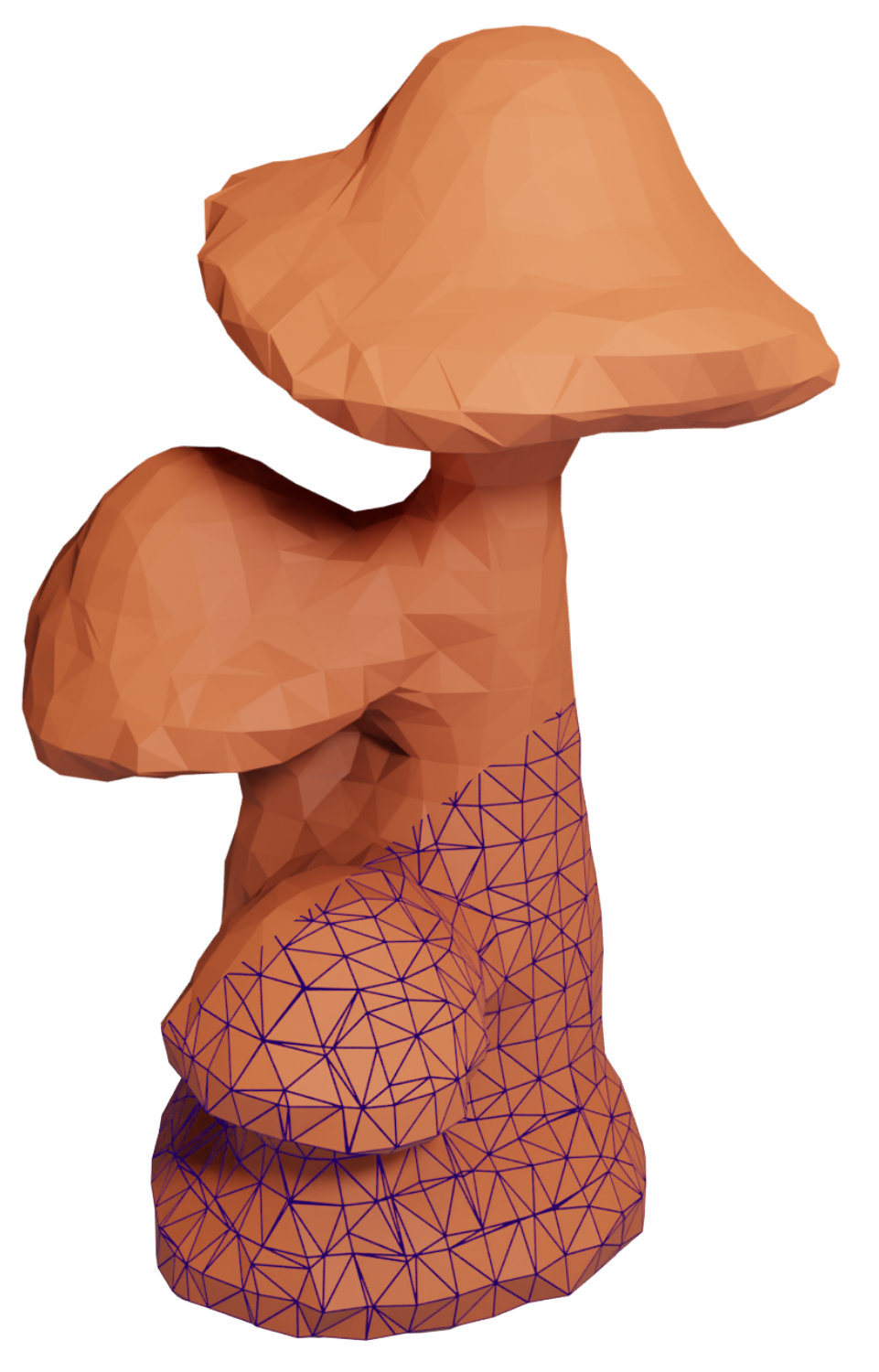}
\end{subfigure}

\centering
 \begin{subfigure}{.15\textwidth}
  \centering
  \includegraphics[width=\linewidth]{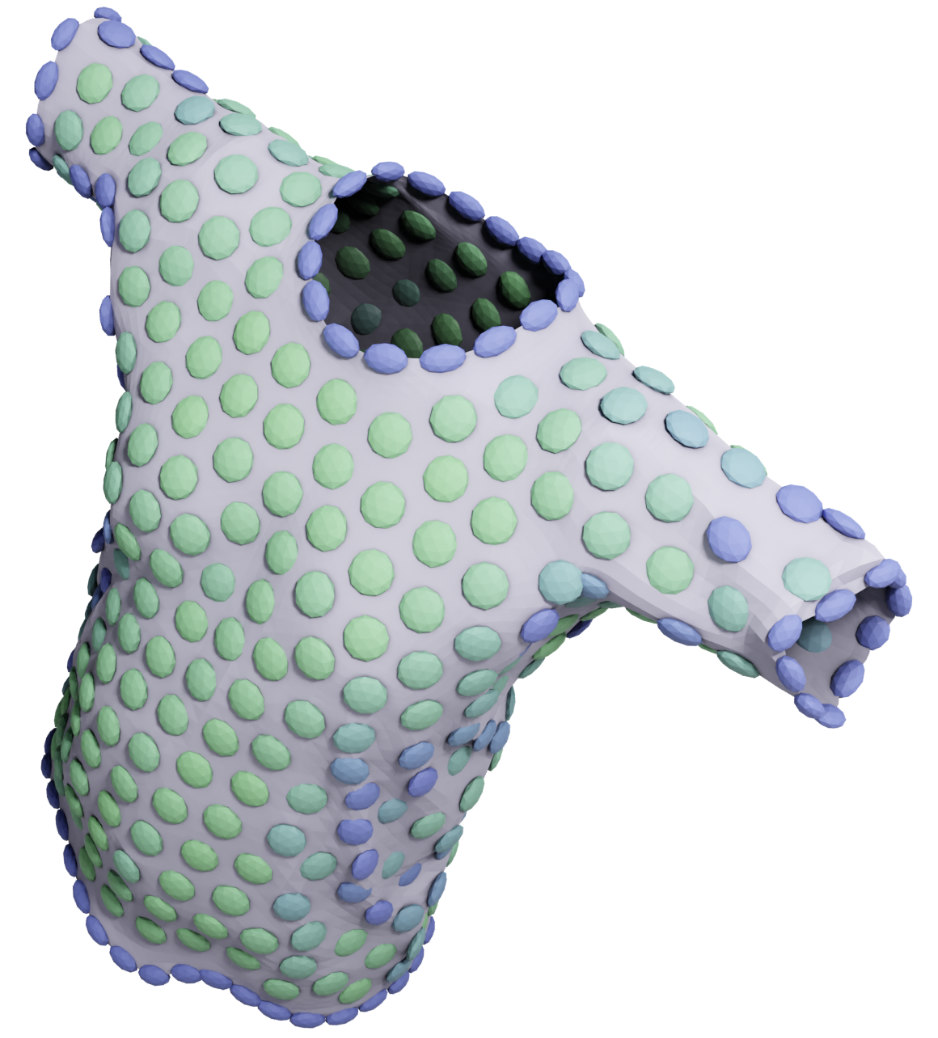} 
    \caption{PoNQ over input}
\end{subfigure}
\begin{subfigure}{.15\textwidth}
  \centering    
  \includegraphics[width=\linewidth]{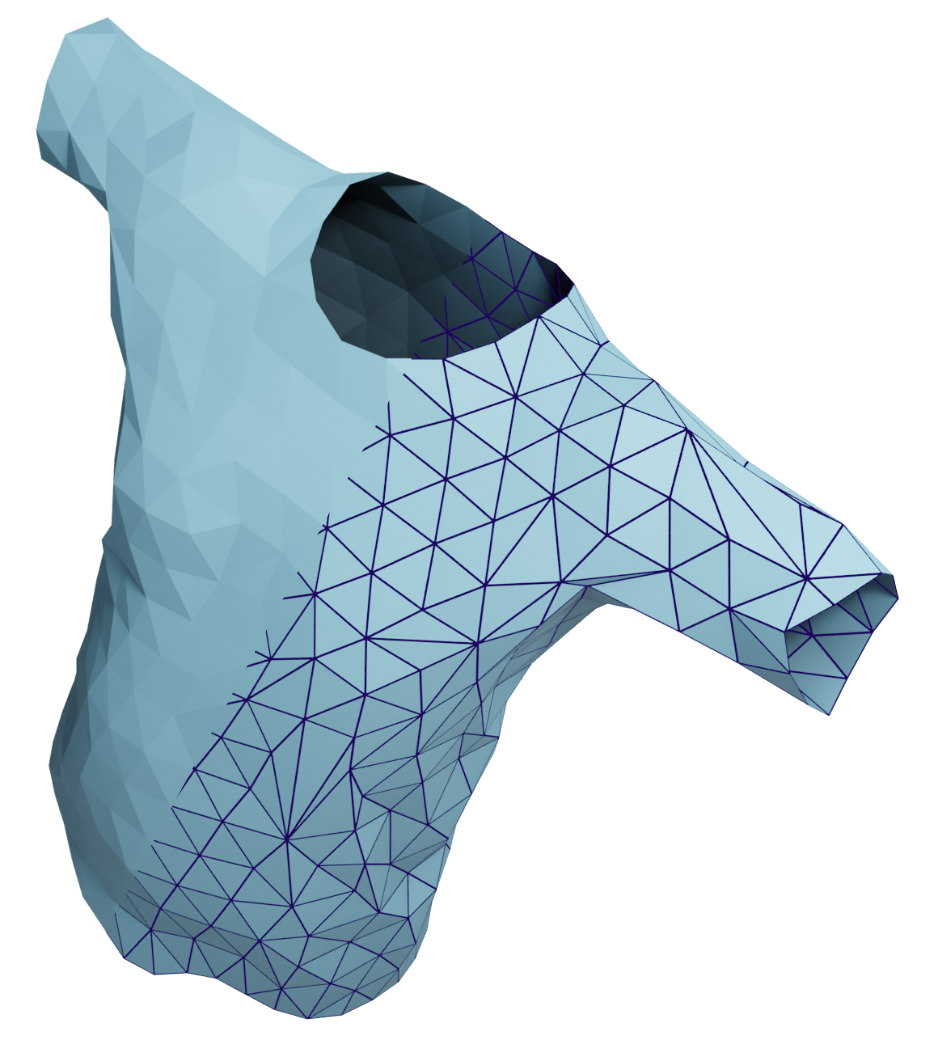}
\caption{PoNQ mesh}
\end{subfigure}
\begin{subfigure}{.15\textwidth}
  \centering    
  \includegraphics[width=\linewidth]{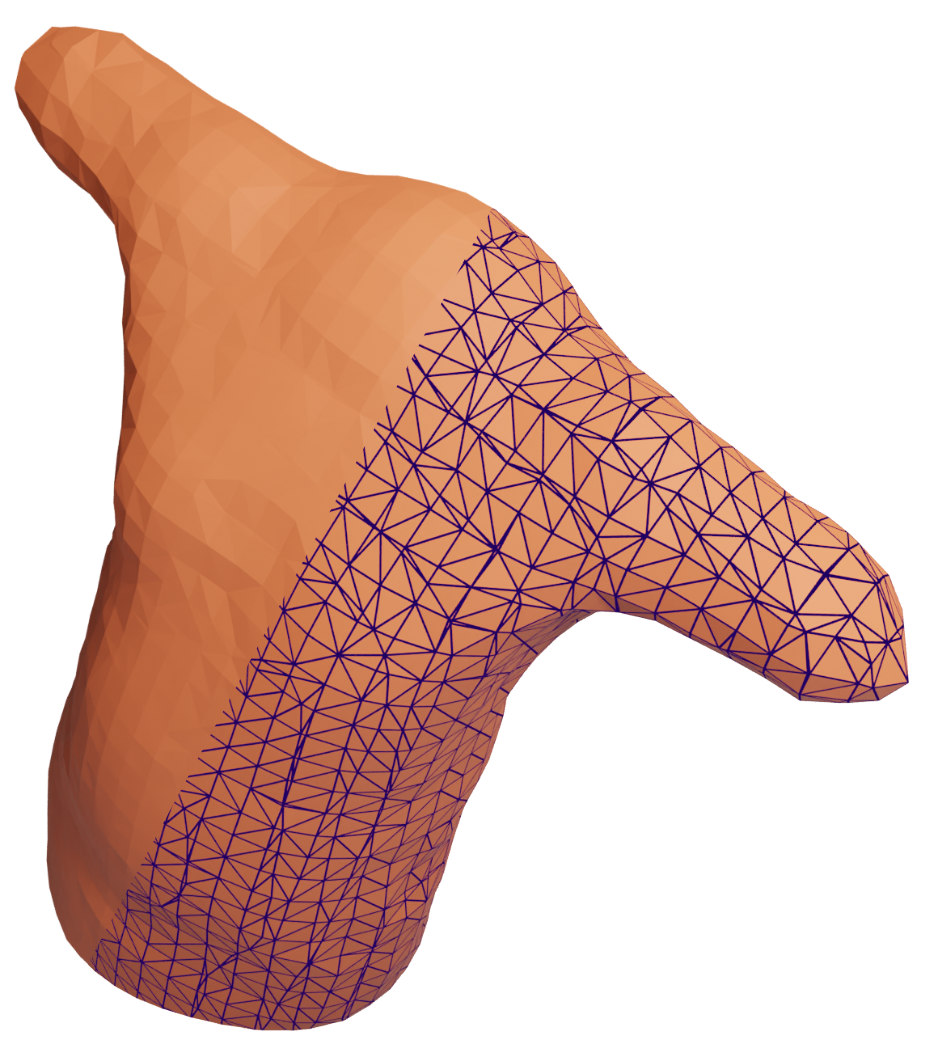}
\caption{Poisson mesh~\cite{kazhdan_poisson_2006}}
\end{subfigure}\vspace*{-1mm}
\caption{\textbf{PoNQ representation.} Quadric error metric (QEM) matrices are fitted and visualized on a given shape (left). Note that their aspect ratios (green to blue) capture the underlying surface information: pancakes for flat regions, cigars for sharp edges and balls for corners. Our mesh extraction outputs a watertight (or, optionally, open, see bottom) and non-self-intersecting mesh that preserves salient features of the input shape (center) and is more concise and faithful than current implicit approaches (right). 
\vspace*{-6mm}
}
  \label{fig:summary}
\end{figure}

In recent years, learning-based methods have shown great promise as an efficient means to handle ill-posed shape processing tasks with complex priors \cite{guo_deep_2021, xie_neural_2022, huang_surface_2022}.
%
Yet, despite the slew of advanced architectures to analyze or process 3D datasets, there is a lack of learnable 3D representations that can capture ridges and corners, while guaranteeing valid output meshes representing real 3D shapes for even the most basic learning tasks.

Early works on shape representation for learning predominantly relied on implicit volumetric representations~\cite{xie_neural_2022,yang_geometry_2021}, which, while eventually yielding a mesh through extraction, posed significant challenges. Notably, their training can be costly due to the volumetric nature of these representations. Moreover, the resulting mesh surfaces often lack detail, tending to generate overly smoothed shapes even where sharp features are expected.
In response to these limitations, recent advancements have formulated explicit representations that effectively encode shapes through strategically chosen sample points and other geometric entities such as normals, from which more accurate meshes can be derived~\cite{peng_shape_2021,maruani_voromesh_2023}.  While these  representations offer higher generalizability and greater control over the generated surfaces, they often cannot guarantee important properties such as \emph{watertightness} (i.e., the mesh must be the boundary of a non-degenerated volume), 
or the \emph{absence of self-intersections} (i.e., the surface is not just immersed, but embedded in $\mathbb{R}^3$). Only very recently an approach has been proposed that enforces these properties ~\cite{maruani_voromesh_2023}. However, 
it can suffer from overly complex output meshes (presence of many spurious polygonal facets) and, at times, wrong occupancy guesses create dents on the resulting shapes. 

In this work, we propose PoNQ (short for point-normal-QEM), a new learnable 3D representation which encodes a shape through discrete points and other \emph{local} geometric quantities to ensure efficient training and sharp reconstructions. As we demonstrate, a \emph{global} mesh can be extracted from our PoNQ representation through a robust  approach that leverages the available geometric data so as to better capture ridges and thin structures. 
Our key contributions are as follows: 

\begin{itemize}
    \item our neural representation is the first to exploit the quadric error metric (QEM), which has been instrumental in classical geometry processing tasks; consequently, PoNQ excels at capturing sharp features and boundaries, 
    preserving the intricate details that are often lost in existing representations (see Fig.~\ref{fig:summary});
    \item PoNQ meshes are guaranteed to be watertight 
    and free of self-intersections, 
    thus broadening their applicability and utility in downstream applications; 
    \item our PoNQ representation can also easily reduce the element count of the output shapes while preserving sharp features due to its reliance on QEM;
    \item finally, our neural representation outperforms state-of-the-art methods in mesh reconstruction from SDF grids as measured with both surface and edge-based metrics. 
\end{itemize}

\section{Related Work}
\label{sec:related}

We begin with a review of related works, including popular methods in shape learning, but also covering relevant parts of shape reconstruction and shape simplification. \smallskip

\noindent \textbf{Shape Reconstruction.} The literature abounds with shape reconstruction methods (typically from input pointsets), each distinguished by their unique priors -- necessary to regularize the ill-posed nature of the task -- and output representations.
In so-called continuous approaches, the reconstructed shape is the fixed-point of a projection operator~\cite{alexa_point_2001}, an algebraic point set surface (APSS~\cite{guennebaud_algebraic_2007}), or the isolevel of a 3D distance-like scalar field~\cite{kazhdan_poisson_2006, kazhdan_screened_2013}, to mention a few popular approaches. 
Approaches from the latter category (often referred to as implicit methods) further require a mesh extraction from the scalar field, denoted isosurfacing, typically through Marching Cubes (MC~\cite{lorensen_marching_1987}, or its improved neural variant NMC~\cite{chen_neural_2021}), Dual Contouring (DC~\cite{ju_dual_2002}, or its neural variant NDC~\cite{chen_neural_2022}), or through the recently-proposed Reach-for-the-Spheres (RTS) technique~\cite{sellan_reach_2023} that leverages geometric properties of a signed distance field to improve isosurfacing. 

Combinatorial methods, popular in Computational Geometry, rely instead on 3D Delaunay triangulations or their duals, Voronoi diagrams. The popular Crust approach, for instance, proceeds by local filtering of (facets of) 3D Delaunay triangulations~\cite{amenta_new_1998}. However, it requires dense point samples to correctly capture thin features, and cannot handle sharp edges. Its Powercrust~\cite{amenta_power_2001} and Tight Cocone~\cite{dey_tight_2003} variants add filtering of the triangulation to improve various properties, like the capture of holes in the reconstructed surface, but they share the same limitations. A global graph-cut extraction of the faces of the Delaunay triangulation of the input samples~\cite{kolluri_spectral_2004} was also shown to be more robust to noisy inputs, just like 
a recent Delaunay-based approach that alternates between filmsticking and sculpting over a 3D triangulation~\cite{wang_restricted_2022}.

\noindent \textbf{Implicit Field Learning.} Most pioneering 3D learning methods~\cite{park_deepsdf_2019, mescheder_occupancy_2019, peng_convolutional_2020} represented a shape as a scalar field (very often, a signed distance field in fact), later converted into a mesh via Marching Cubes~\cite{lorensen_marching_1987} or directly rendered. These methods need to process the global shape, limiting their generalizability beyond ShapeNet~\cite{chang_shapenet_2015}. 
More recent implicit-based approaches~\cite{gropp_implicit_2020, sitzmann_implicit_2020} showed that neural networks can be overfitted to single-shape Signed Distance Fields (SDF), and as such, they cannot learn from multiple shapes. Adding point-based convolutions~\cite{boulch_poco_2022}, hash tables~\cite{muller_instant_2022}, octree structures~\cite{tang_octfield_2021, wang_dual_2022, liu_deep_2021} or kernel methods~\cite{huang_neural_2023} have also been proposed to further refine these implicit methods, while the use of unsigned distance fields was leveraged to represent surfaces with boundaries~\cite{guillard_meshudf_2022, christiansen_neural_2023, zhang_surface_2023}. Although using continuous fields provides topological guarantees, the lack of control of the final locations of mesh vertices yields a distinctively ``blobby'' aspect that smooths out sharp features and small details. Furthermore, implicit fields are trained over the whole \emph{volume} around and inside the shape, while most applications are only interested in the reconstructed \emph{surface}.

\noindent \textbf{Explicit Shape Learning.} To overcome the limitations of implicit fields and offer better control over the elements of the output mesh, end-to-end differential iso-surfacing~ approaches were proposed~\cite{liao_deep_2018, remelli_meshsdf_2020, shen_flexible_2023}. Yet these approaches still rely on regular grids and often do not guarantee intersection-free output meshes. 
Methods that best fit 
a set of canonical geometric primitives~\cite{chen_bsp-net_2020, deng_cvxnet_2020, wu_quadricsnet_2023} or an explicit mesh~\cite{nash_polygen_2020} to an input shape manage to produce concise meshes and preserve sharp features if these primitives (planes, quadric patches, etc) are diverse enough to describe the input mesh. However, they are rapidly compute-intensive even for a limited number of primitives. 
Deforming template shapes~\cite{groueix_papier-mache_2018, tan_variational_2018} or regular tetrahedral meshes~\cite{gao_learning_2020, shen_deep_2021} often constrains the output mesh genus and connectivity, and cannot guarantee intersection-free meshes. Another line of contributions approaches the design of learnable 3D representations from a different standpoint: provided a known reconstruction algorithm, what \emph{geometric estimates} should the neural network predict in order to extract a mesh off of it? Methods relying upon the popular Poisson surface reconstruction algorithm to extract a mesh~\cite{peng_shape_2021, prokudin_dynamic_2023} for instance inherit the topological advantages of an implicit surface representation, but they also suffer from their limitations by exhibiting overly-smoothed sharp features and a large amount of mesh elements. The Neural Marching Cubes (NMC~\cite{chen_neural_2021}) and Neural Dual Contouring (NDC~\cite{chen_neural_2022}) methods reduce some of these limitations, at the price of losing topological guarantees. Recently, VoroMesh~\cite{maruani_voromesh_2023}, computes a Voronoi diagram so that a subset of the 3D Voronoi facets best fit the input mesh, guaranteeing watertight results; 
but VoroMeshes are littered with small faces with spurious normal orientations, causing both visual artifacts and large mesh element counts. A few learning methods rely on a Delaunay triangulation instead~\cite{luo_deepdt_2021, rakotosaona_learning_2021, sulzer_scalable_2021, zhang_dmnet_2023}, but they are limited to the case of a fixed input pointset. We recap in Tab.~\ref{tab:related} some typical properties of explicit learnable 3D representations.

\begin{table}[htb]
\vspace*{-2mm}
  \begin{center}
    \resizebox{.85\linewidth}{!}{


\renewcommand{\arraystretch}{0.85}
\begin{tabular}{r|cccc}
& Arbitrary           & Sharp  & Watertight,  & Open \\
&   connectivity &  features & no self-int.&  surfaces
    \\ \hline
NDC~\cite{chen_neural_2022}     &     &     \color{black}\checkmark   &     &   \color{black}\checkmark \\ 
DMT~\cite{shen_deep_2021} &       &     \color{black}\checkmark   &     &    \\ 
SAP~\cite{peng_shape_2021}      &      &    &  \color{black}\checkmark   &   \\
DPF~\cite{prokudin_dynamic_2023}      &  \color{black}\checkmark       &    &  \color{black}\checkmark   &    \\
VoroMesh~\cite{maruani_voromesh_2023} &  \color{black}\checkmark       &    \color{black}\checkmark    &  \color{black}\checkmark   &    \\
PoNQ        & \color{black}\checkmark & \color{black}\checkmark  &  \color{black}\checkmark   & \color{black}\checkmark \\ \hline
\end{tabular}


}
    \vspace*{-2mm}
    \caption{Properties of explicit learnable 3D representations}
    \label{tab:related}
  \end{center}
  \vspace*{-6mm}
\end{table}

\noindent \textbf{Quadric Error Metrics.} Finally, we review a workhorse of surface approximation. In the context of mesh decimation, Garland and Heckbert~\cite{garland_surface_1997} introduced a Quadric Error Metric (QEM), encoded as a $4\!\times\!4$ matrix Q, which evaluates the sum of squared distances to a (possibly large) set of tangent planes. The use of QEM for their application was particularly appropriate: the sum $Q$ of two matrices $Q_i$ and $Q_j$ represent the sum of the squared distances to the union of the tangent planes used for $Q_i$ and those used for $Q_j$ -- a very economical proxy to all these tangent planes. Consequently, they proposed to use these QEM matrices to perform an efficient mesh decimation: the updated matrices encode the squared distances of all nearby removed facets, thus offering a very fast and low-memory evaluation of the distance to the original mesh throughout the decimation. After a few rounds of simplifications, the $\epsilon$-isovalue of QEM error near a vertex $v_i$ (i.e., the set of 3D positions $x$ such that $[x, 1]^t Q_i [x, 1] =\epsilon$) thus encodes the local region around $v_i$ of the initial surface: it will look like a \emph{pancake} if the original region was flat, an elongated ellipsoid if the region was a sharp feature (and the ellipsoid will precisely match the alignment of the sharp feature), or a sphere if the original region was near a corner. 
While QEM was adapted to deal with colors and texture coordinates~\cite{hoppe_new_1999}, spherical distances~\cite{thiery_sphere-meshes_2013}, mesh filtering~\cite{legrand_filtered_2019} and even variational mesh reconstruction from 3D pointsets~\cite{zhao_variational_2023}, the authors of~\cite{trettner_fast_2020} recently suggested a probabilistic version of QEM as a potential representation for learning-based tasks, while~\cite{agarwal_learning_2019} used QEM as a loss function. As we will see next, we will use a QEM matrix per discrete sample point in our learnable \emph{representation} for 3D shapes instead, to better learn the shape of local regions. 

\section{Method}
\label{sec:method}

We now delve into our PoNQ representation by first offering a more in-depth introduction to QEM and how we use it in Sec.~\ref{sec:motivation}. We then describe our representation in Sec.~\ref{sec:methodNQ} along with its use in learning tasks, and finally introduce our PoNQ mesh extraction approach in Sec.~\ref{sec:method_meshing}. 


\subsection{Motivation: Quadric Error Metrics}
\label{sec:motivation}

Given a sample point $s_k  \in  \mathbb{R}^3$ and a normal $n(s_k) \in \mathbb{R}^3$, the signed distance from a given location $x$ to the plane passing through $s_k$ that is normal to $n(s_k)$ is given simply as $d_{s_k, n(s_k)}(x) = (x - s_k)^t n(s_k)$. The squared distance can thus be rewritten as $d_{s_k, n(s_k)}(x)^2 = x^t A_k x - 2 b_k^t x + c_k$, where $3 \times 3$ matrix $A_k$, vector $b_k$, and scalar $c_k$ are:\vspace*{-.5mm}
\begin{equation}
    A_k = n(s_k) n(s_k)^t, \quad b_k = A_k s_k, \quad c_k = s_k^t A_k s_k. \vspace*{-.5mm}
\end{equation}
The sum of squared distances to a \emph{set of planes} thus yields: 
\vspace*{-.1mm}
\begin{align}
 \operatorname{QEM}(x) &= \sum_{k} d^2_{s_k, n(s_k)}(x) \nonumber\\
    & = x^t \underbracket{\bigl(\sum_k A_k \bigr)}_{\coloneqq A}  x - 2 {\underbracket{\bigl(\sum_k b_k\bigr)}_{\coloneqq b}} ^t x + \underbracket{\bigl(\sum_k c_k\bigr)}_{\coloneqq c} \label{eq:sumsAbc}\\
    & =  [x, 1]^t 
        \underbracket{\begin{pNiceArray}{cc|c}[margin,columns-width=auto]
  \Block{2-2}{A} &  & \Block{2-1}{-b}\\
\\
\hline
\Block{1-2}{-b^t} &  & \Block{1-1}{c}\\
\end{pNiceArray}}_{\coloneqq Q} [x, 1]. \label{eq:qem} \vspace*{-1mm}
\end{align}

The summed distances to a fine sampling $\{ s_k \}_k\!\coloneqq \! S$ of a small surface patch can then be concisely encoded by the QEM matrix $Q  \!\in\!  \mathbb{R}^{4 \times 4}$ from Eq.~\eqref{eq:qem}. Moreover, $Q$ can also indicate the best possible position to represent this surface patch with a single vertex: assuming the submatrix $A$ to be invertible (which is true in the general case~\cite{trettner_fast_2020}), the quadratic $\operatorname{QEM}$ function reaches its unique minimum at a location $v^* =  A^{-1} b$.
We leverage these quantities as part of a powerful 3D \emph{representation} suitable for machine learning.

\subsection{PoNQ representation}
\label{sec:methodNQ}

The PoNQ representation consists in a set  $\mathbf{P} \!=\!\{ \mathbf{p}_i \!\in\! \mathbb{R}^{3} \} $ of points, augmented with their local normals $\mathbf{N} \!=\!\{ \mathbf{n}_i \!\in\! \mathbb{R}^{3} \} $  and quadrics $\mathbf{Q} \!=\!\{ \mathbf{Q}_i \!\in\! \mathbb{R}^{4 \times 4} \} $  -- hence its name. In order to remove possible ambiguities, we will use a bold font to refer to the different quantities involved in the PoNQ representation, i.e., what will be \emph{optimized} or \emph{learned}.



\subsubsection{QEM-based representation via optimization}
\label{sec:method_rep}


In a pure optimization-based setting,  we want PoNQ to fit a watertight and non-self-intersecting input shape finely discretized by samples $s_k \!\in\! S$ with their local normals $n(s_k)$. We initialize the points $\mathbf{P}$ on a regular grid around the input shape (other initializations work equally well), and optimize their position to minimize the bi-directional Chamfer Distance $\operatorname{CD}$:
\vspace*{-.5mm}
\begin{align}
    \operatorname{CD}(\boldsymbol{ \mathbf{P}},S)=&\frac{1}{|\boldsymbol{ \mathbf{P}}|} \sum_{\mathbf{p}_i\in \boldsymbol{ \mathbf{P}}} \min_{s_k\in S} \|\mathbf{p}_i-s_k\|^2\\ &+  \frac{1}{| S|} \sum_{s_k\in S} \min_{\mathbf{p}_i\in\boldsymbol{ \mathbf{P}}} \|\mathbf{p}_i-s_k\|^2,
    \vspace*{-1mm} 
    \label{eq:CD}
\end{align}
The resulting point set will thus lie close to (and spread out over) the target surface. These points define a partition of $\mathbb{R}^3$ into Voronoi cells $V(\mathbf{p}_i)$ for which any location $x\!\in\!V(\mathbf{p}_i)$ has $\mathbf{p}_i$ as its closest point from $\mathbf{P}$, see~\cite{aurenhammer_voronoi_1991}.

\begin{figure}[b]
    \begin{subfigure}{.21\textwidth}
      \centering
      \includegraphics[width=\linewidth]{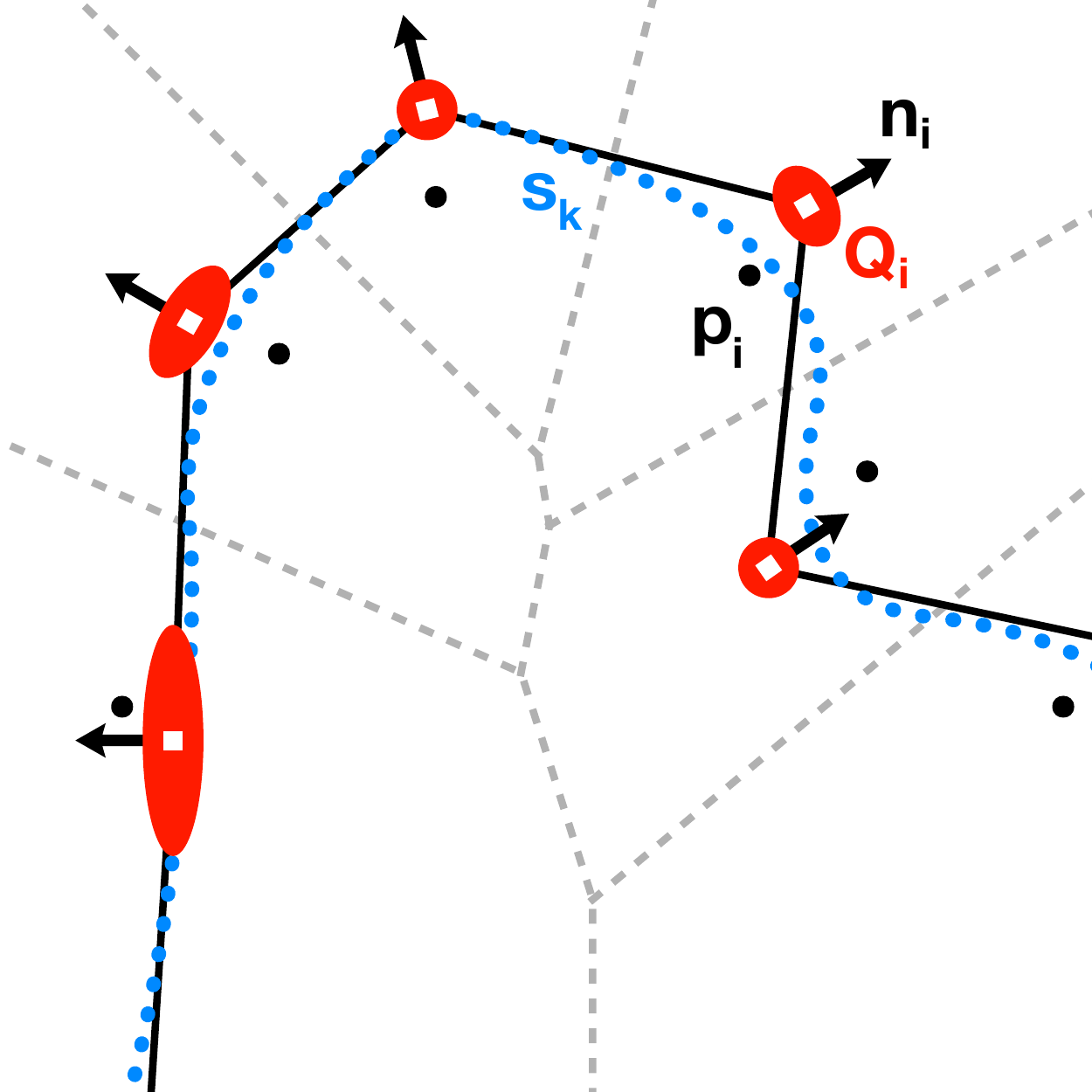} 
    \caption{\vspace*{2mm}\textit{Optimized PoNQ point set}}
    \end{subfigure}
    \hfill
    \begin{subfigure}{.21\textwidth}
      \centering    
      \includegraphics[width=\linewidth]{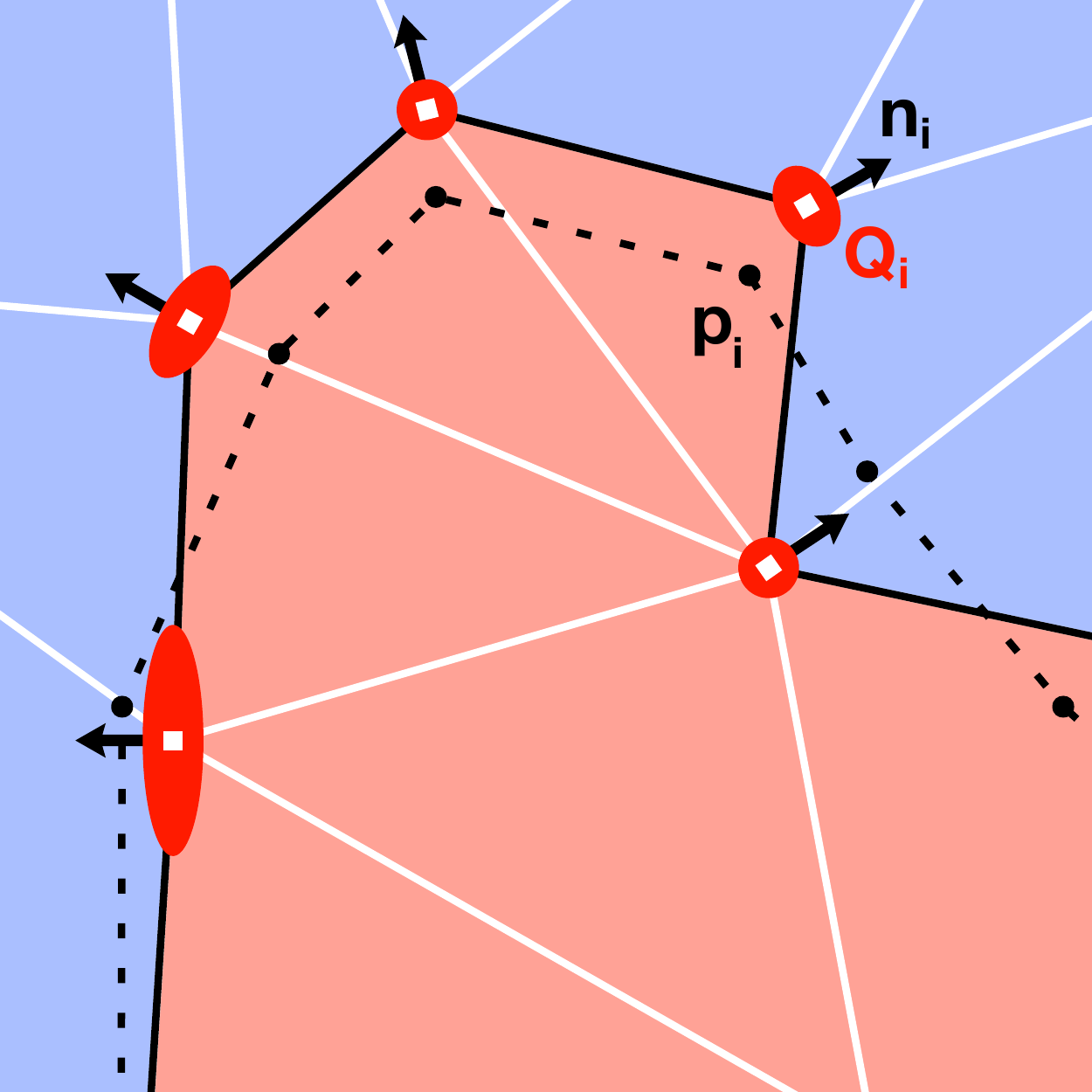}
    \caption{\vspace*{2mm}\textit{PoNQ meshing}}
    \end{subfigure}
    \vspace*{-3mm}
\caption{2D illustration of PoNQ: (a) a sampled ground-truth shape $S$ (blue dots) is represented by PoNQ as points $\mathbf{p}_i$ (whose Voronoi diagram (dotted lines) partitions the input samples), along with normals $\mathbf{n}_i$ and quadrics $\mathbf{Q}_i$ encoding the shape within each Voronoi cell. (b) The PoNQ mesh (black solid lines) is the boundary of the union of labeled tetrahedra from the Delaunay triangulation of the QEM-optimal vertices, providing a better fit than simply interpolating the points (black dotted lines). \vspace*{0mm}}
  \label{fig:2D_summary_method}
\end{figure}

Once the positions are optimized, we enrich each point $\mathbf{p}_i \!\in\! \boldsymbol{ \mathbf{P}}$ with a normal $\mathbf{n}_i$ and a QEM matrix $\textbf{Q}_i$, where  $\mathbf{n}_i$ represents the average of the sample normals $n(s_k)$ for all the samples $s_k$ contained within $V(\mathbf{p}_i)$ (see Fig.~\ref{fig:2D_summary_method}), i.e., 
\vspace*{-1mm}
\begin{equation}
    \mathbf{n}_i = \frac{1}{| S \cap  V (\mathbf{p}_i)|} \sum_{s_k \in \, S\cap V (\mathbf{p}_i)} n(s_k).
    \vspace*{-1mm}
\end{equation}
Similarly, $\mathbf{Q}_i$ is assembled using Eqs.~\eqref{eq:sumsAbc} and~\eqref{eq:qem} using the tangent planes implied by each sample (and its normal) within $V(\mathbf{p}_i)$, i.e., $\textbf{Q}_i$ is the QEM matrix such that 
\vspace*{-1mm}
\begin{equation}
    [x, 1]^t \,\mathbf{Q}_i\, [x, 1] = \sum_{s_k \in  S\cap V(\mathbf{p}_i)} d^2_{s_k, n(s_k)}(x). 
\vspace*{-2mm}
\end{equation}
Note that these additional variables are thus proxies for the local geometry around each point $\mathbf{p}_i$; points with no input samples within their Voronoi cell are simply discarded.
As explained in Section~\ref{sec:related}, each matrix $\mathbf{Q}_i$ implies, through its sub-matrices $\mathbf{A}_i$ and $\mathbf{b}_i$, an optimal location $\mathbf{v^*}\!\coloneqq\!\mathbf{A}_i^{-1}\mathbf{b}_i$ with respect to the input surface, hinting at the fact that storing $\mathbf{v^*}$ instead of  $\mathbf{b}_i$ is a possible alternative, which we will use in the learning context in the next section. We then rely on $\mathbf{Q}$ and $\mathbf{N}$ to extract the connectivity of the optimal vertices $\mathbf{v}^*$, which will be explained in Sec.~\ref{sec:method_meshing}.

\subsubsection{Learning with PoNQ} 
\label{sec:method_learning}

\begin{figure}[hbt]
  \centering
  \vspace*{-4mm} 
  \includegraphics[width=\linewidth]{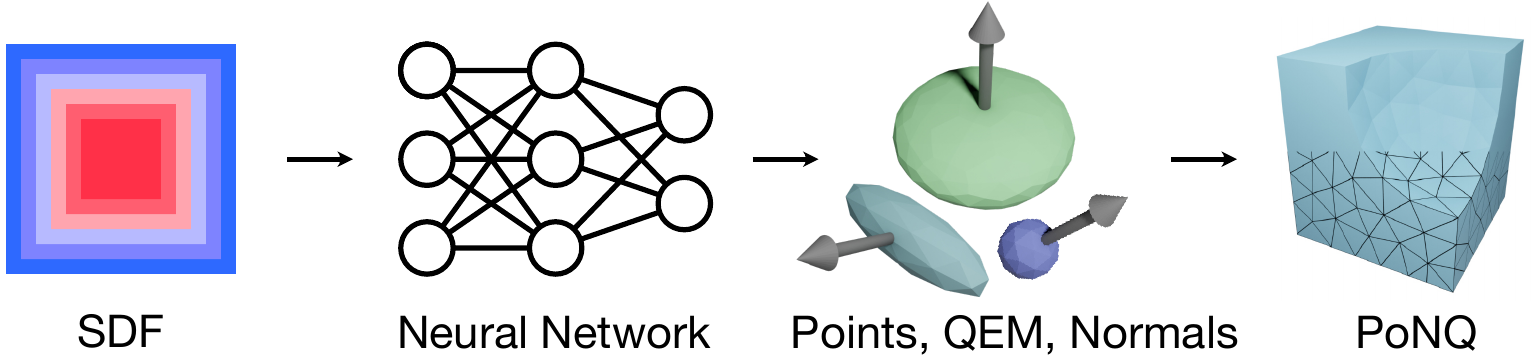}
\caption{Overview of our learning pipeline with PoNQ.
\vspace*{-2mm}
} 
  \label{fig:cnn_summary}
\end{figure}

We demonstrate the benefits of PoNQ in a learning context by applying it to reconstruction from Signed Distance Fields (SDF) (see Fig.~\ref{fig:cnn_summary}). As shown in several recent works~\cite{chen_neural_2021, chen_neural_2022, maruani_voromesh_2023}, this task is especially interesting as it enables the training of a \textit{local} model that can truly generalize to novel shapes outside the training set.

We use an architecture similar to NDC \cite{chen_neural_2022}, consisting of several convolutional neural networks (CNN). Unlike NMC~\cite{chen_neural_2021} (which uses eight points per voxel) and NDC (which uses only one), we can use an arbitrary number $P$ of predicted points per cell since our representation does not depend on a regular grid --- in practice, we found that $P = 4$ can represent sub-voxel details nicely without requiring too large networks.
We use a shared 5-layer encoder that converts the $(N \!+\! 1)^3$ input SDF grid 
into a $N^3$-sized grid of 128 features.
These features are processed by five separate 6-layer decoders (which do not share weights) to predict a PoNQ, i.e.,  $P \times N^3$ points $ \mathbf{p}_i$, along with their associated local normals $\mathbf{n}_i$ and QEM matrices $\mathbf{Q}_i$, to which we add a set $\mathbf{O}$ of $N^3$ binary ``occupancies'' $\mathbf{o}_i$ to mark the voxels containing the surface at inference time.

To stabilize the learning process and avoid having to perform matrix inversions, we do not store QEM matrices directly but only use the quadratic form $\mathbf{A}_i$ and the QEM-optimized vertex location $\mathbf{v^*}$ instead, from which one can reconstruct $\mathbf{Q}_i  =   \bigl(  \begin{smallmatrix}\scriptstyle 
\mathbf{A}_i & -\mathbf{A}_i \mathbf{v}_i^*\\
- (\mathbf{A}_i \mathbf{v}_i^*)^t & 0   
\end{smallmatrix}\bigr) .$
Furthermore, we store instead of $\mathbf{A}_i$ a $3  \times  3$ upper triangular matrix $\mathbf{U}_i$ such that $\mathbf{A}_i  =  \mathbf{U}_i\mathbf{U}_i^t$ is the reversed Cholesky decomposition of $\mathbf{A}_i,$ guaranteeing that $\mathbf{A}_i$ remains invertible.

We supervise our training with a collection of watertight shapes, each converted into a dense sampling $ S$. We preprocess the data by computing the ground-truth occupancy set $O_{gt}$ of the $m$ voxels containing samples, to which we restrict our loss terms (except for $L_\text{occ}$, which checks how well the trained occupancy set matches the ground-truth one).
At training time, we apply the sum of the following losses and backpropagate their gradients w.r.t. the bold variables: \vspace*{-5mm} 

\begin{align}
&L_{\operatorname{CD}} = \operatorname{CD}(\boldsymbol{ \mathbf{P}},  S) \quad (\text{\emph{see Eq.~\eqref{eq:CD}}})\\
 &L_\mathbf{n} = \frac{1}{m} \sum_{i=1}^m  \sum_{s_k\in  S\cap V(\mathbf{p}_i)} \|\mathbf{n}_i - n(s_k)\|^2 \\
 &L_\mathbf{A}  =  \frac{1}{m}   \sum_{i=1}^m  \sum_{s_k\in  S\cap V(\mathbf{p}_i)}   \| \mathbf{U}_i \mathbf{U}_i^t -  n(s_k) n(s_k)^t  \|^2 \\
 &L_{\mathbf{v^*}} = \frac{1}{m}   \sum_{i=1}^m 
  \sum_{s_k\in  S\cap V(\mathbf{p}_i)} 
  \left( n(s_k)^t (\mathbf{v}_i^\textbf{*} - s_k)  \right) ^2 \\
&L_\text{reg}=\frac{1}{m}   \sum_{i=1}^m\| \mathbf{v}_i^\textbf{*}-p_i\|^2 \ \ \text{($\mathbf{p}_i$ assumed fixed here)} \\
 &L_\text{occ} =\| \boldsymbol{\mathbf{O}} - O_\text{gt}\|^2 \vspace*{-2.5mm}
\end{align}

In the order in which they are listed above, these losses were designed to: help spread around the pointset $ \mathbf{P}$ and best fit the inputs; enforce that each normal is the mean normal of the local sample normals; enforce that the quadratic forms $\mathbf{A}_i$ correspond to the proper submatrix of the QEM of the local samples and normals around point $\mathbf{p}_i$; enforce, similarly, that $\mathbf{v}_i^\textbf{*}$ minimizes the sum of squared distance to the local tangent planes around point $\mathbf{p}_i$; regularize (with a very small coefficient) the positions of the optimal points $\mathbf{v}_i^\textbf{*}$ in flat regions --- as in this case, any point on this flat region is optimal in theory, so we force it to stay close to $\mathbf{p}_i$; and make sure that we match the ground-truth occupancy set. See \S 2.3 of the supplementary material for ablation studies.

\subsection{Meshing our representation}
\label{sec:method_meshing}

Given a PoNQ representation (either optimized or produced by a trained network), 
one can easily extract a mesh
by combining two approaches from computational geometry~\cite{amenta_new_1998, kolluri_spectral_2004} to ensure robustness.
Note that we do not even use the pointset $\mathbf{P}$ which only served to build a shape-adapted partition: we construct a mesh whose vertices are the QEM-optimal positions $\mathbf{v}_i^\textbf{*}$ as they capture features best.  We present here a concise overview of our meshing method; see our supplemental material for details.



\tightpara{Pre-processing} We first compute the Delaunay tetrahedralization of the optimal vertices $\mathbf{v}_i^\textbf{*}$ deriving from the QEM matrices $\mathbf{Q}$. The next two steps will tag each tetrahedron as either \emph{inside} or \emph{outside} based on local information, so that \emph{our final PoNQ mesh will be simply the triangle mesh forming the inside/outside boundary} --- ensuring watertightness and no self-intersections by design. 

\tightpara{Tagging obvious inside/outside tetrahedra.} We leverage the circumcenter criterion put forth in the Crust algorithm~\cite{amenta_new_1998}. In our case, each vertex $\mathbf{v}^*_i$ and its assigned normal $\mathbf{n}_i$ define an oriented plane: we tag a tetrahedron as \emph{outside} (resp., \emph{inside}) if both its circumcenter and barycenter are determined to be in the \emph{outside} (resp., \emph{inside}) half-space of each of the four vertices. Considering that the shape is contained within the convex hull of the $\mathbf{v}^*_i$ and that each vertex of the Delaunay tetrahedralization must be part of the final surface allows us to further tag a series of tetrahedra where there is no ambiguity; see our supplemental material for a detailed rationale. 

\tightpara{Tagging remaining tetrahedra.}
Delaunay-based meshing approach (like \emph{Crust}~\cite{amenta_new_1998}) require a dense point sampling (formally, an $\epsilon-$sampling), which is not compatible with our desire to deal with thin structures, sharp features and corners --- and this is the main reason why our earlier phase can end up not providing a tag for \emph{every} tetrahedra. To finish our tetrahedron tagging based on the ones we already have, we use a graph cut approach, inspired by (but simpler than) an existing spectral graph partitioning~\cite{kolluri_spectral_2004}. For each Delaunay triangle $T$, we compute a likelihood score $S(T)$ that evaluates how confident we are that this triangle is to appear on the final output mesh: this  triangle score evaluates the fitness of $T$ based on the local normals $\mathbf{N}$ and quadric matrices $\mathbf{Q}$ as explained in the supplemental material. We now tag the remaining undetermined tetrahedra through a \emph{minimum cut} of the Voronoi graph (in which each dual of a tetrahedron is a node, and each dual of a Delaunay triangle $T$ is an edge with weight $S(T)$) using the already-tagged "inside" ones as a source and the "outside" one as a drain.

\tightpara{Surface extraction.} We extract the final \textit{PoNQ mesh} as the triangle mesh forming the boundary between the inside and outside tetrahedra (see Fig.~\ref{fig:2D_summary_method}). As mentioned above, this automatically enforces by design the fact that our mesh is watertight and intersection-free.



\section{Experimental Results}
\label{sec:experiments}

\begin{figure}[t] 
\centering
 \vspace*{-1mm}
 \hspace*{-1mm}
 \begin{subfigure}{.09\textwidth}
  \centering
  \includegraphics[width=\linewidth]{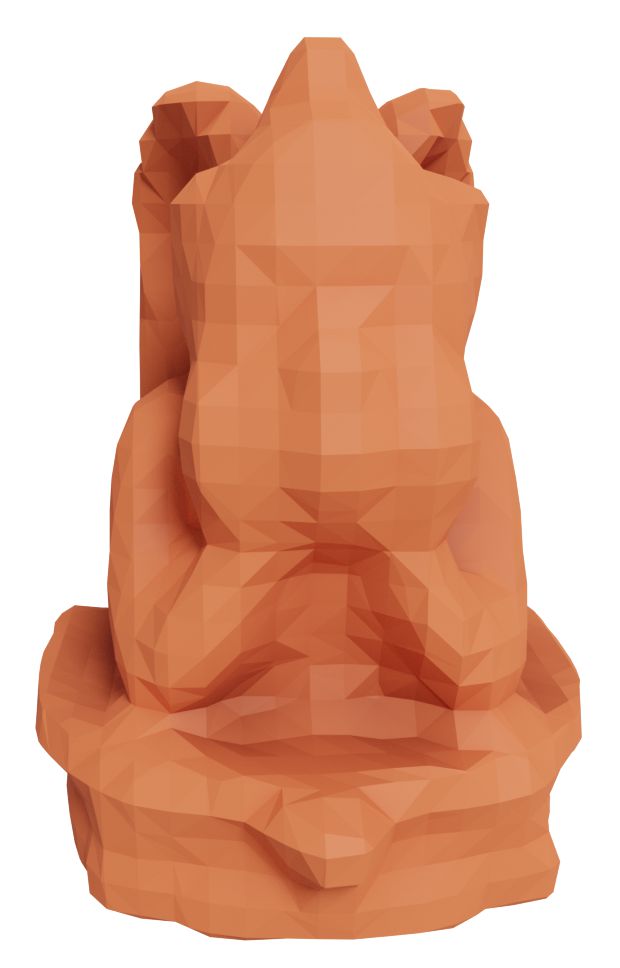} 
\end{subfigure}
 \begin{subfigure}{.09\textwidth}
  \centering
  \includegraphics[width=\linewidth]{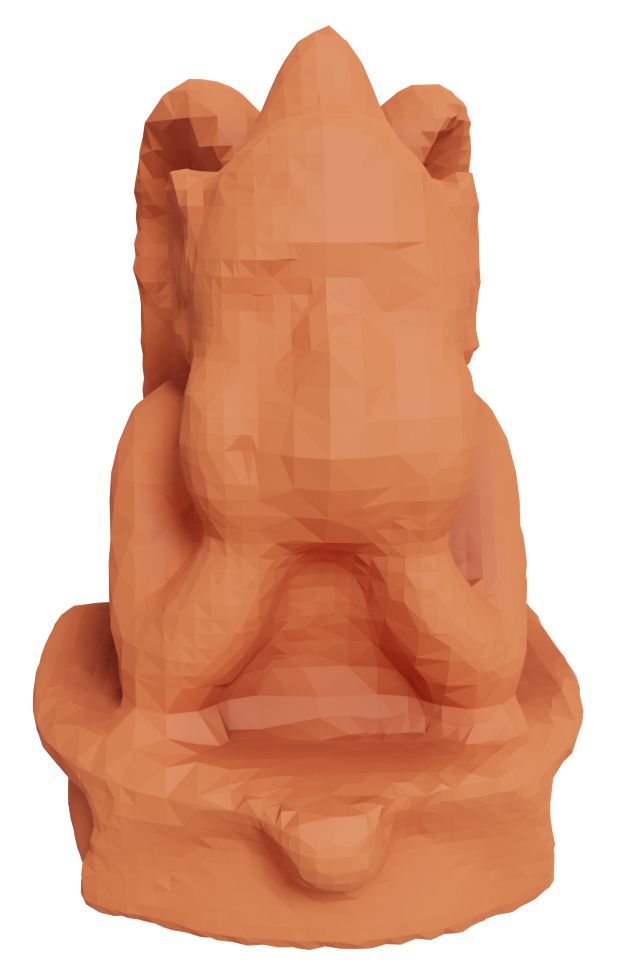} 
\end{subfigure}
 \begin{subfigure}{.09\textwidth}
  \centering
  \includegraphics[width=\linewidth]{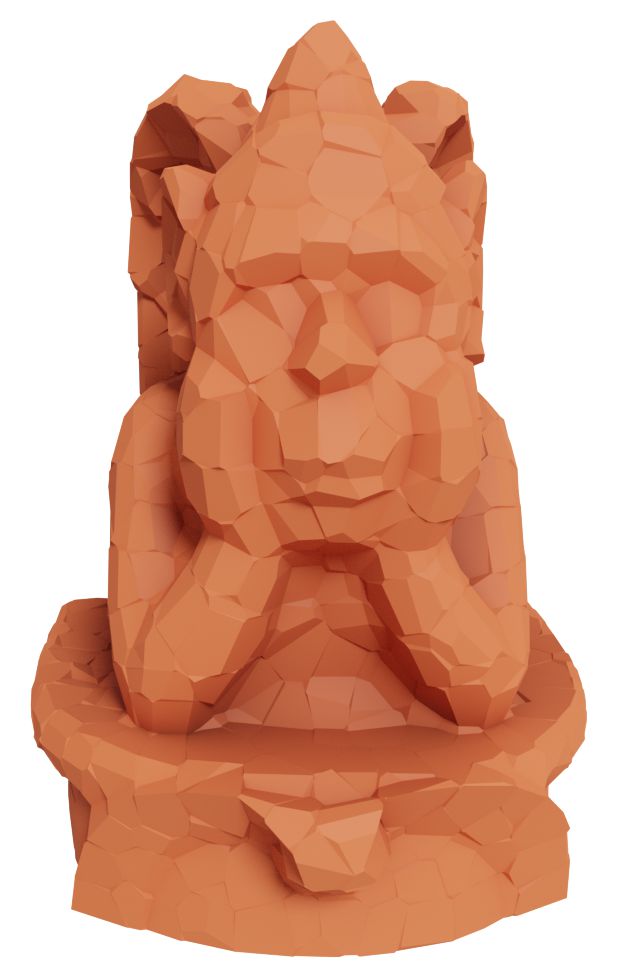} 
\end{subfigure}
 \begin{subfigure}{.09\textwidth}
  \centering
  \includegraphics[width=\linewidth]{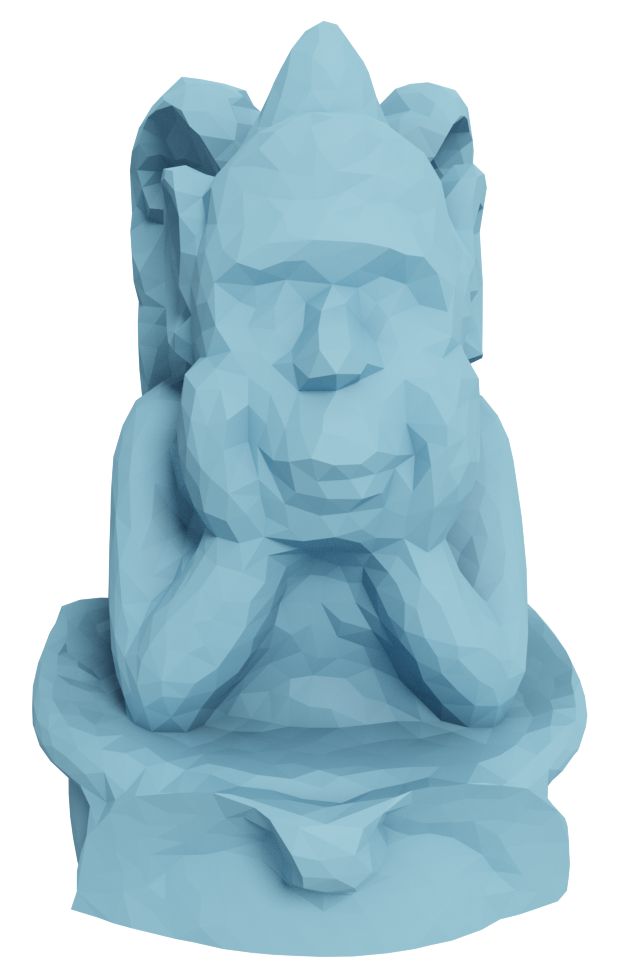} 
\end{subfigure}
 \begin{subfigure}{.09\textwidth}
  \centering
  \includegraphics[width=\linewidth]{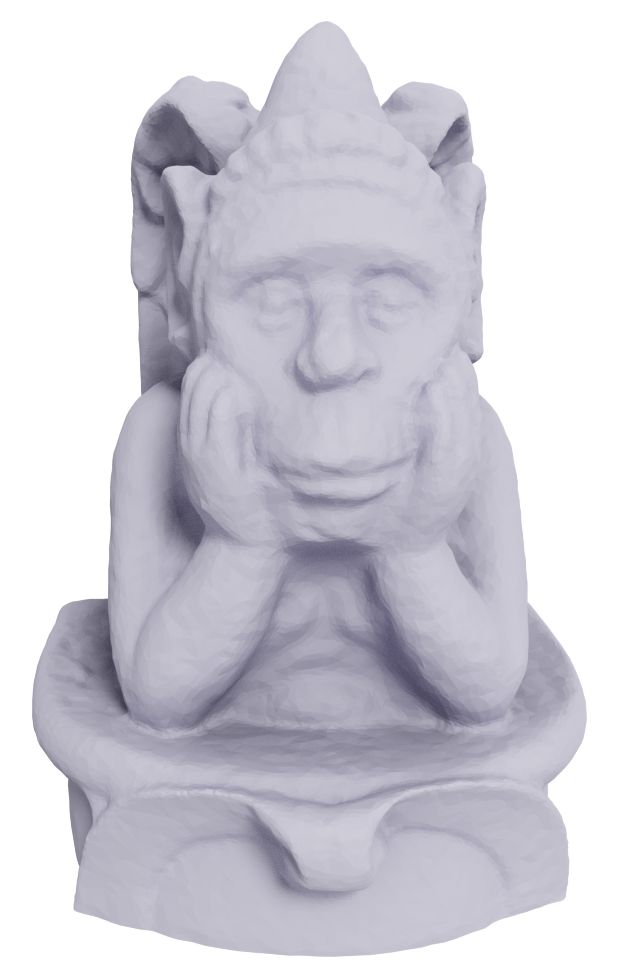} 
\end{subfigure}\\[-1mm]
 \hspace*{-1mm}
\begin{subfigure}{.09\textwidth}
  \centering
  \includegraphics[width=\linewidth]{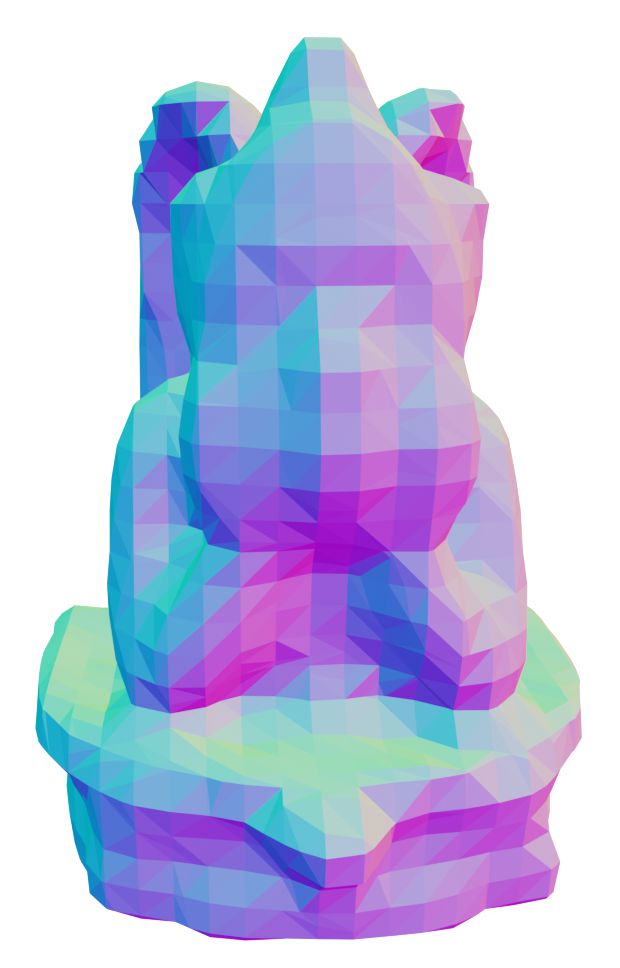} 
  \caption{SAP}
\end{subfigure}
 \begin{subfigure}{.09\textwidth}
  \centering
  \includegraphics[width=\linewidth]{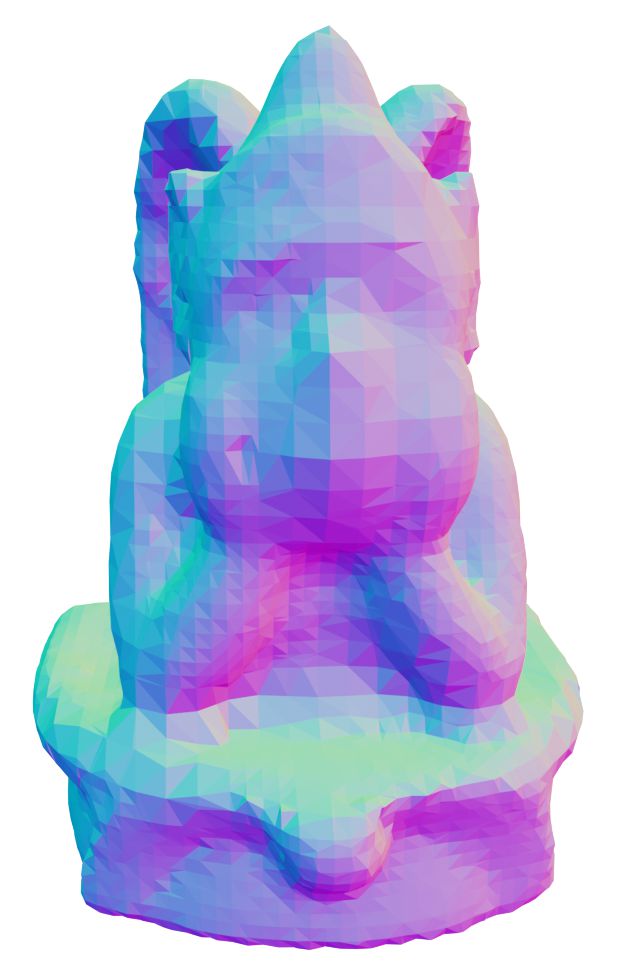} 
  \caption{DPF}
\end{subfigure}
 \begin{subfigure}{.09\textwidth}
  \centering
  \includegraphics[width=\linewidth]{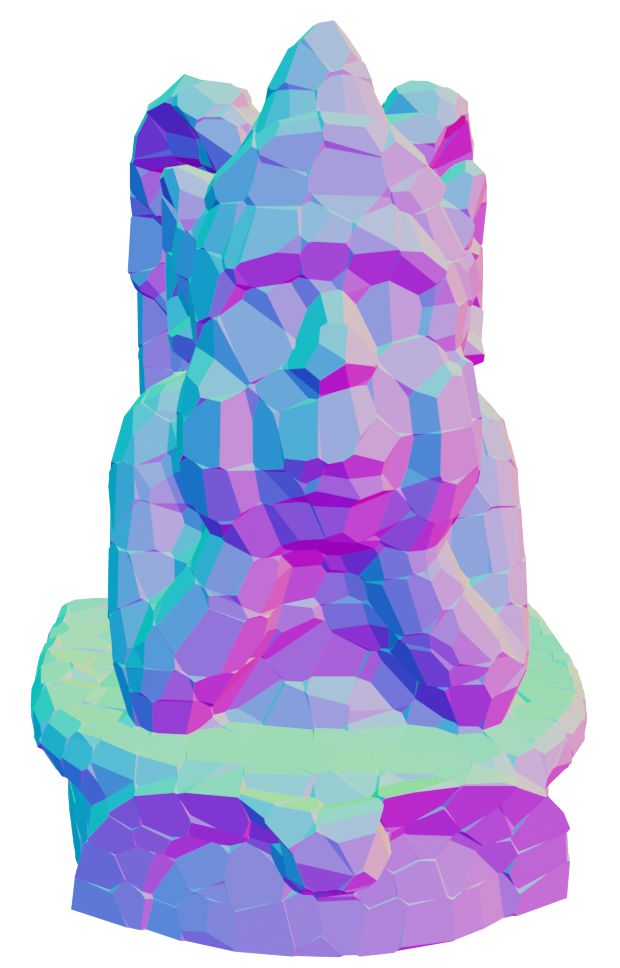}
  \caption{VoroMesh}
\end{subfigure}
 \begin{subfigure}{.09\textwidth}
  \centering
  \includegraphics[width=\linewidth]{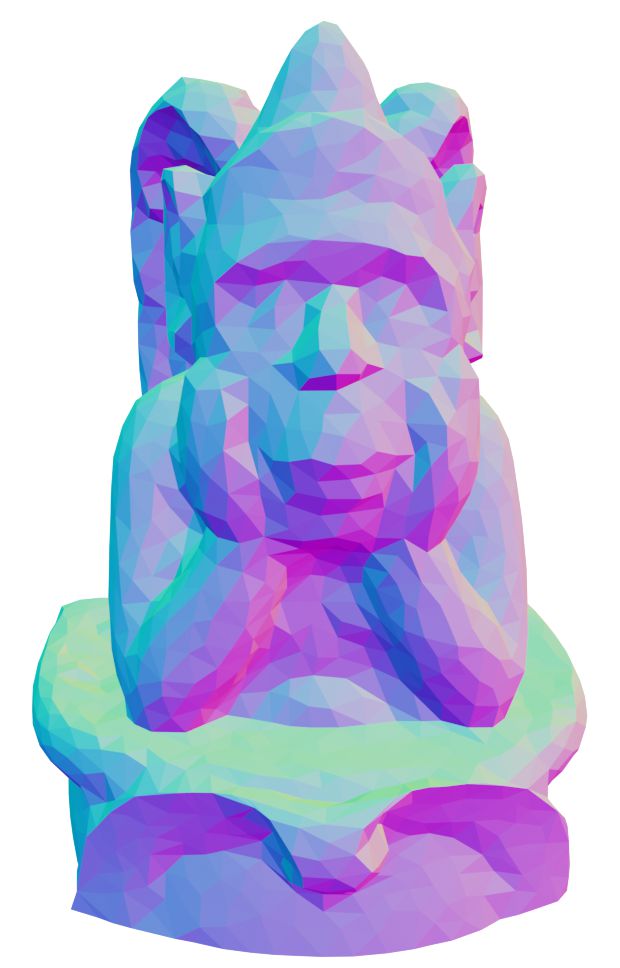}
  \caption{PoNQ}
\end{subfigure}
 \begin{subfigure}{.09\textwidth}
  \centering
  \includegraphics[width=\linewidth]{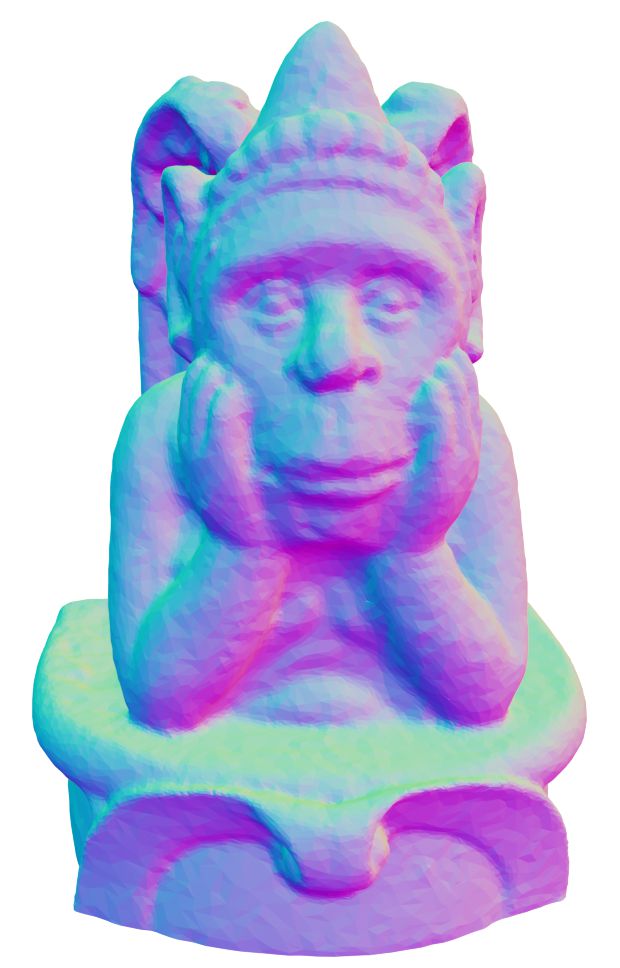} 
  \caption{Gr. Truth}
\end{subfigure}
 \vspace*{-3mm}
\caption{Optimization-based results $(32^3)$. \vspace*{-4mm}}
  \label{fig:optim}
\end{figure}

We implemented PoNQ for both optimization and learning tasks in Python, using PyTorch~\cite{paszke_pytorch_2019}, 
SciPy~\cite{virtanen_scipy_2020} and libigl~\cite{jacobson_libigl_nodate} --- our code is available on \href{https://nissmar.github.io/projects/ponq/}{our project page}. All timings were computed on a single workstation with 54 cores, a NVidia A6000 GPU and 512 GB of RAM.  

\subsection{Optimization-based 3D Reconstruction}
\label{sec:exp_optim}

\begin{table}[b] 
\vspace{-3mm} 
  \begin{center}
    \resizebox{1.\linewidth}{!}{

\begin{tabular}{l c c c c c c c c c  }
  \hline
  \textbf{Method}            & Grid & CD $\downarrow$               & F1  $\uparrow$  & NC $\uparrow$ & ECD $\downarrow$ & EF1 $\uparrow$  & \# V & \# F & Time \\
                             & Size & ($\times 10^{-5}$) &                            &                & ($\times 10^3$)  & ($\times 10^3$) & ($\times 10^3$)  & ($\times 10^3$) & (s)  \\
    \hline
  SAP~\cite{peng_shape_2021},  & $32^3$    &  6.475  &  0.589  &  0.894 & 0.235 & 0.058  & 1.6 & 3.2 &   51.2  \\
  DPF~\cite{prokudin_dynamic_2023}   & $32^3$    & 2.256  &  0.724  &  0.935 & 0.147 & 0.115  &  5.8 & 11.6 & 45.5 \\
  $\text{DPF}_{\text{chamfer}}$~\cite{prokudin_dynamic_2023}   & $32^3$    &   2.077  &  0.717  &  0.933 & 0.160 & 0.104 & 5.7 & 11.3 & 2.5 \\
  VoroMesh~\cite{maruani_voromesh_2023} & $32^3$    & \textbf{0.802}  &   \textbf{0.919}  &  0.957 & 0.257 & 0.242  & 5.9 & 11.8 &  2.0 \\
  PoNQ         & $32^3$    & 0.972  &  0.892  &  \textbf{0.961} & \textbf{0.106} & \textbf{0.447} & 2.3 & 4.6 &     2.6       \\
  
  \hline
  \hline
   SAP~\cite{peng_shape_2021},  & $64^3$    &  1.912  &  0.858  &  0.949 & 0.119 & 0.267 & 7.0 & 13.9 &  106.3   \\
  DPF~\cite{prokudin_dynamic_2023}   & $64^3$    &    0.795  &  0.909  &  0.971 & 0.104 & 0.435 & 22.6 & 45.1 & 51.5 \\
  $\text{DPF}_{\text{chamfer}}$~\cite{prokudin_dynamic_2023}   & $64^3$    & 0.797  &  0.907  &  0.970 & 0.094 & 0.415 & 24.1 & 48.3 & 4.0 \\
  VoroMesh~\cite{maruani_voromesh_2023} & $64^3$    & \textbf{0.645}  &  \textbf{0.938}  &  0.975 & 0.251 & 0.249 & 23.4 & 46.9 & 4.1    \\
  PoNQ         & $64^3$   &0.655  &  0.936  &  \textbf{0.981} & \textbf{0.061} & \textbf{0.645} & 10.1 & 20.1 &   4.1         \\
  
  \hline
  \hline
     SAP~\cite{peng_shape_2021},  & $128^3$    &    0.671  &  0.934  &  0.978 & 0.060 & 0.619 & 28.8 & 57.7 & 191.3 \\
  DPF~\cite{prokudin_dynamic_2023}   & $128^3$    &    0.644  &  0.938  &  0.986 & 0.089 & 0.665 & 90.0 & 180.0 & 71.0 \\
  $\text{DPF}_{\text{chamfer}}$~\cite{prokudin_dynamic_2023}   & $128^3$    &   0.644  &  0.938  &  0.986 & 0.086 & 0.664 & 97.3 & 194.5 & 17.7 \\
  VoroMesh~\cite{maruani_voromesh_2023} & $128^3$    & \textbf{0.634}  &  \textbf{0.939}  &  0.982 & 0.264 & 0.213  & 91.3 & 182.6 &  36.3 \\
  PoNQ         & $128^3$    & 0.637  &  \textbf{0.939}  &  \textbf{0.988} & \textbf{0.039} & \textbf{0.795} & 42.3 & 84.6 & 17.9           \\
  
  \hline
\end{tabular}}
    \vspace*{-2mm}
    \caption{Optimization-based results. Quantitative comparisons of Chamfer distance (CD), F1 score, and normal consistency (NC) on the Thingi30 dataset for three different grid resolutions. }
    \label{tab:direct}
    \vspace*{-3mm}
  \end{center}
\end{table}

For our tests in optimization-based 3D reconstruction, we use the 30 watertight shapes from the Thingi10k~\cite{zhou_thingi10k_2016} dataset chosen in VoroMesh~\cite{maruani_voromesh_2023} as it is arguably the most related and most recent neural representation with which to compare. 
We consider three grid resolutions $\text{res} \!\in\! [32^3, 64^3 , 128^3]$, and sample $1024 \times \text{res}^{2/3}$ surface samples. As described in Sec. \ref{sec:method_rep}, we use the chamfer distance as a loss to optimize the points for 400 epochs with the Adam optimizer before computing the mean normals and quadrics with our GPU-based implementation. We then extract the PoNQ representation as explained in Sec.~\ref{sec:method_meshing}.

\textbf{Baselines.} 
Besides VoroMesh, we also compare our results with Shape As Points (SAP)~\cite{peng_shape_2021} and Dynamic Point Field (DPF)~\cite{prokudin_dynamic_2023} as these three point-based neural representations all guarantee watertight outputs. We add a variant of DPF, trained with the Chamfer distance only (i.e., without the image-based loss), which we denote $\text{DPF}_{\text{chamfer}}$. For each method, we use the same number of optimized points. 

\textbf{Metrics.} We use the most common surface-based metrics, i.e.,  chamfer distance (CD), F-score (F1, with a threshold of $0.003$), and normal consistency (NC). In order to assess the reconstruction quality of sharp edges, we sample $10^5$ points on the edges featuring a dihedral angle larger than $\frac{\pi}{6}$, and compute the edge-chamfer distance (ECD) and edge-F-score (EF1, with a threshold of $0.005$) between the ground truth and the reconstruction samples (see supplemental material for details). We also report the number of triangles and faces of the extracted 3D models. Finally, we provide timings of the optimization step for all methods, as it is systematically the most time-consuming phase; see supplemental material for additional timings. 

\textbf{Results.} As SAP and DPF both rely on Poisson surface reconstruction, they smooth out sharp edges and small details. DPF does not optimize the extracted field, but rather leverages local information from which the mesh is assembled without supervision, leading to faster convergence and better scores than SAP for all metrics. As the evaluation point cloud is sampled from the reconstructed Poisson mesh (rather than the predicted points), the image-based loss does not significantly improve results, and $\text{DPF}_{\text{chamfer}}$ is faster than DPF with comparable performance.

VoroMesh exhibits finer details and better sharpness overall. With the same number of points as SAP, DPF and PoNQ, it has the best overall surface fitting scores due to the large number of Voronoi vertices and faces. However many of these faces have spurious normals and create undesired sharp edges, thus impacting the NC, ECD, and EF1 scores.

Qualitatively, PoNQ is better at dealing with sharp features than SAP and DPF and does not generate surface artifacts like VoroMesh (see Figure~\ref{fig:optim}). Quantitatively, our method yields the best normal consistency and sharp edge fitting scores (see Tab.~\ref{tab:direct}). In terms of surface fitting, it matches the considered baselines at resolution $128^3$, and comes close second behind VoroMesh at resolutions $32^3$ and $64^3$, but with significantly lower face counts. We will discuss in Sec.~\ref{sec:extensions} how one can further lower the face count of PoNQ meshes at very little cost on the scores (see Fig.~\ref{fig:lite}), setting PoNQ apart even more prominently. 

\subsection{Learning-based 3D reconstruction}

\subsubsection{Reconstruction from SDF}
We now assess the behavior of our PoNQ representation in the learning-based task of 3D shape reconstruction from SDF grids. We train and evaluate our method on the CAD shapes of the ABC dataset~\cite{koch_abc_2019}. We also assess the generalizability of our method on the free-form shapes of the Thingi10k~\cite{zhou_thingi10k_2016} dataset, without any fine-tuning. For fairness, we use the train/test split provided in VoroMesh~\cite{maruani_voromesh_2023}: 3,843/962 in the training and testing set for ABC, with 30 watertight validation shapes for Thingi10k. We train our network for 600 epochs while increasing the number of sampled points and decreasing the learning rate and regularization -- see supplemental material for additional details.

\textbf{Baselines.} We compare our method against the three most closely related baselines: NMC~\cite{chen_neural_2021}, NDC~\cite{chen_neural_2022} and VoroMesh~\cite{maruani_voromesh_2023} using the authors' code and their best pre-trained model (we also discuss comparisons with RTS~\cite{sellan_reach_2023} and DMTet~\cite{shen_deep_2021} in the supplementary material). 

\textbf{Metrics.} We use the same evaluation metrics that were already mentioned in Sec.~\ref{sec:exp_optim}.

\begin{figure}[t]
\centering
\hspace*{-1mm}
 \begin{subfigure}{.09\textwidth}
  \centering
  \includegraphics[width=\linewidth]{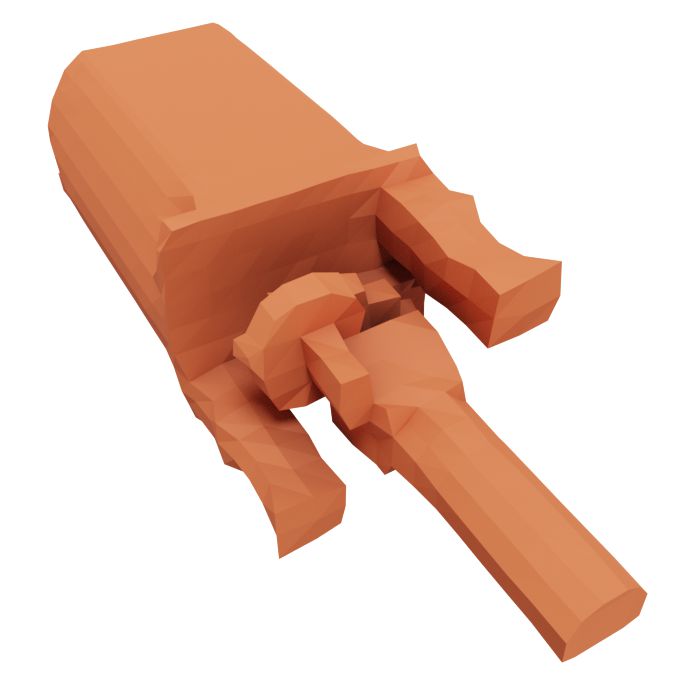} 
\end{subfigure}
 \begin{subfigure}{.09\textwidth}
  \centering
  \includegraphics[width=\linewidth]{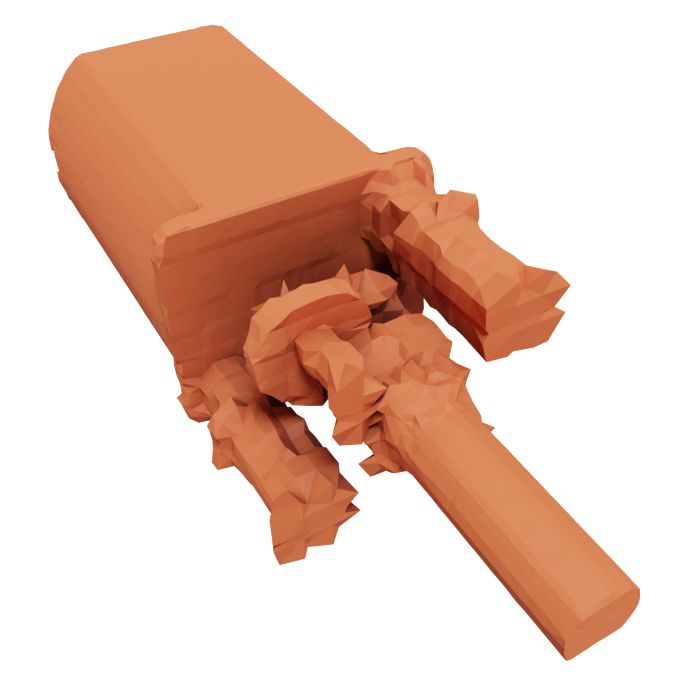} 
\end{subfigure}
 \begin{subfigure}{.09\textwidth}
  \centering
  \includegraphics[width=\linewidth]{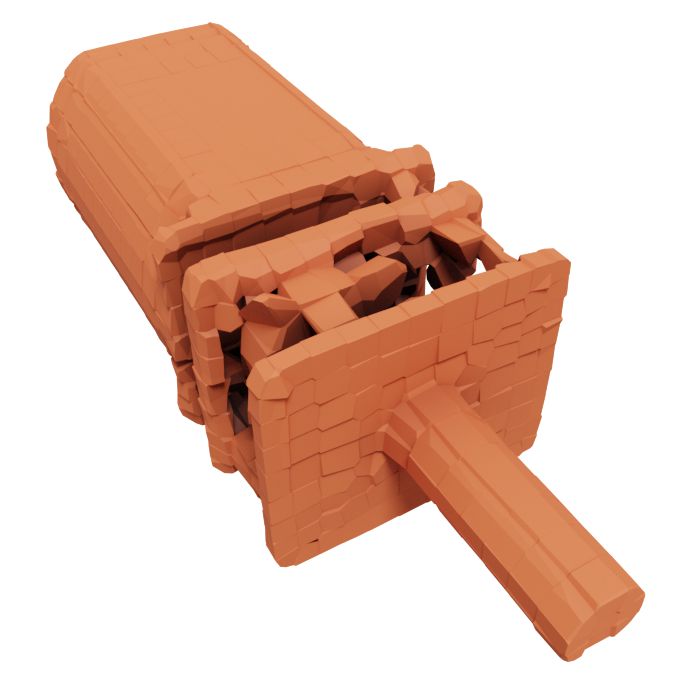} 
\end{subfigure}
 \begin{subfigure}{.09\textwidth}
  \centering
  \includegraphics[width=\linewidth]{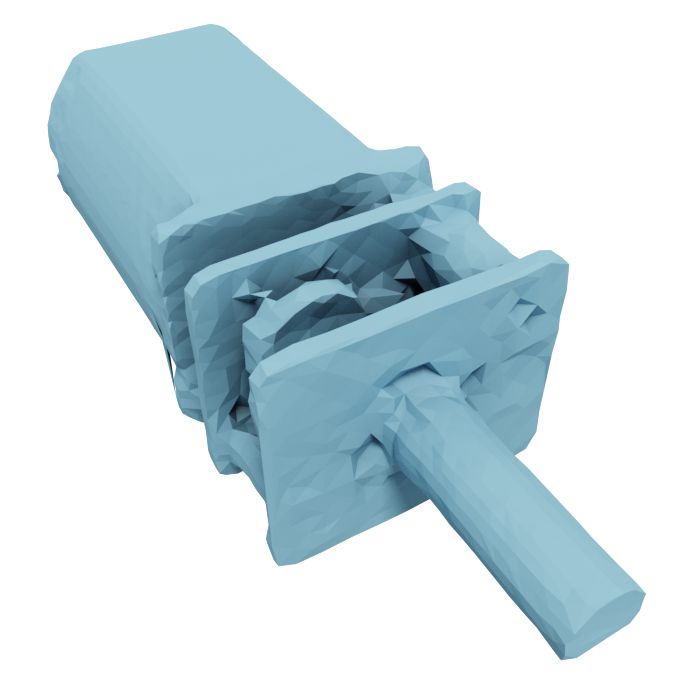} 
\end{subfigure}
 \begin{subfigure}{.09\textwidth}
  \centering
  \includegraphics[width=\linewidth]{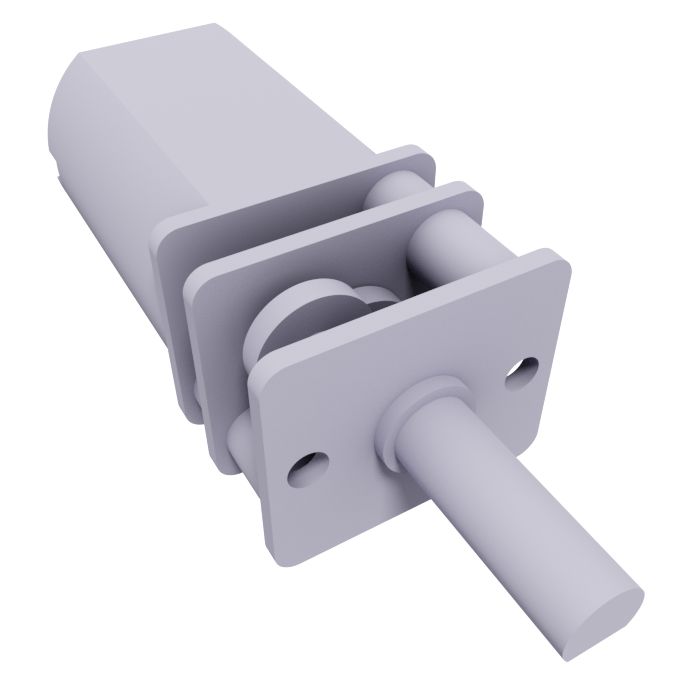} 
\end{subfigure}
 \hspace*{-1mm}
 \begin{subfigure}{.09\textwidth}
  \centering
  \includegraphics[width=\linewidth]{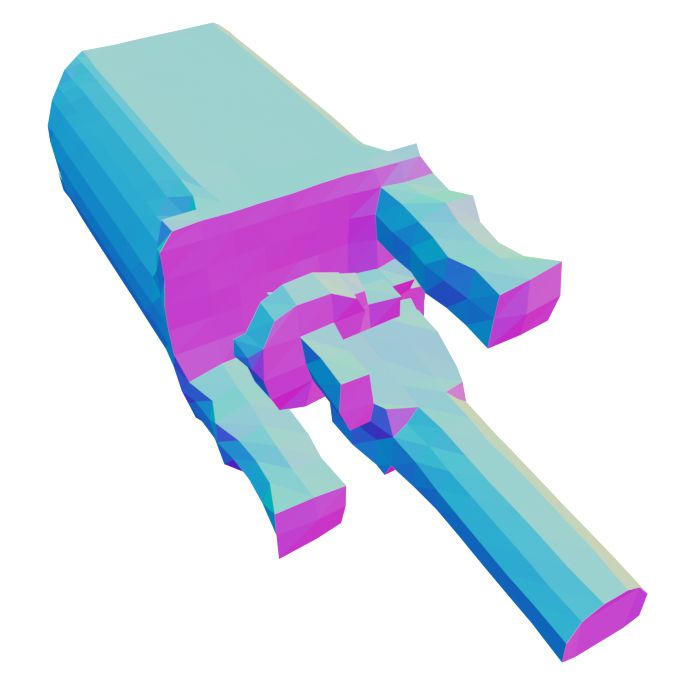} 
\end{subfigure}
 \begin{subfigure}{.09\textwidth}
  \centering
  \includegraphics[width=\linewidth]{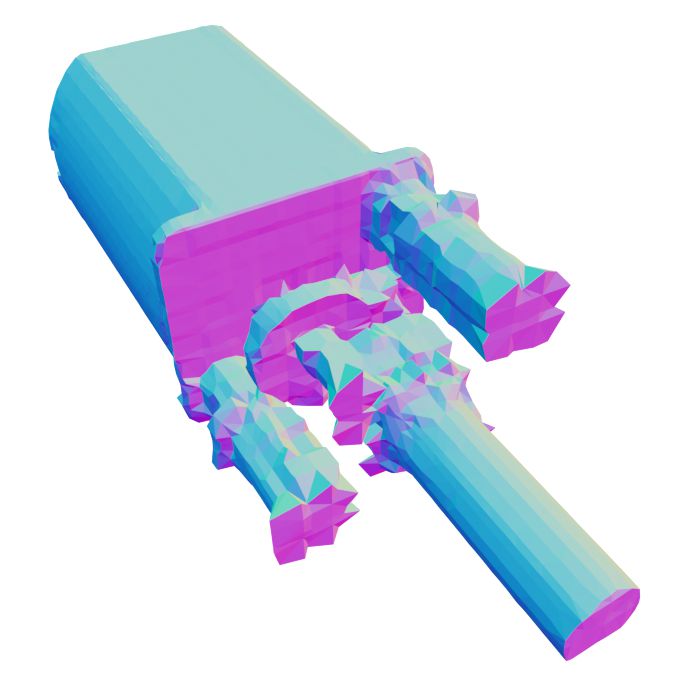} 
\end{subfigure}
 \begin{subfigure}{.09\textwidth}
  \centering
  \includegraphics[width=\linewidth]{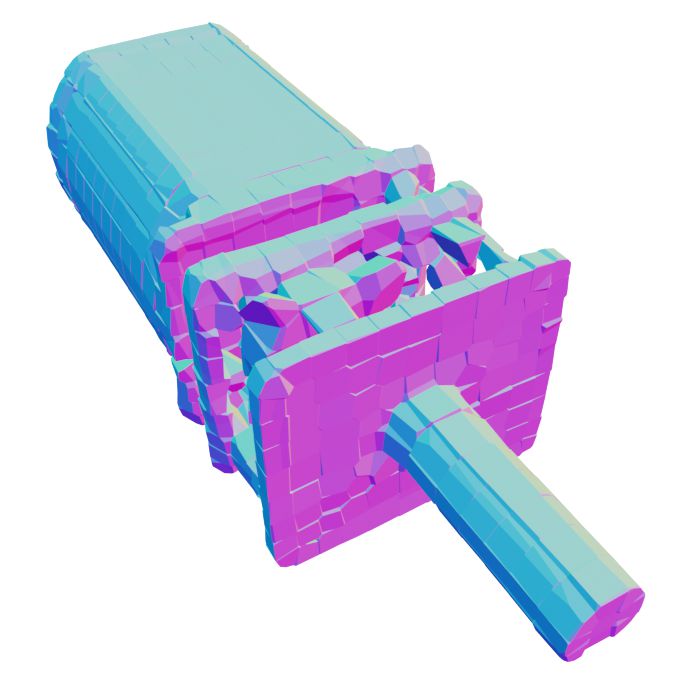} 
\end{subfigure}
 \begin{subfigure}{.09\textwidth}
  \centering
  \includegraphics[width=\linewidth]{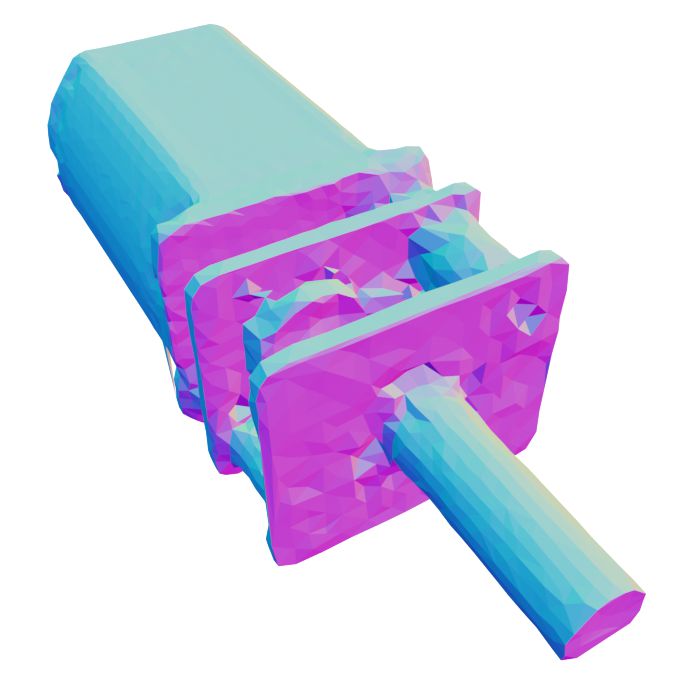} 
\end{subfigure}
 \begin{subfigure}{.09\textwidth}
  \centering
  \includegraphics[width=\linewidth]{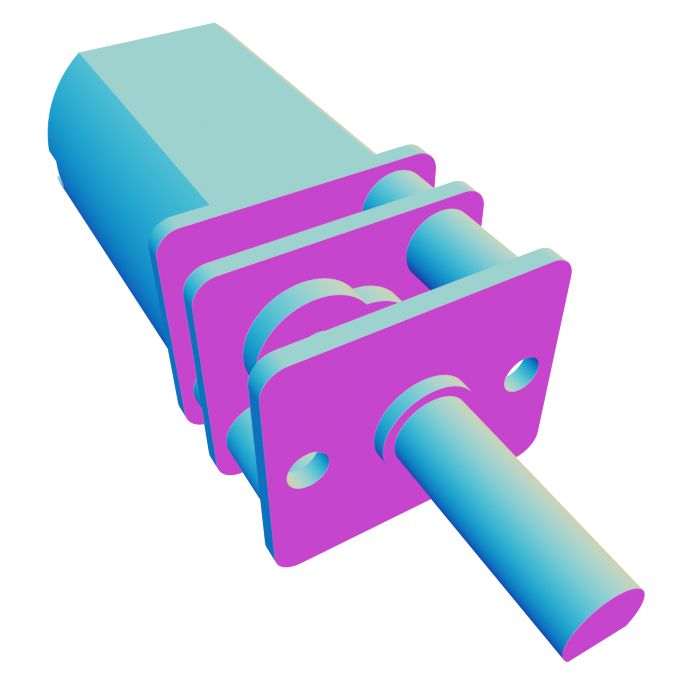} 
\end{subfigure}
 \begin{subfigure}{.09\textwidth}
  \centering
  \includegraphics[width=\linewidth]{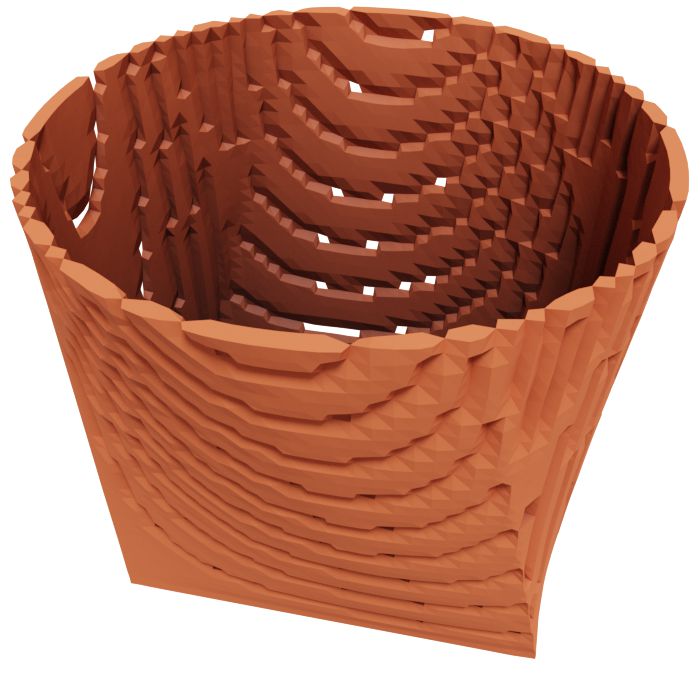} 
\end{subfigure}
 \begin{subfigure}{.09\textwidth}
  \centering
  \includegraphics[width=\linewidth]{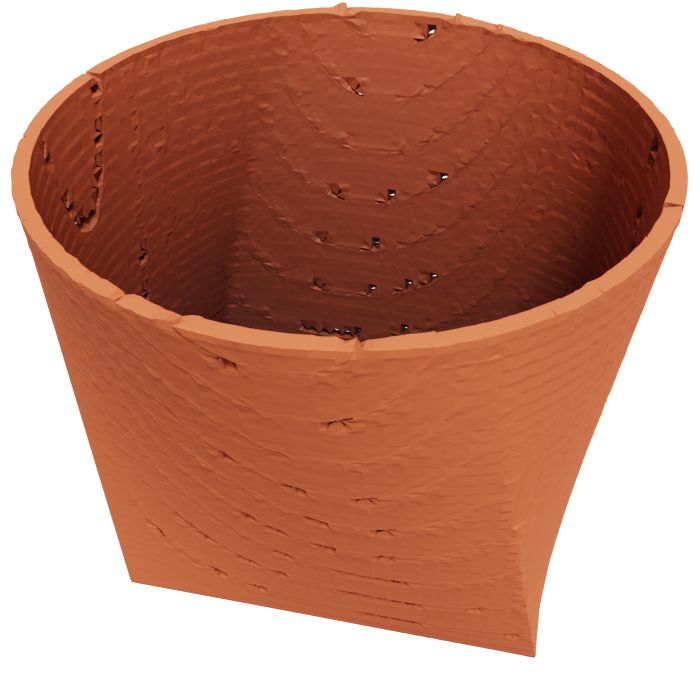} 
\end{subfigure}
 \begin{subfigure}{.09\textwidth}
  \centering
  \includegraphics[width=\linewidth]{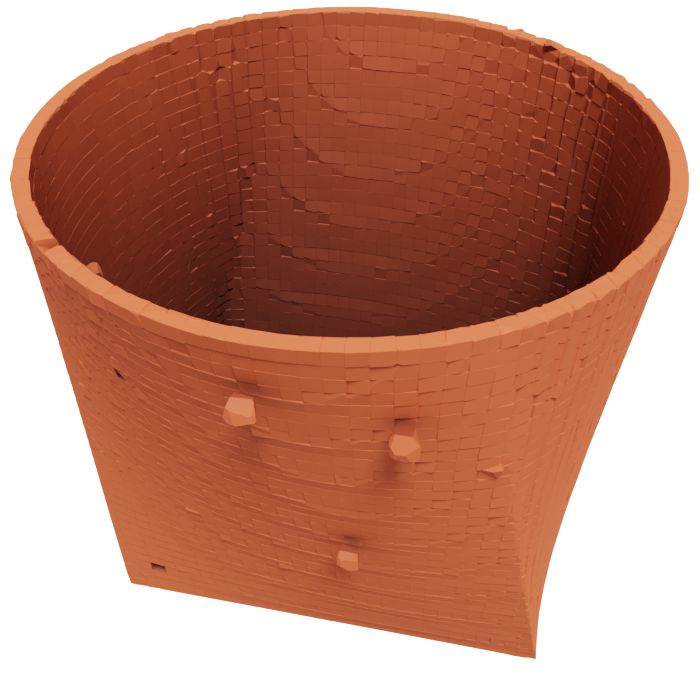} 
\end{subfigure}
 \begin{subfigure}{.09\textwidth}
  \centering
  \includegraphics[width=\linewidth]{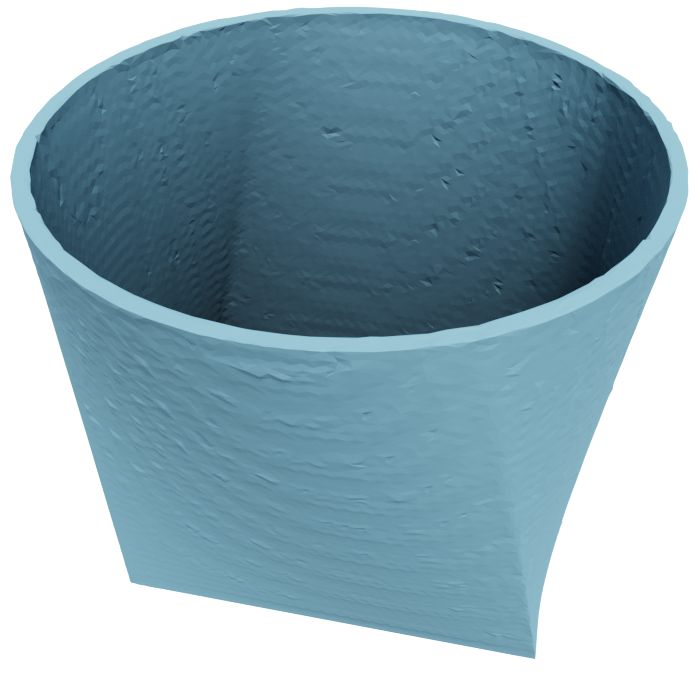} 
\end{subfigure}
 \begin{subfigure}{.09\textwidth}
  \centering
  \includegraphics[width=\linewidth]{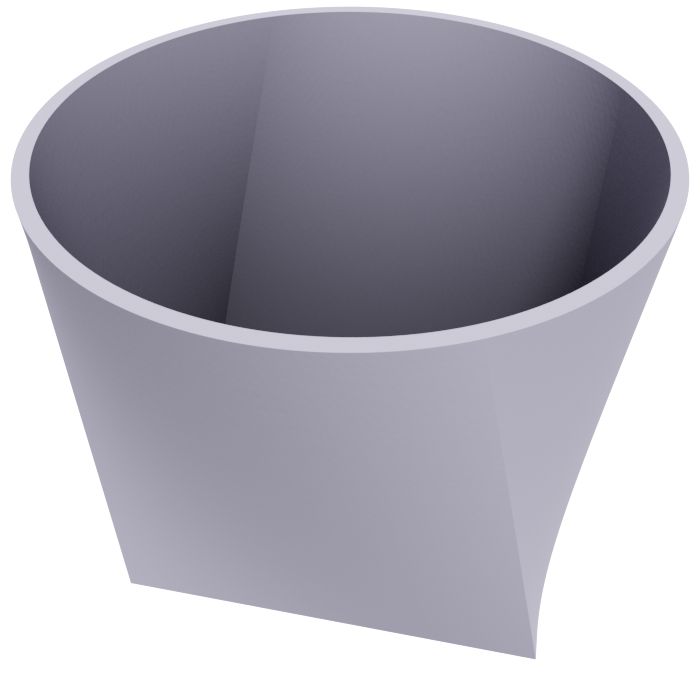} 
\end{subfigure}
 \begin{subfigure}{.09\textwidth}
   \hspace*{-1mm}
  \includegraphics[width=\linewidth]{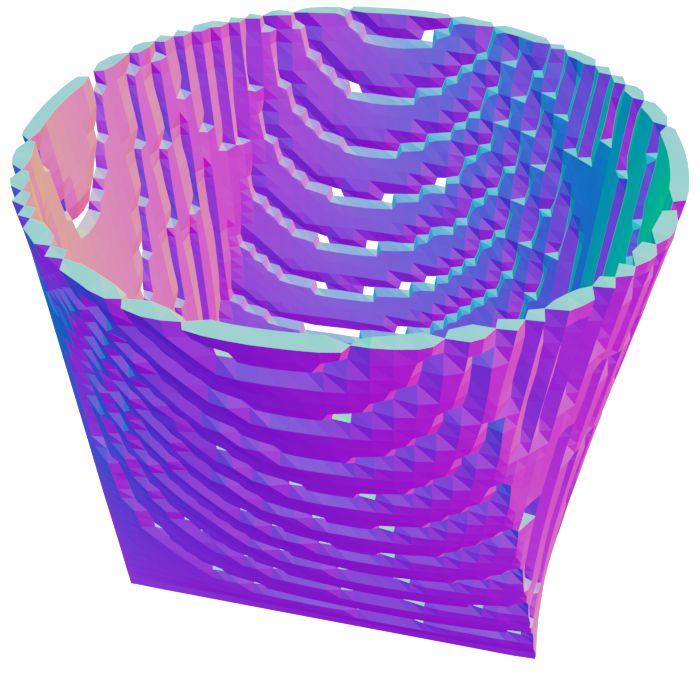}
  \vspace*{-5mm}
   \caption{NDC}
\end{subfigure}
 \begin{subfigure}{.09\textwidth}
  \centering
  \includegraphics[width=\linewidth]{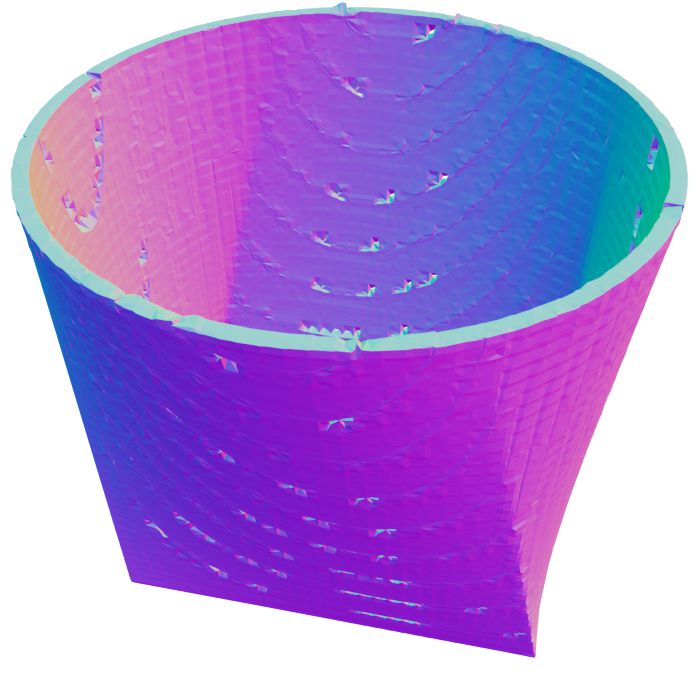}
  \vspace*{-5mm}
    \caption{NMC}
\end{subfigure}
 \begin{subfigure}{.09\textwidth}
  \centering
  \includegraphics[width=\linewidth]{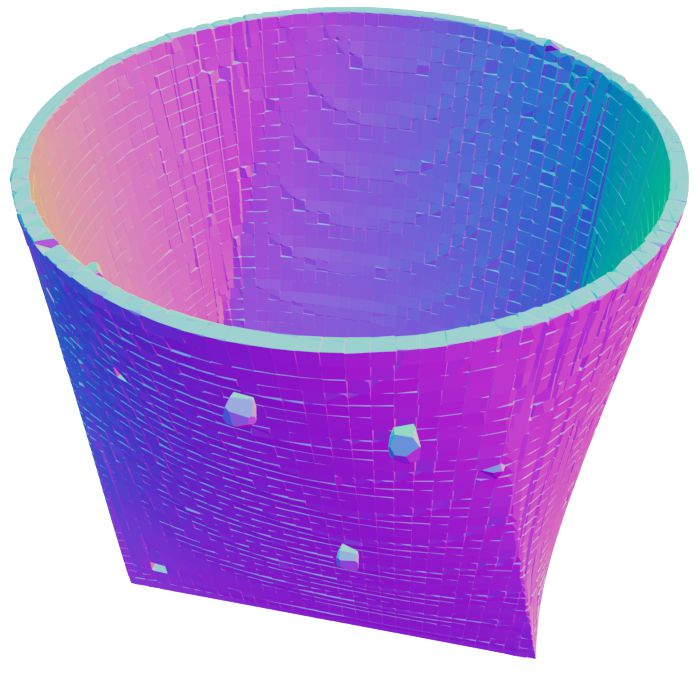}
  \vspace*{-5mm}
  \caption{VoroMesh}
\end{subfigure}
 \begin{subfigure}{.09\textwidth}
  \centering
  \includegraphics[width=\linewidth]{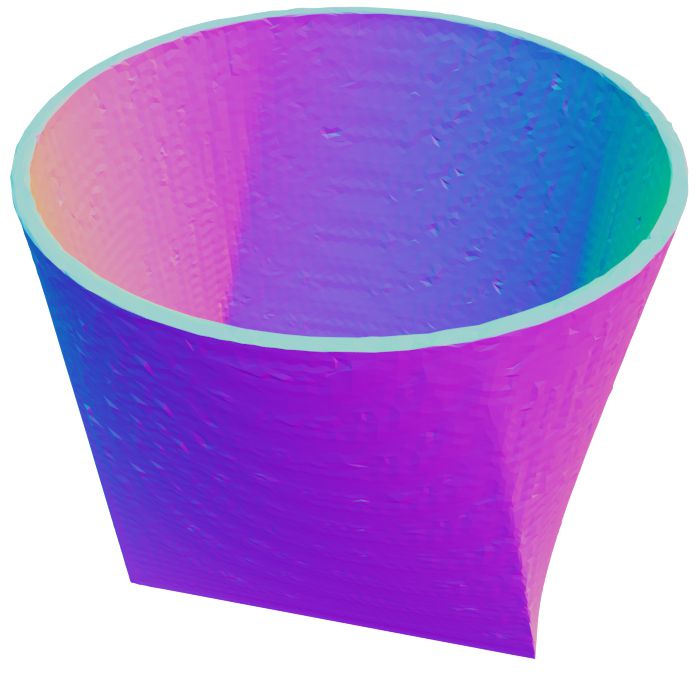}
  \vspace*{-5mm}
   \caption{PoNQ}
\end{subfigure}
 \begin{subfigure}{.09\textwidth}
  \centering
  \includegraphics[width=\linewidth]{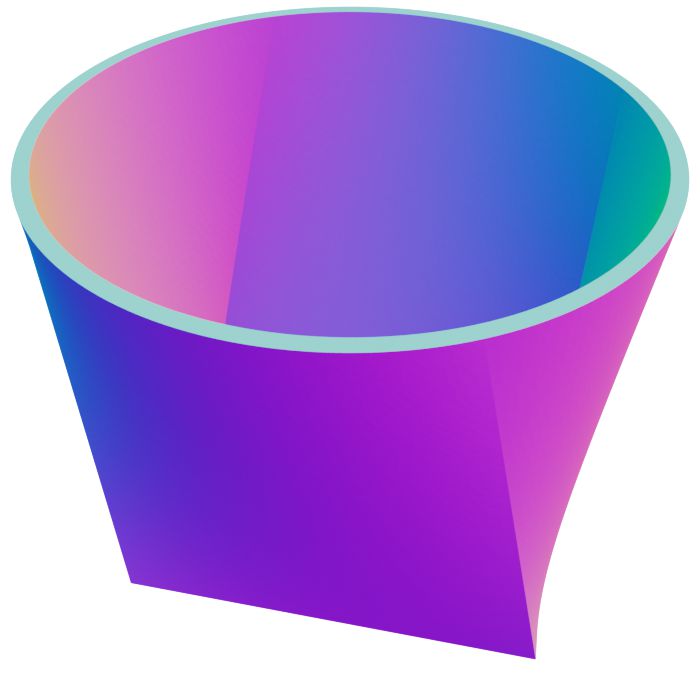} 
  \vspace*{-5mm}
   \caption{Gr. Truth}
\end{subfigure} 
\vspace*{-1mm}
\caption{Learning results (top: $32^3$; bottom: $64^3$) on ABC.
 \vspace*{-1mm}}
  \label{fig:sdf_abc}
\end{figure}

\begin{table}[t] \vspace*{0mm}
  \begin{center}
    \resizebox{\linewidth}{!}{\begin{tabular}{l c c c c c c c c c }
  \hline
  \textbf{Method}            & Grid & CD $\downarrow$               & F1  $\uparrow$  & NC $\uparrow$ & ECD $\downarrow$ & EF1 $\uparrow$ & Watertight $\uparrow$ & \# V & \# F   \\
                             & Size & ($\times 10^{-5}$) &  &   &   &           & no self-int. & ($\times 10^3$)  & ($\times 10^3$)  \\

  \hline

  NDC~\cite{chen_neural_2022}  & $32^3$    & 66.004  &  0.787  &  0.941 & 0.445 & 0.658  & 44\%  & 1.3 & 2.6  \\
  NMC~\cite{chen_neural_2021}  & $32^3$    & 60.755  &  0.833  &  0.954 & 0.350 & 0.693 & 26\% & 9.7 & 19.3 \\

  VoroMesh~\cite{maruani_voromesh_2023}          & $32^3$    & 2.228  &  0.835  &  0.941 & 0.802 & 0.232 & \textbf{100}\%    & 10.0 & 20.0       \\
  PoNQ-lite & $32^3$    & 3.539  &  0.810  &  0.953 & 0.296 & 0.658 & \textbf{100}\%  & 1.3 & 2.6 \\

  PoNQ & $32^3$    & \textbf{1.514}  &  \textbf{0.852}  &  \textbf{0.964} & \textbf{0.184} & \textbf{0.713} & \textbf{100}\%  & 5.1 & 10.2     \\
  
  \hline
  NDC~\cite{chen_neural_2022}  & $64^3$    & 2.211  &  0.882  &  0.975 & 0.223 & 0.855    & 23\%  & 5.5 & 11.0     \\
  NMC~\cite{chen_neural_2021}  & $64^3$    & 2.138  &  0.891  &   \textbf{0.980} & 0.254 & 0.854   & 18\% & 42.8 & 85.5 \\

  VoroMesh  ~\cite{maruani_voromesh_2023}        & $64^3$    & 1.219  &  0.886  &  0.966 & 0.796 & 0.207   & \textbf{100}\%  & 38.4 & 76.9     \\
  PoNQ-lite & $64^3$    & 1.074  &  0.888  &  0.978 & 0.128 & 0.858 & \textbf{100}\%  &  5.5 & 10.9  \\
PoNQ   & $64^3$   & \textbf{0.886}  &  \textbf{0.892}  &  \textbf{0.980} & \textbf{0.109} & \textbf{0.866} & \textbf{100}\%  &  21.2 & 42.3   \\

 \hline

NDC~\cite{chen_neural_2022}  & $128^3$  & 1.889  &   \textbf{0.896}  &  0.983 & 0.095 &  \textbf{0.947} &     14\%  &  22.1 & 44.2 \\
NMC~\cite{chen_neural_2021}  & $128^3$ 	  & 1.888  &   \textbf{0.896}  &   \textbf{0.984} & 0.349 & 0.859 &   12\%   &  175.9 & 351.9 \\
VoroMesh~\cite{maruani_voromesh_2023} & $128^3$                         & 1.069  &  0.894  &  0.974 & 0.792  & 0.189 &         \textbf{100}\% & 149.1 & 298.2 \\
PoNQ-lite			& $128^3$ 			  & 1.007  &   \textbf{0.896}  &   \textbf{0.984} &  \textbf{0.043} & 0.933 &     \textbf{100}\%  &  21.9 & 43.8 \\
PoNQ				& $128^3$  		          &  \textbf{0.920}  &   \textbf{0.896}  &   \textbf{0.984} & 0.191 & 0.878 &     \textbf{100}\%    &   85.7 & 171.2  \\

  \hline\label{tab:sdf_abc}
\end{tabular}}  
  \end{center}\vspace*{-8mm}
    \caption{Results on ABC with our network trained on ABC.
    } \vspace*{-3.5mm}
    \label{tab:sdf_abc_tab}
\end{table}

\begin{figure}[t]
\centering
  \hspace*{-1mm}
 \begin{subfigure}{.09\textwidth}
  \centering
  \includegraphics[width=\linewidth]{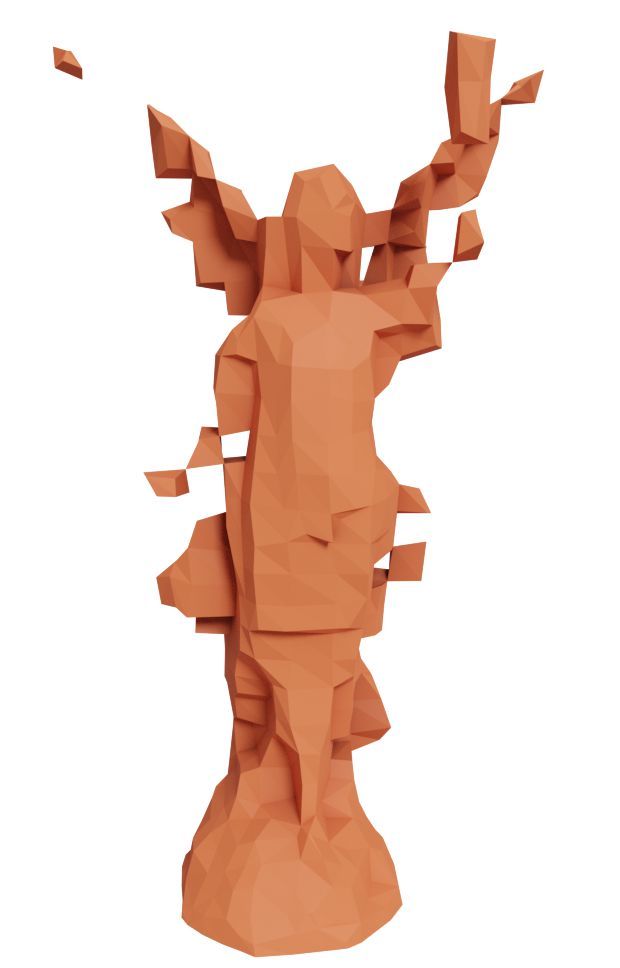} 
\end{subfigure}
 \begin{subfigure}{.09\textwidth}
  \centering
  \includegraphics[width=\linewidth]{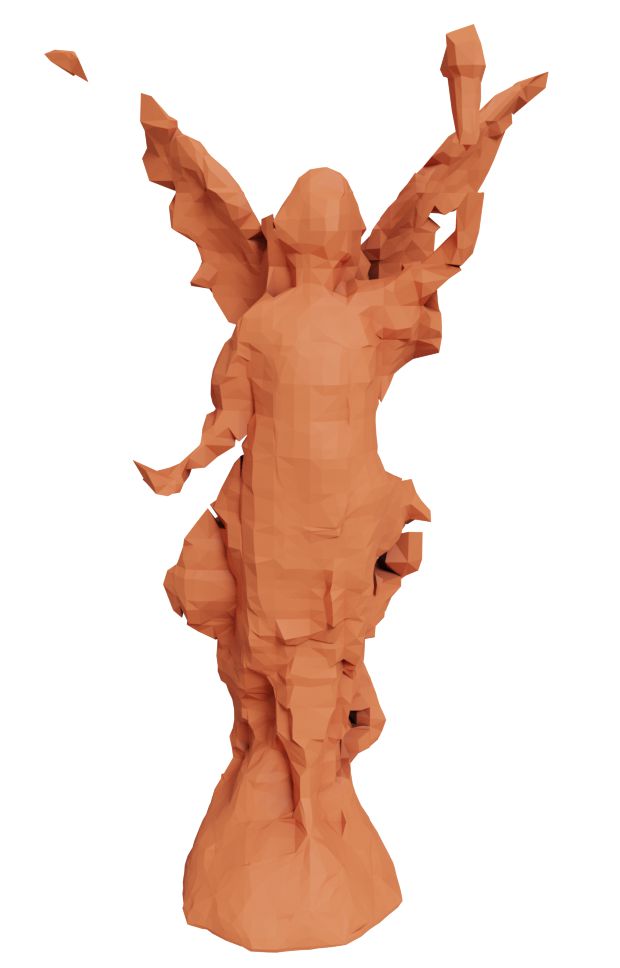} 
\end{subfigure}
 \begin{subfigure}{.09\textwidth}
  \centering
  \includegraphics[width=\linewidth]{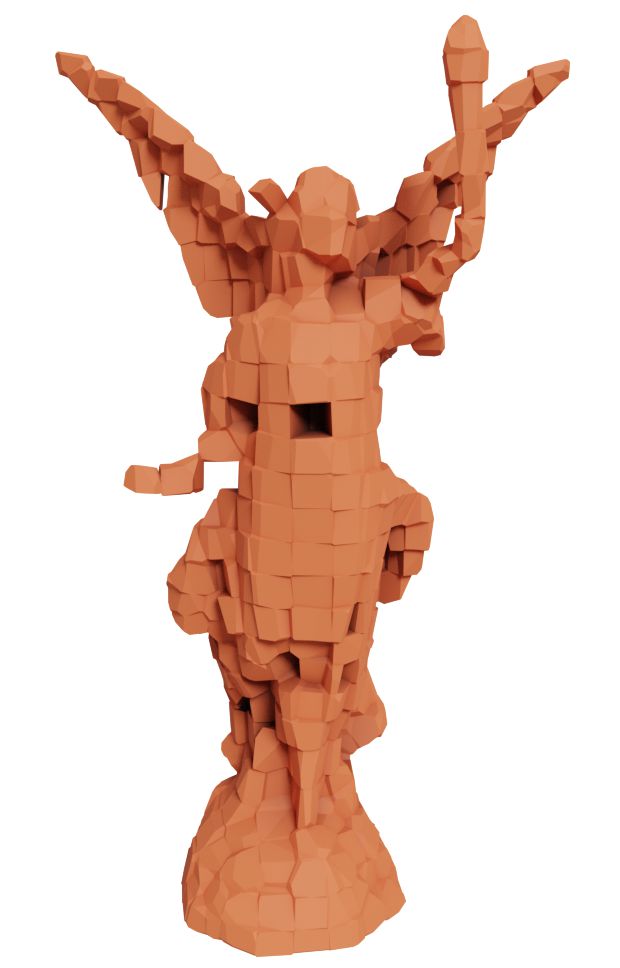} 
\end{subfigure}
 \begin{subfigure}{.09\textwidth}
  \centering
  \includegraphics[width=\linewidth]{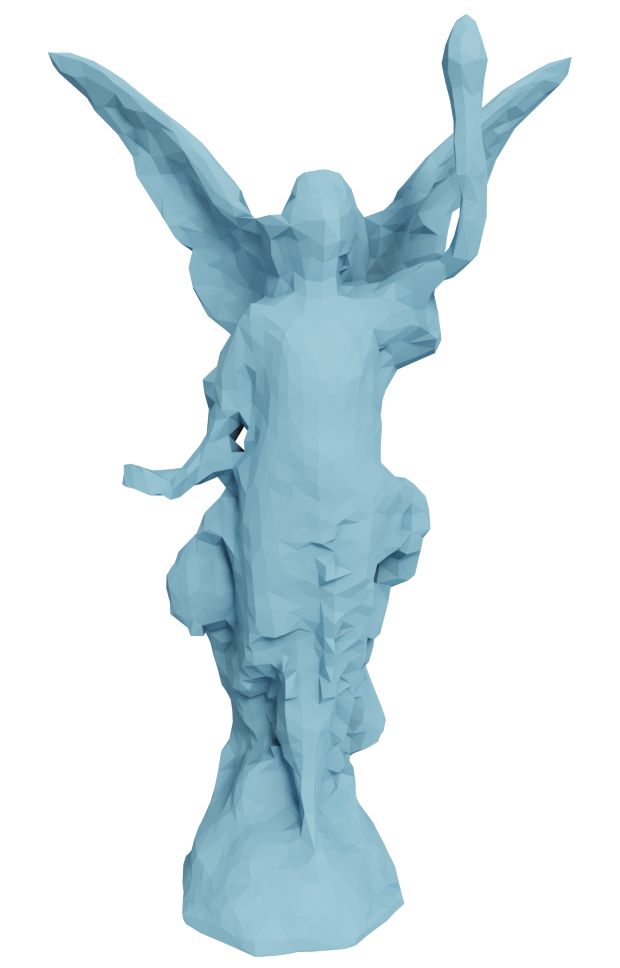} 
\end{subfigure}
 \begin{subfigure}{.09\textwidth}
  \centering
  \includegraphics[width=\linewidth]{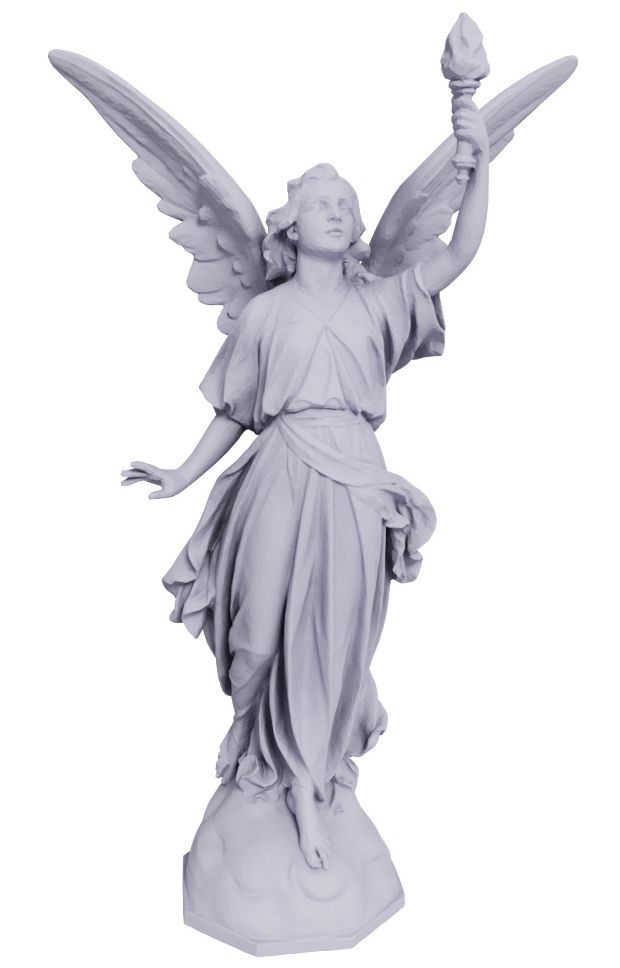} 
\end{subfigure}
 \hspace*{-1mm}
 
\begin{subfigure}{.09\textwidth}
  \centering
  \includegraphics[width=\linewidth]{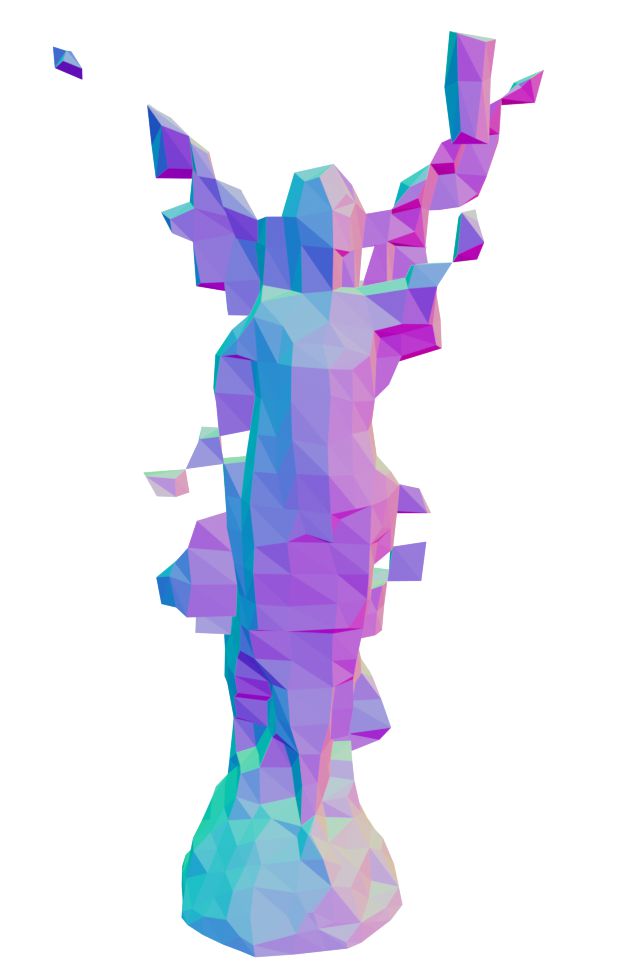} 
\end{subfigure}
 \begin{subfigure}{.09\textwidth}
  \centering
  \includegraphics[width=\linewidth]{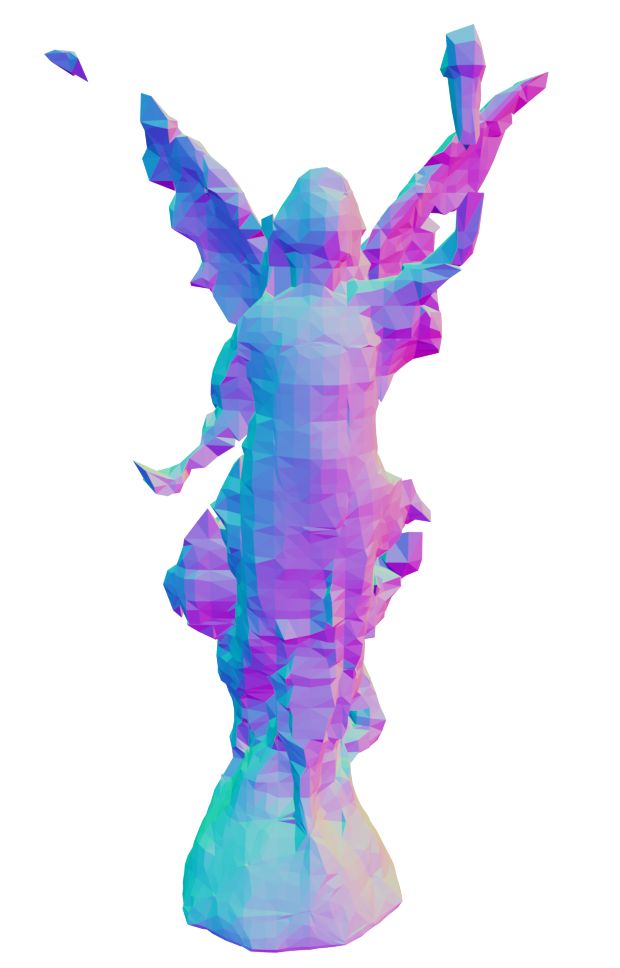} 
\end{subfigure}
 \begin{subfigure}{.09\textwidth}
  \centering
  \includegraphics[width=\linewidth]{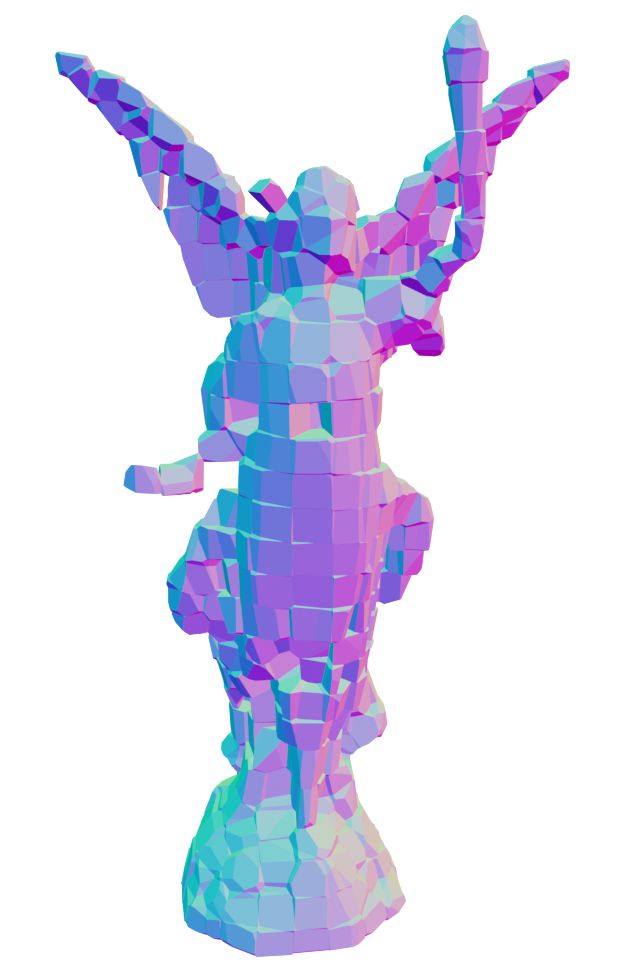} 
\end{subfigure}
 \begin{subfigure}{.09\textwidth}
  \centering
  \includegraphics[width=\linewidth]{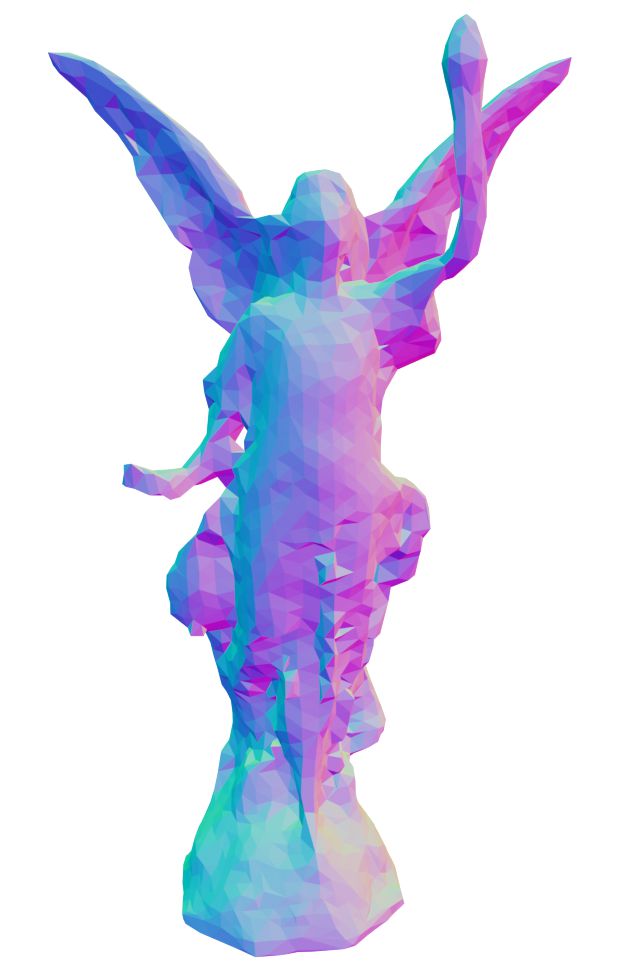} 
\end{subfigure}
 \begin{subfigure}{.09\textwidth}
  \centering
  \includegraphics[width=\linewidth]{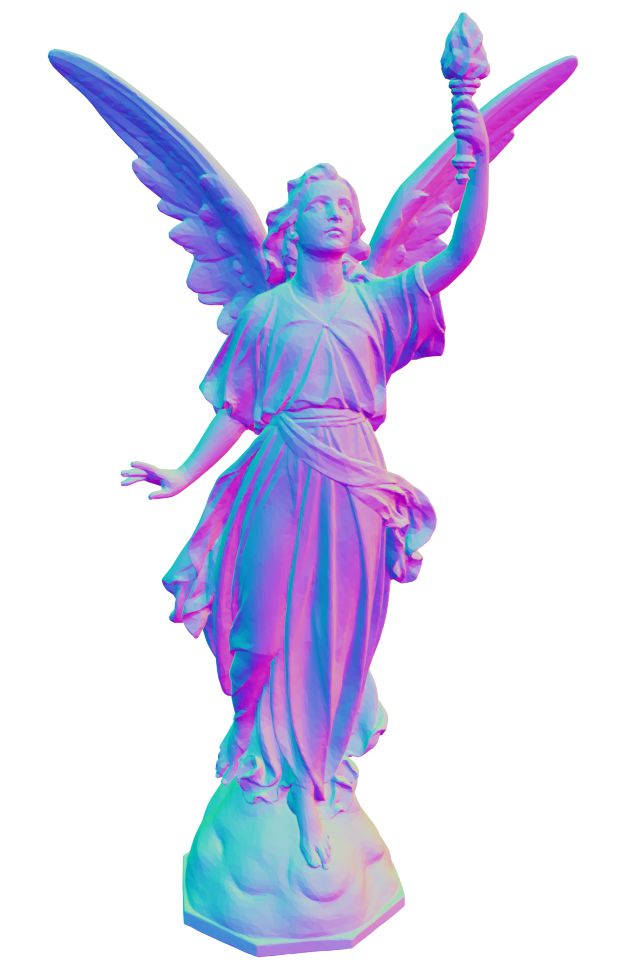} 
\end{subfigure}
 \hspace*{-1mm}
 \begin{subfigure}{.09\textwidth}
  \centering
  \includegraphics[width=\linewidth]{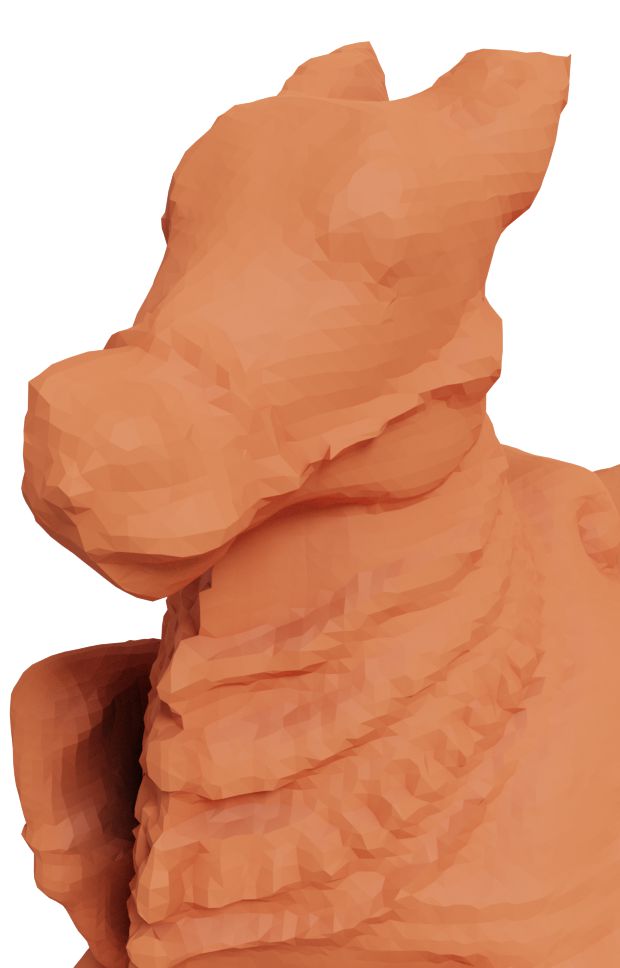} 
\end{subfigure}
 \begin{subfigure}{.09\textwidth}
  \centering
  \includegraphics[width=\linewidth]{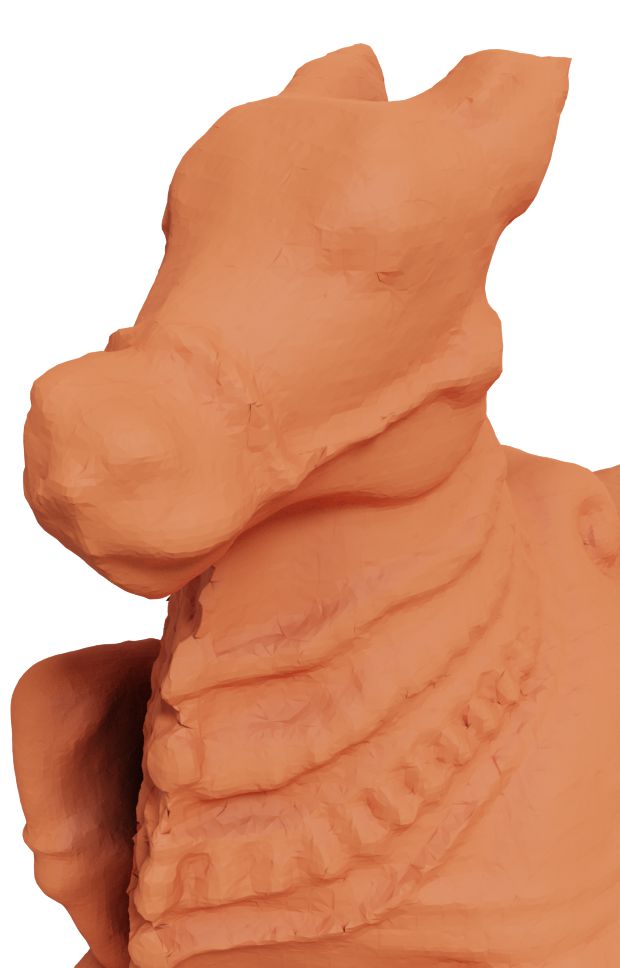} 
\end{subfigure}
 \begin{subfigure}{.09\textwidth}
  \centering
  \includegraphics[width=\linewidth]{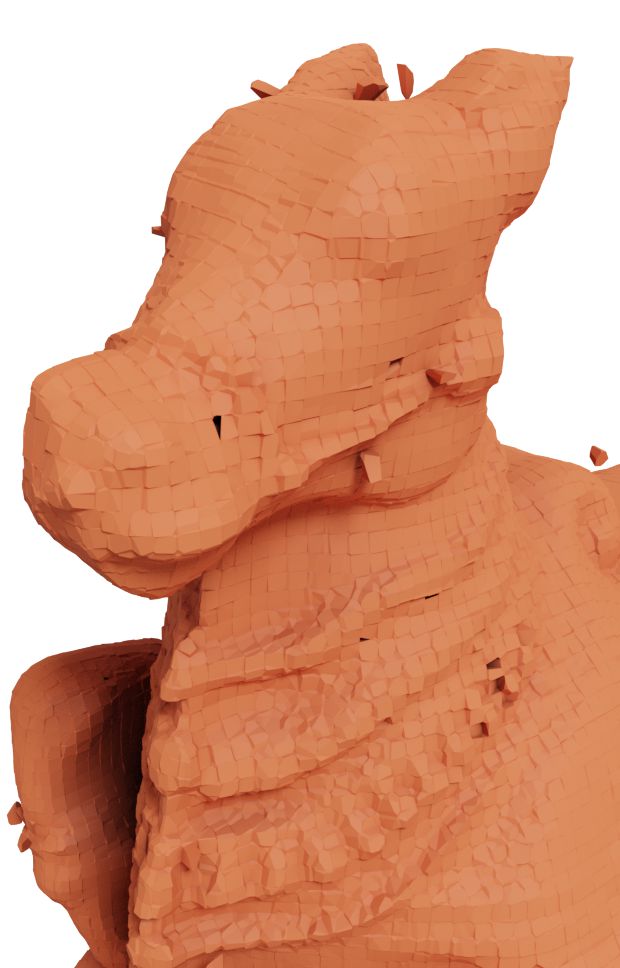} 
\end{subfigure}
 \begin{subfigure}{.09\textwidth}
  \centering
  \includegraphics[width=\linewidth]{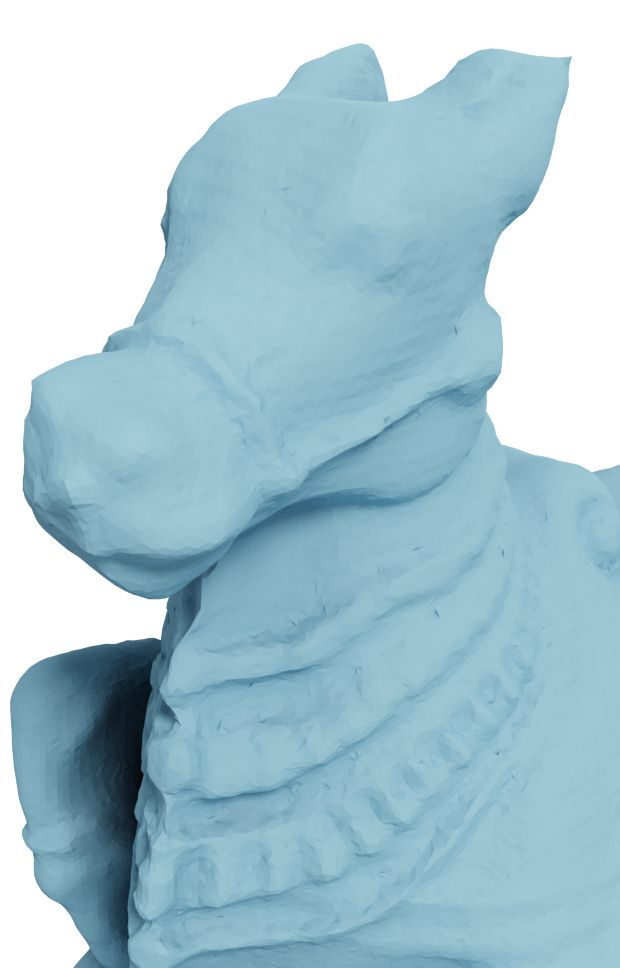} 
\end{subfigure}
 \begin{subfigure}{.09\textwidth}
  \centering
  \includegraphics[width=\linewidth]{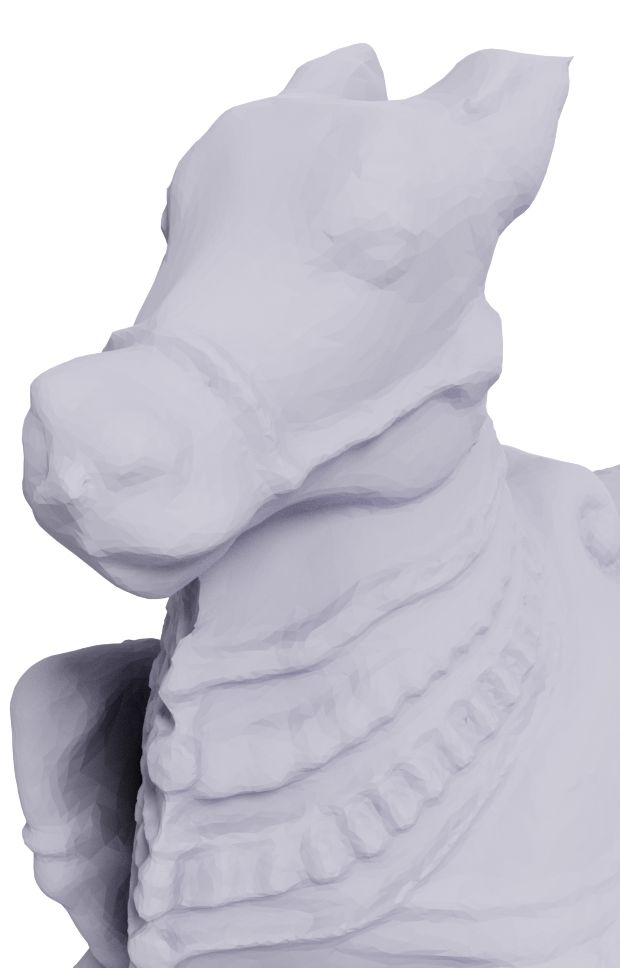} 
\end{subfigure}
 \hspace*{-1mm}
\begin{subfigure}{.09\textwidth}
  \centering
  \includegraphics[width=\linewidth]{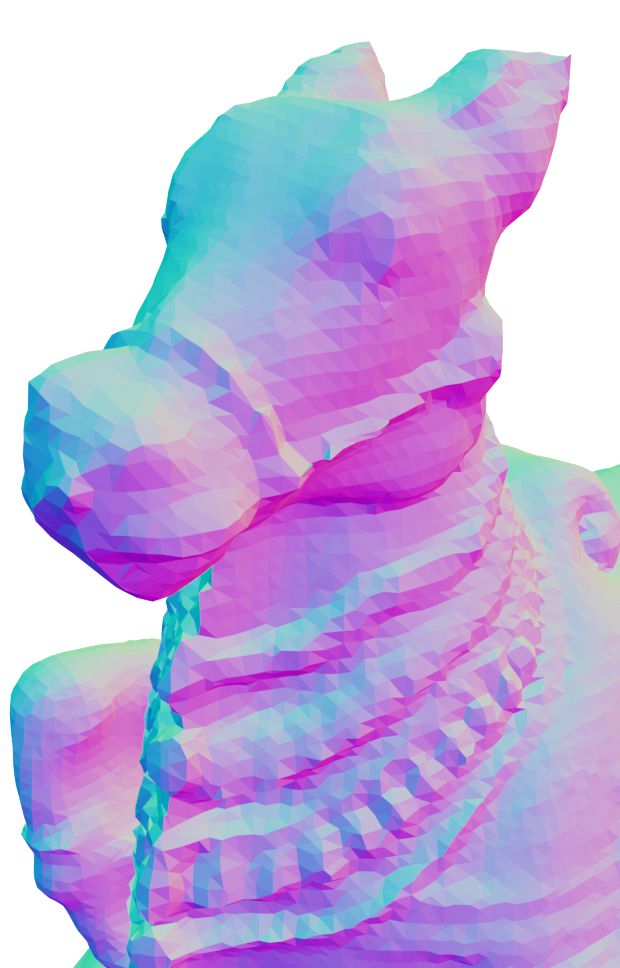} 
  \caption{NDC}
\end{subfigure}
 \begin{subfigure}{.09\textwidth}
  \centering
  \includegraphics[width=\linewidth]{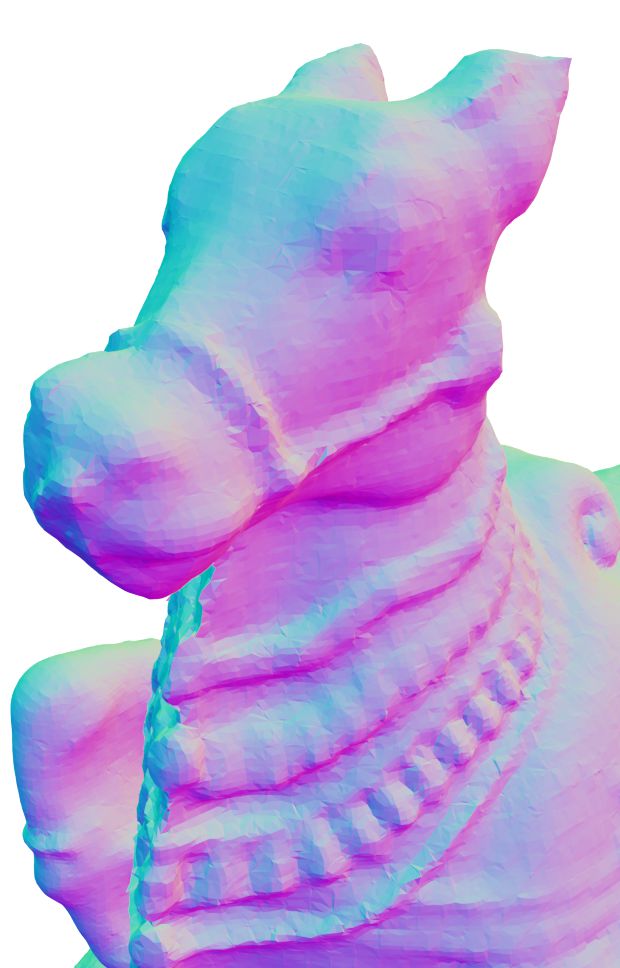} 
  \caption{NMC}
\end{subfigure}
 \begin{subfigure}{.09\textwidth}
  \centering
  \includegraphics[width=\linewidth]{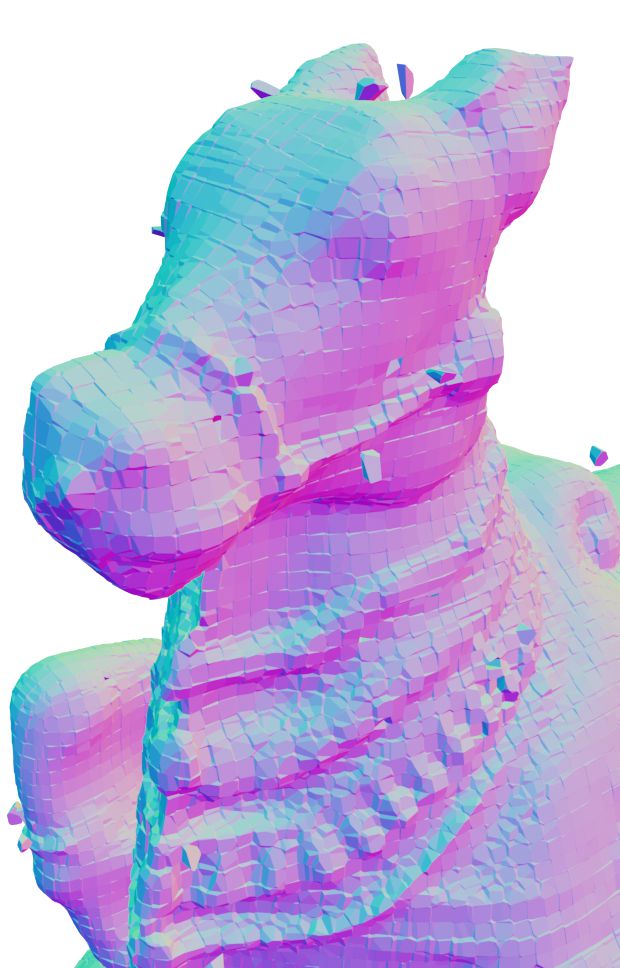} 
  \caption{VoroMesh}
\end{subfigure}
 \begin{subfigure}{.09\textwidth}
  \centering
  \includegraphics[width=\linewidth]{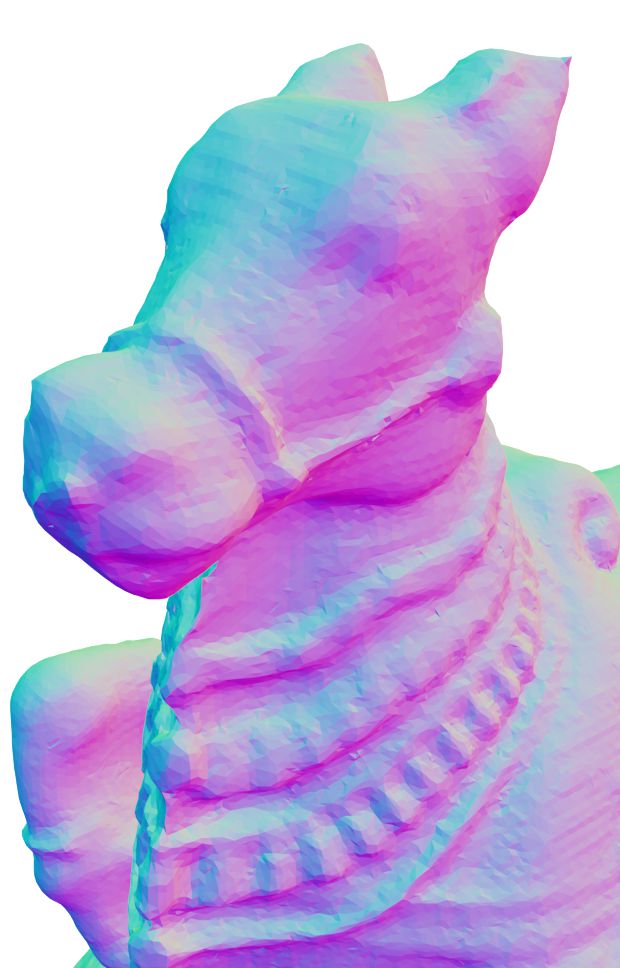} 
        \caption{PoNQ}
\end{subfigure}
 \begin{subfigure}{.09\textwidth}
  \centering
  \includegraphics[width=\linewidth]{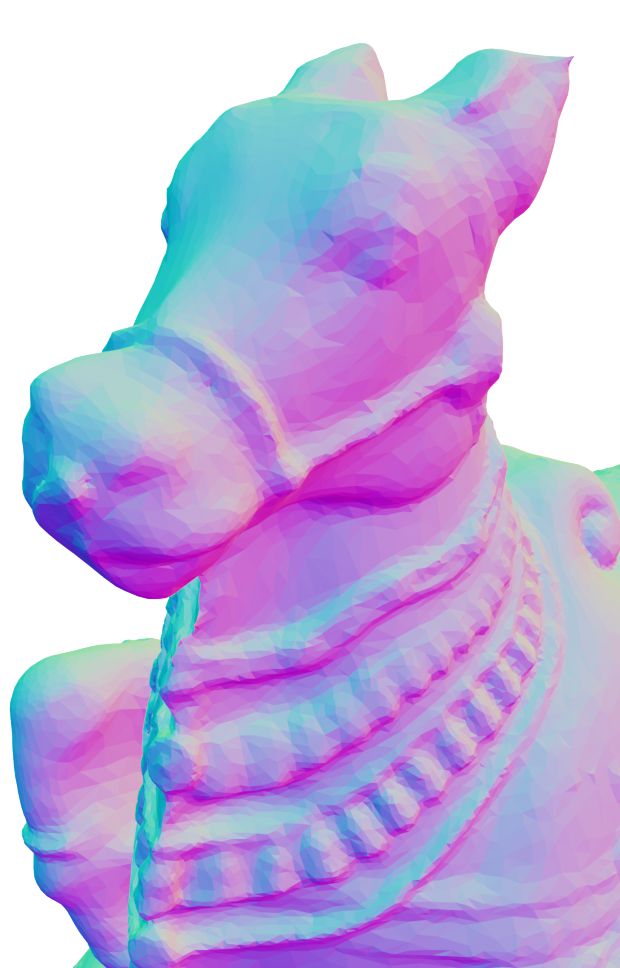} 
        \caption{G. Truth}
\end{subfigure} 
 \vspace*{-1mm}
\caption{Learning results (top: $32^3$; bottom: $128^3$) on Thingi30.
\vspace*{-1mm}}
  \label{fig:sdf_thingi}
\end{figure}

\textbf{Results.} 
Our key results are reported in Tabs.~\ref{tab:sdf_abc_tab} and~\ref{tab:sdf_thingi_tab}, where PoNQ outperforms state-of-the-art methods \textit{on every resolution, dataset, and metric while guaranteeing watertight output meshes that are devoid of self-intersections}. 
We note that NMC and NDC both rely on a regular-grid based meshing. As a result, they fail to capture thin structures and exhibit aliasing artifacts (see Figs.~\ref{fig:sdf_abc} and~\ref{fig:sdf_thingi}), which impacts their surface, normal and sharp-edge fitting scores.

VoroMesh does not rely on a regular grid, so its ability to capture thin surfaces leads to better results than NMC and NDC for the complex ABC dataset or on low-resolution models of Thingi30. However, as noticed in Tabs.~\ref{tab:sdf_abc_tab} and~\ref{tab:sdf_thingi_tab}, it suffers from local surface artifacts leading to sharp edges and spurious normals, resulting in worst NC, ECD and EF1. Moreover, its two-stage training implies that its encoder is not trained for occupancy prediction; as a result, it can mislabel  Voronoi generators, leading to either missing parts or floating volumes around the shapes (see Figs.~\ref{fig:sdf_abc} and~\ref{fig:sdf_thingi}). 

In contrast, our method is able to capture fine details due to our use of QEM, which helps to capture surface characteristics, and it does not visually exhibit any of the artifacts of other approaches due to our PoNQ mesh being extracted from a Delaunay tetrahedralization. Ultimately, PoNQ leads to more faithful reconstructions while guaranteeing $100\%$ watertight and intersection-free results.

\begin{table}[t] \vspace*{-1mm}
  \begin{center}
    \resizebox{\linewidth}{!}{\begin{tabular}{l c c c c c c c c c }
  \hline
  \textbf{Method}            & Grid & CD $\downarrow$               & F1  $\uparrow$  & NC $\uparrow$ & ECD $\downarrow$ & EF1 $\uparrow$ & Watertight $\uparrow$ & \# V & \# F   \\
                             & Size & ($\times 10^{-5}$) &  &   &   &           & no self-int. & ($\times 10^3$)  & ($\times 10^3$)  \\

  \hline
  NDC~\cite{chen_neural_2022}  & $32^3$  & 6.390  &  0.745  &  0.920 & 0.172 & 0.245 &  40\%     & 1.3 &  2.6\\
  NMC~\cite{chen_neural_2021}  & $32^3$  &5.188  &  0.796  &  0.936 & 0.148 & 0.271 & 0\%  & 8.7 & 17.3\\
  VoroMesh~\cite{maruani_voromesh_2023}        & $32^3$  & 2.825  &  0.758  &  0.902 & 0.263 & 0.156 & \textbf{100}\%     & 9.9  & 19.9   \\
    PoNQ-lite  & $32^3$  &  1.705  &  0.754  &  0.934 & 0.154 & 0.270 &\textbf{100}\%   &   1.3 & 2.6  \\

    PoNQ  & $32^3$  &   \textbf{1.344}  &  \textbf{0.810}  &  \textbf{0.942} & \textbf{0.137} & \textbf{0.314} &\textbf{100}\%    &   4.8 & 9.7 \\

  \hline
  
  NDC~\cite{chen_neural_2022}  & $64^3$  & 0.849  &  0.908  &  0.961 & 0.106 & 0.441  & 3\% &  5.4 & 10.8 \\
  NMC~\cite{chen_neural_2021}  & $64^3$  & 0.776  &  0.923  &  0.969 & 0.115 & 0.467 & 0\%  & 36.8 & 73.6\\
  VoroMesh~\cite{maruani_voromesh_2023}        & $64^3$  & 1.021  &  0.906  &  0.939 & 0.259 & 0.192 & \textbf{100}\%     & 39.4 & 78.9  \\
    PoNQ-lite  & $64^3$  &  0.769  &  0.914  &  0.968 & \textbf{0.090} & 0.495 &\textbf{100}\%   & 5.3 & 10.6 \\

    PoNQ & $64^3$  &  \textbf{0.758}  &  \textbf{0.924}  &  \textbf{0.971} & 0.100 & \textbf{0.511} & \textbf{100}\%    & 19.9 & 39.9    \\
    
  \hline
  NDC~\cite{chen_neural_2022}  & $128^3$ &  0.650  &  0.937  &  0.980 & 0.065 &  0.644 & 0\%    & 22.0 &  44.1\\
  NMC~\cite{chen_neural_2021}  & $128^3$ & 0.642  &  \textbf{0.939}  &  \textbf{0.984} & 0.131 & 0.574 & 0\%  & 151.7 & 303.3\\
  VoroMesh~\cite{maruani_voromesh_2023}        & $128^3$ & 0.731  &  0.932  &  0.959 & 0.260 & 0.198  & \textbf{100}\%    & 157.2 & 314.5 \\
     PoNQ-lite & $128^3$  &  0.644  &  0.938  &  \textbf{0.984} & \textbf{0.055} & \textbf{0.699} &\textbf{100}\%   &     21.7 & 43.3   \\
    PoNQ  & $128^3$  &  \textbf{0.641}  &  \textbf{0.939} &  \textbf{0.984} & 0.123 & 0.592 & \textbf{100}\%  &    80.8 & 161.8  \\
    
  \hline \label{tab:sdf_thingi}
\end{tabular}}\vspace*{-4mm}
    \caption{Results on Thingi30 with our network trained on ABC.\vspace*{-3mm} }
    \label{tab:sdf_thingi_tab}
  \end{center}
\end{table}

\subsection{Additional extensions}
\label{sec:extensions}

We conclude this section with other examples leveraging the unique nature of our neural representation.

\textbf{Surfaces with boundary.}
With minor modifications, PoNQ can output surfaces with boundaries as well (Fig.~\ref{fig:summary} and supplementary material). To optimize or train our representation for this case, we simply compute a \emph{boundary sampling} $S_b$ of the input surface boundaries, and duplicate it to create a sampling $S'_b$ where we just rotate all the normals by $\pi/2$ around the boundary. Changing $S$ into $S\cup S_b\cup S_b'$ will thus enforce that all the QEM matrices on the boundaries will generate elongated ellipsoids aligned with the local boundary. Meshing can proceed as before; but the output mesh will automatically close the holes. We further cull any triangle $T$ of the extracted mesh for which the anisotropy of the QEM of each of its vertices is above 40\%. In practice, we measure the anisotropy through the ratio $r_i  \!=\!  \lambda_2/\lambda_1$ of the two largest eigenvalues of the matrix $\mathbf{A}_i$. Obviously, the surface no longer bounds a volume since we cut holes in the original PoNQ mesh extraction, but it remains devoid of self-intersections. Note that this PoNQ variant cannot handle special cases such as a single square sheet. 

\begin{figure}[h!] \vspace*{-2mm}
 \centering
 \begin{subfigure}{.47\columnwidth}
  \centering
  \includegraphics[width=.48\linewidth]{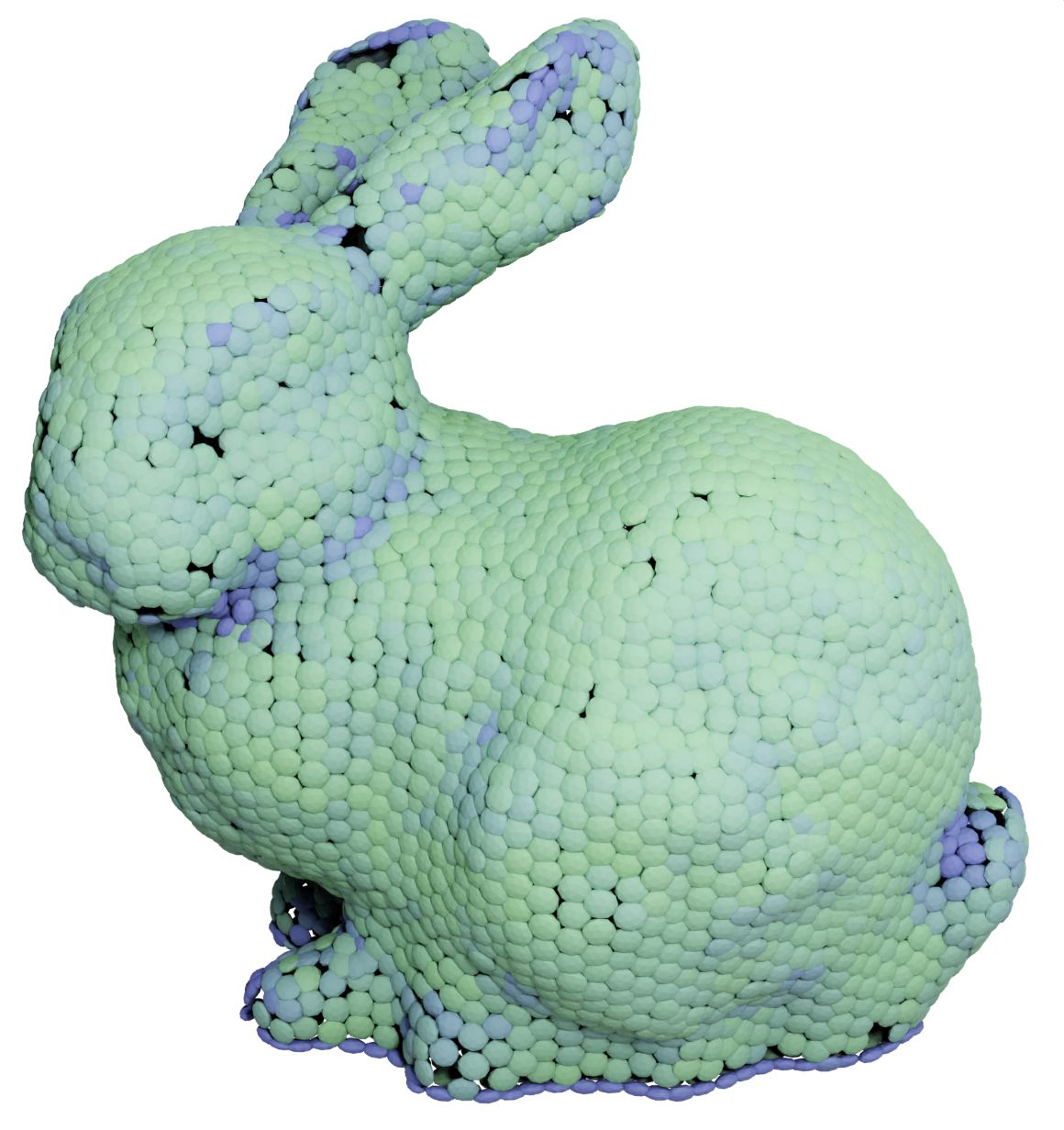} 
    \includegraphics[width=.48\linewidth]{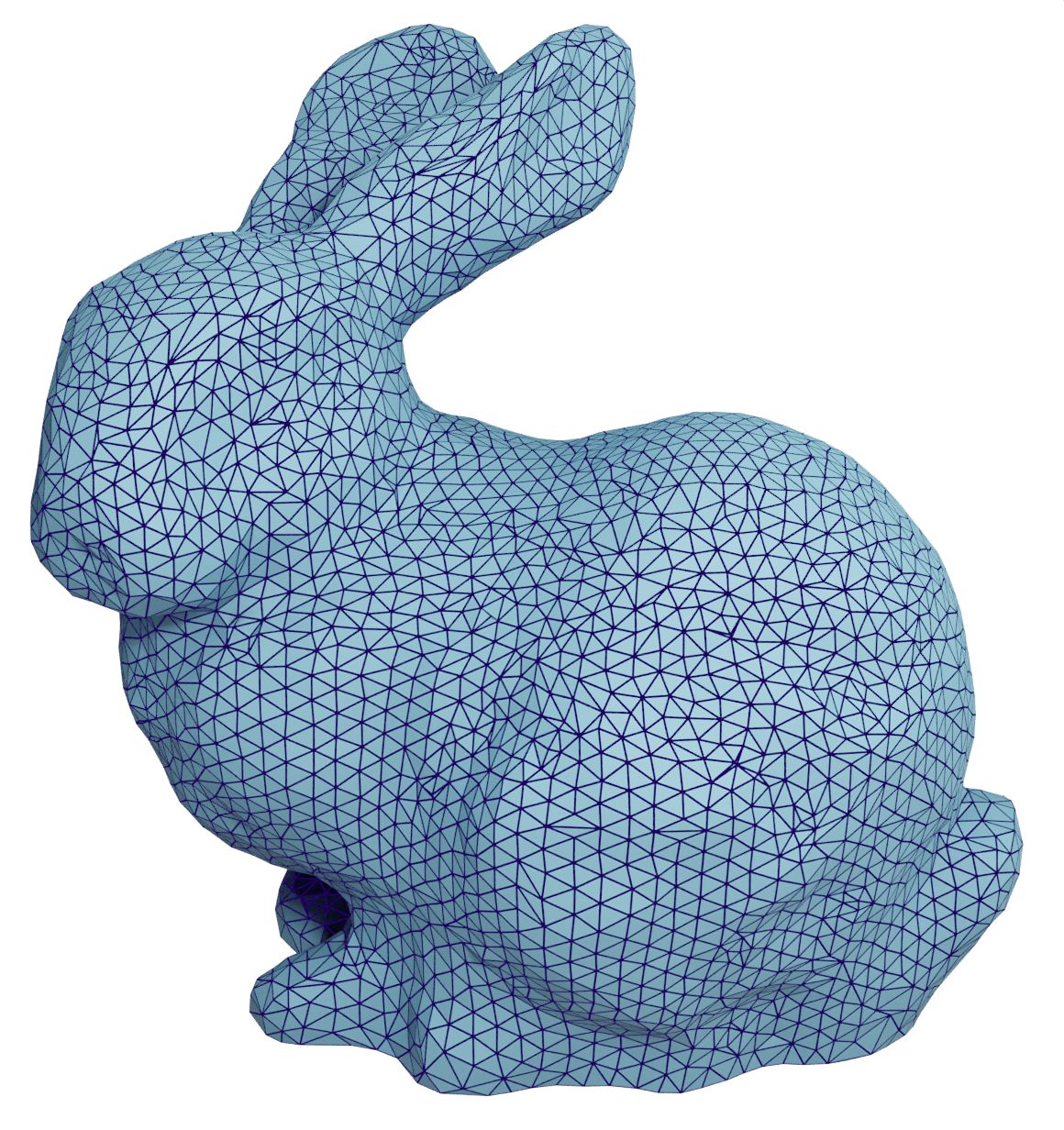} 
\end{subfigure}
 \begin{subfigure}{.47\columnwidth}
  \centering
    \includegraphics[width=.48\linewidth]{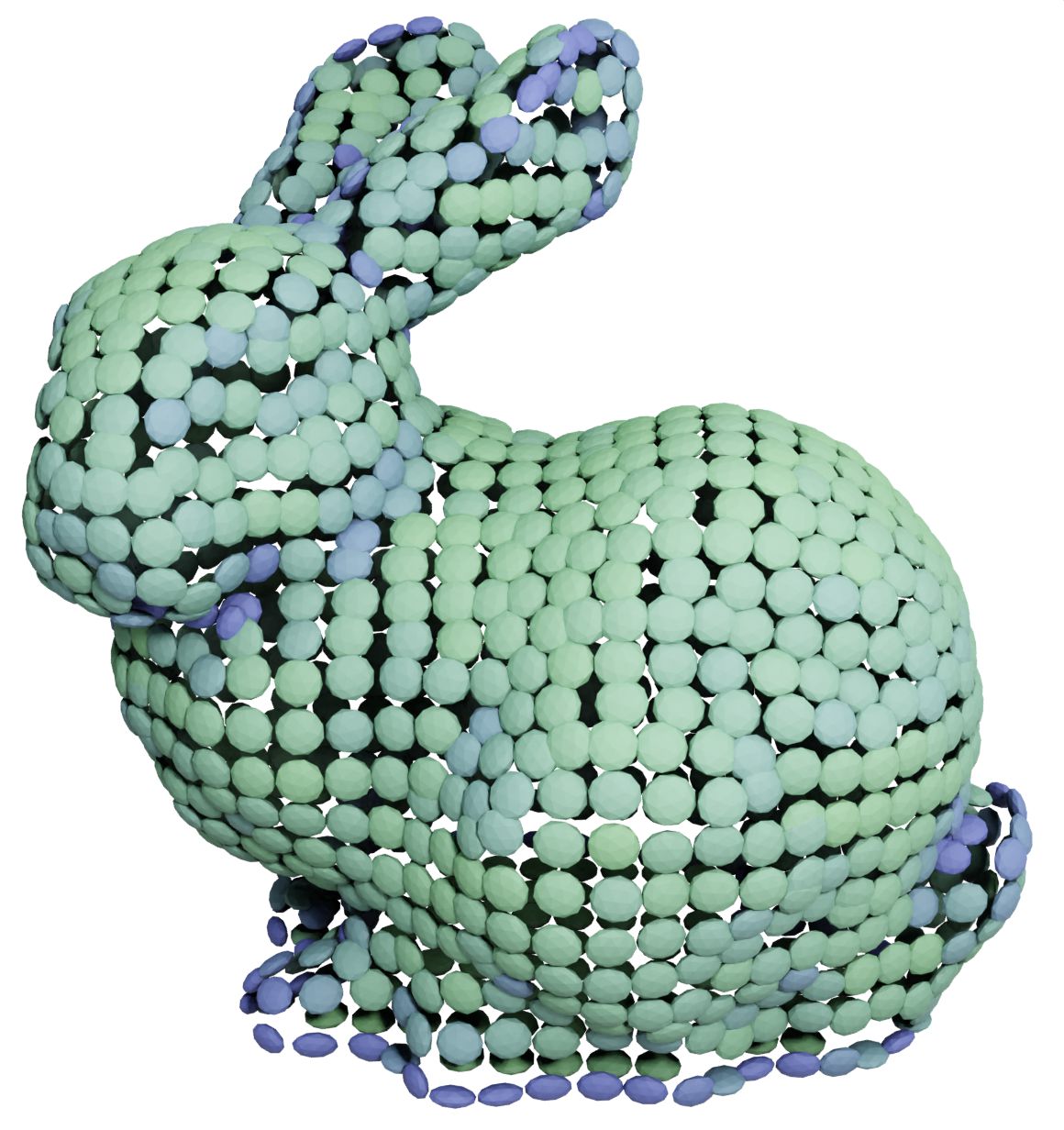} 
  \includegraphics[width=.48\linewidth]{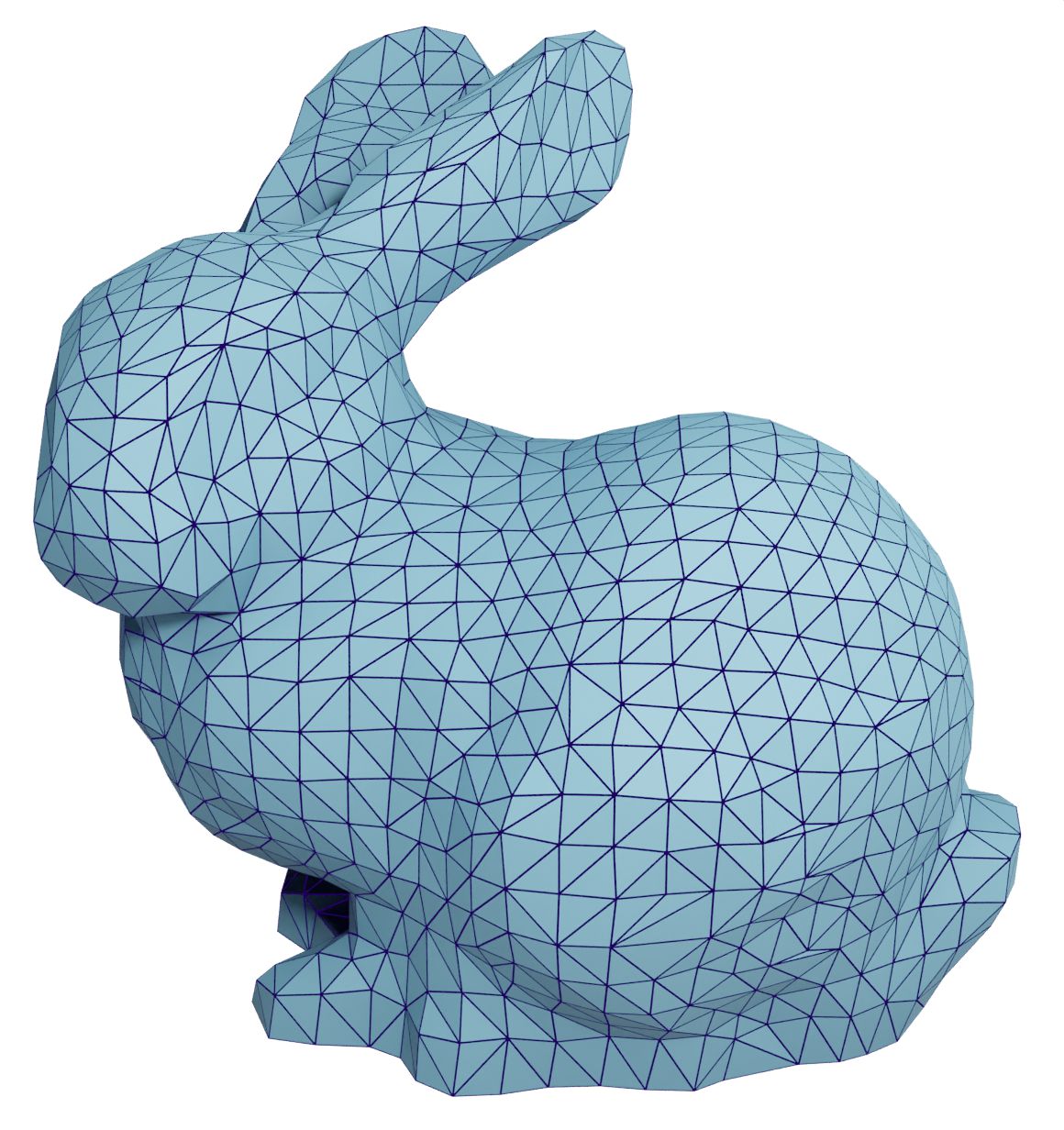} 
    \end{subfigure}
 \begin{subfigure}{.47\columnwidth}
  \centering
  \includegraphics[width=.42\linewidth]{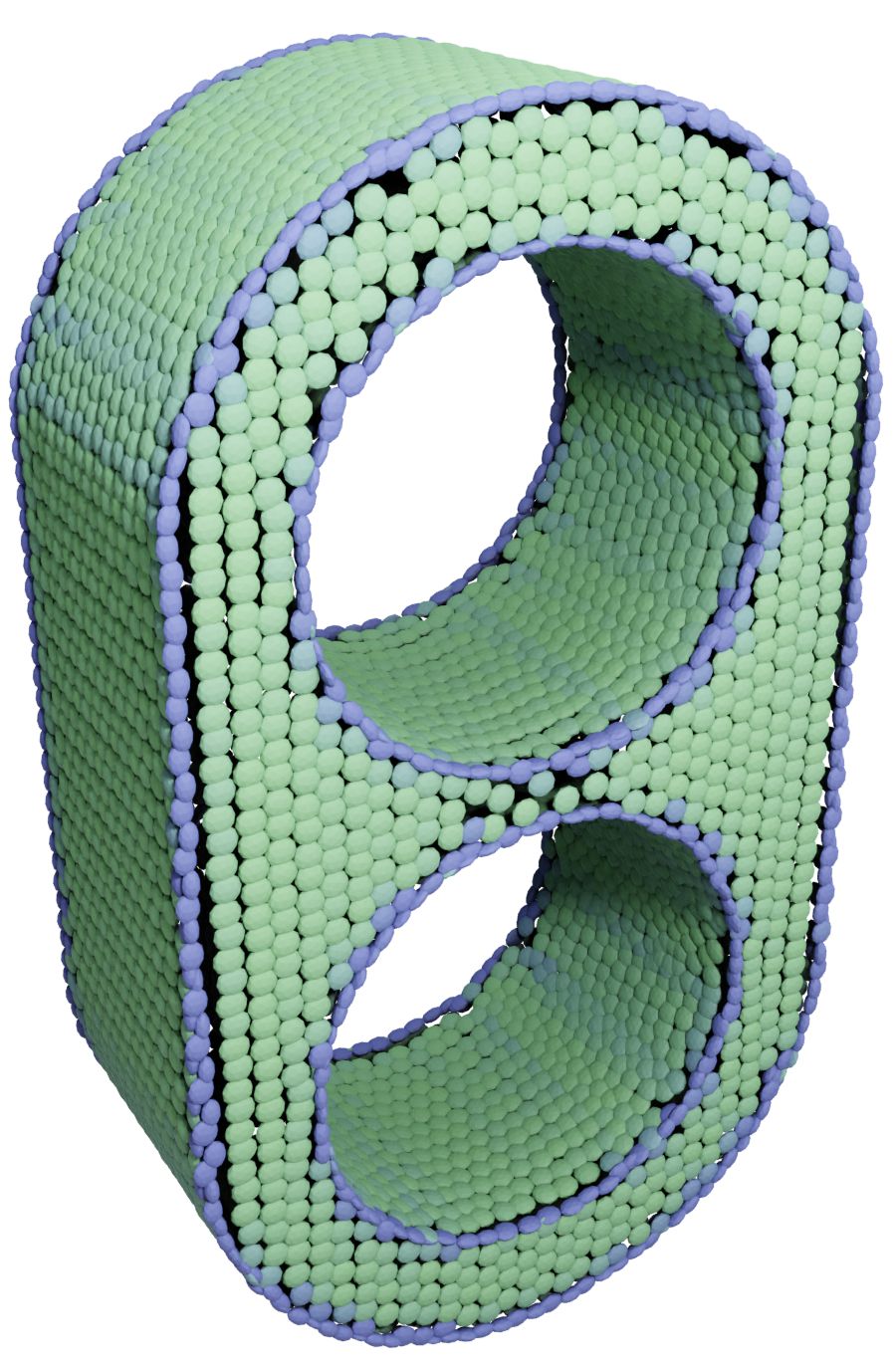} 
    \includegraphics[width=.42\linewidth]{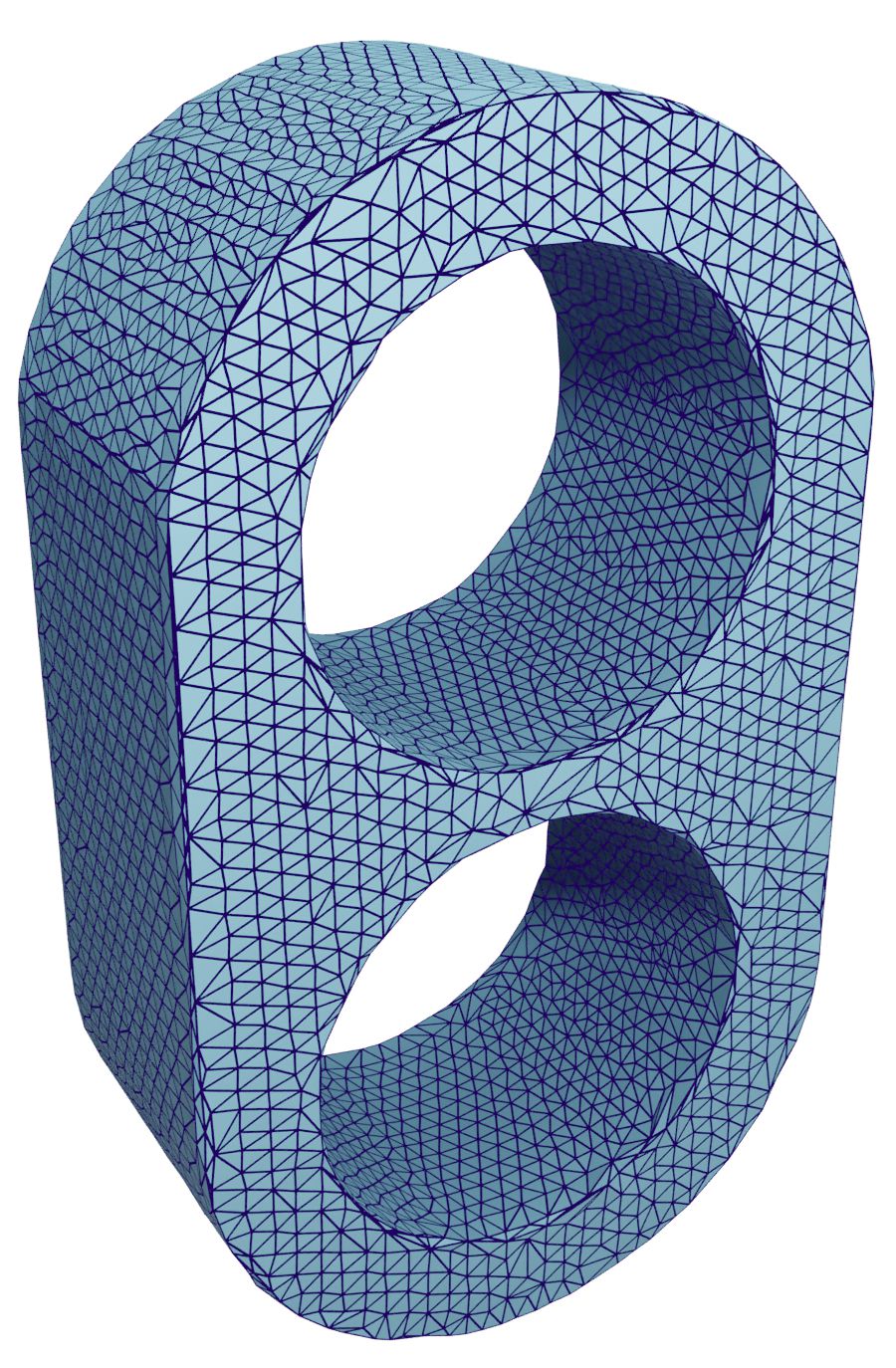} 
  \caption{PoNQ}
\end{subfigure}
 \begin{subfigure}{.47\columnwidth}
  \centering
    \includegraphics[width=.42\linewidth]{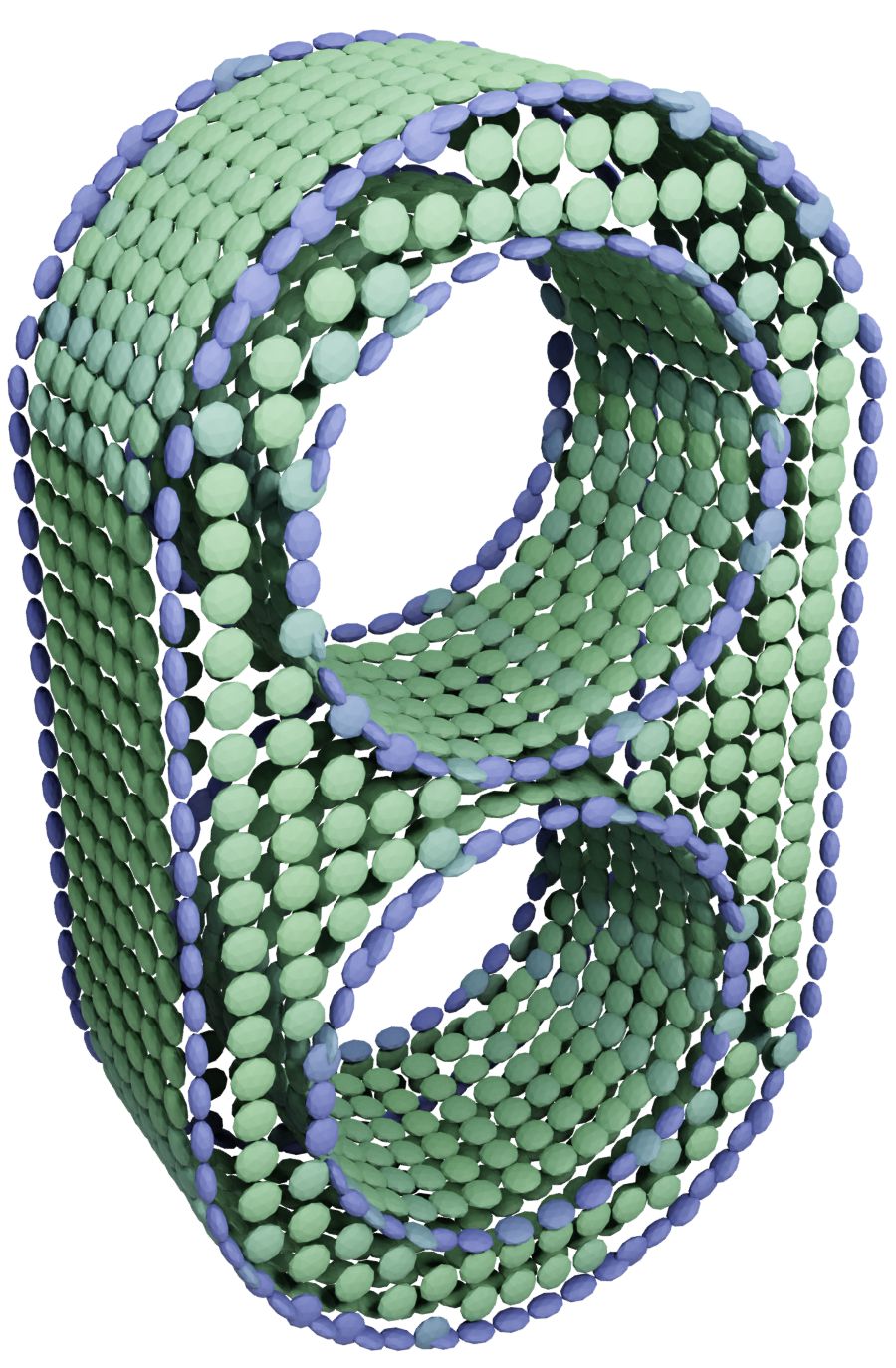} 
  \includegraphics[width=.42\linewidth]{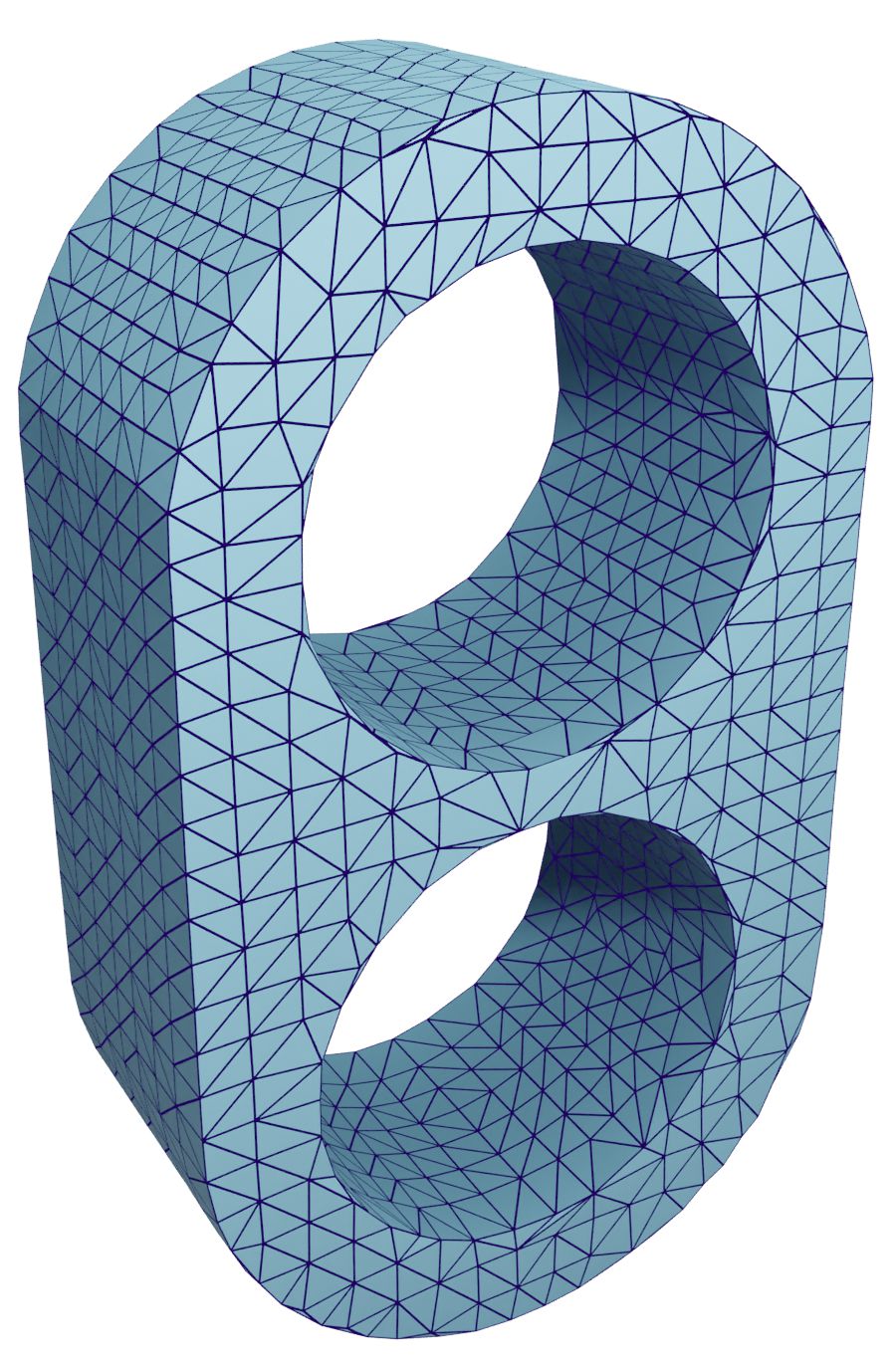} 
    \caption{PoNQ-lite} 
\end{subfigure}\vspace*{-3mm}
\caption{For a network trained on ABC with $P\!=\!4$ predicted QEM matrices per cell (left), one can sum these matrices -- i.e., via average pooling -- per cell to produce a more compact mesh (right) that still naturally preserves the detected sharp features.\vspace*{-2mm}}
  \label{fig:lite}
\end{figure}

\textbf{PoNQ-lite.} While using $P\!=\!4$ predicted points per voxel yielded the best performances in our learning-based tests as mentioned in Sec.~\ref{sec:method_learning}, we can trivially provide a ``lite'' version of our output PoNQ with a single point per cell using the same network: 
due to our reliance on QEM matrices, one can simply sum the $P$ matrices $\mathbf{Q}_i$ within a cell to directly create a single QEM matrix per cell, from which is derived a new optimal position $\mathbf{v}_i^\textbf{*}$ for each cell. 
The normals $\mathbf{n}_i$ are also trivially averaged into the new one. 
Sharp features are still well preserved due to our use of quadric error metrics (see Fig.~\ref{fig:lite}), and these PoNQ-lite meshes in fact outperform NDC for an equal level of element count (see Tabs.~\ref{tab:sdf_abc_tab} and~\ref{tab:sdf_thingi_tab}), proving the superiority of grid-free methods. 
If even coarser meshes are desired, one can also construct, at nearly no cost, a whole hierarchy of PoNQ meshes by first applying PoNQ-lite, and then merging each group of eight cells into a single larger one to form a twice-coarser grid ($2\times 2$ average pool; see Fig.~\ref{fig:hierarchy}).

Although PoNQ outperforms PoNQ-lite on almost all metrics, the latter slightly pulls ahead on ECD and EF1 on Thingi30 at $128^3$ resolution. This opens the door to a series of further investigations, out of scope for this paper: one could potentially simplify a PoNQ output \textit{adaptively} depending on the contents of the cell. 
\section{Discussion}

\begin{figure}[h!]\vspace*{-3mm}
 \centering
  \includegraphics[width=.235\linewidth]{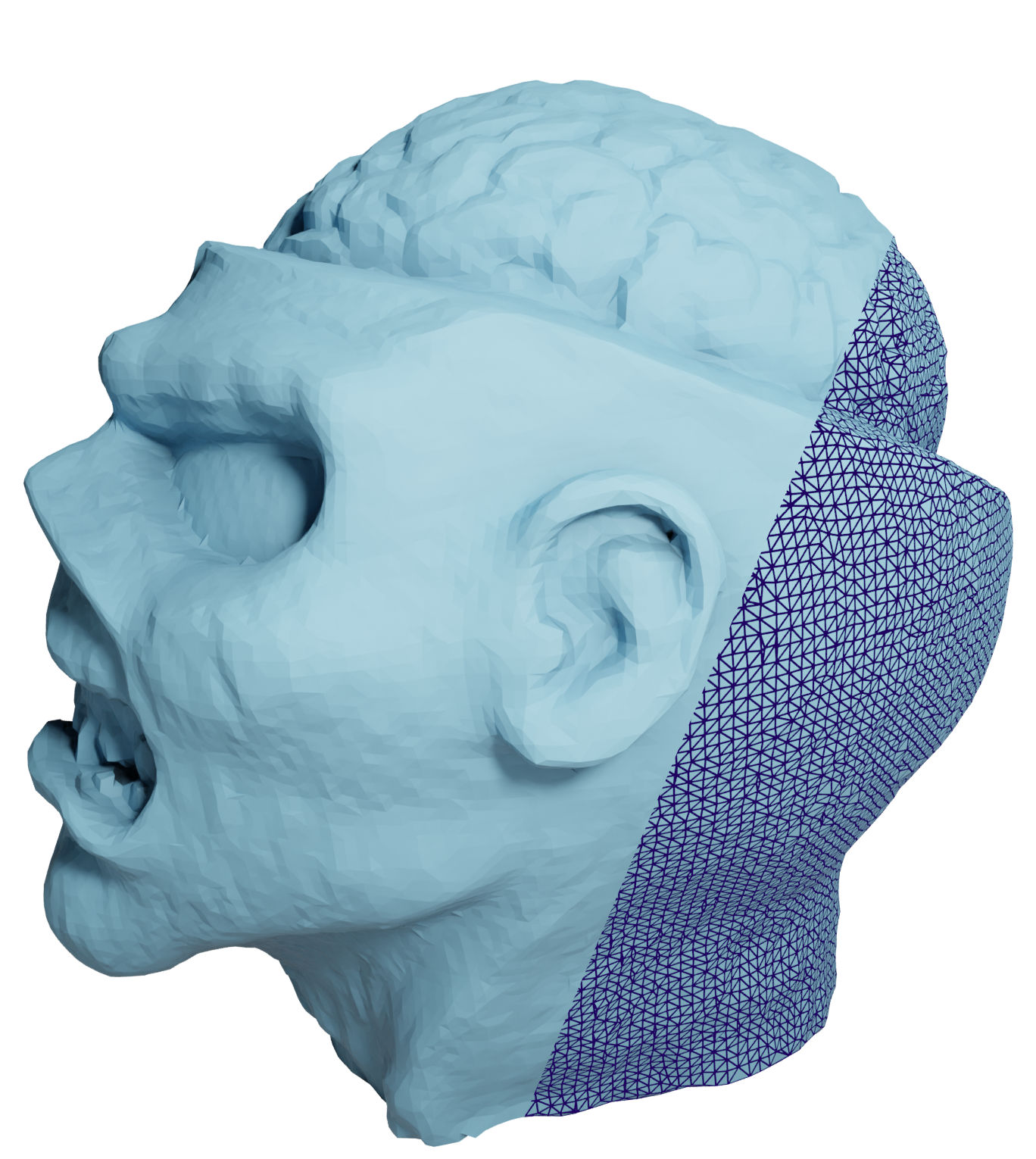} 
  \includegraphics[width=.235\linewidth]{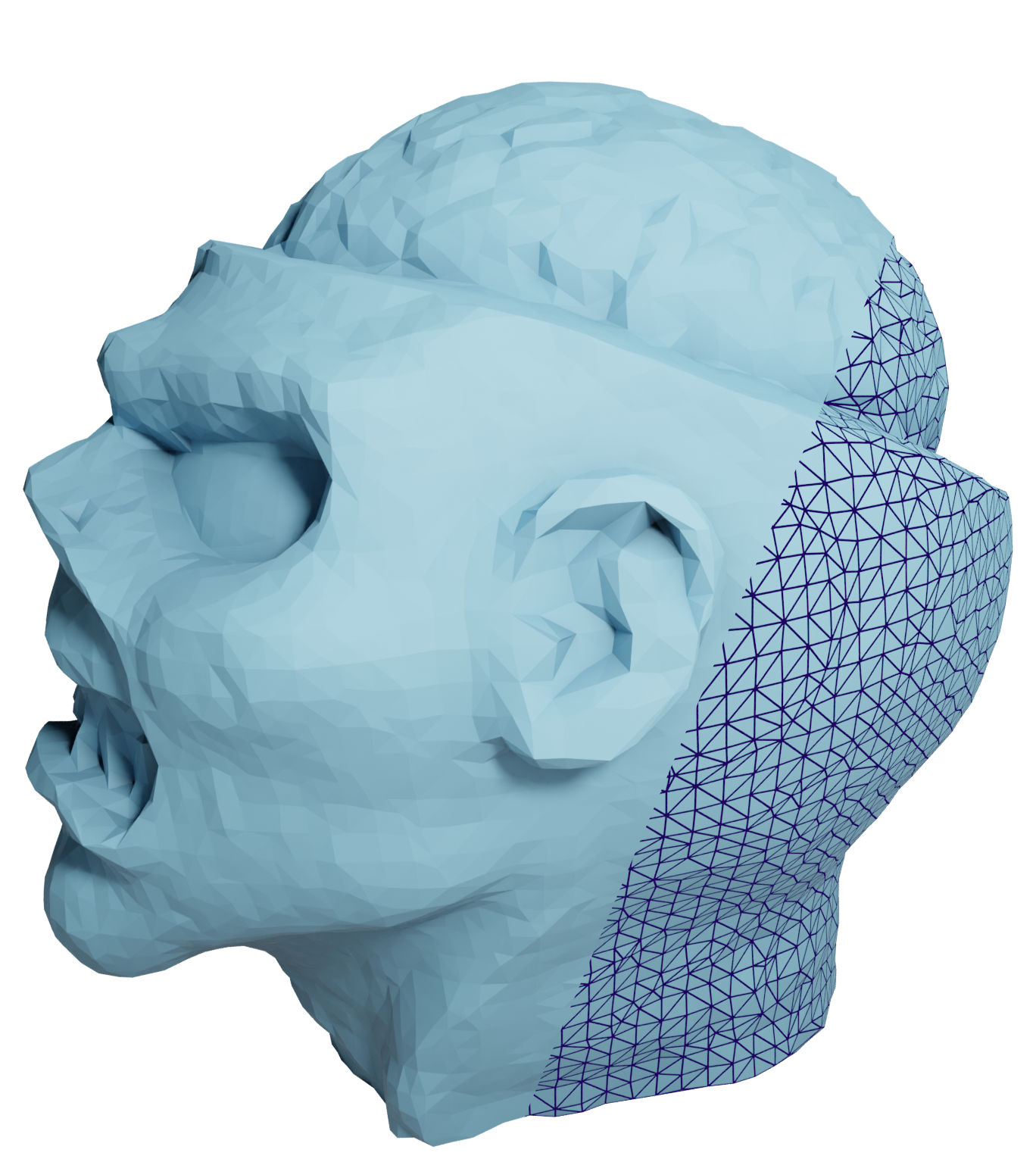} 
  \includegraphics[width=.235\linewidth]{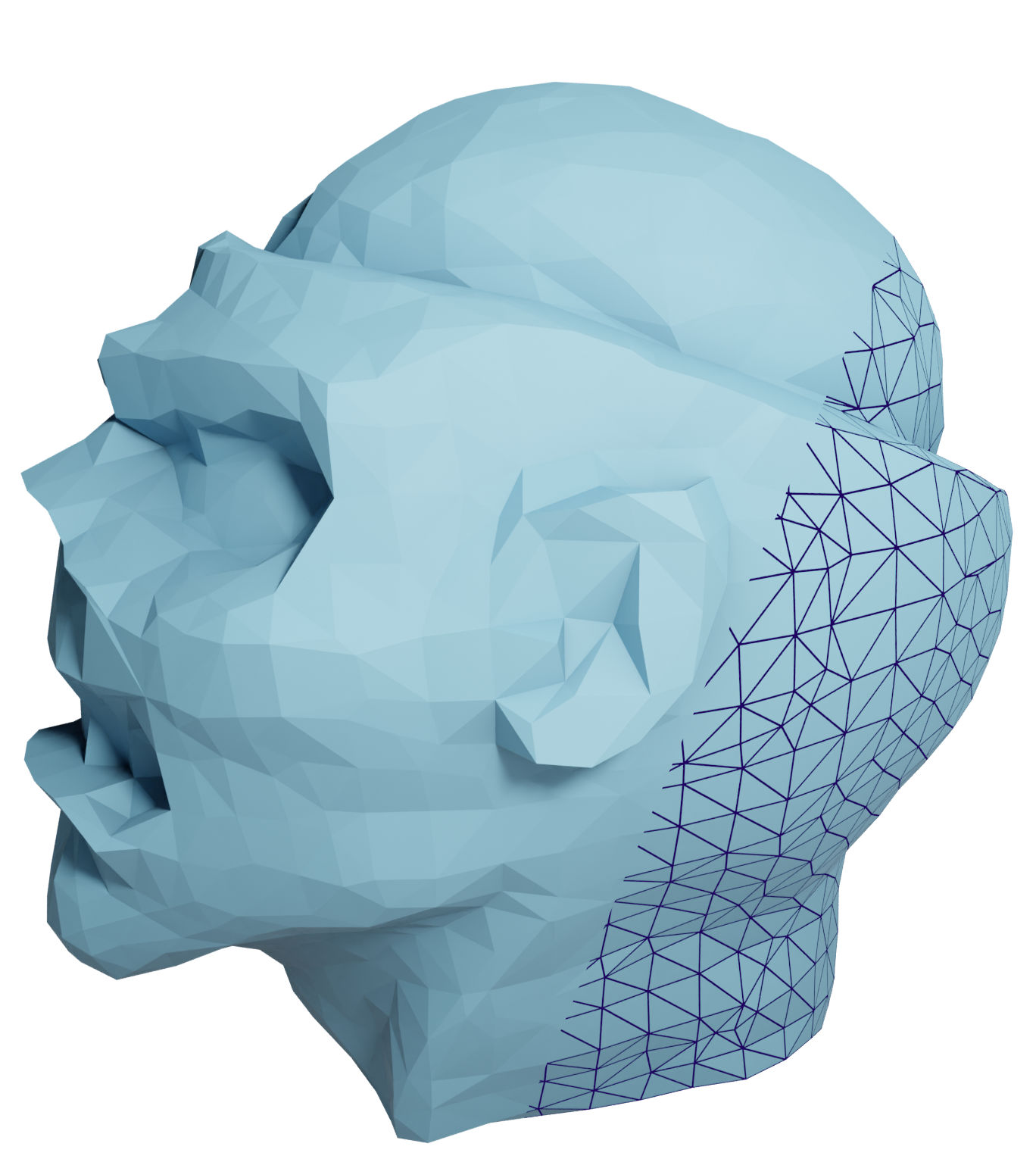} 
  \includegraphics[width=.235\linewidth]{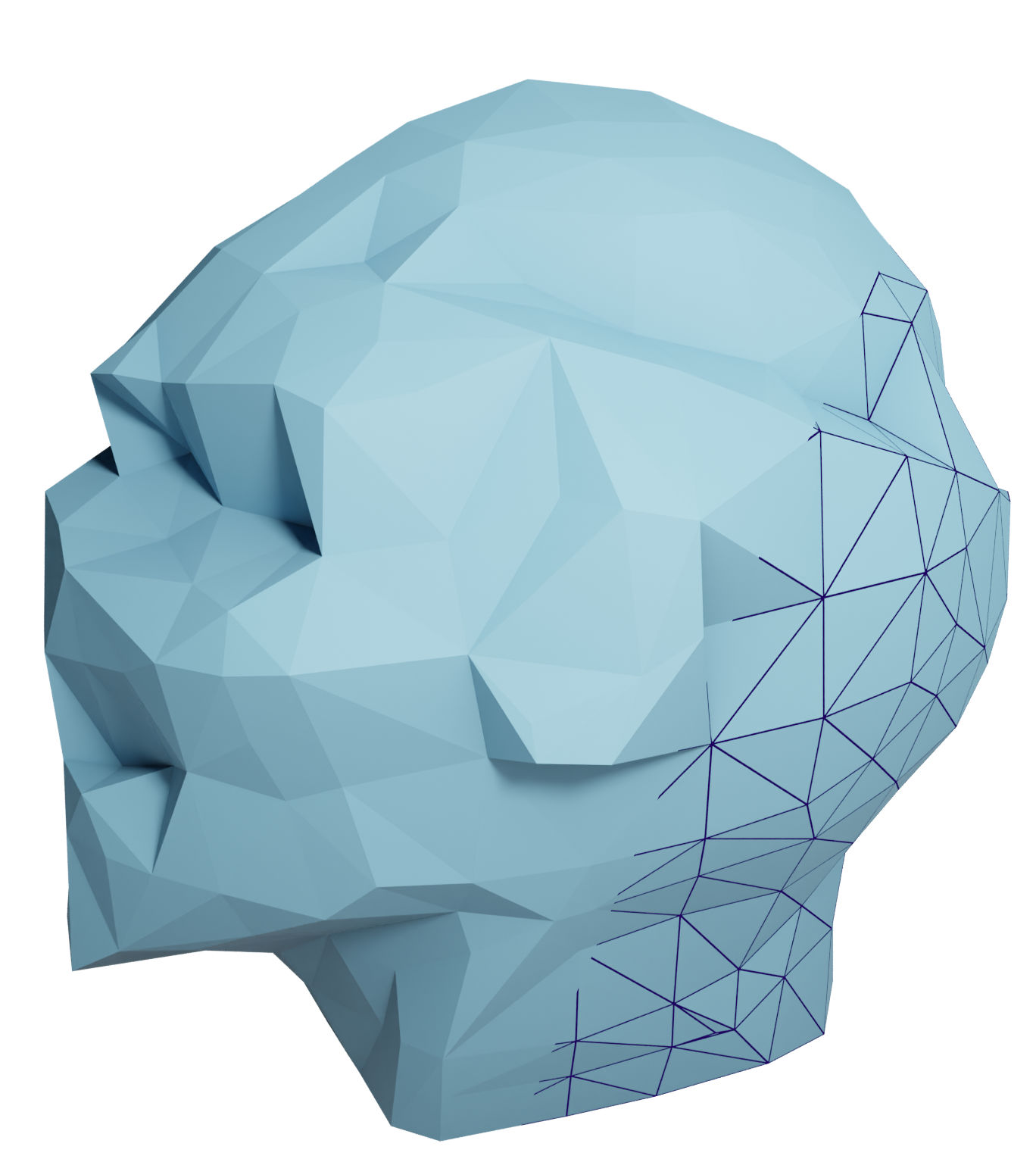} 
  \vspace*{-3mm}
  \caption{Learning-based results. From PoNQ-lite ($128^3$), to power-of-two simplifications down to $16^3$ via average pooling.
  \vspace*{-4mm}}\label{fig:hierarchy}
\end{figure}

The QEM matrices, encoded via points $v_i^*$  
and SPD matrices $A_i$ 
are essential to our work: besides their role in capturing sharp features and second-order shape properties, they (a) disambiguate meshing compared to just point+normals to achieve state-of-the-art results (see supplementary material), (b) allow direct simplification through quadric collapses (PoNQ-lite, Fig.~\ref{fig:lite} \&~\ref{fig:hierarchy}), and (c) allow open boundaries (Fig.~\ref{fig:summary}). Yet, 
compared to SAP or VoroMesh, one may wonder if the added quadric information are worth a higher network size. In fact, producing
QEM information does not significantly affect network size since 80\% of the weights are concentrated in the shared encoder.
Moreover, the task of fitting points, QEM and normals is quite straightforward and does not require overly large networks (e.g., VoroMesh requires 8.4M parameters while PoNQ only requires 2.7M). 

Another possible perceived limitation is that our PoNQ mesh extraction may create non-manifold vertices or edges despite being watertight 
-- for instance, a 1-ring of a vertex may contain two \textit{non-adjacent} tetrahedra both labeled as \emph{inside}. However, one can simply duplicate these few non-manifold elements and connect them to their neighboring mesh elements to enforce manifoldness~\cite{maruani_voromesh_2023}. In addition, final results can be affected by two types of flaws:  wrong estimates of geometric quantities (CNN failure, noise in the motor plate in Fig.~\ref{fig:sdf_abc}), or tetrahedra mislabelling (reconstruction failure, unwanted link elbow-body Fig~\ref{fig:sdf_thingi}).



There are also exciting aspects of PoNQ we have not explored yet. For example, our meshing of open boundaries can potentially be further refined to extract directly the surface through a tagging of boundary edges, instead of relying on a potentially brittle final score-based filtering. A nice extension to our work would be to integrate the PoNQ representation in a differentiable rendering pipeline. Finally, the naturally multiscale nature of PoNQ through average pooling is also bound to be exploitable in a number of contexts.

\section{Conclusion}
\label{sec:conclusion}

We proposed a novel learnable 3D shape representation, coined PoNQ which combines the power of the quadric error metric (QEM) originally devised for mesh decimation and insights from computational geometry. PoNQ relies on points, normals, and QEM matrices to represent local geometric information, which are later leveraged to construct a triangle mesh that is guaranteed to be the boundary of a volume and devoid of self-intersections. We demonstrated the representation power of PoNQ through optimization-based tasks and learning-based reconstruction experiments, showing significant improvement upon previous 3D representations. We thus believe that PoNQ is poised to find many applications and extensions in neural shape processing.

\vspace*{-2mm}

\paragraph*{Acknowledgments.} Work supported by 3IA C\^ote d'Azur (ANR-19-P3IA-0002), ERC Starting Grant 758800 (EXPROTEA), ERC Consolidator Grant 101087347 (VEGA), ANR AI Chair AIGRETTE, Ansys, Adobe Research, and a Choose France Inria chair.


\cleardoublepage
{\small
\bibliographystyle{ieee_fullname}
\bibliography{egbib}
}

\clearpage
\setcounter{page}{1}
\maketitlesupplementary

In this supplemental material,
we discuss further evaluations of our PoNQ representation by comparing it with~\cite{sellan_reach_2023} and~\cite{shen_deep_2021} in Sec.~\ref{sec:additionalComparisons}, before providing  a number of implementation details in Sec.~\ref{sec:ImpDetails} which were not thoroughly covered in the main paper.

\section{Discussing Other Related Works}
\label{sec:additionalComparisons}

To complement the series of comparisons presented in the main paper, we also discuss how PoNQ differs from two other (less) related works. 

\subsection{Reach for The Spheres~\cite{sellan_reach_2023}}

\begin{figure}[h!] \vspace*{-4mm}
 \centering
 \begin{subfigure}{.15\textwidth}
  \centering
  \includegraphics[width=\linewidth]{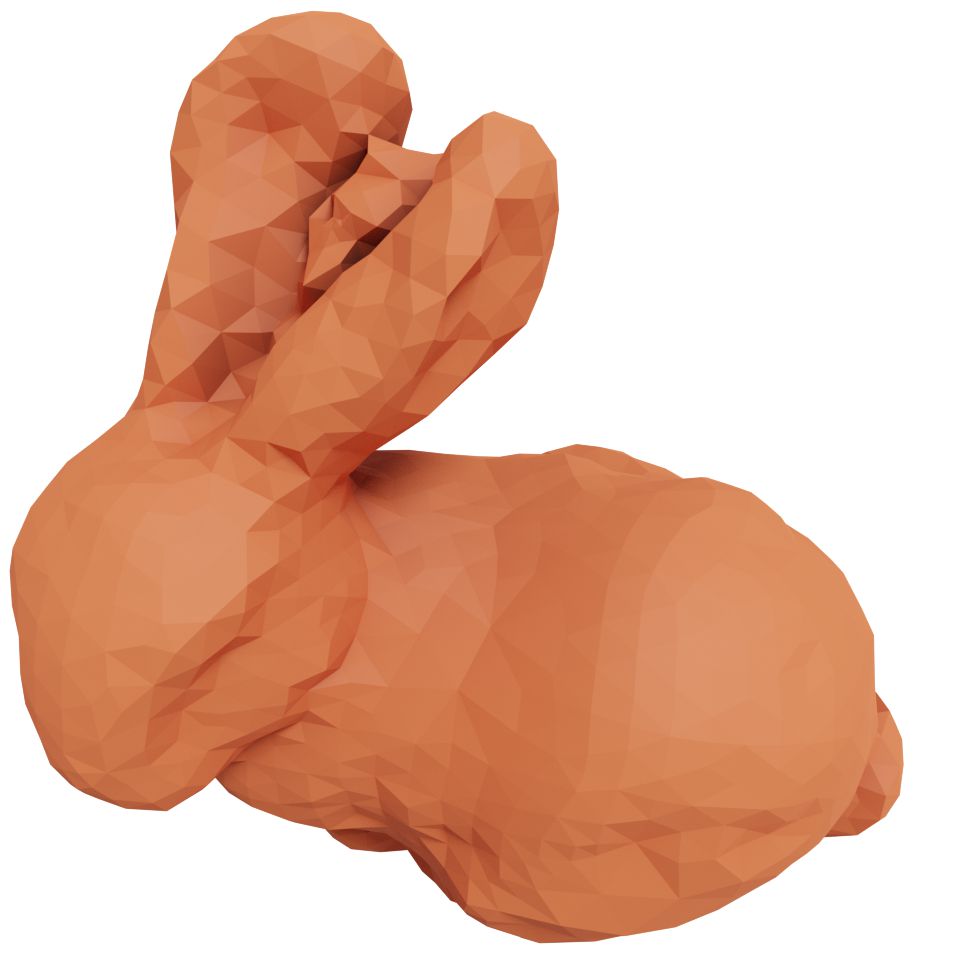} 
\end{subfigure}
 \begin{subfigure}{.15\textwidth}
  \centering
  \includegraphics[width=\linewidth]{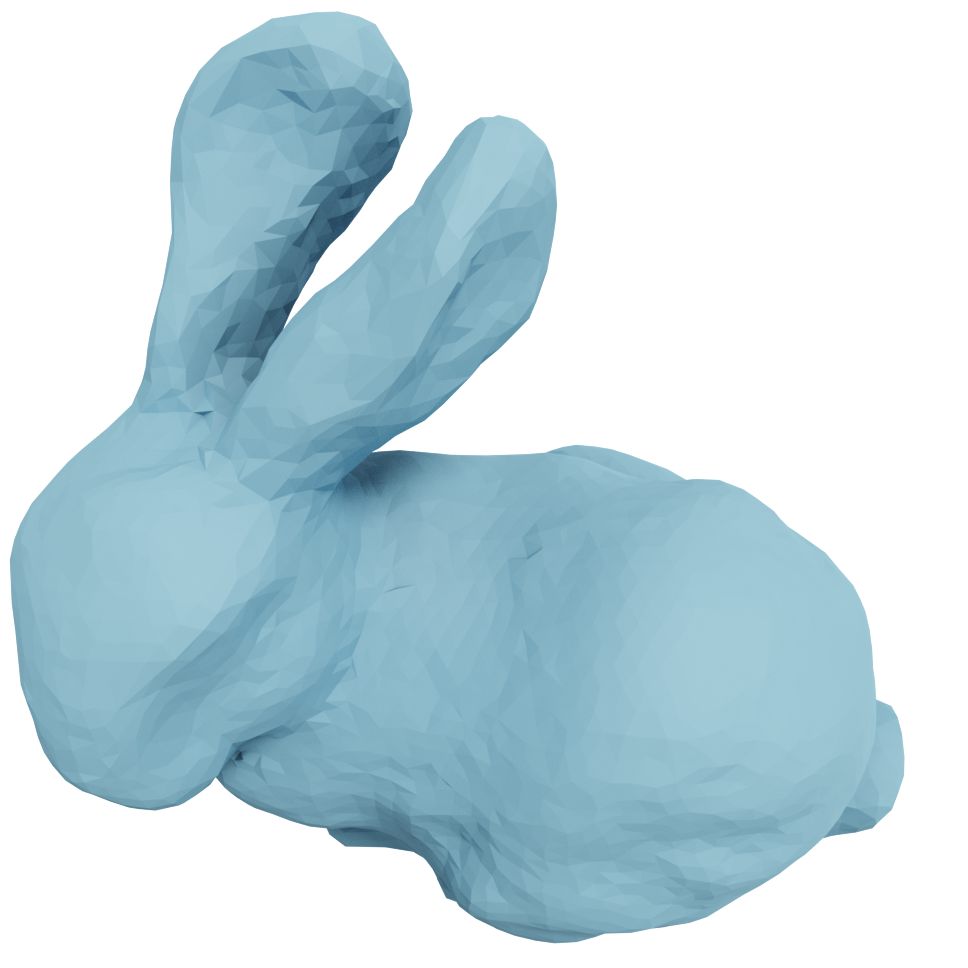} 
\end{subfigure}
 \begin{subfigure}{.15\textwidth}
  \centering
  \includegraphics[width=\linewidth]{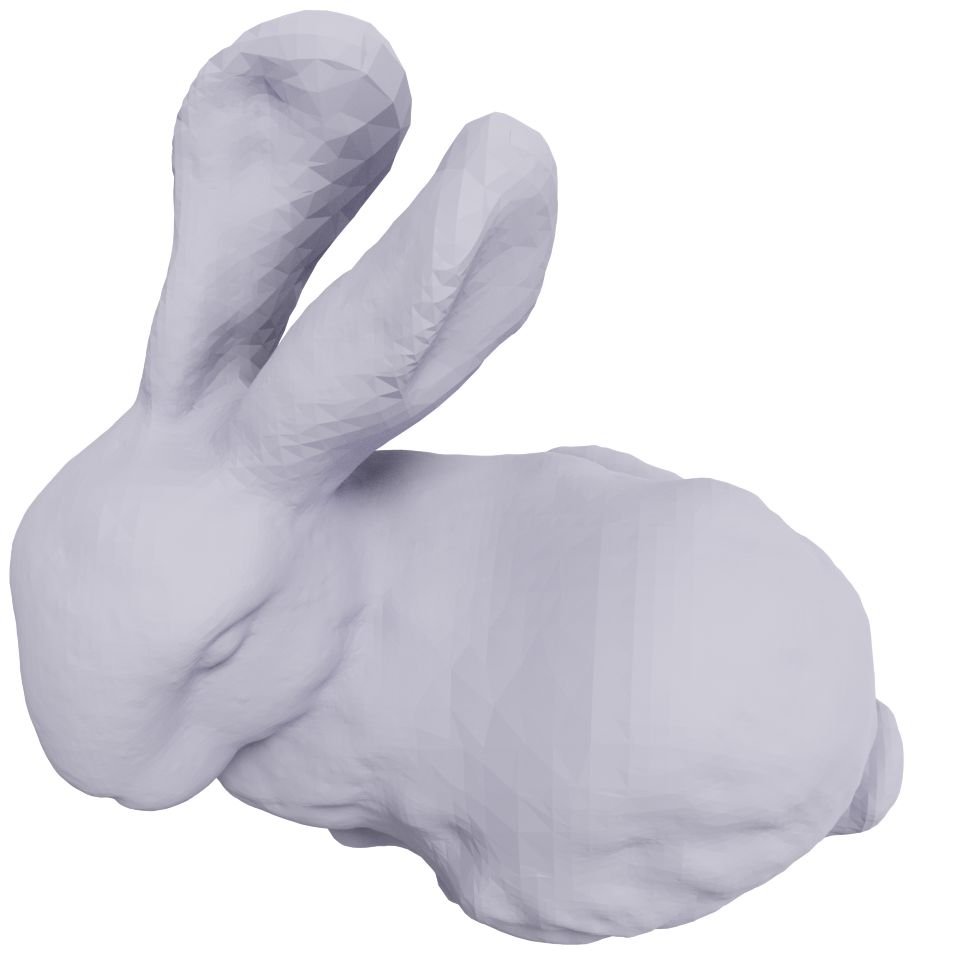} 
\end{subfigure}\\
\vspace*{2mm}
 \begin{subfigure}{.15\textwidth}
  \centering
  \includegraphics[width=\linewidth]{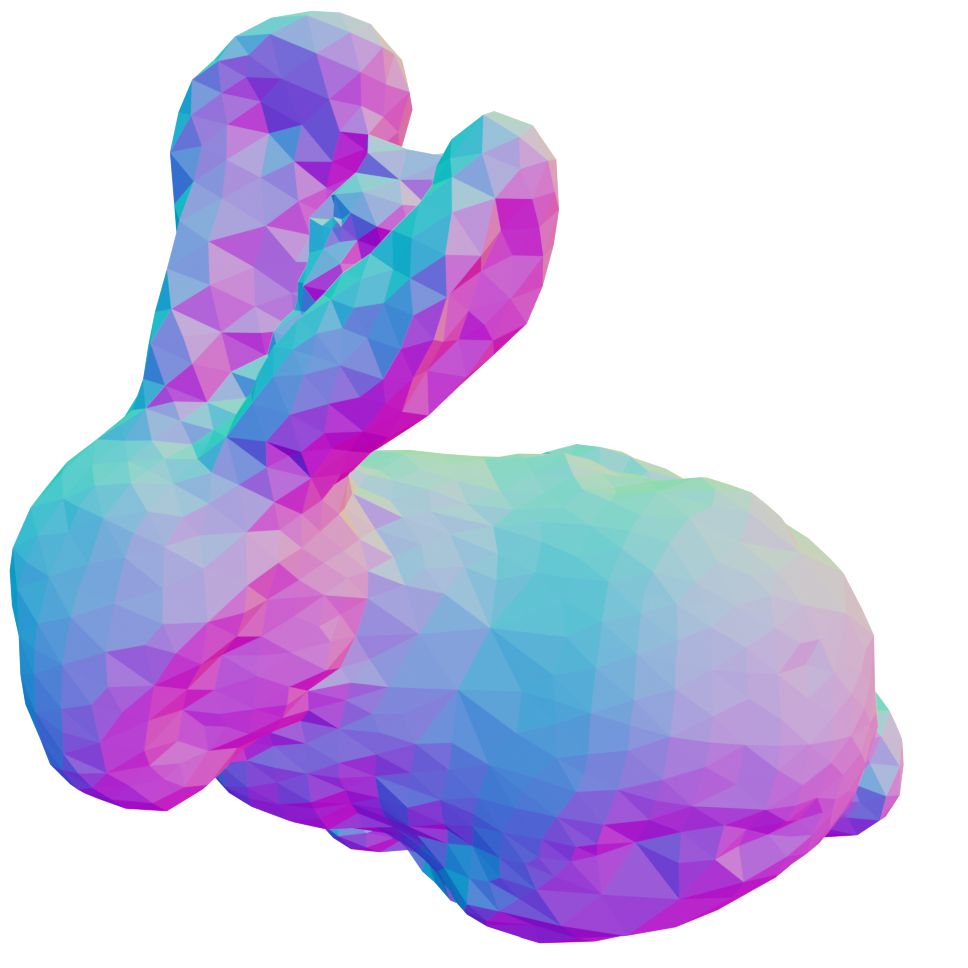} 
\end{subfigure}
 \begin{subfigure}{.15\textwidth}
  \centering
  \includegraphics[width=\linewidth]{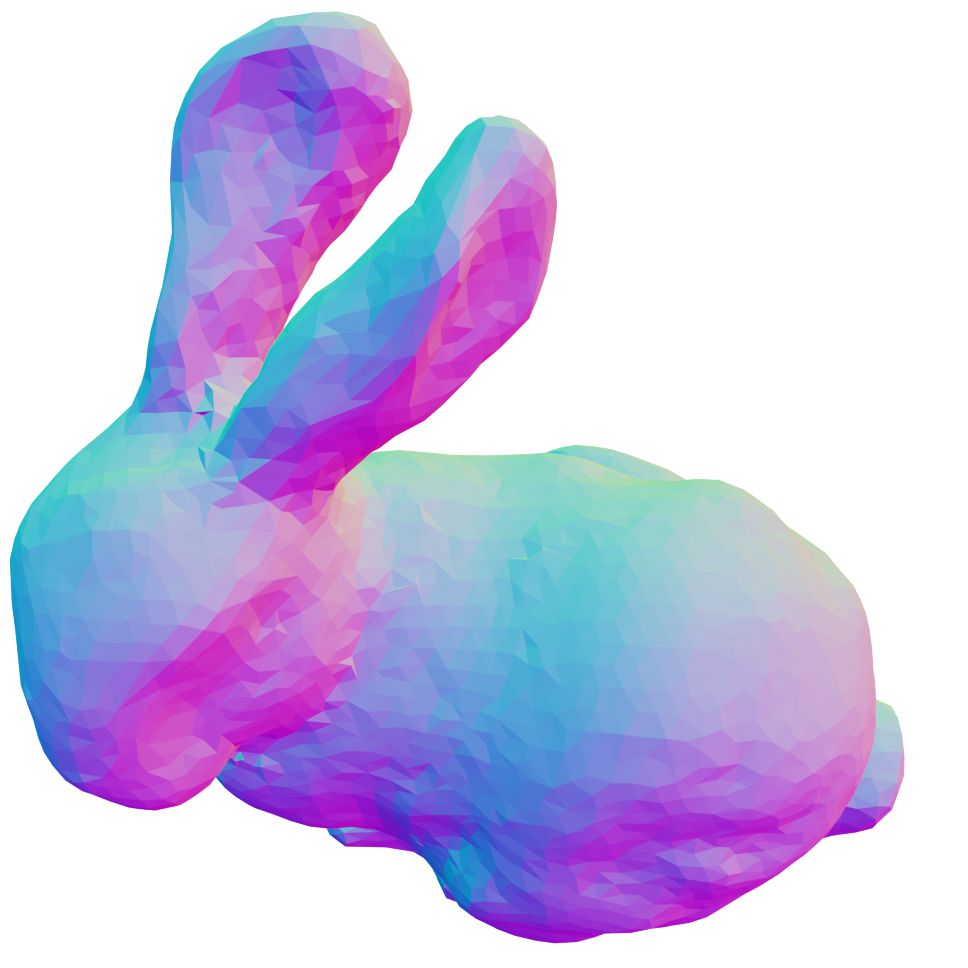} 
\end{subfigure}
 \begin{subfigure}{.15\textwidth}
  \centering
  \includegraphics[width=\linewidth]{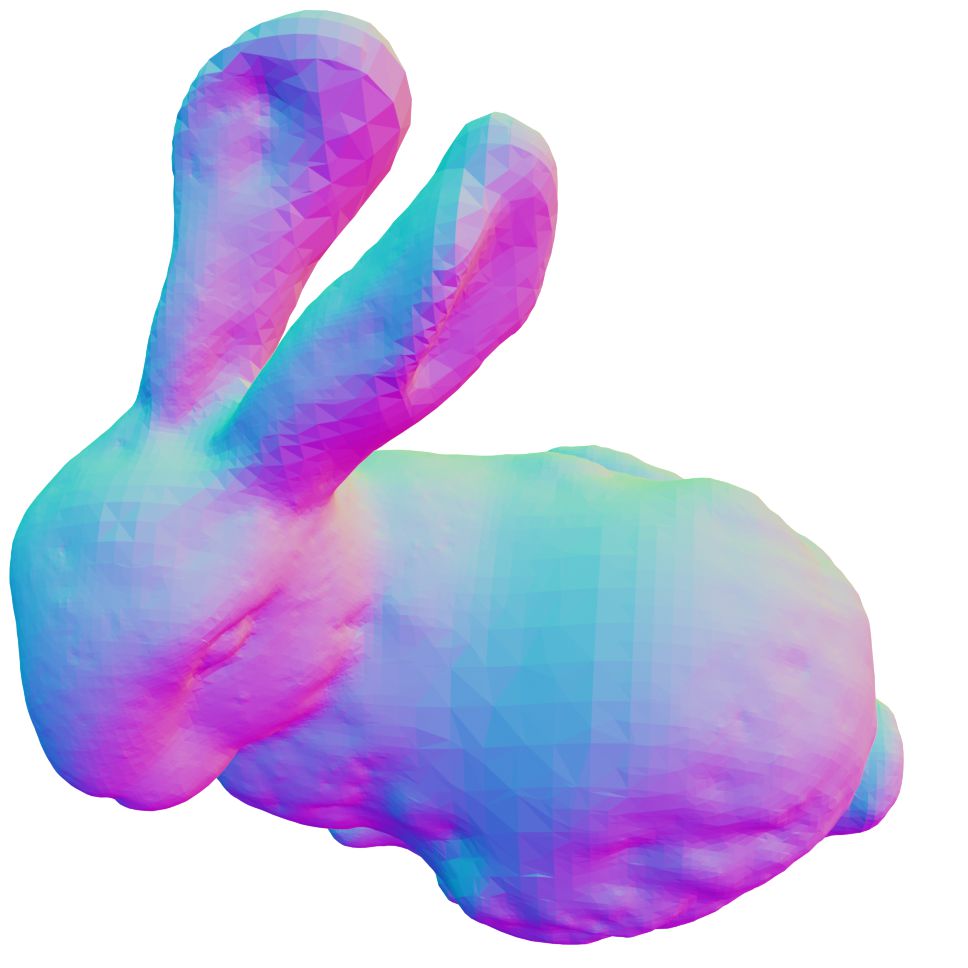} 
\end{subfigure}\\
\vspace*{3mm}
 \begin{subfigure}{.15\textwidth}
  \centering
  \includegraphics[width=\linewidth]{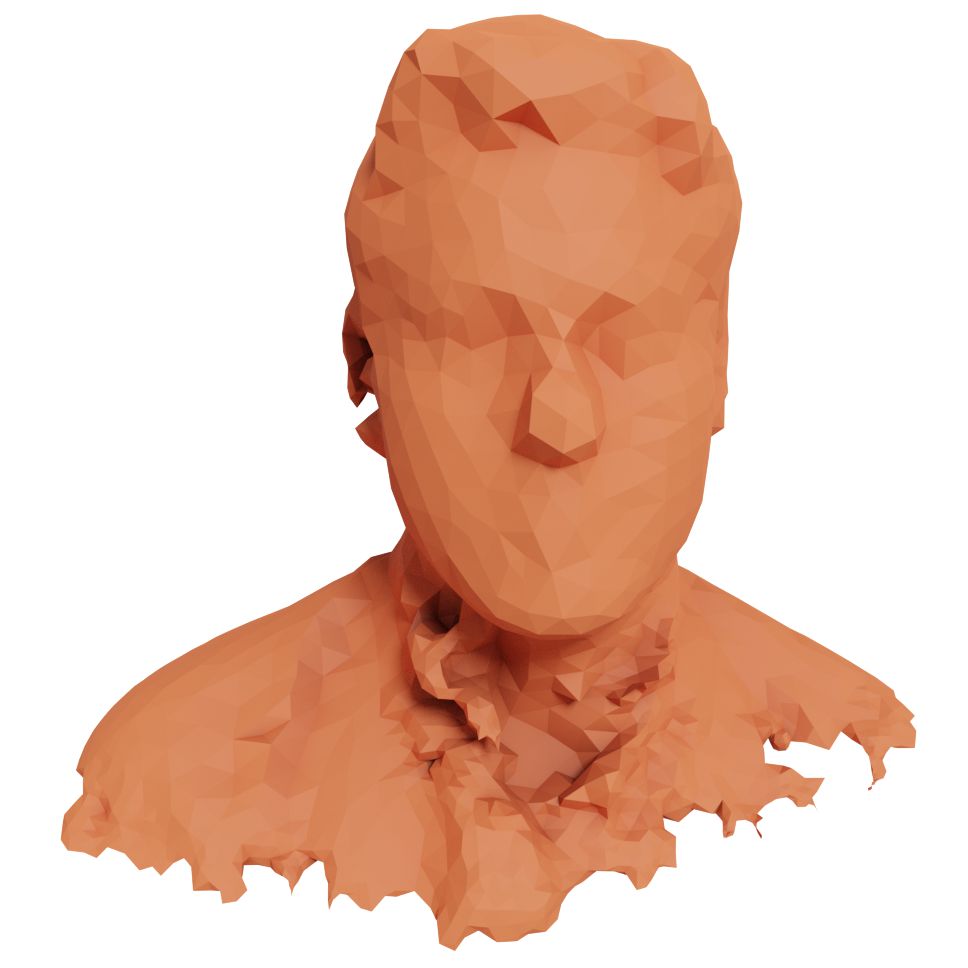} 
\end{subfigure}
 \begin{subfigure}{.15\textwidth}
  \centering
  \includegraphics[width=\linewidth]{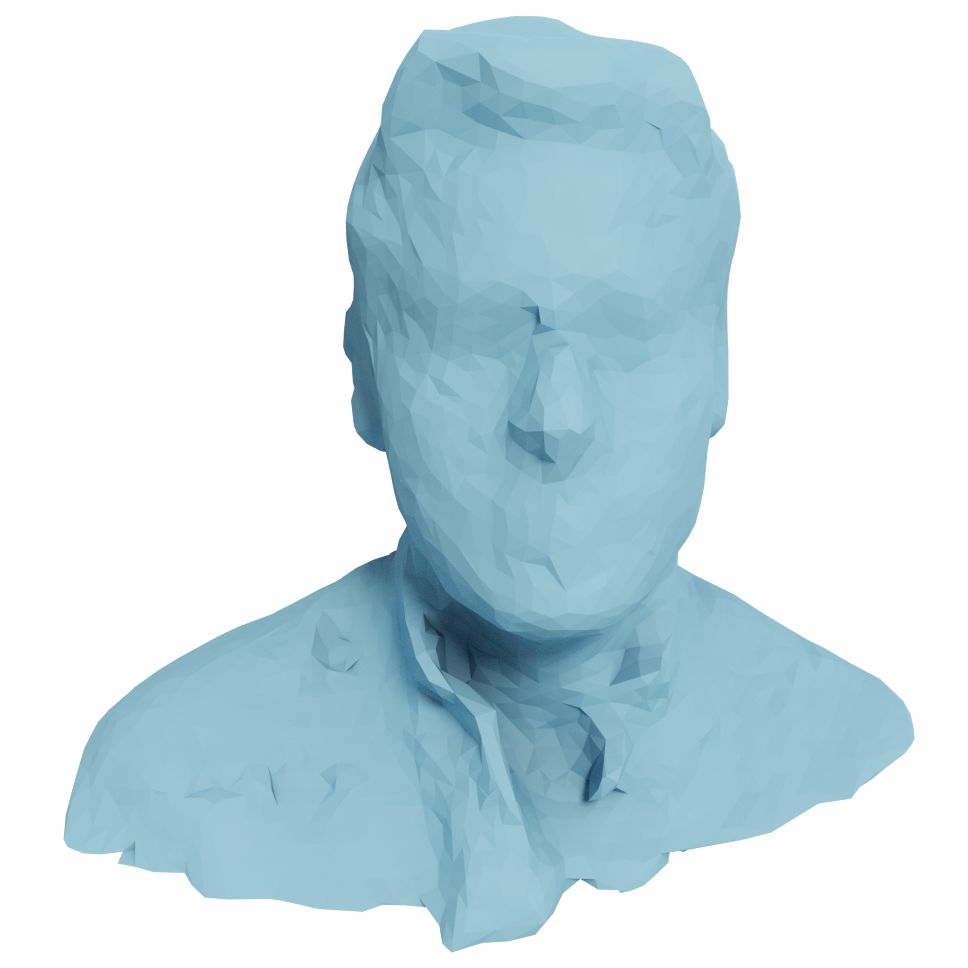} 
\end{subfigure}
 \begin{subfigure}{.15\textwidth}
  \centering
  \includegraphics[width=\linewidth]{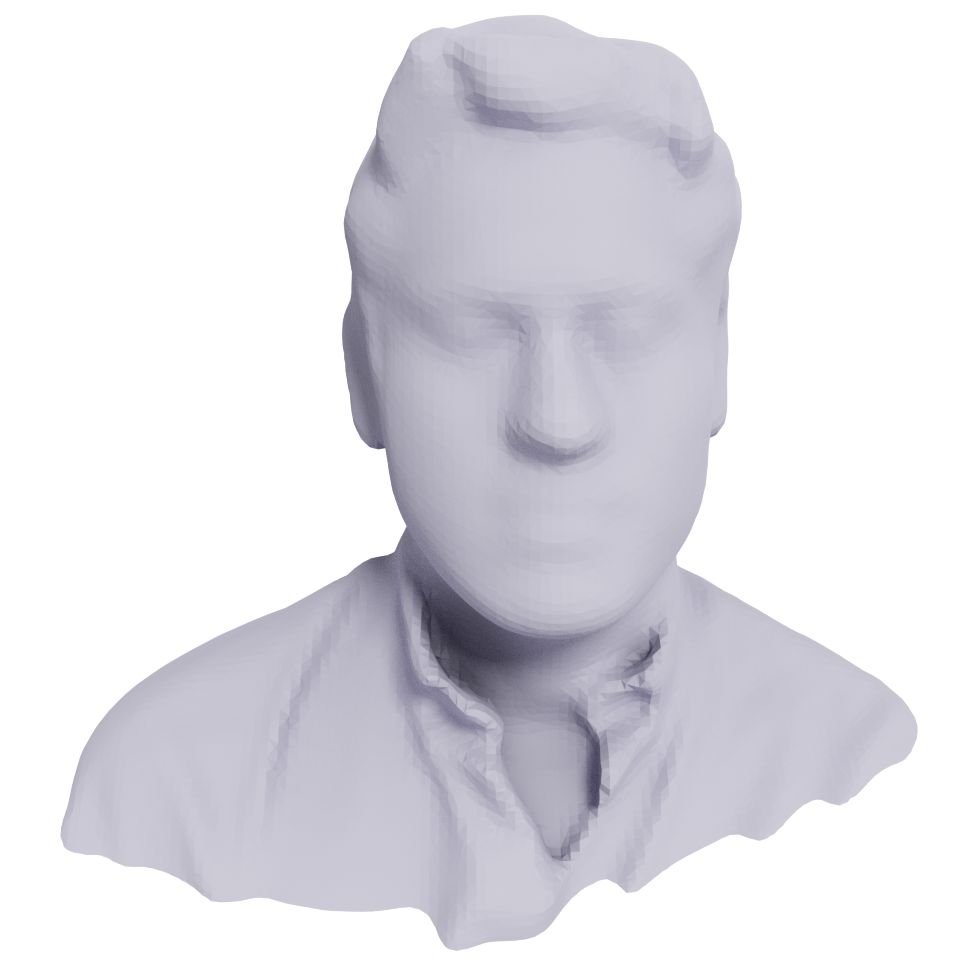} 
\end{subfigure}\\
 \vspace*{3mm}
\begin{subfigure}{.15\textwidth}
  \centering
  \includegraphics[width=\linewidth]{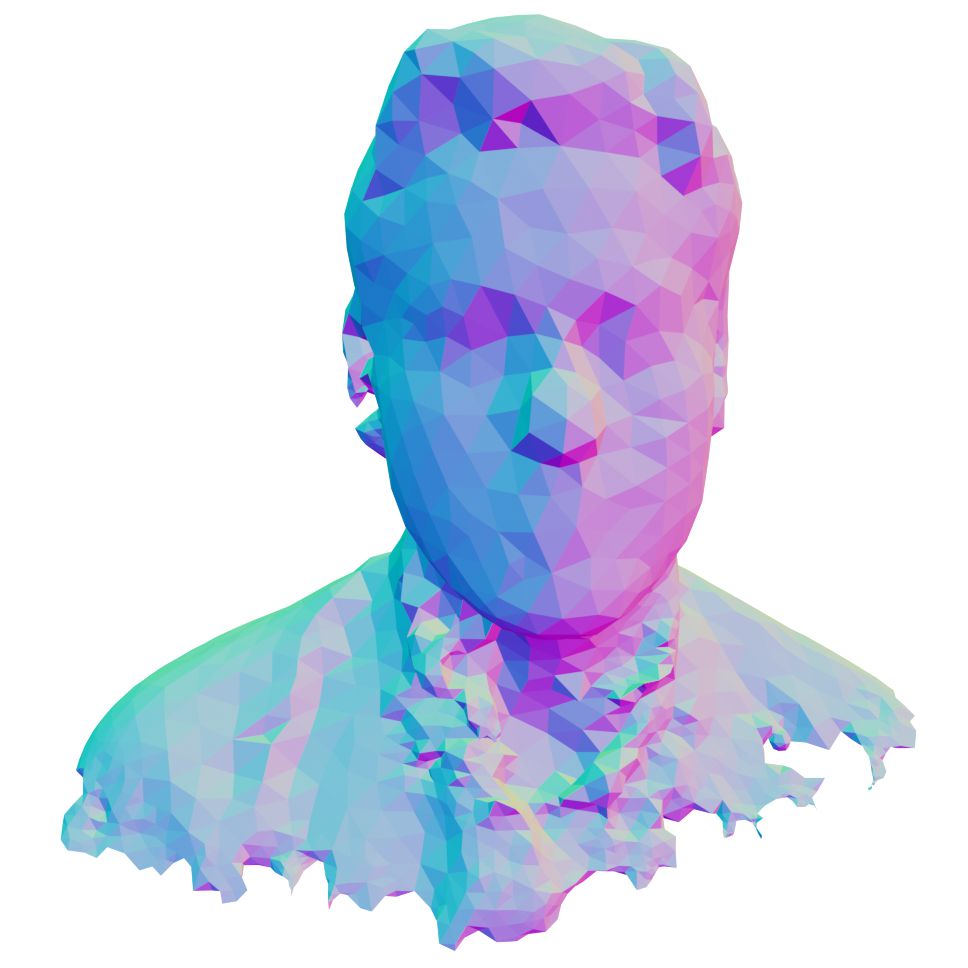} 
  \caption{RTS~\cite{sellan_reach_2023}}
\end{subfigure}
 \begin{subfigure}{.15\textwidth}
  \centering
  \includegraphics[width=\linewidth]{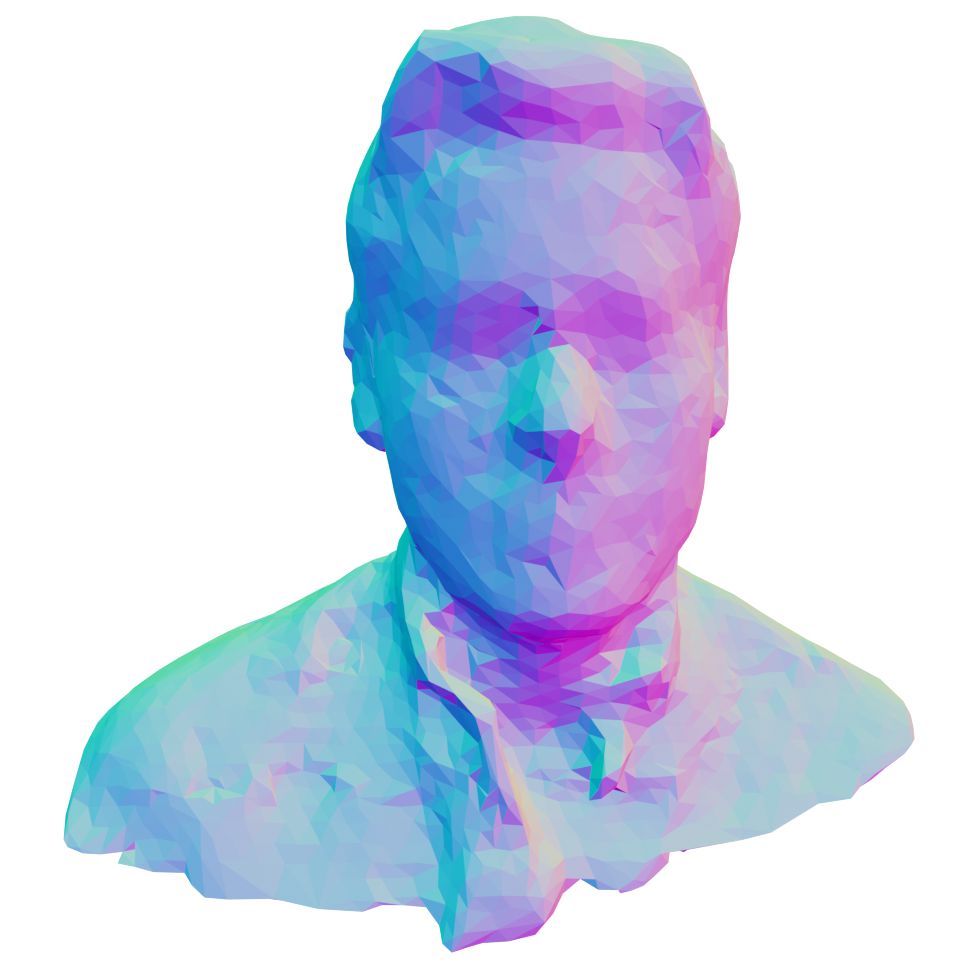} 
  \caption{PoNQ}
\end{subfigure}
 \begin{subfigure}{.15\textwidth}
  \centering
  \includegraphics[width=\linewidth]{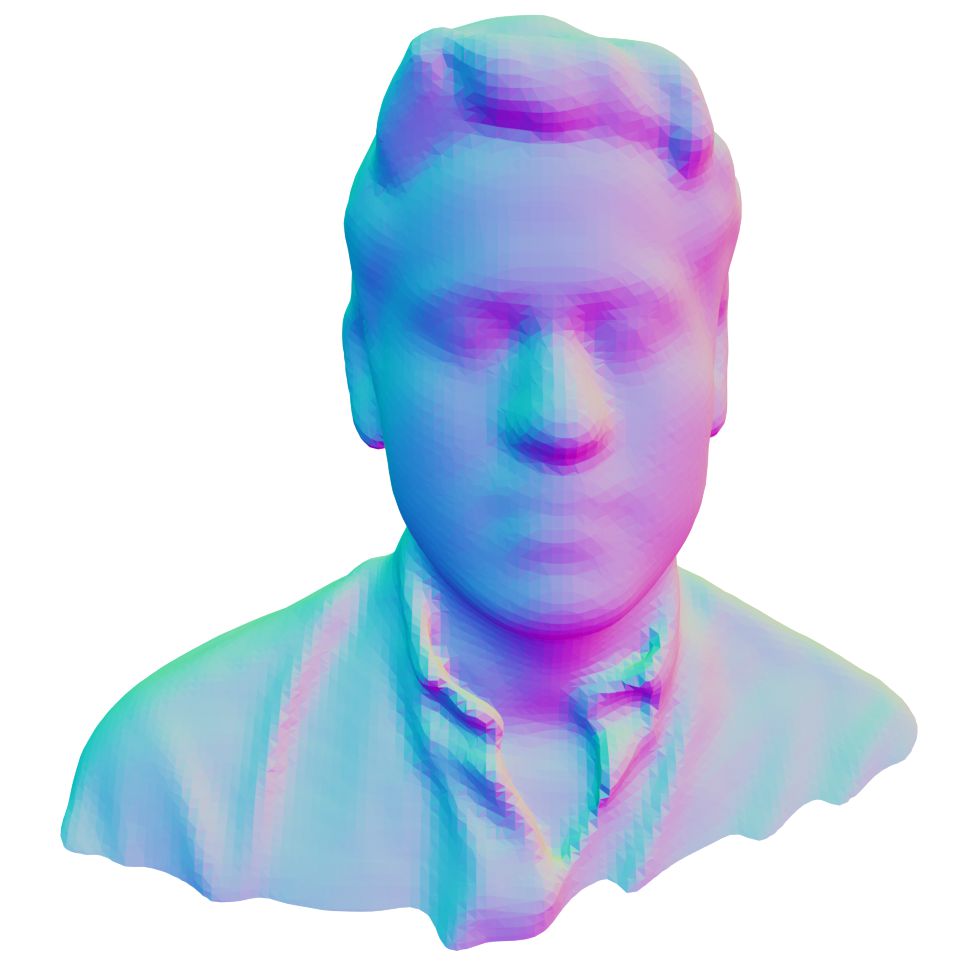} 
  \caption{Ground truth}
\end{subfigure}\vspace*{-2mm}
\caption{Learning-based results for PoNQ. Optimization-based results for Reach For the Spheres~\cite{sellan_reach_2023}. Note that both method rely on the same input, here a $32^3$ SDF grid. RTS can either miss thin structures (bottom rows) or fill thin voids (top rows).\vspace*{-2mm}}
  \label{fig:reach}
\end{figure}

A recent work on surface reconstruction of SDF, called 
Reach for The Spheres (RTS)~\cite{sellan_reach_2023},
relies on the deformation of an initial mesh, thus requiring that its genus matches the genus of the underlying SDF-encoded shape. In practice, several shapes of the Thingi30 have a genus strictly higher than one. Consequently, and despite our best efforts, we could not run the code provided by the authors of Reach for The Spheres (RTS)~\cite{sellan_reach_2023} on the Thingi30 dataset: segmentation faults and inverted elements systematically appeared in the optimization process when starting from a unit sphere --- and using the output of a marching-cube (MC) extraction instead did not solve this problem, since the resulting genus depends on the 
MC table being used. Nevertheless, we provide for completeness a visual comparison of the two methods in Fig.~\ref{fig:reach} for two shapes that RTS could handle. While RTS exhibits many artifacts and PoNQ is far more robust to the genus of shapes, we note that the RTS empty sphere constraint could potentially be added to our loss function.

\subsection{Deep Marching Tetrahedra~\cite{shen_deep_2021}}

Regarding Deep Marching Tetrahedra (DMTet)~\cite{shen_deep_2021}, which relies on the deformation of a regular tetrahedra grid, two critical differences exist between their work and ours: 
\begin{itemize}
    \item strictly speaking, DMTet is not a point-based approach: their hybrid representation requires a neural network to model a continuous field of SDF and displacements;
    \item unless one restricts the amplitude of the deformation, it cannot guarantee intersection-free output meshes.
\end{itemize}
DMTet is thus quite distinct from PoNQ, making it difficult to compare them fairly.
Nonetheless, we tried to provide a comparison; but given the absence of DMTet implementation for surface reconstruction from SDF grids, we found ourselves unable to provide a meaningful visual comparison --- just like VoroMesh~\cite{maruani_voromesh_2023} was unable to compare with DMTet for the simpler case of the optimization-based task.
\vspace*{-2mm}

\section{Implementation Details}
\label{sec:ImpDetails}
We now cover in detail a series of aspects of PoNQ that were not fully detailed in the main paper.

\subsection{Training}

For all the examples we show in our paper, we trained our PoNQ network with the AdamW optimizer and a linear combination of the losses, i.e., \vspace*{-2mm} 
\begin{align*}
    L = \alpha_{\operatorname{CD}} L_{\operatorname{CD}} + \alpha_{\mathbf{n}} L_{\mathbf{n}} + \alpha_{\mathbf{A}} L_{\mathbf{A}} + \alpha_{\mathbf{v}^*} L_{\mathbf{v}^*}  \\ + \alpha_{reg} L_{reg} +\alpha_{occ}  L_{occ} \vspace*{-2.5mm} 
\end{align*}

We rely on three training phases of 200 epochs each, with batches of size 16 with the following learning rates $\gamma$, weights $\boldsymbol{\alpha} = (\alpha_{\operatorname{CD}}, \alpha_{\mathbf{n}}, \alpha_{\mathbf{A}} , \alpha_{\mathbf{v}^*} , \alpha_{reg} , \alpha_{occ})$ and point samples $S$:

\begin{itemize}
    \item $\gamma = 6.4 \cdot 10^{-5}$, $\boldsymbol{\alpha} = (100, .1, .1, 100, 100, .1)$, $S= 5 \cdot 10^5$
    \item $\gamma =3.2\cdot 10^{-5}$, $\boldsymbol{\alpha} = (100, .1, .1, 100, 100, .1)$, $S= 7 \cdot 10^5$
    \item $\gamma = 3.2 \cdot 10^{-5}$, $\boldsymbol{\alpha} = (100, .1, .1, 100, 1, .1)$, $S= 7 \cdot 10^5$.
\end{itemize}

\subsection{Meshing}

As explained in the main paper, the choices we made to design our QEM-based PoNQ representation allow for a simple meshing approach, inspired by computational geometry to ensure robustness: due to the local optimality of the positions $\mathbf{v}_i^\textbf{*}$ (in particular, their ability to capture corners and sharp features), our output mesh is produced by filtering the triangle facets of a 3D Delaunay triangulation of the optimal positions $\mathbf{v}_i^\textbf{*}$.
Given the PoNQ data (produced by a trained network or through optimization), we now describe how a PoNQ mesh is extracted in full detail.

\tightpara{0. Pre-processing} We first normalize the quadrics by dividing them by their largest eigenvalue. While this technically changes their meaning (they now measure Euclidean distances \emph{up to a multiplicative constant}), it also removes the possible bias due to variable sampling density. Additionally, a bounding box $B$ of all the points $\mathbf{v}_i^\textbf{*}$ is computed.

\tightpara{1. \!Triangulation of QEM optimal positions.\!\!}
We then compute the Delaunay tetrahedralization of all points $\mathbf{v}_i^\textbf{*}$, to which we add eight ``protective'' points defined as the corners of the bounding box $B$: all the adjacent tetrahedra to these protective points will be guaranteed to be outside of the shape we wish to reconstruct.  
The next two steps will tag each of the remaining tetrahedra as either \emph{inside} or \emph{outside} based on local geometric information, so that \emph{our final PoNQ mesh will simply be the triangle mesh forming the boundary between the inside and outside regions}, ensuring watertightness and no self-intersections by design. 

\tightpara{2.\! Tagging obvious inside/outside tetrahedra.}
We first tag all the tetrahedra adjacent to the eight protective point as \emph{outside}. Tagging the rest of the tetrahedra seems daunting, but by leveraging the ideas put forth in the Crust algorithm~\cite{amenta_new_1998}, we note that the location of the circumcenter of a Delaunay tetrahedron whose four vertices are on the surface to mesh (which is the case we are in) can often determine the insideness or outsideness of this tetrahedron --- corresponding to whether this circumcenter is on the inside (resp. outside) medial axis. We use a similar approach here, except that our Delaunay vertices have additional information to help us: we also know a local normal $\mathbf{n}_i$ in the vicinity of each vertex $\mathbf{v}_i^\textbf{*}$. Thus, each vertex and its assigned normal defines an oriented plane, forming the boundary between the outside half-space (the one pointed by the normal) and the inside half-plane. Using this local test to determine inside/outside, we tag a tetrahedron as \emph{outside} (resp., \emph{inside}) if both its circumcenter and barycenter are determined to be in the outside (resp., inside) half-space of each of its four vertices. 
We then go through every (non-protective) vertex that already has at least one tagged adjacent tetrahedron. If such a vertex does not have any adjacent \emph{outside} (resp., \emph{inside}) tetrahedron, we pick its untagged adjacent tetrahedron with the smallest edge; if the size of this smallest edge is below a certain large threshold (i.e., we are not in a very sparse region of the domain), we then tag this selected tetrahedron as \emph{outside} (resp., \emph{inside}). 
The rationale behind this last round of tagging is that we know that all vertices of our 3D Delaunay triangulation are \emph{on} the surface, so unless the smallest tetrahedron is too big (in this case, there is clearly a large uncertainty), we can safely tag it to be on the opposite side of the surface than what the other tags had already determined. While this tagging procedure can, on rare occasions, tag \emph{all} tetrahedra, there are often a few remaining untagged tetrahedra (typically caused by the presence of near sharp features or thin structures) that are too ambiguous to tag. We now need to lift the remaining uncertainty based on additional PoNQ data. 
\tightpara{3.\! Finishing up triangle selection.\!}
Delaunay-based meshing approaches (like \emph{Crust}) require a dense point sampling (formally, an $\epsilon-$sampling) to offer topological and geometric guarantees, which is not compatible with our desire to deal with thin structures, sharp features and corners --- and this is the main reason why our earlier phase often ends up not providing a tag for \emph{every} tetrahedron. To finish our tetrahedron tagging based on the ones we already have, we propose to use a graph cut approach, inspired by existing spectral graph partitioning~\cite{kolluri_spectral_2004}. 
For each Delaunay triangle $T$ between two adjacent tetrahedra for which \emph{at least} one of them is still undetermined, we compute a likelihood score $S(T)$ that evaluates how confident we are that this triangle is to appear on the final output mesh. We propose a score that evaluates the  fitness of $T$ based on the local PoNQ normals and the local PoNQ quadrics matrices: \vspace*{-2.5mm}
\[S(T) = S_\mathbf{n}(T) + h S_\mathbf{Q}(T)\vspace*{-2.5mm}\] (with $h$ set as the squared inverse of the edge length of the SDF grid), where: \vspace*{-2.5mm}
\begin{align*}
S_\mathbf{n}(T) &=  \bigl(\tfrac2\pi \textstyle\sum\limits_{\mathbf{v}_i^\textbf{*} \in T} \arccos(n_{T}^t \,\mathbf{n}_i) \bigr)^2,   \\[-1mm]
S_\mathbf{Q}(T) &=  \textstyle\sum\limits_{ \mathbf{v}_i^\textbf{*} \in T}  \textstyle\sum\limits_{\substack{ \mathbf{v}_j^\textbf{*} \in T \\ i \neq j}} [\mathbf{v}_j^\textbf{*}, 1]^t \mathbf{Q}_i [\mathbf{v}_j^\textbf{*}, 1],\\[-9mm]
\label{eq:meshing}
\end{align*}
where $n_{T}$ denotes the normal of triangle $T$. We can now tag the remaining undetermined tetrahedra with a definite  \emph{inside} or \emph{outside} label: we compute a minimum cut of the Voronoi graph (in which each dual of a tetrahedron is a node, and each dual of a Delaunay triangle face is an edge) using the already-tagged ``inside'' ones as a source and the ``outside'' ones as a drain, and each edge weight between two tetrahedra (with a common face $T$)  set to the score $S(T)$.
Since most of tetrahedra are already tagged, we can merge Voronoi edges between marked tetrahedra to reduce the graph size and thus accelerate computations. We now just extract the final \textit{PoNQ mesh} as the triangle mesh forming the boundary between the inside and outside regions. 


\subsection{Ablation studies}\label{sec:ablation}
\begin{figure}[!h]\vspace*{-4mm}
  \centering
   \begin{subfigure}{.35\linewidth}
  \centering
  \includegraphics[width=\linewidth]{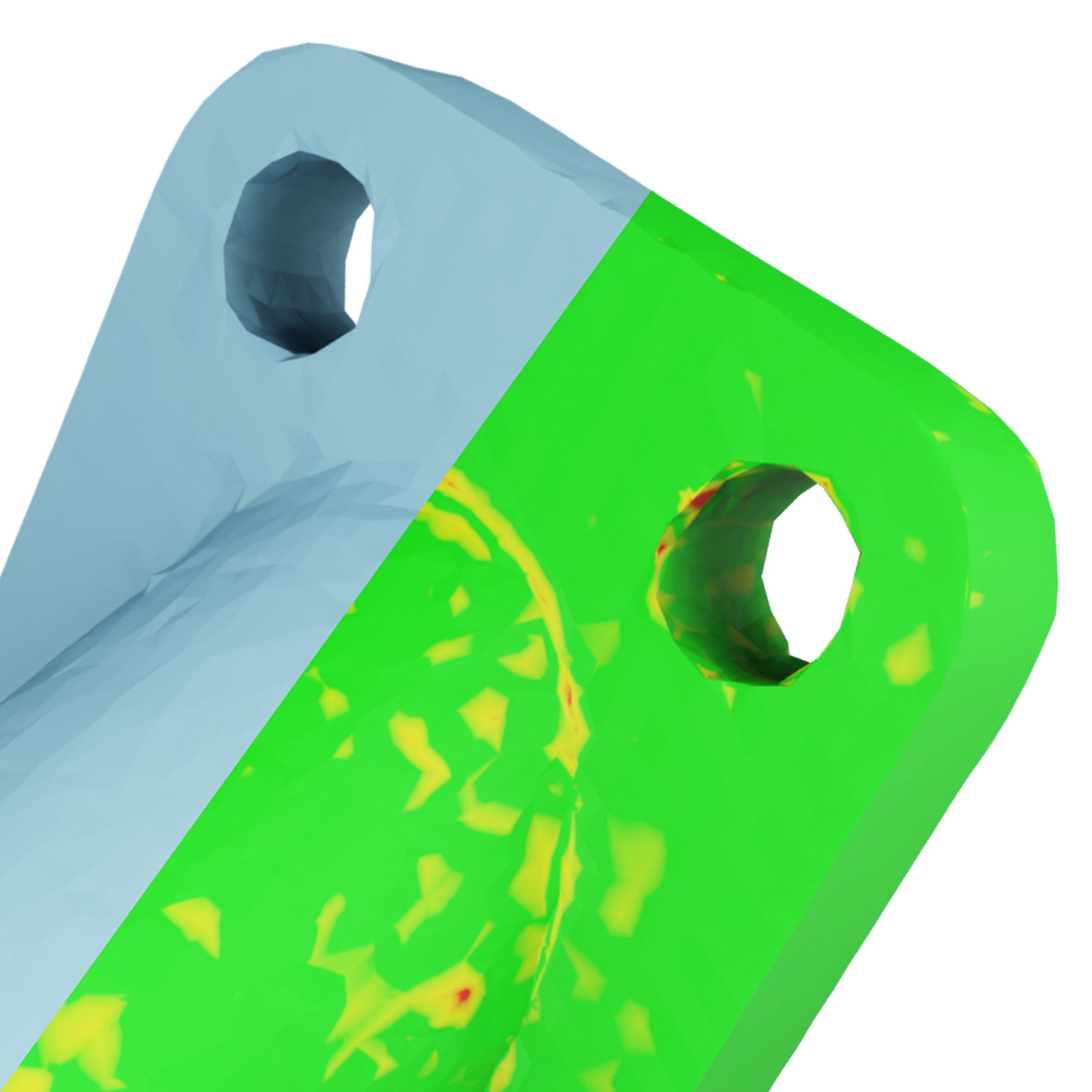} 
    \caption{PoNQ}
\end{subfigure}
  \begin{subfigure}{.35\linewidth}
  \centering
  \includegraphics[width=\linewidth]{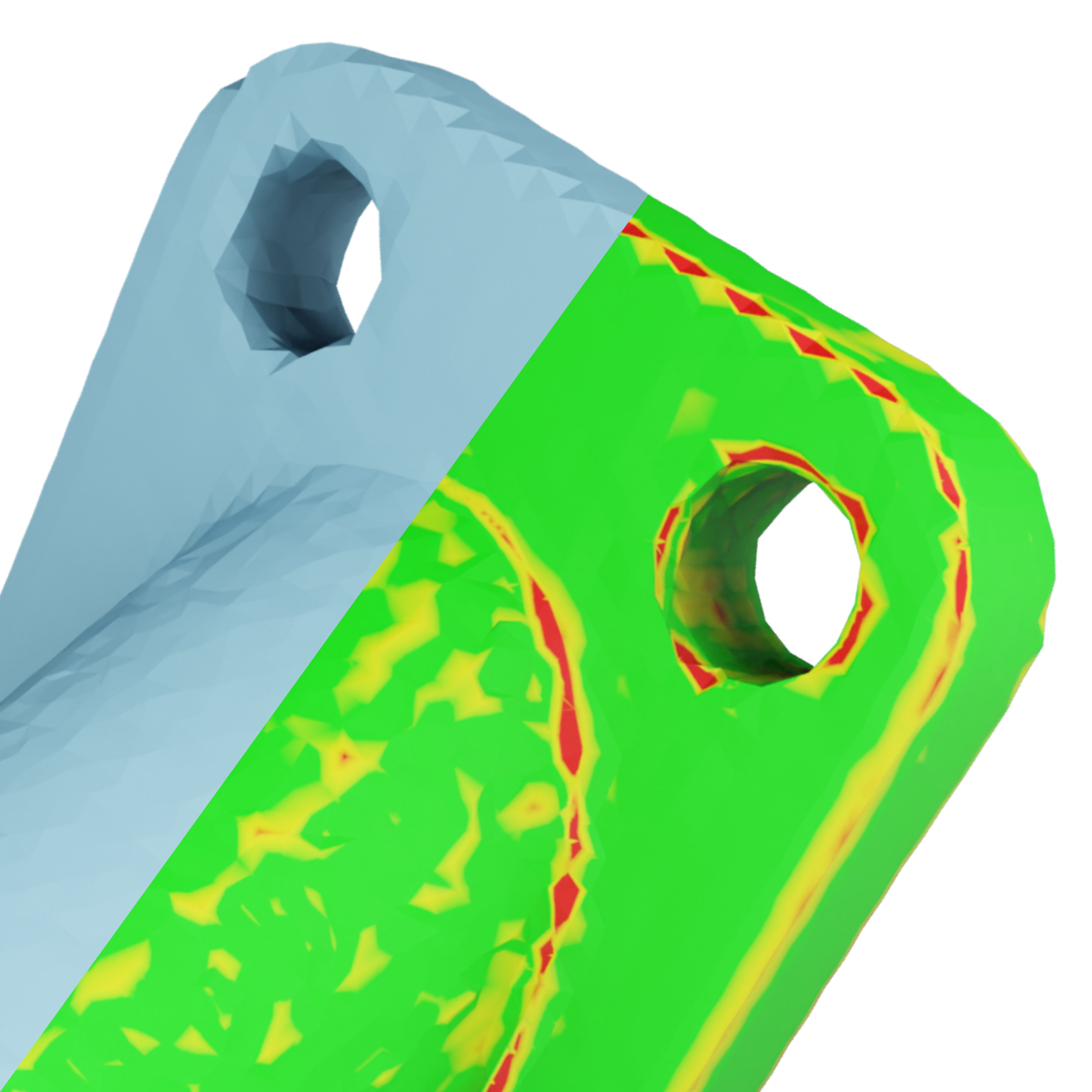} 
    \caption{PoNQ w/out QEM}
\end{subfigure}
  \centering
   \begin{subfigure}{.35\linewidth}
  \centering
  \includegraphics[width=\linewidth]{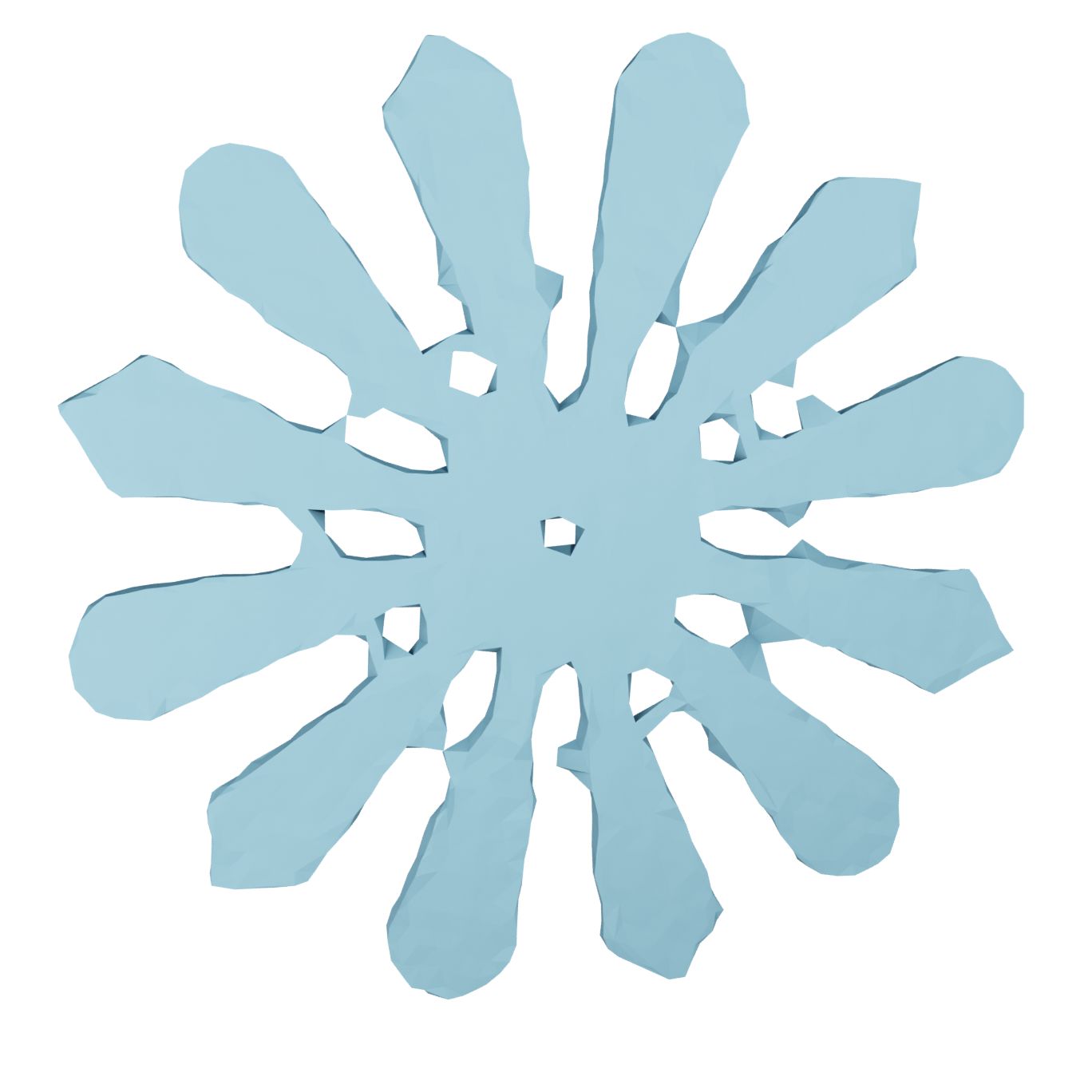} 
    \caption{PoNQ}
\end{subfigure}
  \begin{subfigure}{.35\linewidth}
  \centering
  \includegraphics[width=\linewidth]{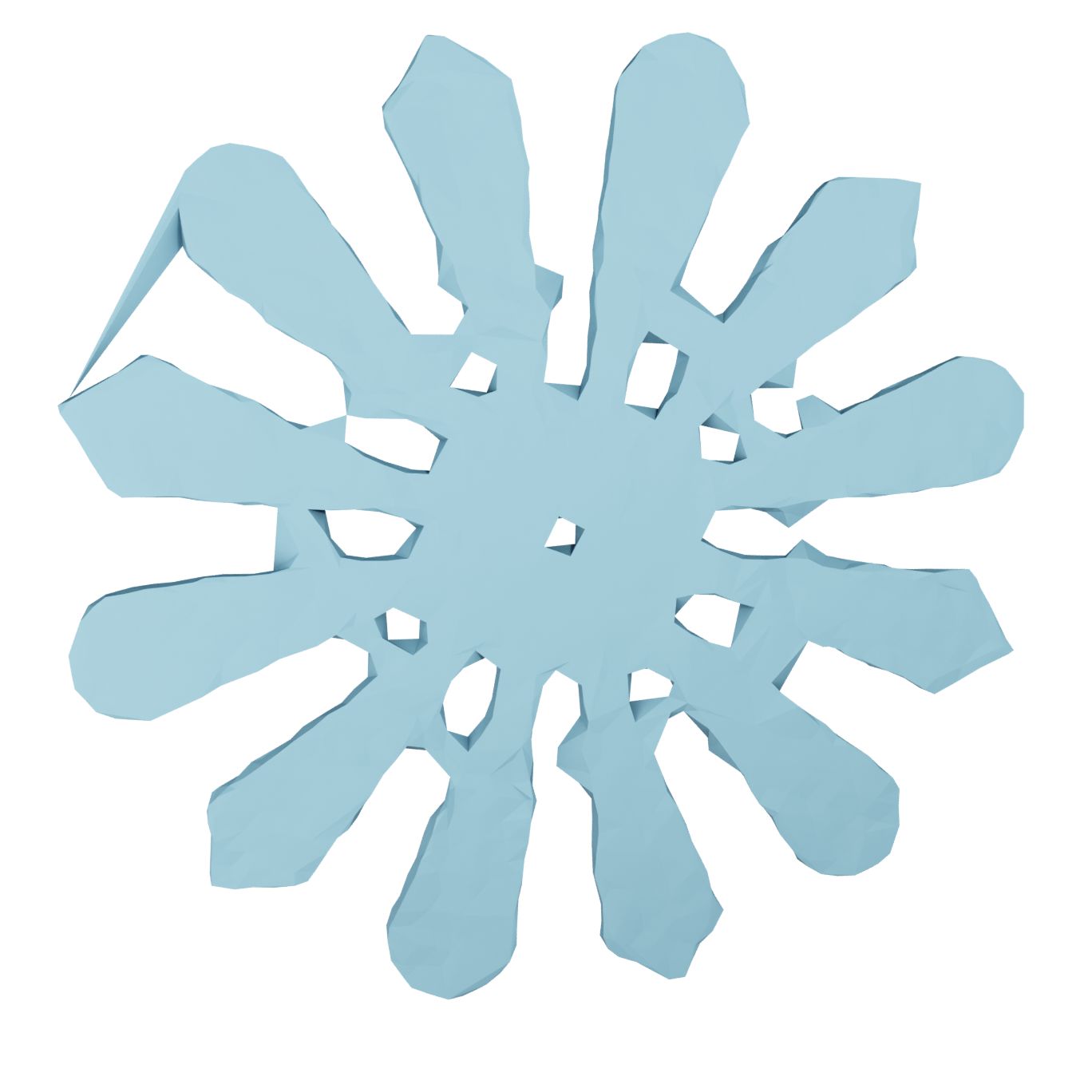} 
    \caption{PoNQ w/out $S_Q$}
\end{subfigure}
    \caption{A network trained without QEM (b) fails to recover sharp edges. The second-order QEM information provided by $S_Q$ helps to disambiguate tetrahedra labeling (c).}
    \label{fig:ablation_quadrics}
\end{figure} 

To further justify the need for QEM, we provide two ablation studies: we re-generated models with a simpler version of our meshing algorithm that does not rely on quadrics (thus removing $S_Q$), and we re-trained a network producing only points, normals, and voxel occupancies (thus removing the QEM part and its associated losses $L_A$, $L_{v^*}$, and $L_{reg}$). Figure~\ref{fig:ablation_quadrics} shows that the QEM-optimal vertex placement is essential to fit sharp features, and that using only $S_n$ can mess up labeling in intricate/thin regions. Quantitatively, non-QEM versions fall behind our competitors in both surface and sharp-edge fitting scores, see Table~\ref{tab:ablation}.

\begin{table}[h] 
\vspace{0mm} 
  \begin{center}
    \resizebox{1.\linewidth}{!}{\begin{tabular}{l c c c c c c }
  \hline
  \textbf{Method}            & Grid & CD $\downarrow$               & F1  $\uparrow$  & NC $\uparrow$ & ECD $\downarrow$ & EF1 $\uparrow$ \\
                             & size & ($\times 10^{-5}$) &  &   &   &            \\

  \hline
retrained w/o QEM & $32^3$    & \textbf{1.327}  &  0.840  &  0.960 & \textbf{0.184} & 0.598   \\
  PoNQ w/o $S_Q$ & $32^3$    & 1.801      &  0.851  &   \textbf{0.964} &  0.191 & \textbf{0.715}    \\
  PoNQ & $32^3$    & 1.514  &  \textbf{0.852}  &  \textbf{0.964} & \textbf{0.184} & 0.713      \\

  \hline
  retrained w/o QEM & $64^3$   & 0.921  &  0.891  &  0.979 & 0.115 & 0.837     \\
    PoNQ w/o $S_Q$ & $64^3$    & 0.931  &  \textbf{0.892}   & \textbf{0.980} & \textbf{0.103} & 0.863    \\
    PoNQ   & $64^3$   & \textbf{0.886}  &  \textbf{0.892}  &  \textbf{0.980} & 0.109 & \textbf{0.866} \\
  \hline
  \hline
retrained w/o QEM & $32^3$    & 1.387  &  0.803  &  0.942 & 0.139 & 0.286   \\
  PoNQ w/o $S_Q$ & $32^3$    & 1.475  &  \textbf{0.810}  &  \textbf{0.943} & \textbf{0.137} & \textbf{0.315}   \\
    PoNQ  & $32^3$  &   \textbf{1.344}  &  \textbf{0.810}  &  0.942 & \textbf{0.137} & 0.314  \\
  \hline
retrained w/o QEM & $64^3$    & 0.784  &  0.922  &  \textbf{0.971}  & 0.102 & 0.489 \\
  PoNQ w/o $S_Q$ & $64^3$    & 0.779  &  \textbf{0.924}  &  \textbf{0.971}  & 0.102 & \textbf{0.511} \\
    PoNQ & $64^3$  &  \textbf{0.758}  &  \textbf{0.924}  &  \textbf{0.971} & \textbf{0.100} & \textbf{0.511}     \\
  \hline
retrained w/o QEM & $128^3$    & 0.645  &  \textbf{0.939}   &  \textbf{0.984} & 0.135 & 0.556  \\
  PoNQ w/o $S_Q$ & $128^3$    & 0.671  &  0.938  &  \textbf{0.984} & 0.126 & 0.584    \\
    PoNQ  & $128^3$  &  \textbf{0.641}  &  \textbf{0.939} &  \textbf{0.984} & \textbf{0.123} & \textbf{0.592}   \\
  \hline

\end{tabular}}
    \vspace*{-2mm}
    \caption{Ablation studies on ABC (top) and Thingi30 (bottom).}
    \label{tab:ablation}
  \end{center}
\end{table}

\subsection{Open surfaces}
\label{sec:boundary}

\begin{figure}[h!] \vspace*{-4mm}
 \centering
 \begin{subfigure}{.11\textwidth}
  \centering
  \includegraphics[width=\linewidth]{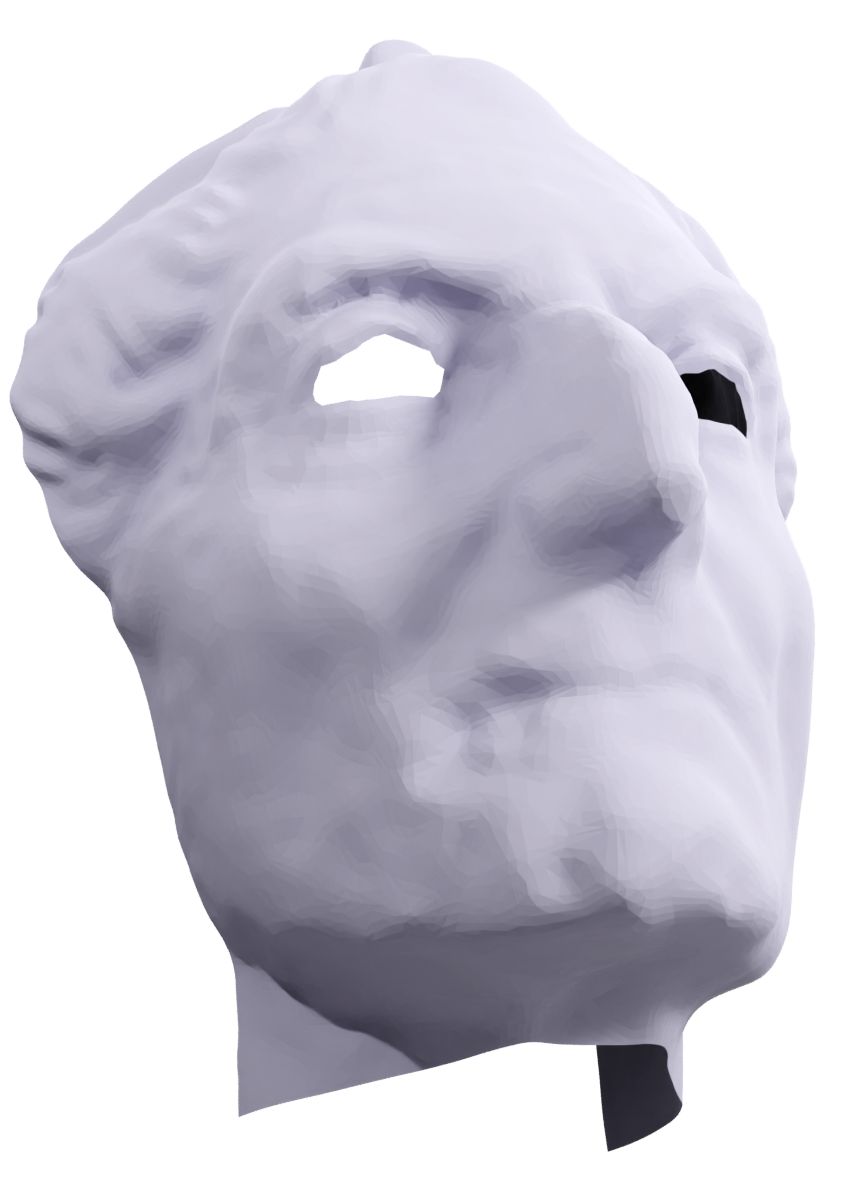} 
\end{subfigure}
 \begin{subfigure}{.11\textwidth}
  \centering
  \includegraphics[width=\linewidth]{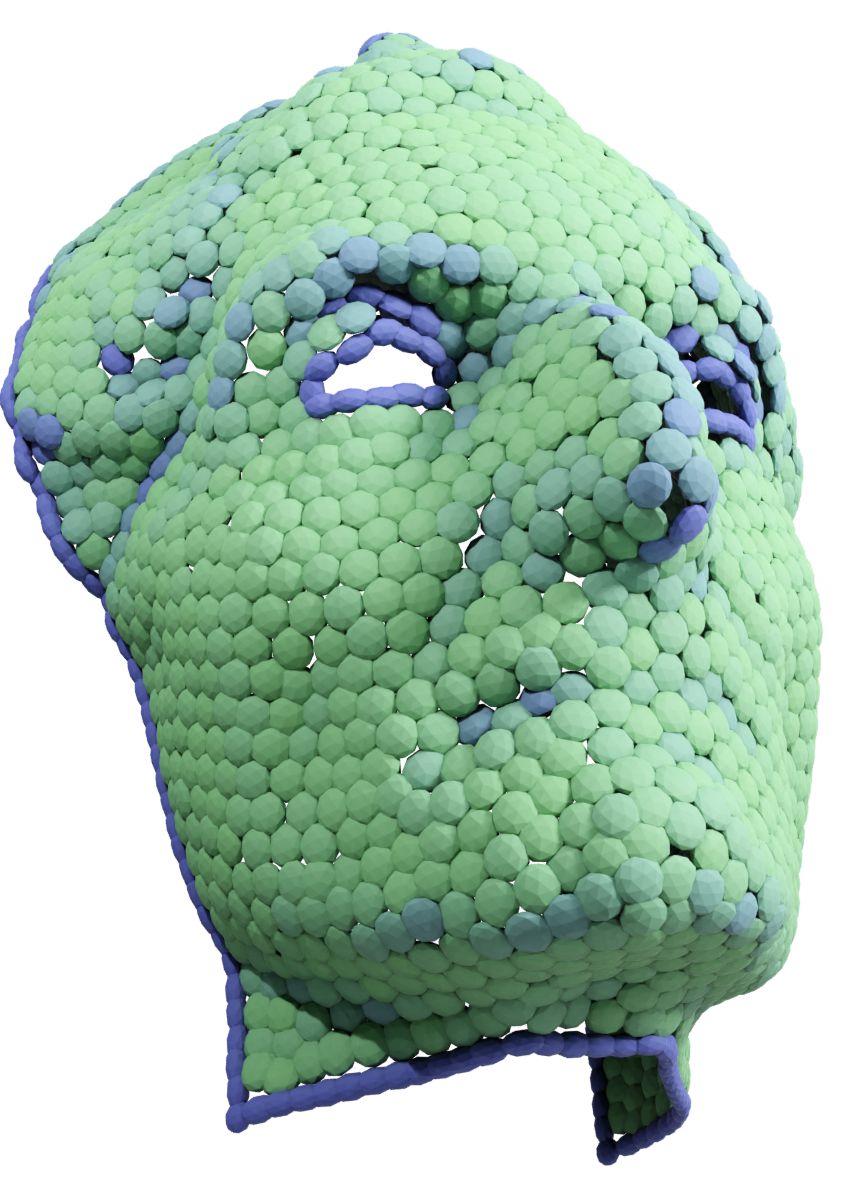} 
\end{subfigure}
 \begin{subfigure}{.11\textwidth}
  \centering
  \includegraphics[width=\linewidth]{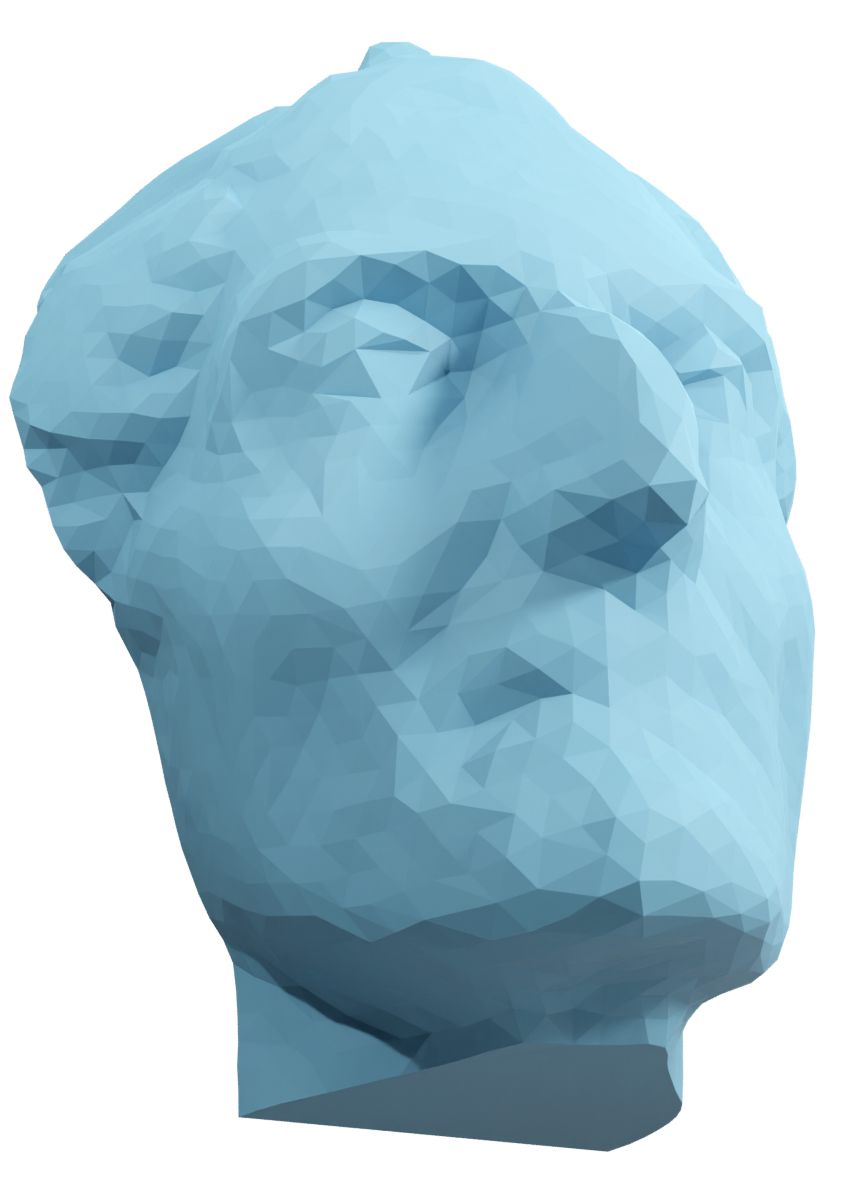} 
\end{subfigure}
 \begin{subfigure}{.11\textwidth}
  \centering
  \includegraphics[width=\linewidth]{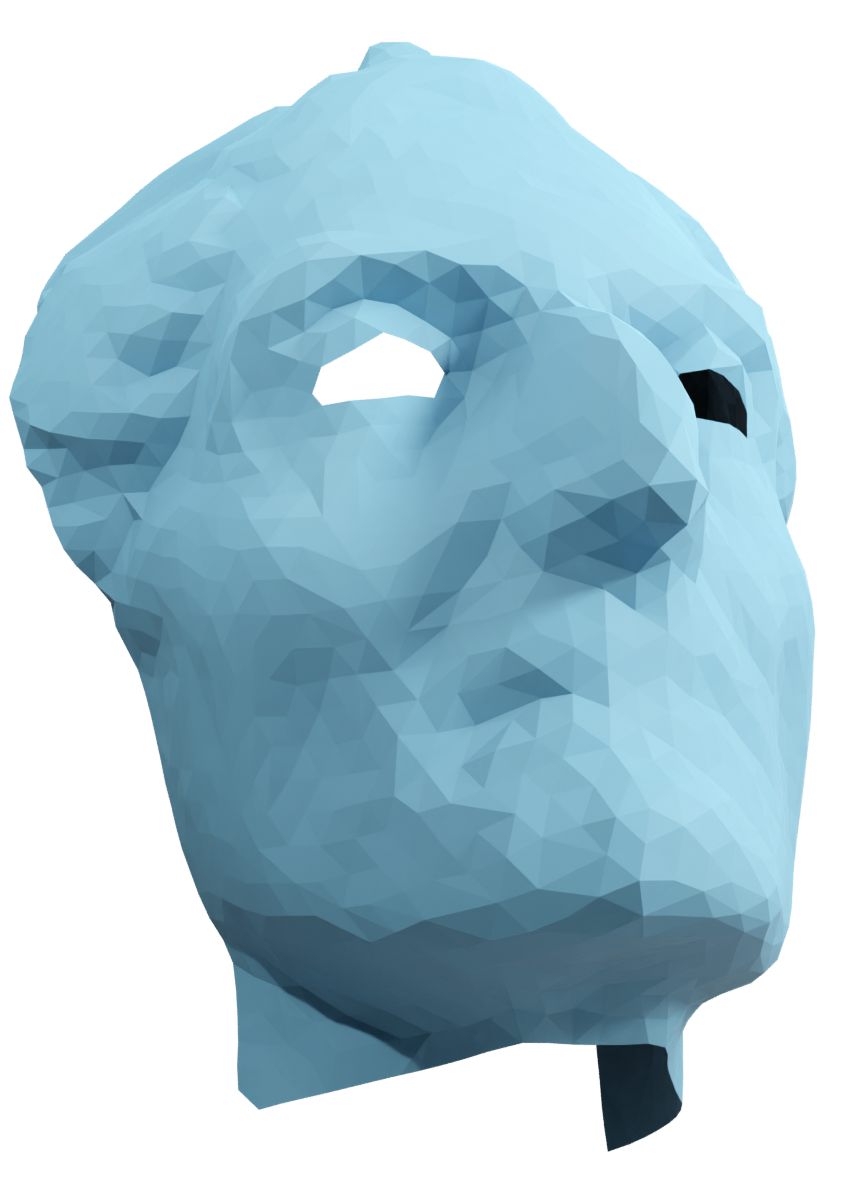} 
\end{subfigure}

 \begin{subfigure}{.11\textwidth}
  \centering
  \includegraphics[width=\linewidth]{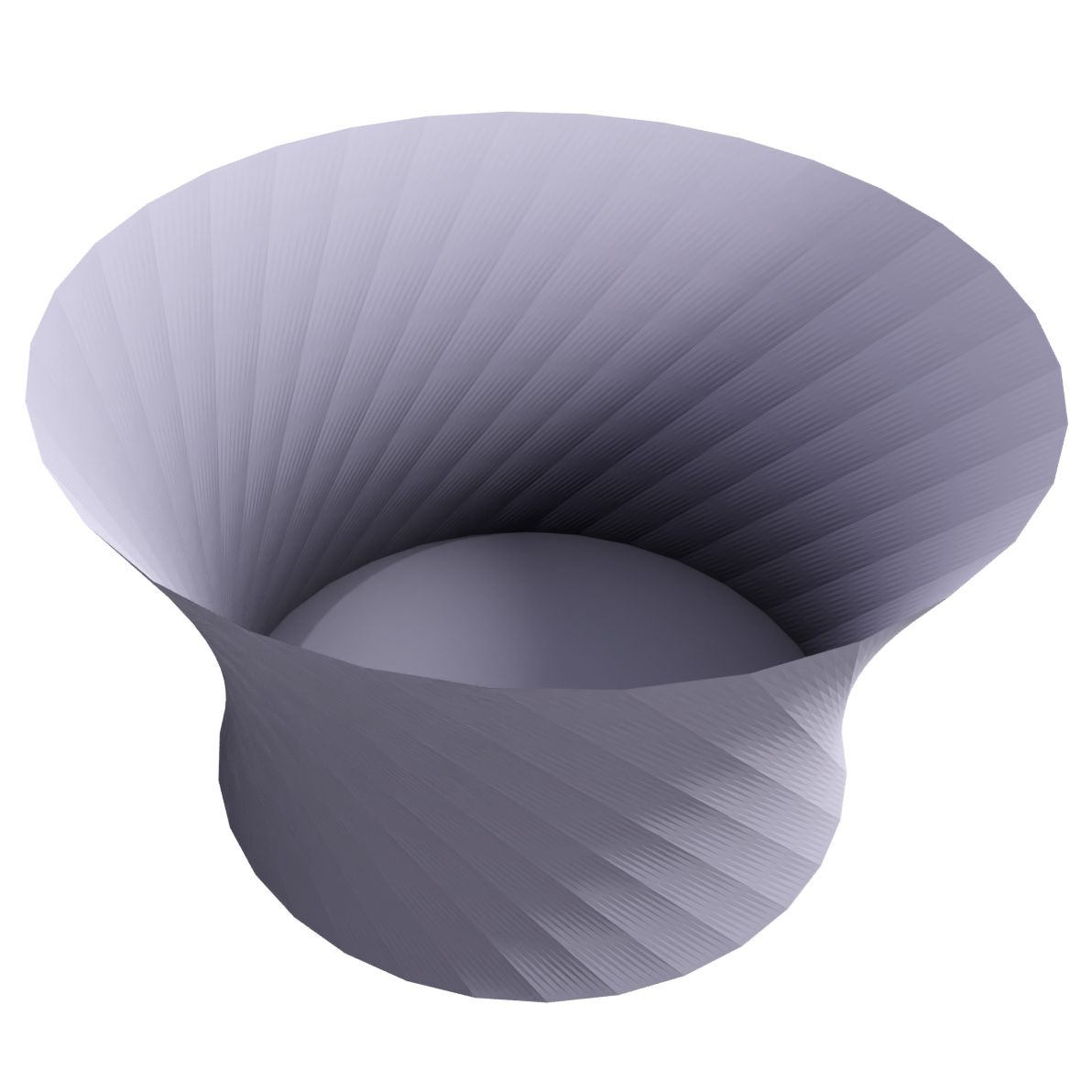} 
  \caption{Gr. Truth}
\end{subfigure}
 \begin{subfigure}{.11\textwidth}
  \centering
  \includegraphics[width=\linewidth]{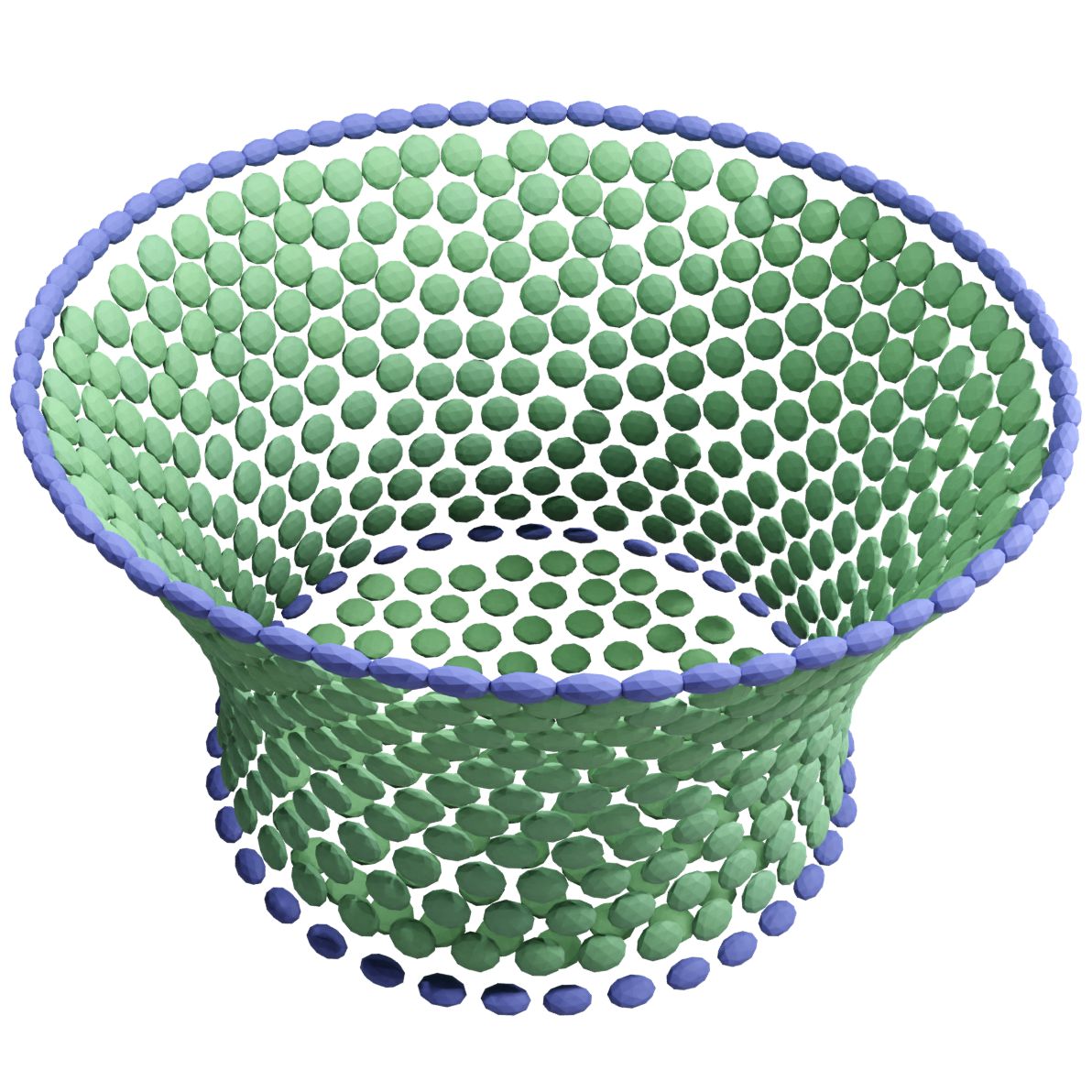} 
  \caption{PoNQ}\label{fig:open_quadric}
\end{subfigure}
 \begin{subfigure}{.11\textwidth}
  \centering
  \includegraphics[width=\linewidth]{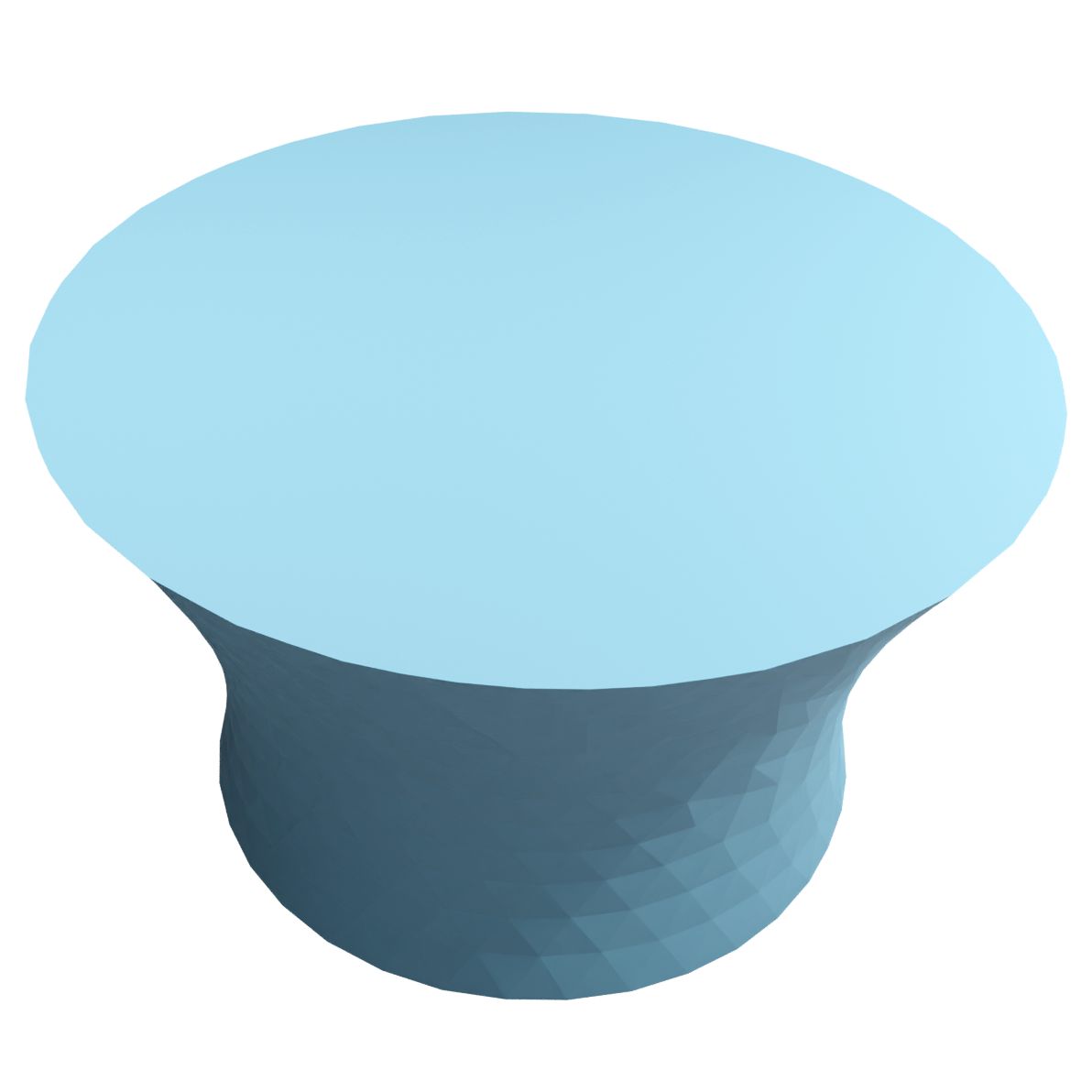} 
  \caption{Closed mesh}\label{fig:open_closed}
\end{subfigure}
 \begin{subfigure}{.11\textwidth}
  \centering
  \includegraphics[width=\linewidth]{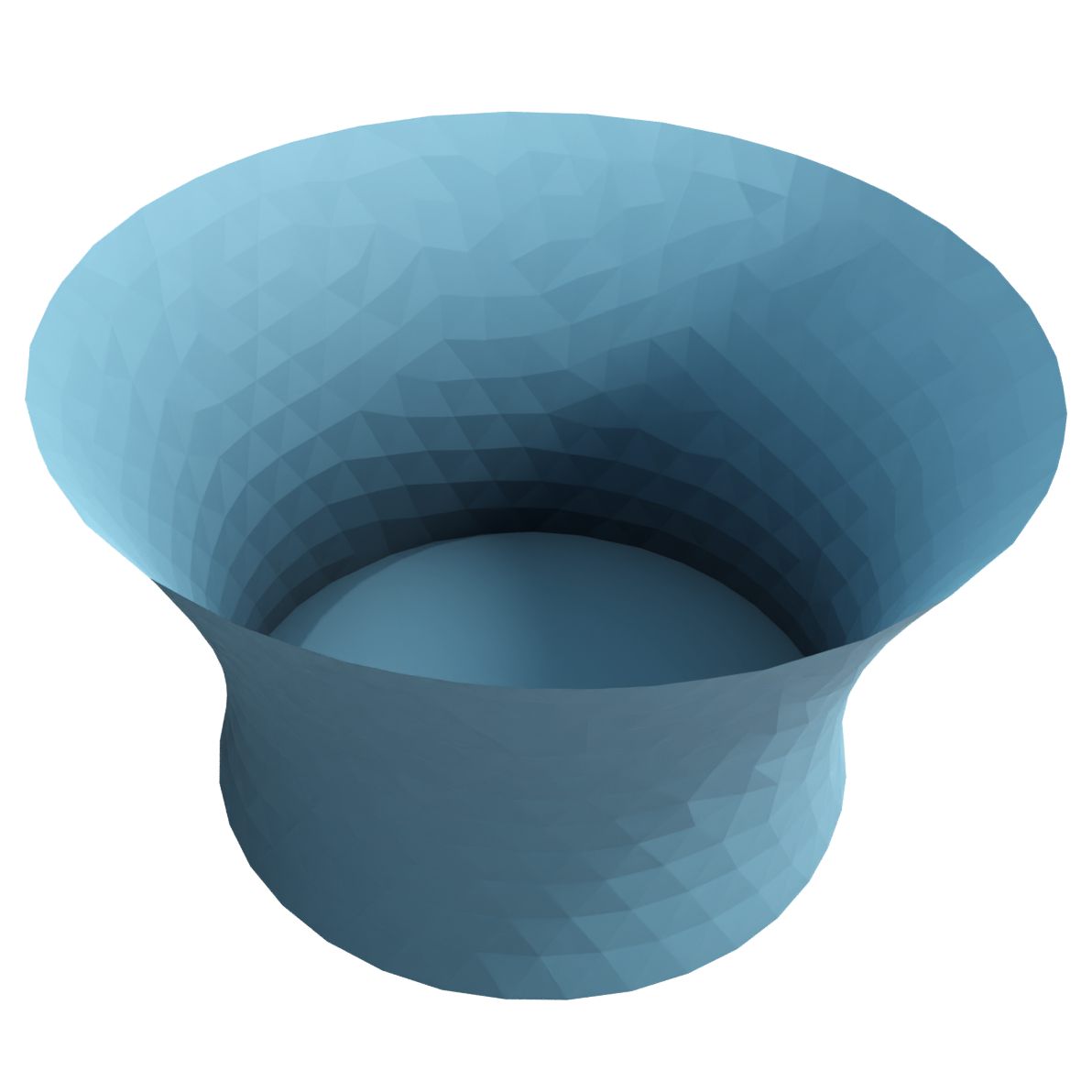} 
  \caption{Open mesh}
\end{subfigure}
\caption{Optimization-based results for PoNQ when open boundaries are used.}
  \label{fig:open}
\end{figure}

We provide more results in Fig.~\ref{fig:open}, this time for open surfaces. Note that the QEM-optimized vertices $\mathbf{v^*}$ snap to sharp and boundary features, allowing for clean reconstructions. As explained in the main paper, we currently transit via a closed mesh (see Fig.~\ref{fig:open_closed}) that is then filtered to create holes between open boundaries, thus limiting our representation of open surfaces. However, since the PoNQ representation provides a precise fit of the target surface (see Fig.~\ref{fig:open_quadric}), we believe a more sophisticated meshing could allow for arbitrary open surfaces.

\section{Edge-based CD: Discussion}

\begin{figure}[h!]
 \centering
 \begin{subfigure}{.48\columnwidth}
  \centering
  \includegraphics[width=\linewidth]{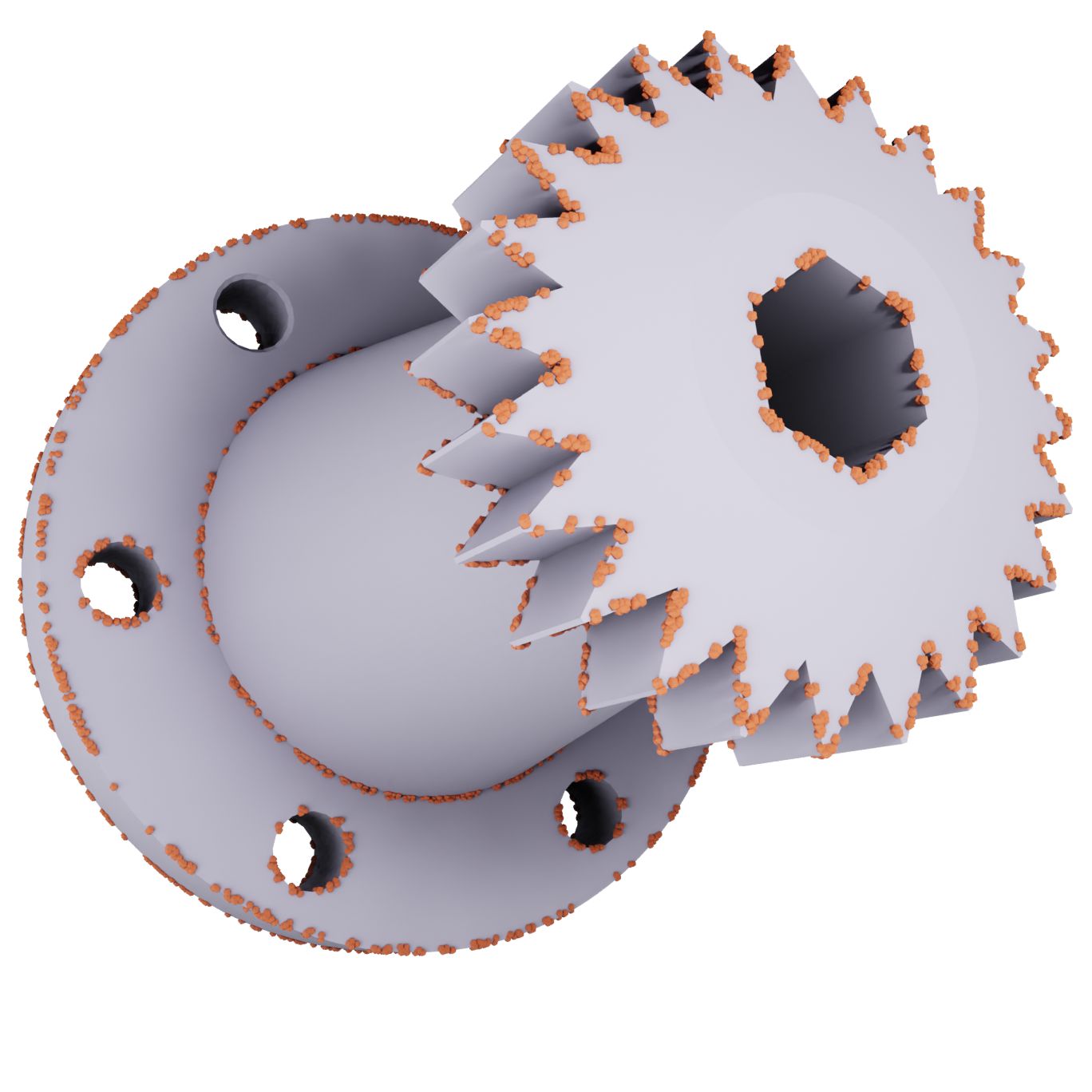}
  \vspace*{-7.5mm}
  \caption{Surface-based ECD}\label{fig:ecd_surface}
\end{subfigure}\hfill
 \begin{subfigure}{.48\columnwidth}
  \centering
  \includegraphics[width=\linewidth]{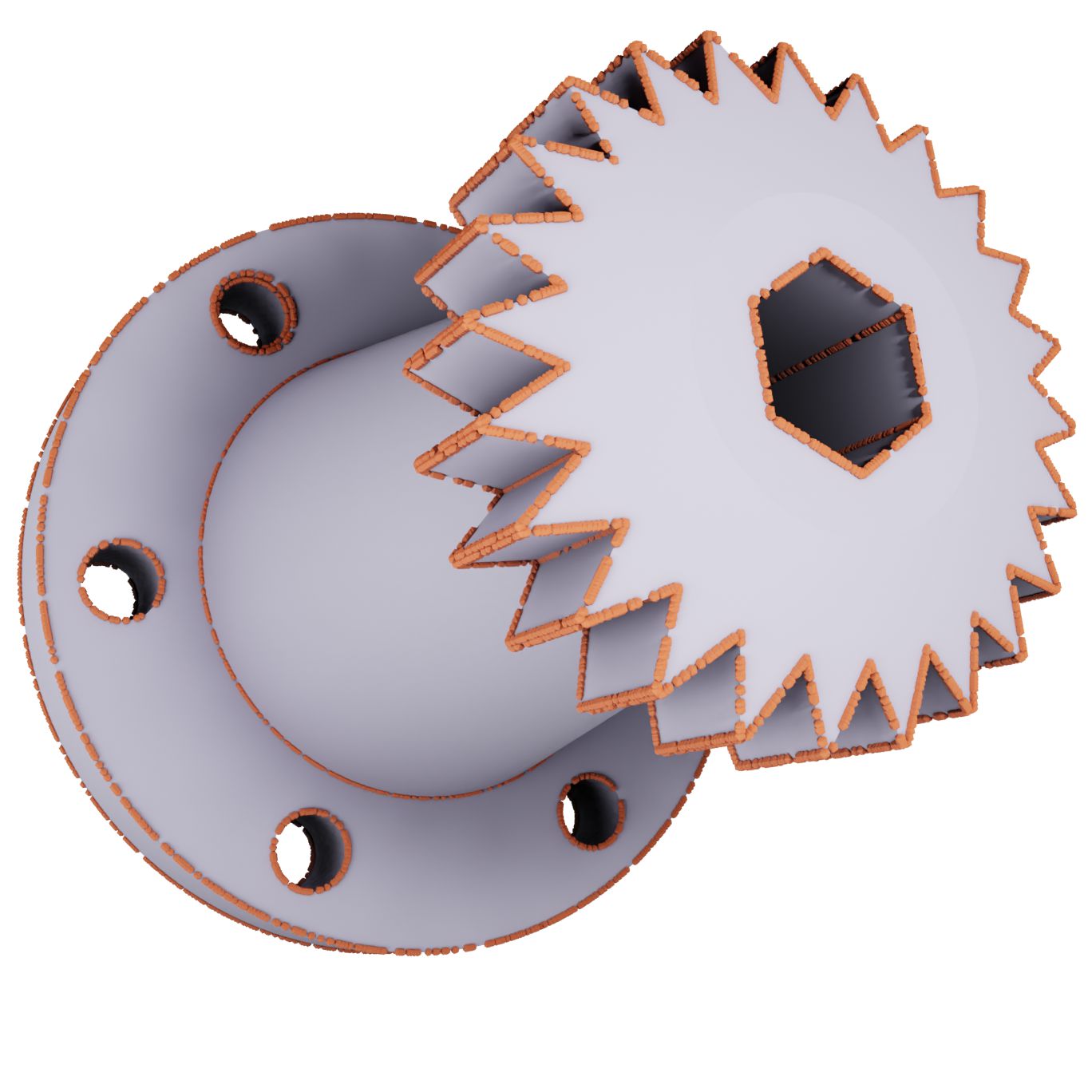} 
    \vspace*{-7.5mm}
    \caption{Edge-based ECD}\label{fig:ecd_edge}
\end{subfigure} \vspace*{-2mm}
\caption{Comparison between two different ECD evaluation methods: sampling surface points and identifying right-angle normal changes between neighbors (left image) is less accurate than sampling sharp edges (right image), and fails to capture sharp edges with angles different than 90 degrees.}
  \label{fig:ecd_omp}
\end{figure}

The Edge Chamfer Distance was introduced in BSP-Net~\cite{chen_bsp-net_2020} to evaluate sharp reconstructions on ShapeNet~\cite{chang_shapenet_2015}. It is based on a sampling of the surface, and only captures sharp edges featuring an angle close to 90 degrees, see Fig.~\ref{fig:ecd_surface}. While this metric might be sufficient for ShapeNet, in which most sharp angles lie on the edges of boxy objects (chairs, televisions...), it is no longer true for ABC~\cite{koch_abc_2019} where sharp edges can have a variety of dihedral angles. As a result, we observed a large variance in this metric evaluation depending on the initial random surface sampling; moreover, it also failed to capture the spurious sharp edges of VoroMesh~\cite{maruani_voromesh_2023} due to faces of small area not being sampled. For these two reasons, and since most meshes of the ABC dataset have clean and well-defined sharp edges, we found that sampling the sharp edges (defined as those with a dihedral angle larger than $\frac{\pi}{6}$) directly provided a more robust and faithful metric (see cog teeth in Fig.~\ref{fig:ecd_edge}). 

\subsection{Additional timings}

VoroMesh~\cite{maruani_voromesh_2023} provides two implementations for their mesh extraction, based on SciPy and CGAL. Since our mesh extraction is also based on SciPy, we use their SciPy one out of fairness in Tab.~\ref{tab:timings} to evaluate and compare timings; we also added the timings of the meshing phase of NMC~\cite{chen_neural_2021} and NDC~\cite{chen_neural_2022} for comparison purposes. While our code is not yet optimized, an implementation of PoNQ in CGAL would undoubtedly allow for an even faster extraction. 

\begin{table}[h!]
  \begin{center}
    \resizebox{.91\linewidth}{!}{\begin{tabular}{l c c c}
  \hline
  \textbf{Method}            & Grid & Optimization (s) & Meshing (s) \\                   
  \hline
  NDC~\cite{chen_neural_2022}  & $32^3$ & - & 0.001 \\
  NMC~\cite{chen_neural_2021}  & $32^3$ & - & 0.1 \\
  VoroMesh~\cite{maruani_voromesh_2023}  & $32^3$ & 2.0 & 0.3 \\
  PoNQ & $32^3$    & 2.6 & 0.3 \\
  \hline
    NDC~\cite{chen_neural_2022}  & $64^3$ & - & 0.02 \\
    NMC~\cite{chen_neural_2021}  & $64^3$ & - & 0.9 \\
  VoroMesh~\cite{maruani_voromesh_2023}  & $64^3$ & 4.2 & 1.1 \\
   PoNQ & $64^3$    & 4.1 & 1.3 \\
   \hline
   NDC~\cite{chen_neural_2022}  & $128^3$ & - & 0.2 \\
   NMC~\cite{chen_neural_2021}  & $128^3$ & - & 7.3 \\
    VoroMesh~\cite{maruani_voromesh_2023}  & $128^3$ & 36.2 & 4.8 \\
    PoNQ & $128^3$    & 17.9 & 7.8 \\
     \hline
\end{tabular}}
    \vspace*{-2mm}
    \caption{Timings for optimization-based experiments.}
    \label{tab:timings}
  \end{center}
\end{table}

\section{Additional Renders}

\begin{figure*}[t]
\centering
 \begin{subfigure}{.13\textwidth}
  \centering
  \includegraphics[width=\linewidth]{figs/optim/sap_64764.jpeg} 
\end{subfigure}
 \begin{subfigure}{.13\textwidth}
  \centering
  \includegraphics[width=\linewidth]{figs/optim/dpf_64764.jpeg} 
\end{subfigure}
 \begin{subfigure}{.13\textwidth}
  \centering
  \includegraphics[width=\linewidth]{figs/optim/voro_64764.jpeg} 
\end{subfigure}
 \begin{subfigure}{.13\textwidth}
  \centering
  \includegraphics[width=\linewidth]{figs/optim/quadric_64764.jpeg} 
\end{subfigure}
 \begin{subfigure}{.13\textwidth}
  \centering
  \includegraphics[width=\linewidth]{figs/groundtruth/64764gt.jpeg} 
\end{subfigure}

 \begin{subfigure}{.13\textwidth}
  \centering
  \includegraphics[width=\linewidth]{figs/optim/sap_64764_normal.jpeg} 
\end{subfigure}
 \begin{subfigure}{.13\textwidth}
  \centering
  \includegraphics[width=\linewidth]{figs/optim/dpf_64764_normal.jpeg} 
\end{subfigure}
 \begin{subfigure}{.13\textwidth}
  \centering
  \includegraphics[width=\linewidth]{figs/optim/voro_64764_normal.jpeg} 
\end{subfigure}
 \begin{subfigure}{.13\textwidth}
  \centering
  \includegraphics[width=\linewidth]{figs/optim/quadric_64764_normal.jpeg} 
\end{subfigure}
 \begin{subfigure}{.13\textwidth}
  \centering
  \includegraphics[width=\linewidth]{figs/groundtruth/64764_normal.jpeg} 
\end{subfigure}

\begin{subfigure}{.13\textwidth}
  \centering
  \includegraphics[width=\linewidth]{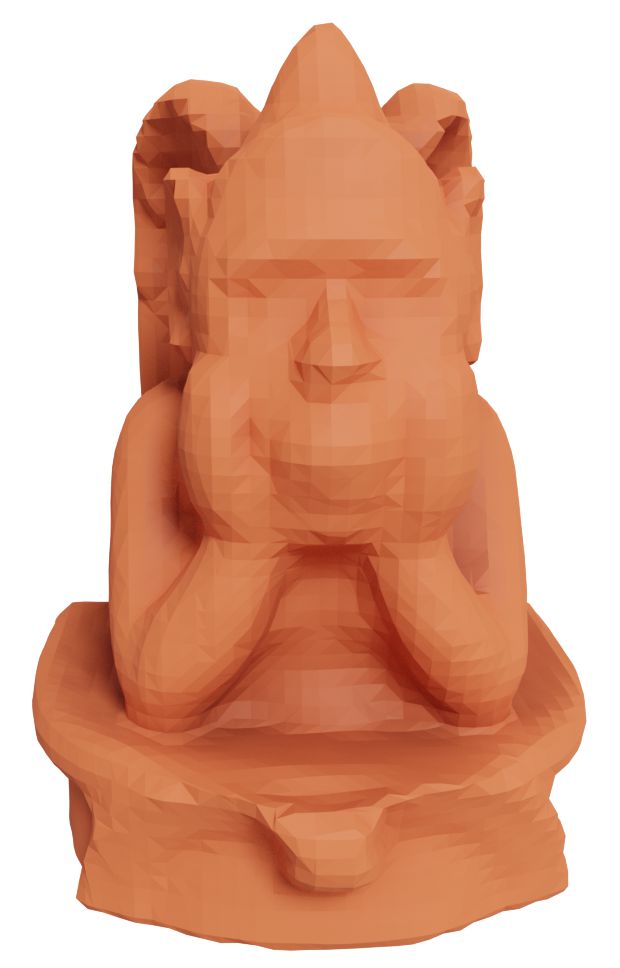} 
\end{subfigure}
 \begin{subfigure}{.13\textwidth}
  \centering
  \includegraphics[width=\linewidth]{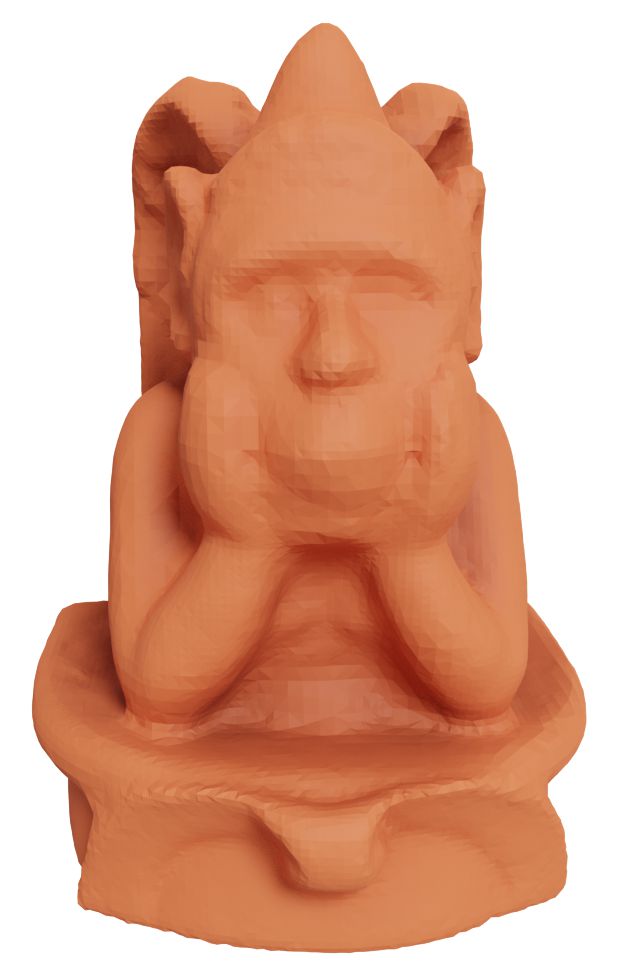} 
\end{subfigure}
 \begin{subfigure}{.13\textwidth}
  \centering
  \includegraphics[width=\linewidth]{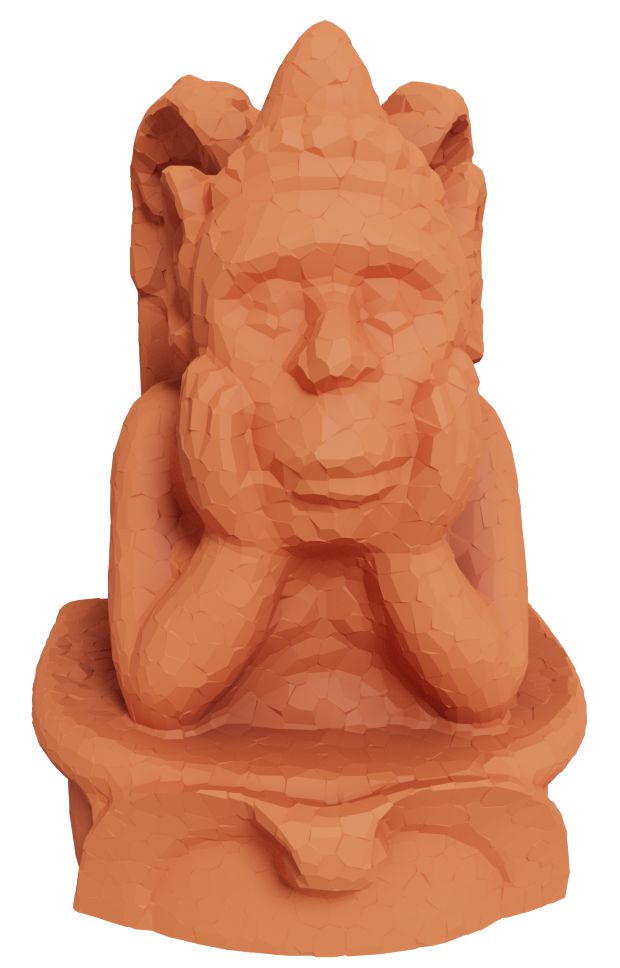} 
\end{subfigure}
 \begin{subfigure}{.13\textwidth}
  \centering
  \includegraphics[width=\linewidth]{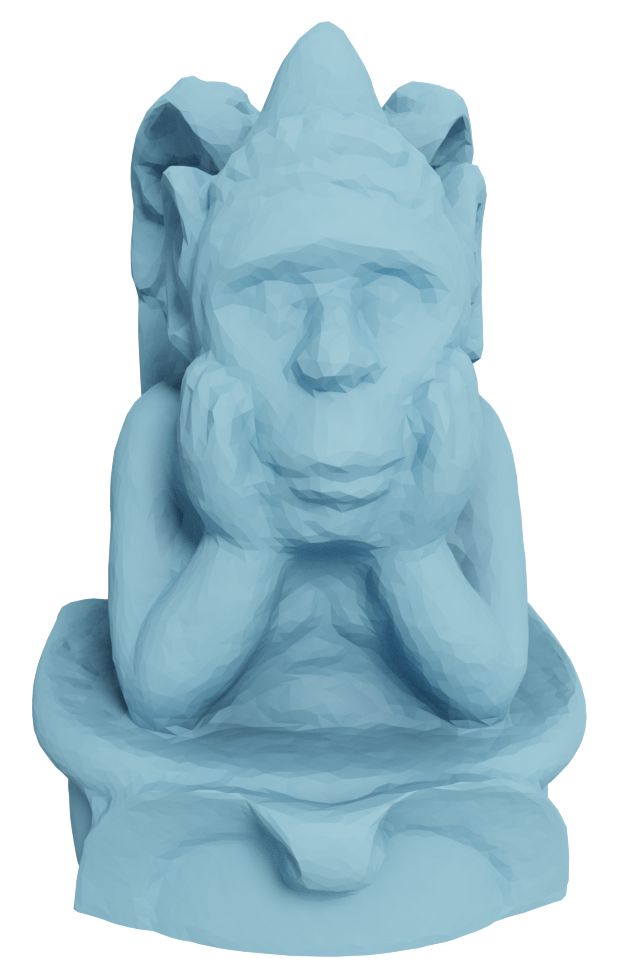} 
\end{subfigure}
 \begin{subfigure}{.13\textwidth}
  \centering
  \includegraphics[width=\linewidth]{figs/groundtruth/64764gt.jpeg} 
\end{subfigure}

 \begin{subfigure}{.13\textwidth}
  \centering
  \includegraphics[width=\linewidth]{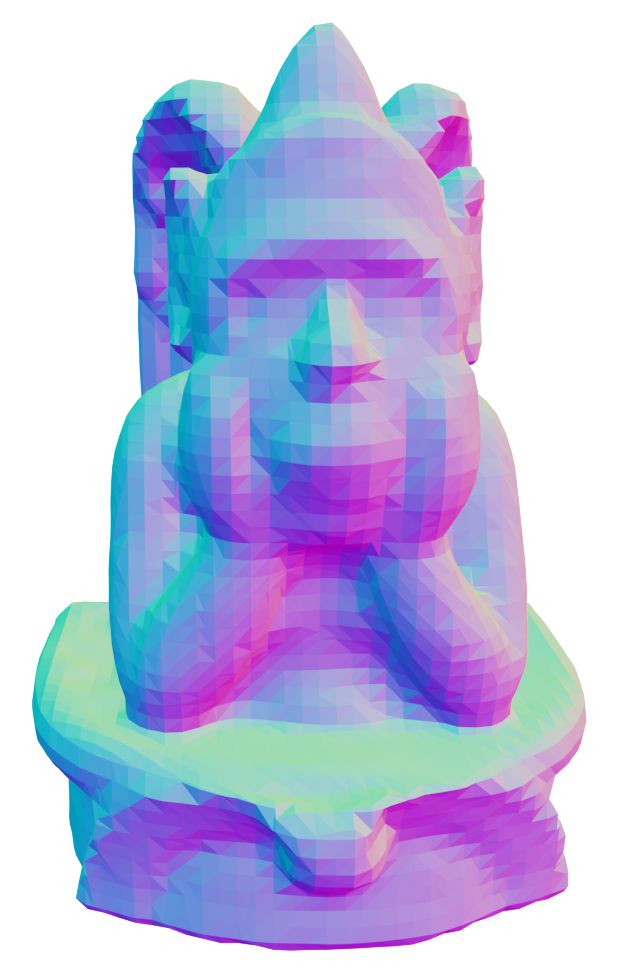} 
\end{subfigure}
 \begin{subfigure}{.13\textwidth}
  \centering
  \includegraphics[width=\linewidth]{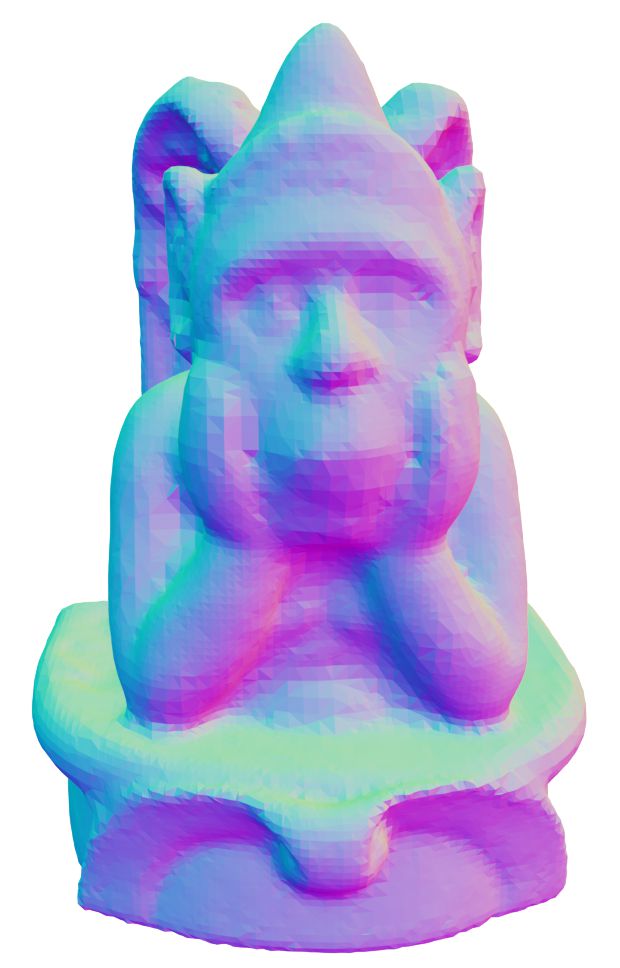} 
\end{subfigure}
 \begin{subfigure}{.13\textwidth}
  \centering
  \includegraphics[width=\linewidth]{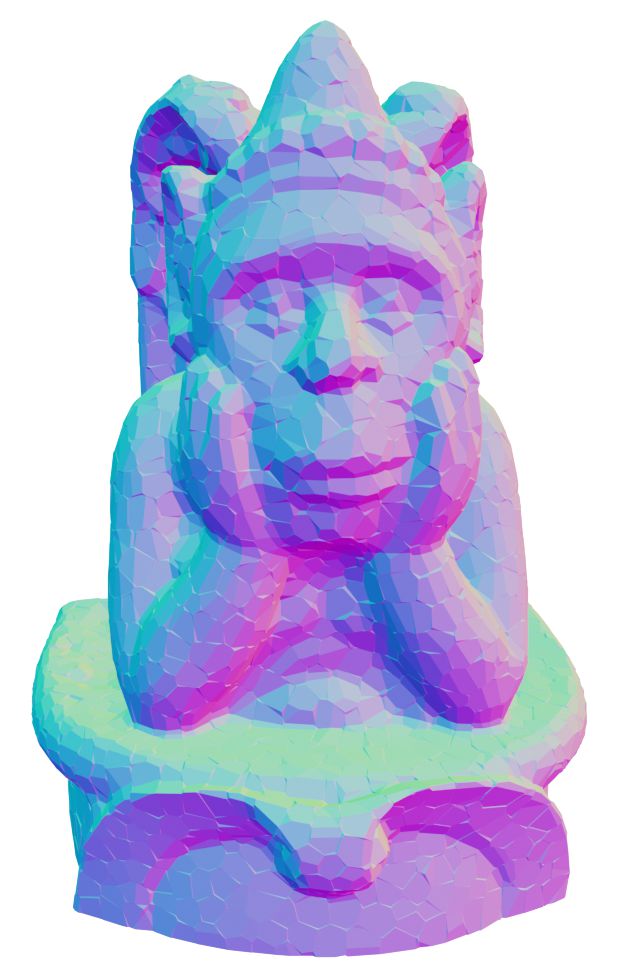} 
\end{subfigure}
 \begin{subfigure}{.13\textwidth}
  \centering
  \includegraphics[width=\linewidth]{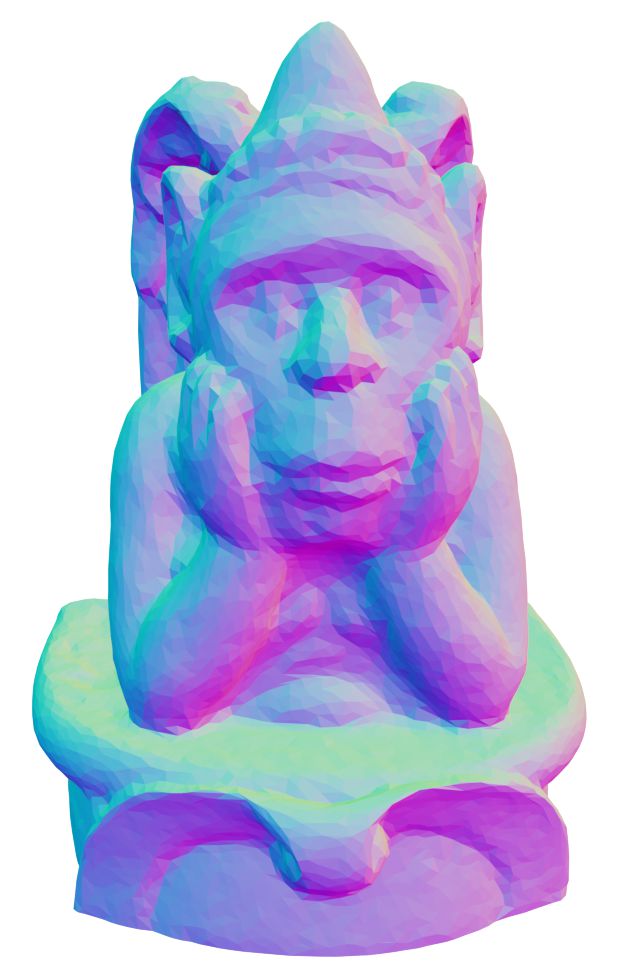} 
\end{subfigure}
 \begin{subfigure}{.13\textwidth}
  \centering
  \includegraphics[width=\linewidth]{figs/groundtruth/64764_normal.jpeg} 
\end{subfigure}

\begin{subfigure}{.13\textwidth}
  \centering
  \includegraphics[width=\linewidth]{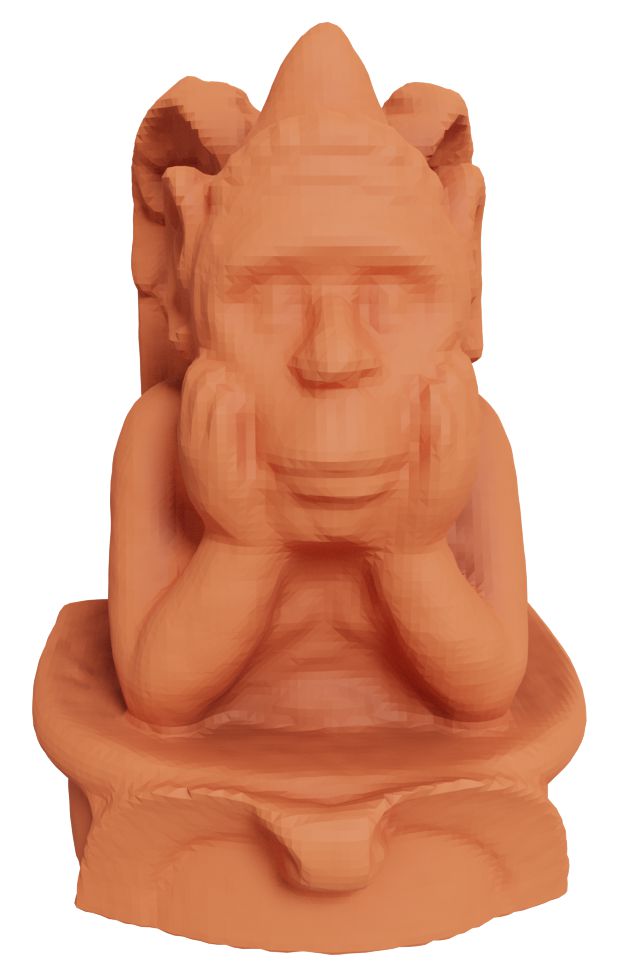} 
\end{subfigure}
 \begin{subfigure}{.13\textwidth}
  \centering
  \includegraphics[width=\linewidth]{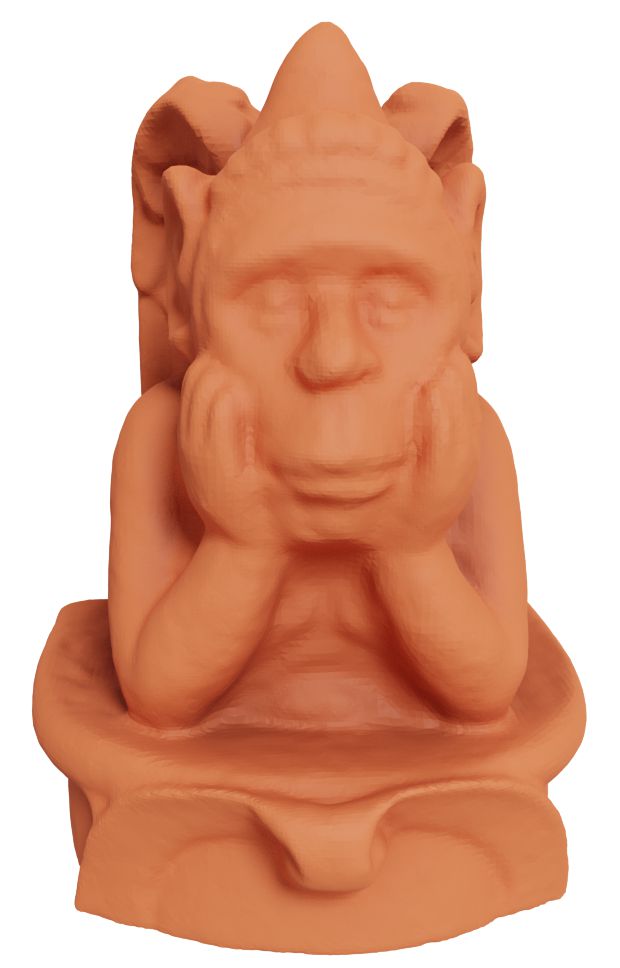} 
\end{subfigure}
 \begin{subfigure}{.13\textwidth}
  \centering
  \includegraphics[width=\linewidth]{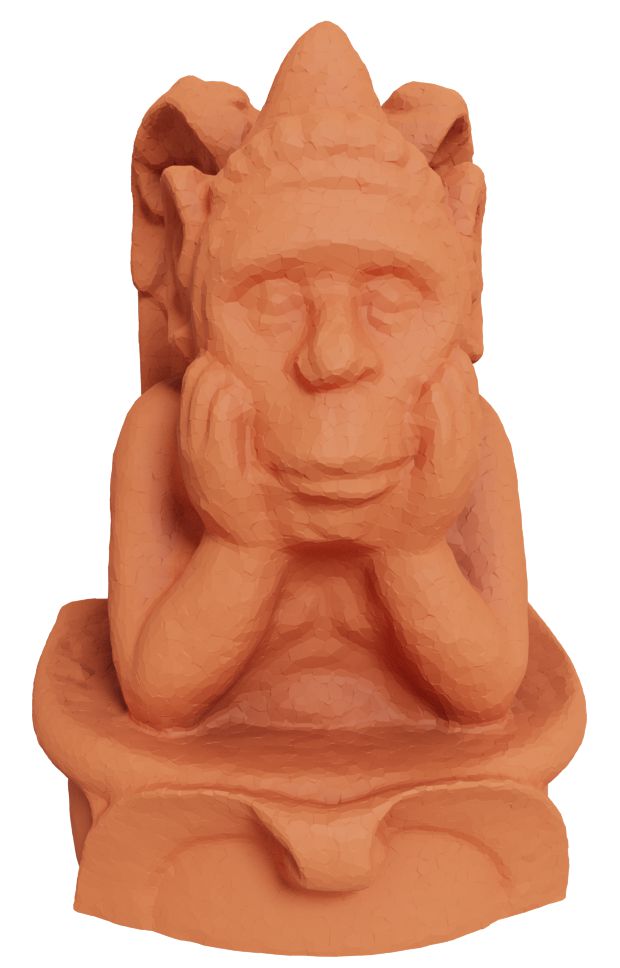} 
\end{subfigure}
 \begin{subfigure}{.13\textwidth}
  \centering
  \includegraphics[width=\linewidth]{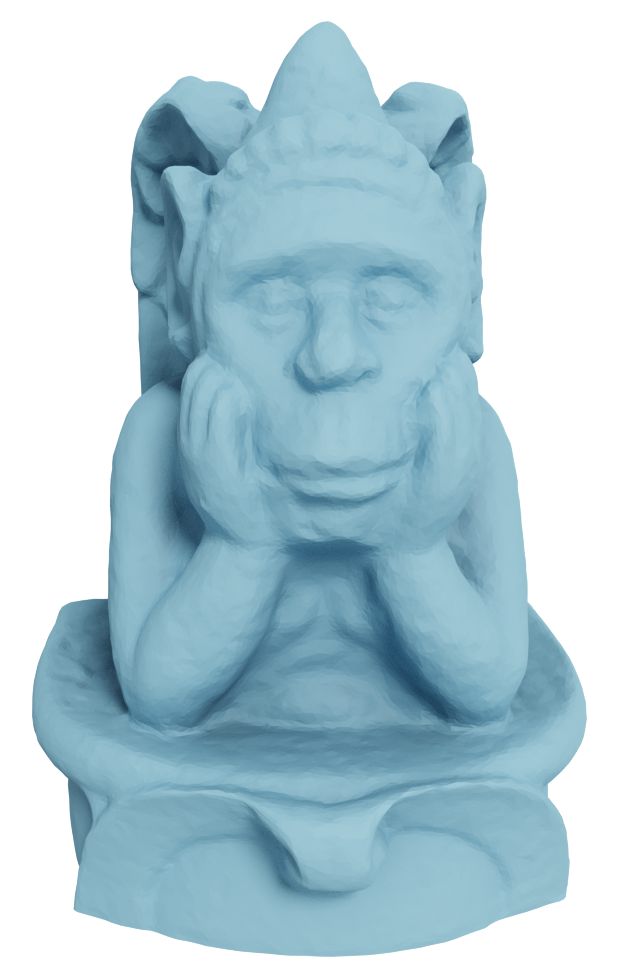} 
\end{subfigure}
 \begin{subfigure}{.13\textwidth}
  \centering
  \includegraphics[width=\linewidth]{figs/groundtruth/64764gt.jpeg} 
\end{subfigure}

 \begin{subfigure}{.13\textwidth}
  \centering
  \includegraphics[width=\linewidth]{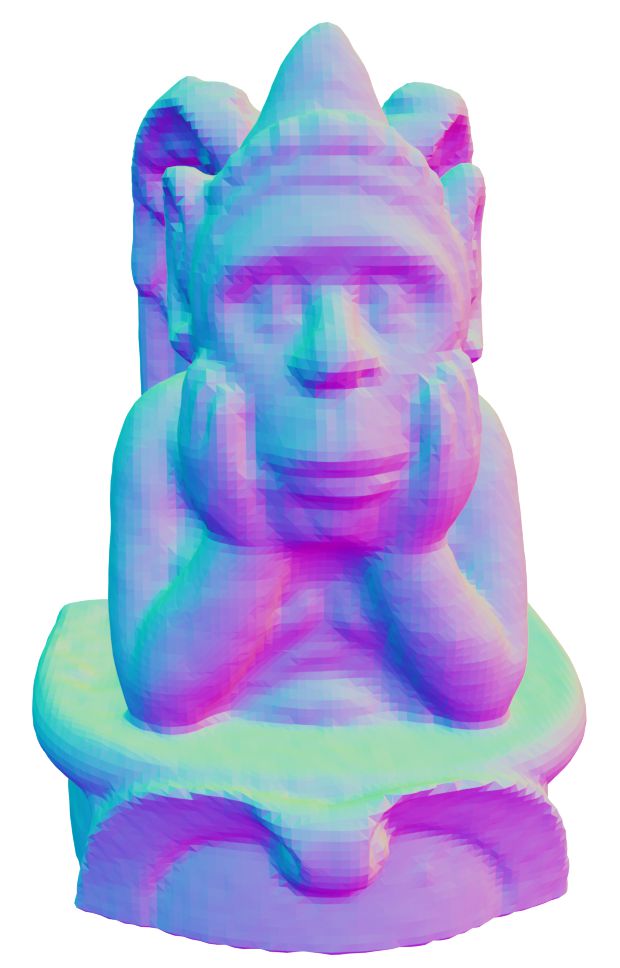} 
  \caption{SAP}
\end{subfigure}
 \begin{subfigure}{.13\textwidth}
  \centering
  \includegraphics[width=\linewidth]{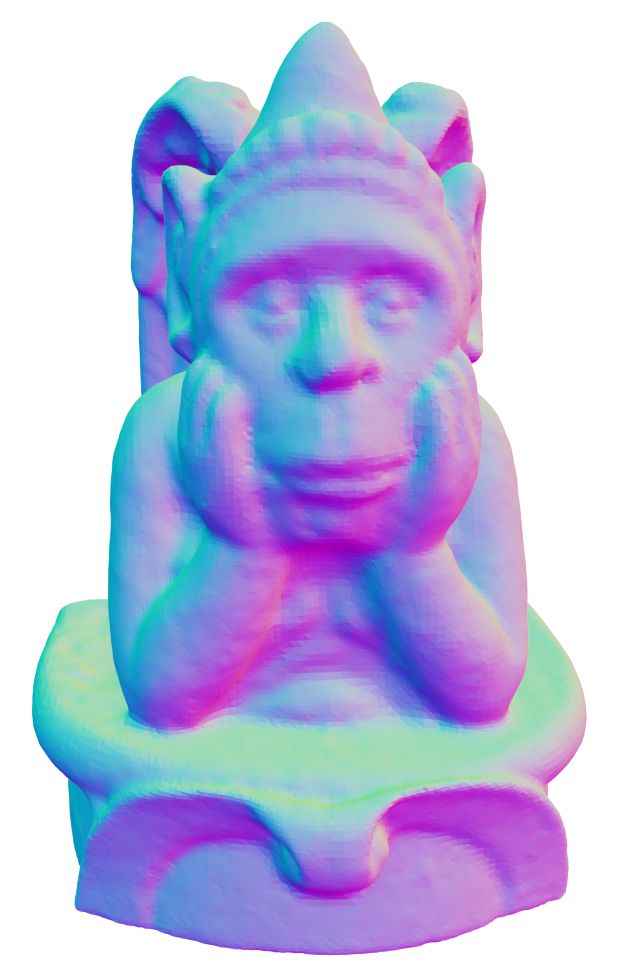} 
  \caption{DPF}
\end{subfigure}
 \begin{subfigure}{.13\textwidth}
  \centering
  \includegraphics[width=\linewidth]{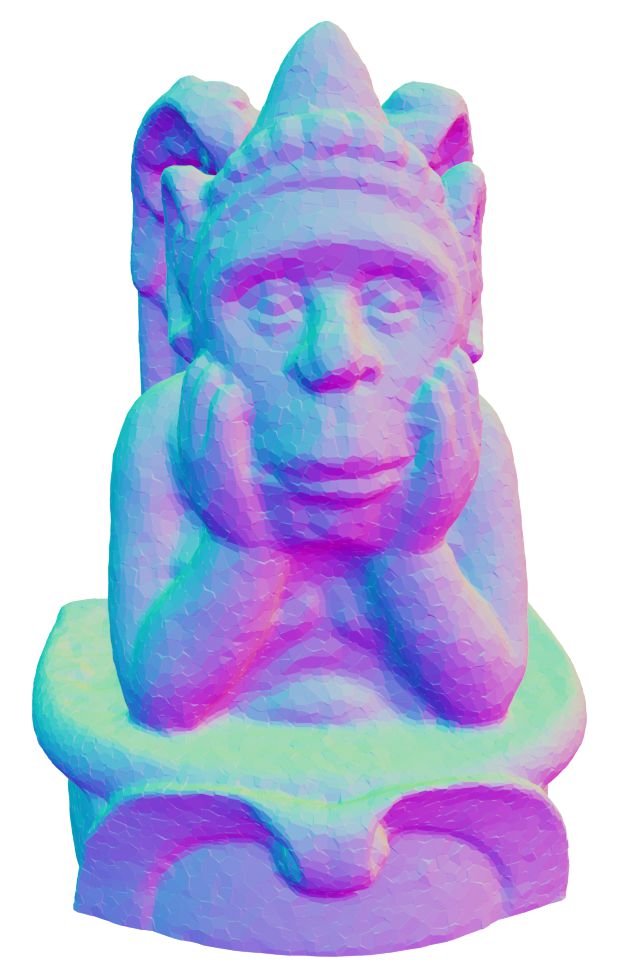} 
  \caption{VoroMesh}
\end{subfigure}
 \begin{subfigure}{.13\textwidth}
  \centering
  \includegraphics[width=\linewidth]{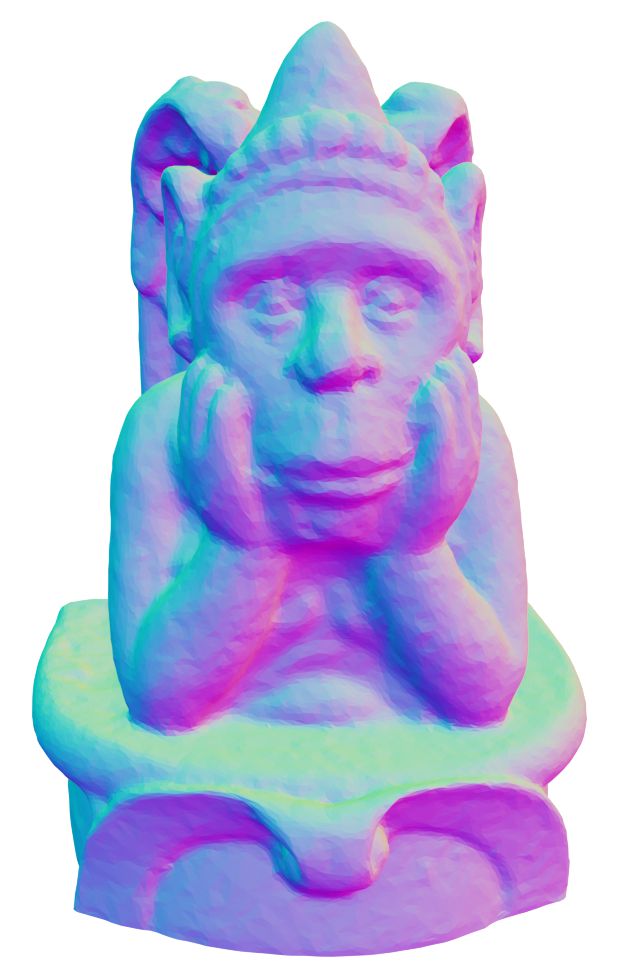} 
  \caption{PoNQ}
\end{subfigure}
 \begin{subfigure}{.13\textwidth}
  \centering
  \includegraphics[width=\linewidth]{figs/groundtruth/64764_normal.jpeg} 
  \caption{Gr. Truth}
\end{subfigure}

\caption{Optimization-based results (top to bottom: $32^3$, $64^3$, $128^3$) on Thingi30.
}
  \label{fig:optim_supl}
\end{figure*}

\begin{figure*}[t]
\centering
 \begin{subfigure}{.18\textwidth}
  \centering
  \includegraphics[width=\linewidth]{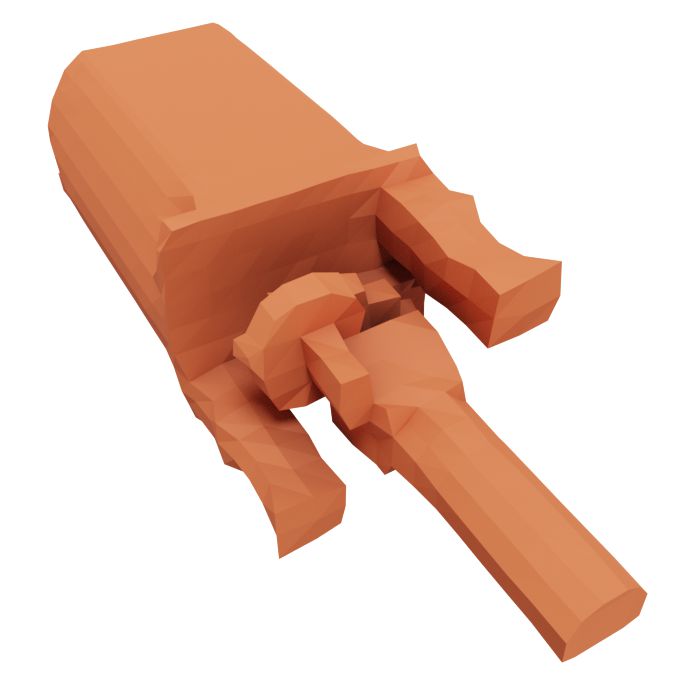} 
\end{subfigure}
 \begin{subfigure}{.18\textwidth}
  \centering
  \includegraphics[width=\linewidth]{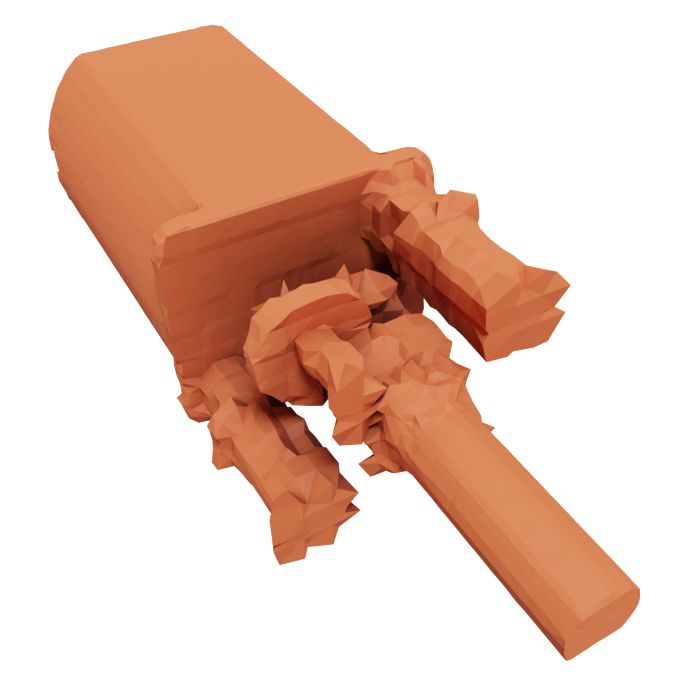} 
\end{subfigure}
 \begin{subfigure}{.18\textwidth}
  \centering
  \includegraphics[width=\linewidth]{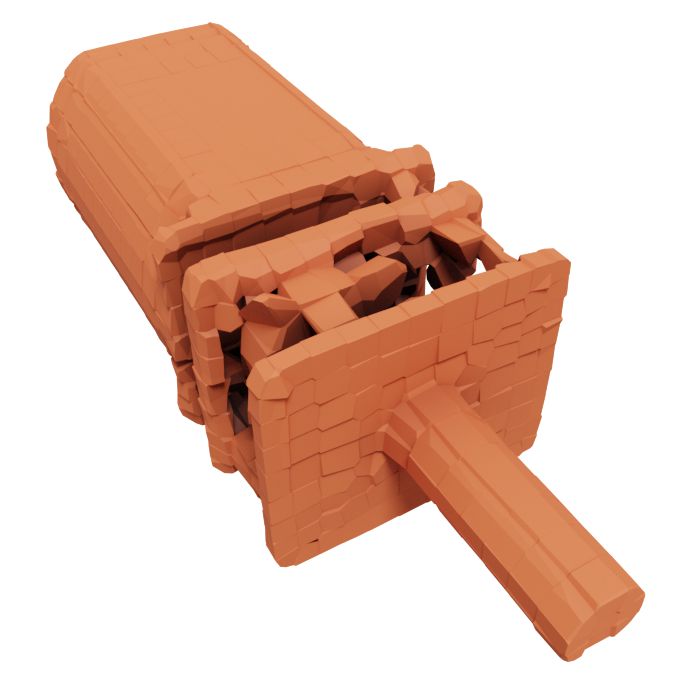} 
\end{subfigure}
 \begin{subfigure}{.18\textwidth}
  \centering
  \includegraphics[width=\linewidth]{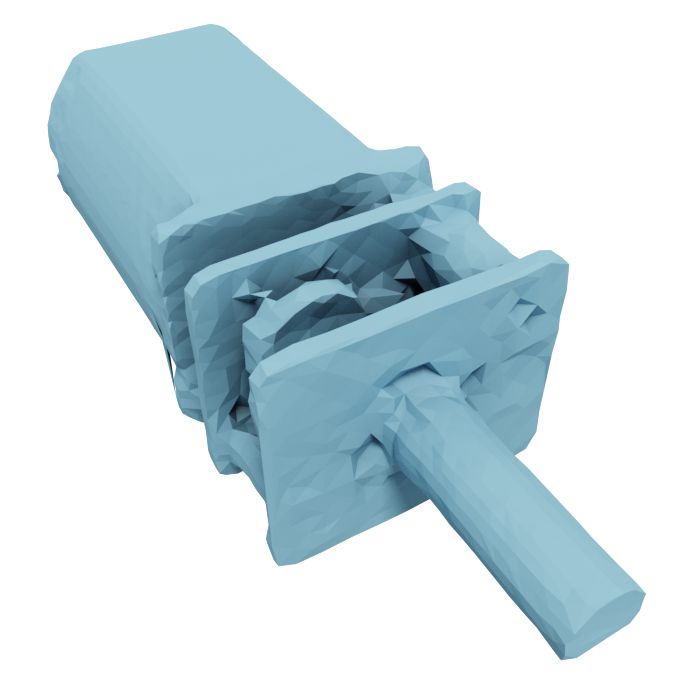} 
\end{subfigure}
 \begin{subfigure}{.18\textwidth}
  \centering
  \includegraphics[width=\linewidth]{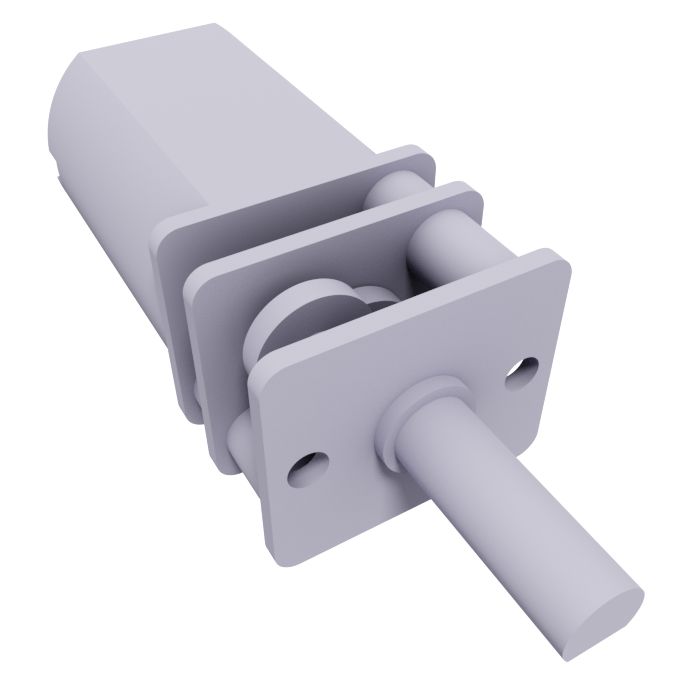} 
\end{subfigure}
 \begin{subfigure}{.18\textwidth}
  \centering
  \includegraphics[width=\linewidth]{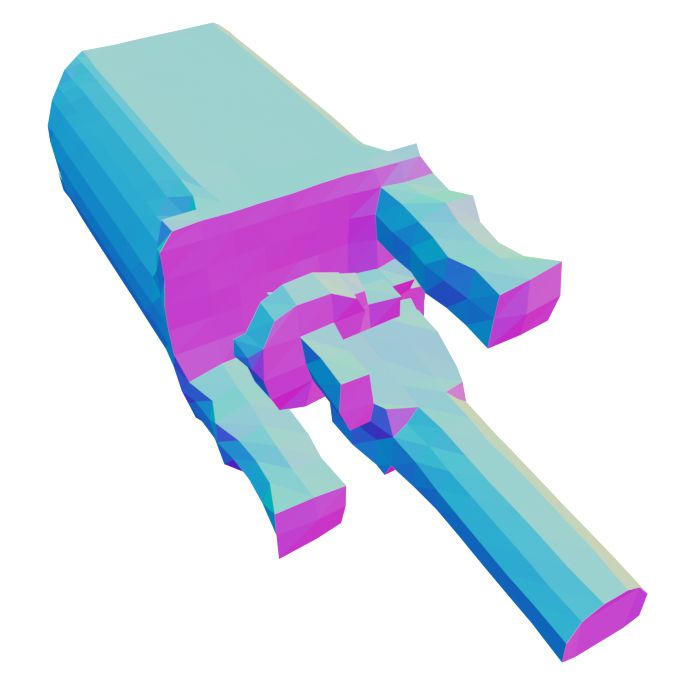} 
\end{subfigure}
 \begin{subfigure}{.18\textwidth}
  \centering
  \includegraphics[width=\linewidth]{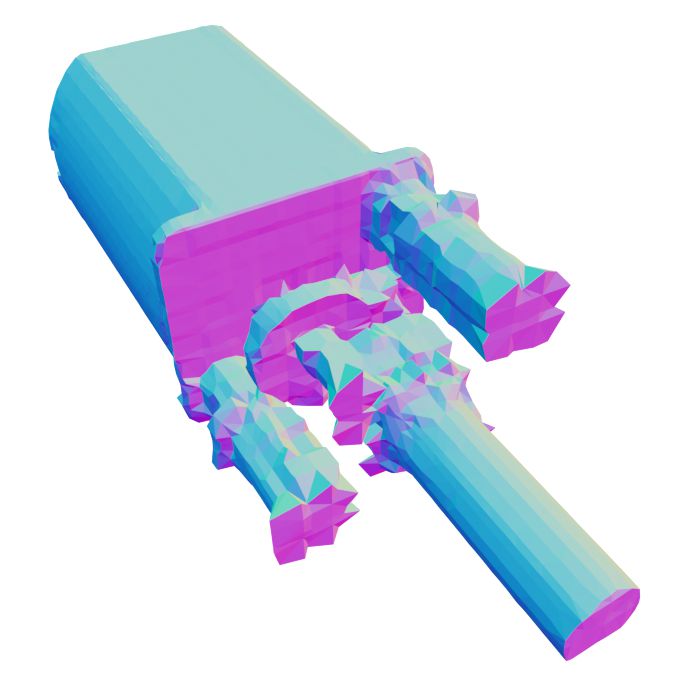} 
\end{subfigure}
 \begin{subfigure}{.18\textwidth}
  \centering
  \includegraphics[width=\linewidth]{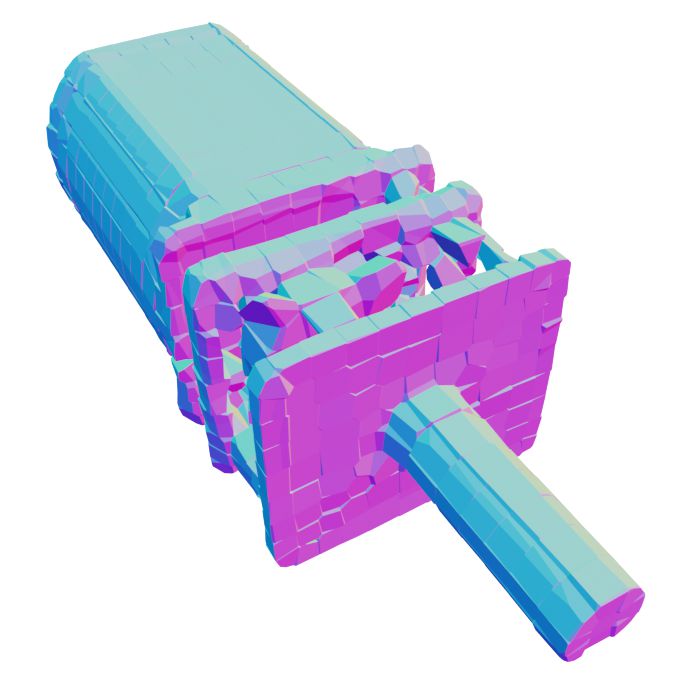} 
\end{subfigure}
 \begin{subfigure}{.18\textwidth}
  \centering
  \includegraphics[width=\linewidth]{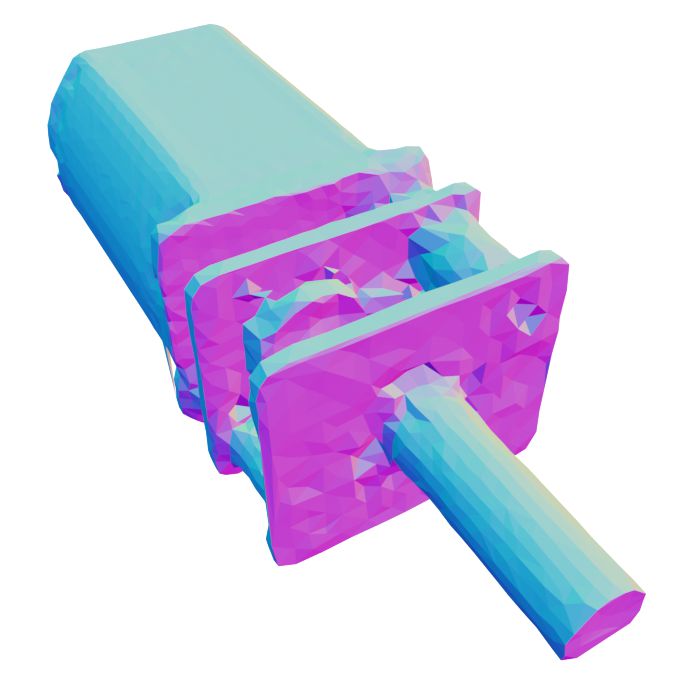} 
\end{subfigure}
 \begin{subfigure}{.18\textwidth}
  \centering
  \includegraphics[width=\linewidth]{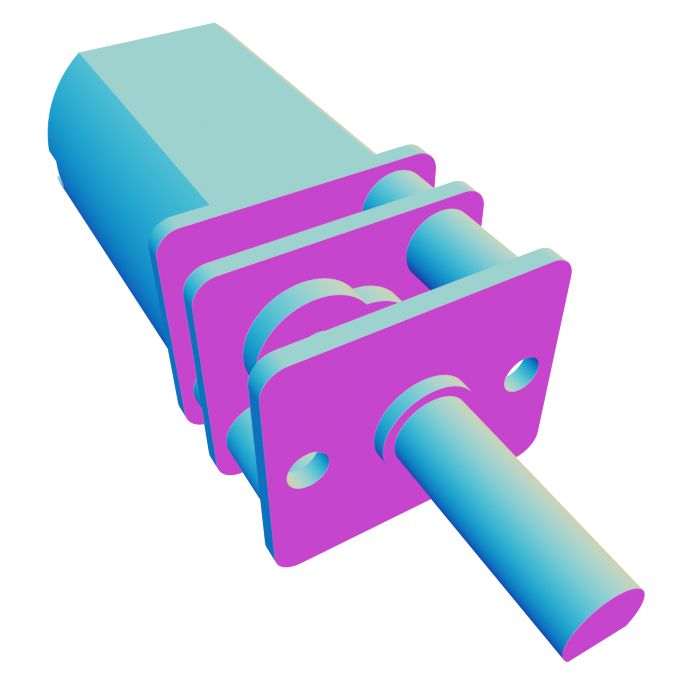} 
\end{subfigure}
\begin{subfigure}{.18\textwidth}
  \centering
  \includegraphics[width=\linewidth]{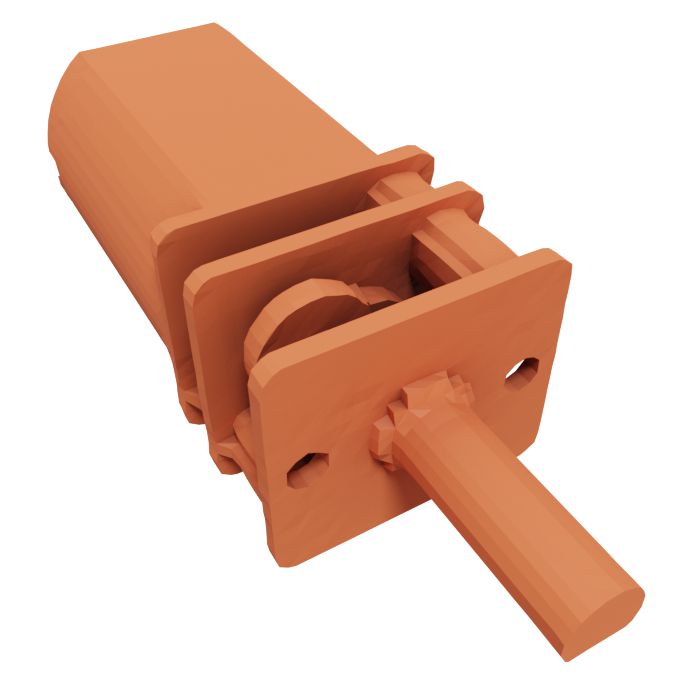} 
\end{subfigure}
 \begin{subfigure}{.18\textwidth}
  \centering
  \includegraphics[width=\linewidth]{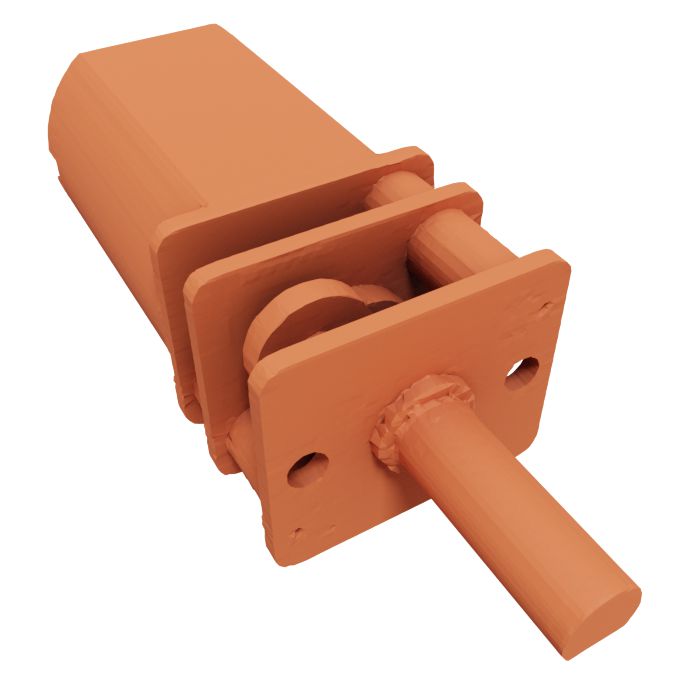} 
\end{subfigure}
 \begin{subfigure}{.18\textwidth}
  \centering
  \includegraphics[width=\linewidth]{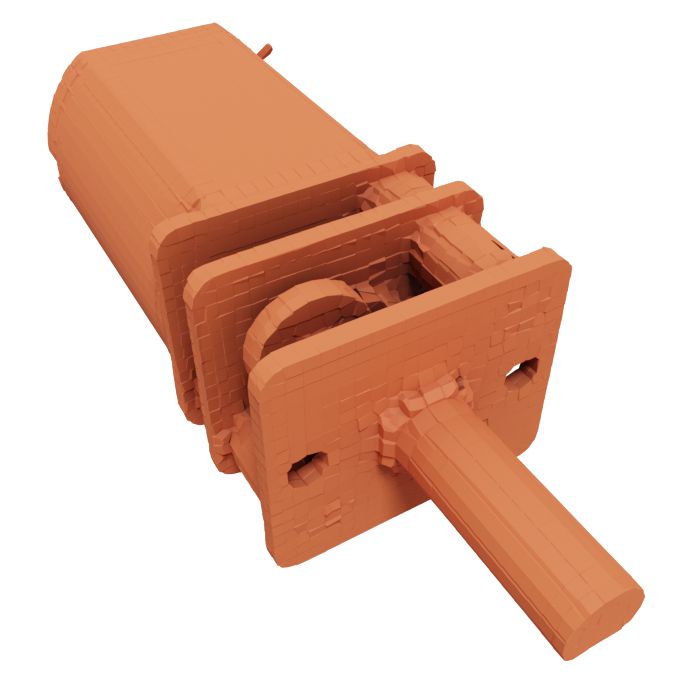} 
\end{subfigure}
 \begin{subfigure}{.18\textwidth}
  \centering
  \includegraphics[width=\linewidth]{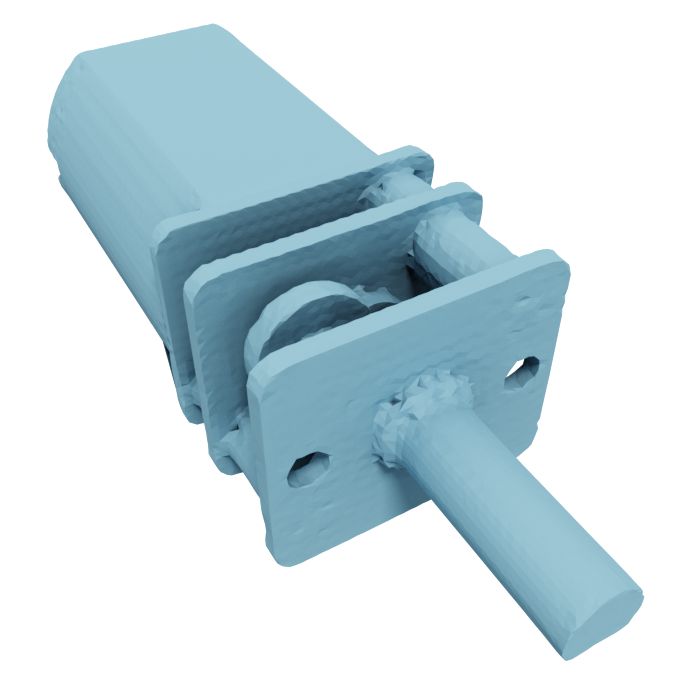} 
\end{subfigure}
 \begin{subfigure}{.18\textwidth}
  \centering
  \includegraphics[width=\linewidth]{figs/learning/gt_test_8.jpeg} 
\end{subfigure}
 \begin{subfigure}{.18\textwidth}
  \centering
  \includegraphics[width=\linewidth]{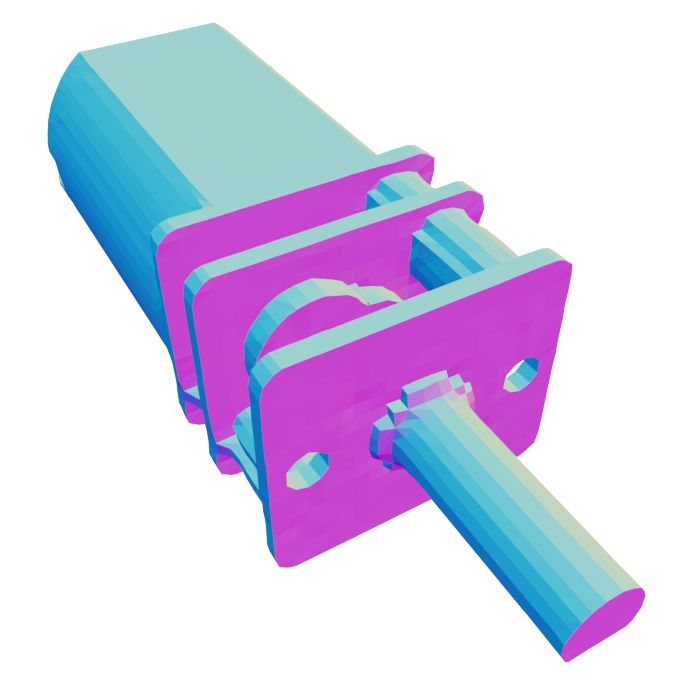} 
     \caption{NDC}

\end{subfigure}
 \begin{subfigure}{.18\textwidth}
  \centering
  \includegraphics[width=\linewidth]{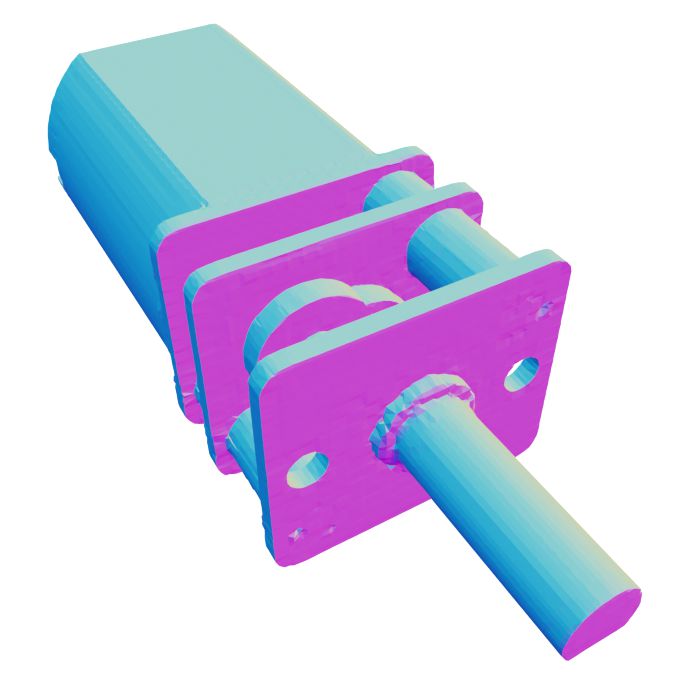} 
     \caption{NMC}

\end{subfigure}
 \begin{subfigure}{.18\textwidth}
  \centering
  \includegraphics[width=\linewidth]{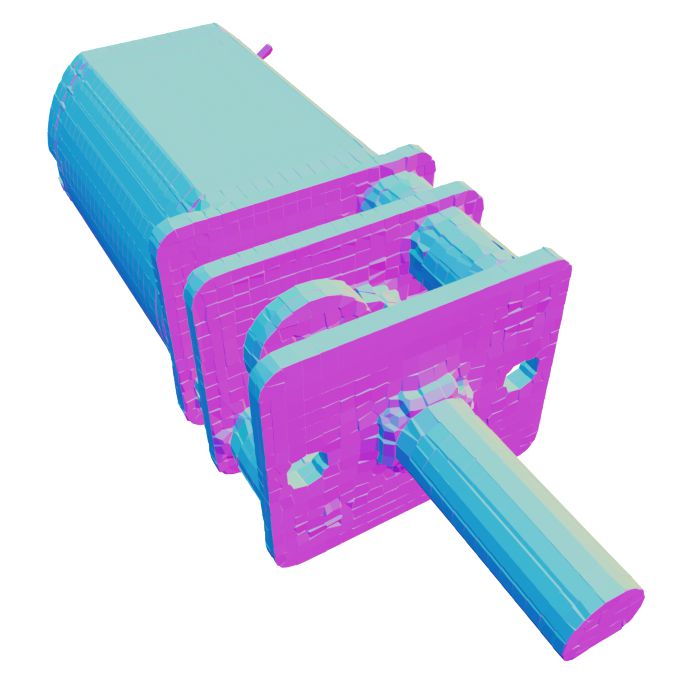} 
     \caption{VoroMesh}

\end{subfigure}
 \begin{subfigure}{.18\textwidth}
  \centering
  \includegraphics[width=\linewidth]{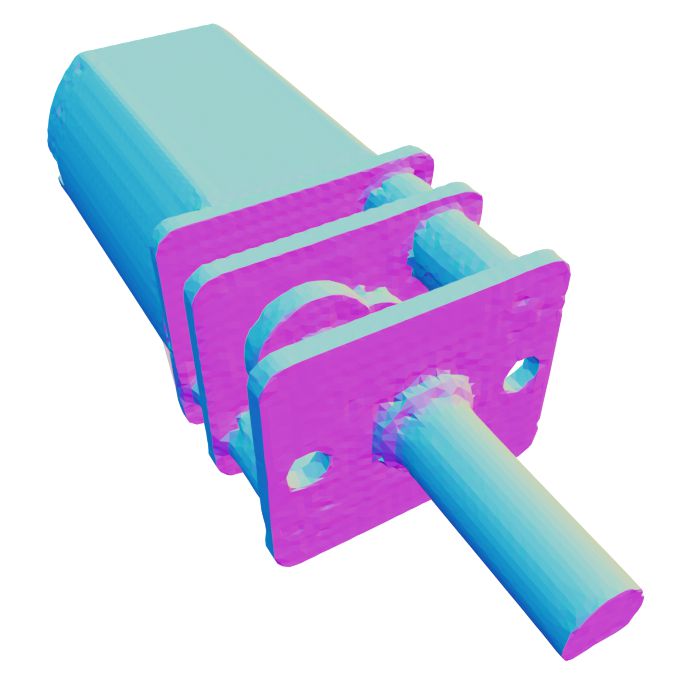} 
     \caption{PoNQ}

\end{subfigure}
 \begin{subfigure}{.18\textwidth}
  \centering
  \includegraphics[width=\linewidth]{figs/learning/gt_test_8_normal.jpeg}  
       \caption{Gr. Truth}

\end{subfigure}\\
\caption{Learning results (top: $32^3$; bottom: $64^3$) on ABC.
}\label{fig:supl_sdf_abc}
\end{figure*}

\begin{figure*}[t]
\centering
\begin{subfigure}{.18\textwidth}
  \centering
  \includegraphics[width=\linewidth]{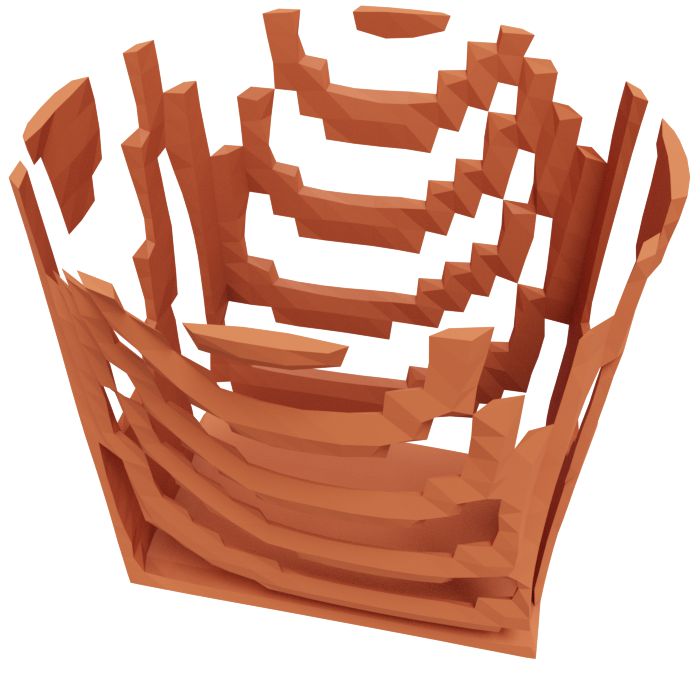} 
\end{subfigure}
 \begin{subfigure}{.18\textwidth}
  \centering
  \includegraphics[width=\linewidth]{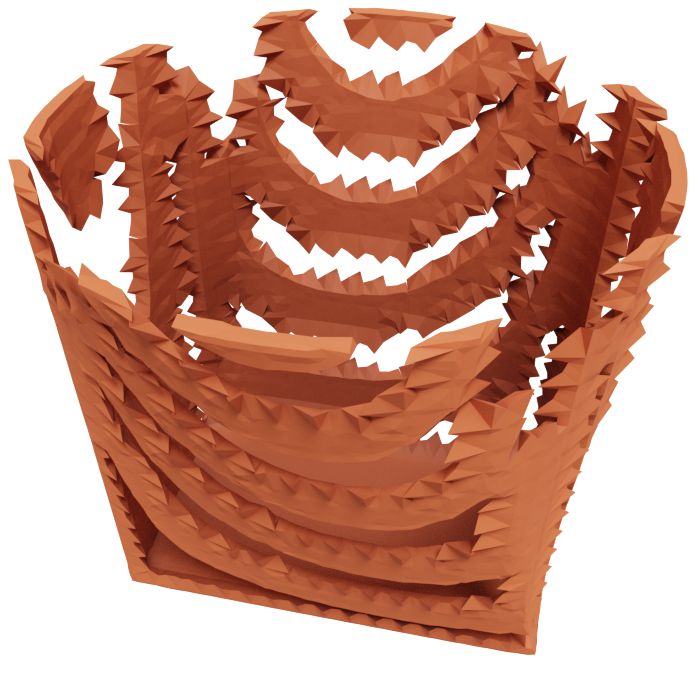} 
\end{subfigure}
 \begin{subfigure}{.18\textwidth}
  \centering
  \includegraphics[width=\linewidth]{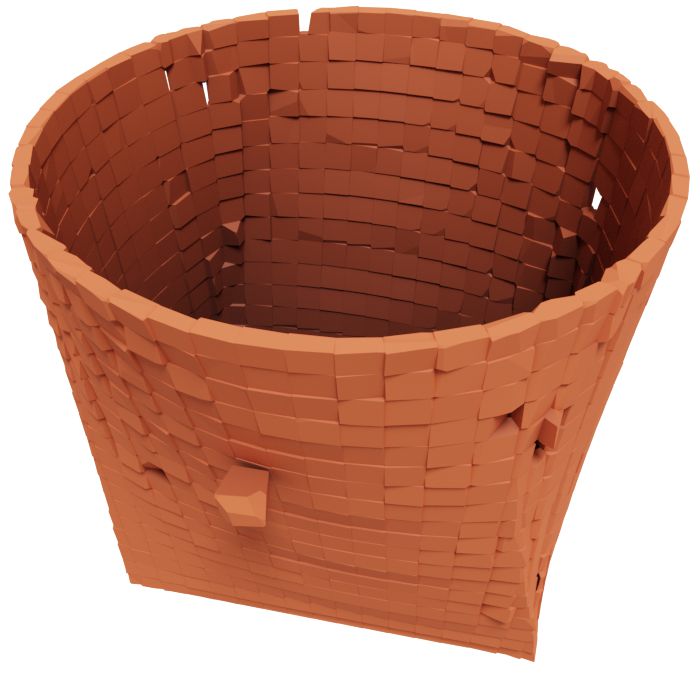} 
\end{subfigure}
 \begin{subfigure}{.18\textwidth}
  \centering
  \includegraphics[width=\linewidth]{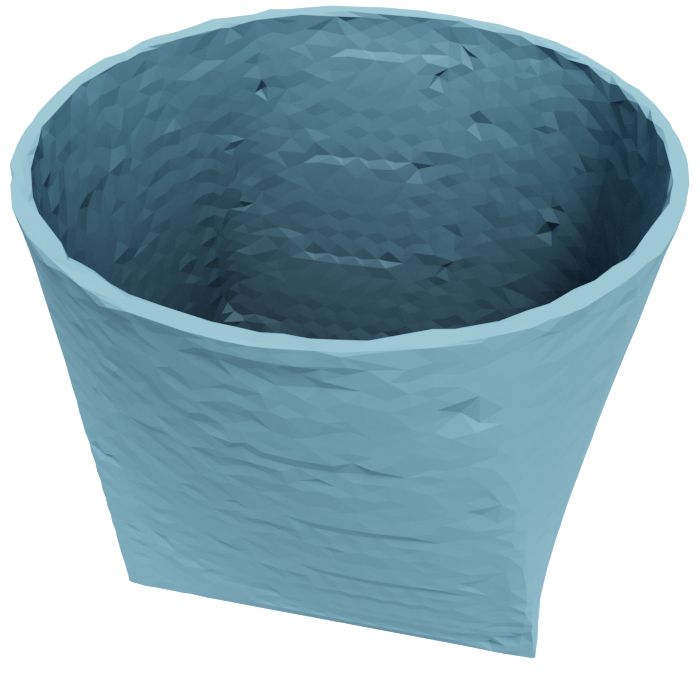} 
\end{subfigure}
 \begin{subfigure}{.18\textwidth}
  \centering
  \includegraphics[width=\linewidth]{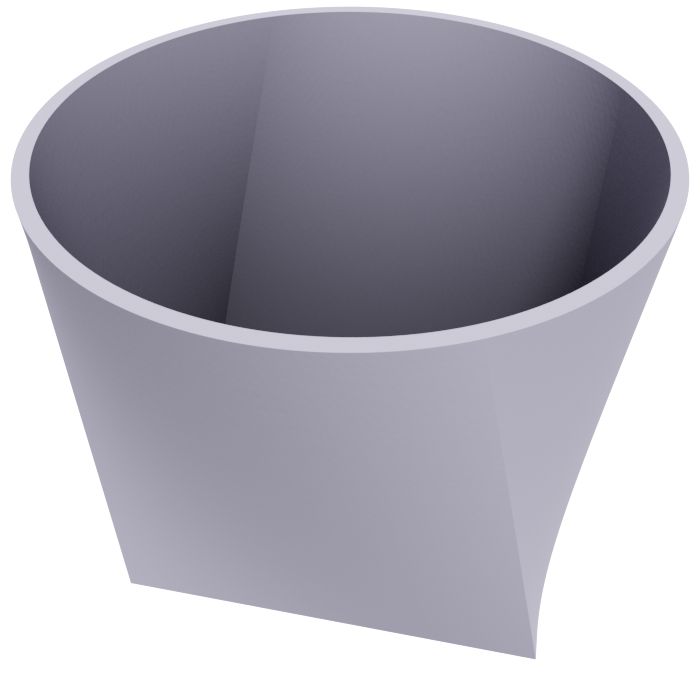} 
\end{subfigure}
 \begin{subfigure}{.18\textwidth}
  \includegraphics[width=\linewidth]{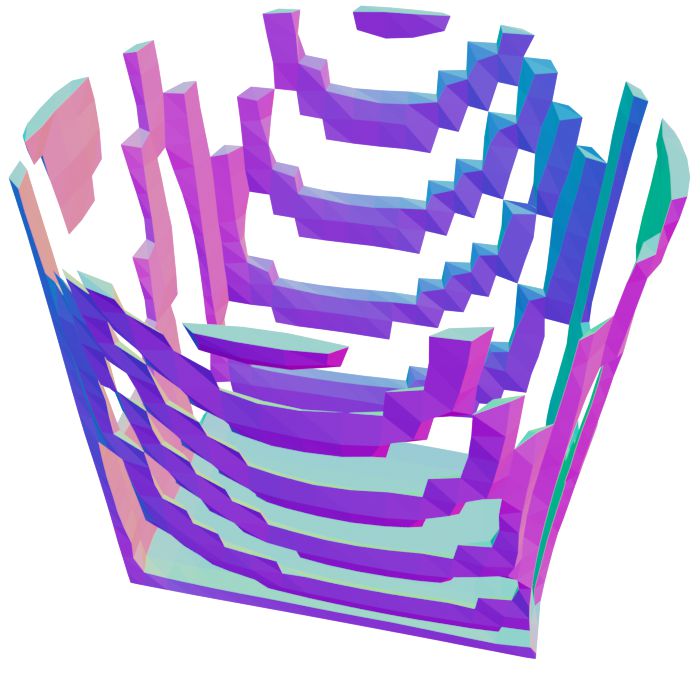}
\end{subfigure}
 \begin{subfigure}{.18\textwidth}
  \centering
  \includegraphics[width=\linewidth]{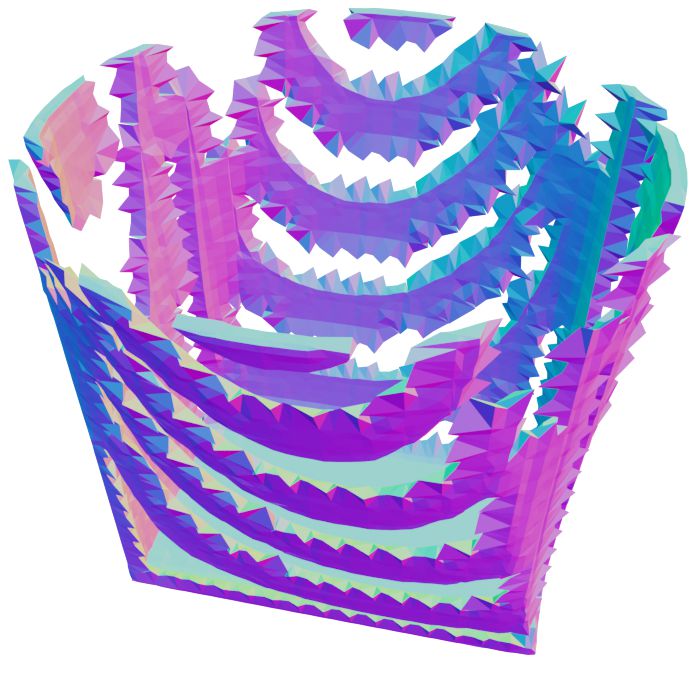}
\end{subfigure}
 \begin{subfigure}{.18\textwidth}
  \centering
  \includegraphics[width=\linewidth]{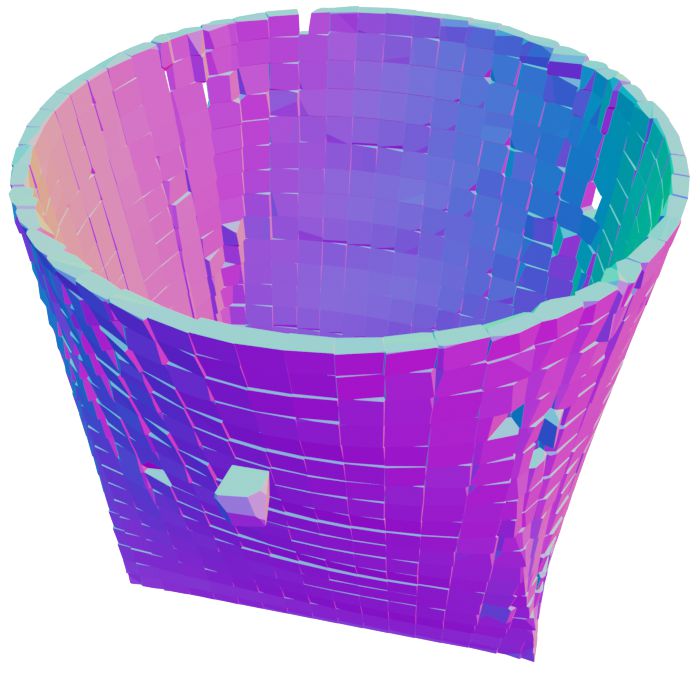}
\end{subfigure}
 \begin{subfigure}{.18\textwidth}
  \centering
  \includegraphics[width=\linewidth]{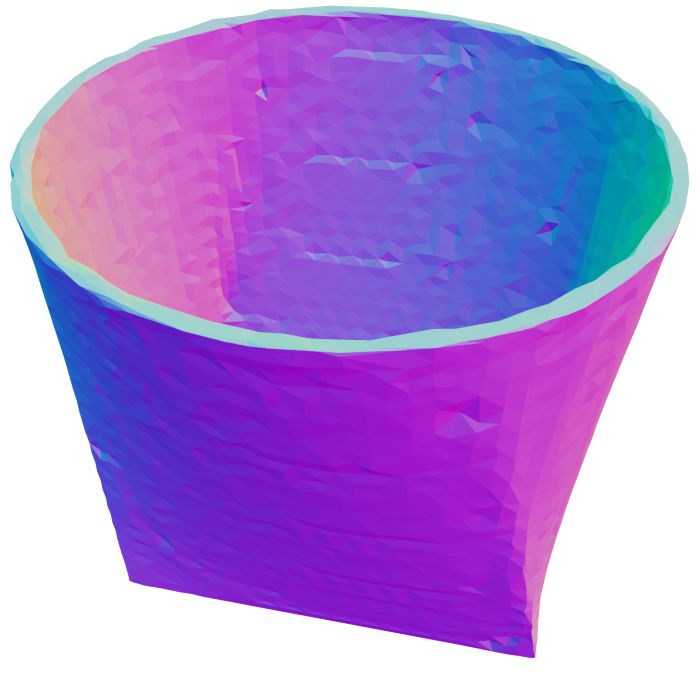}
\end{subfigure}
 \begin{subfigure}{.18\textwidth}
  \centering
  \includegraphics[width=\linewidth]{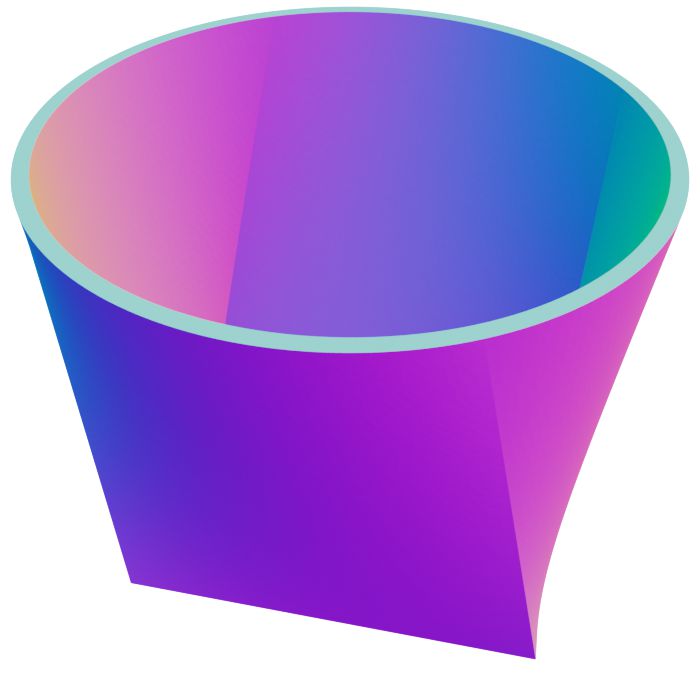} 
\end{subfigure} 

 \begin{subfigure}{.18\textwidth}
  \centering
  \includegraphics[width=\linewidth]{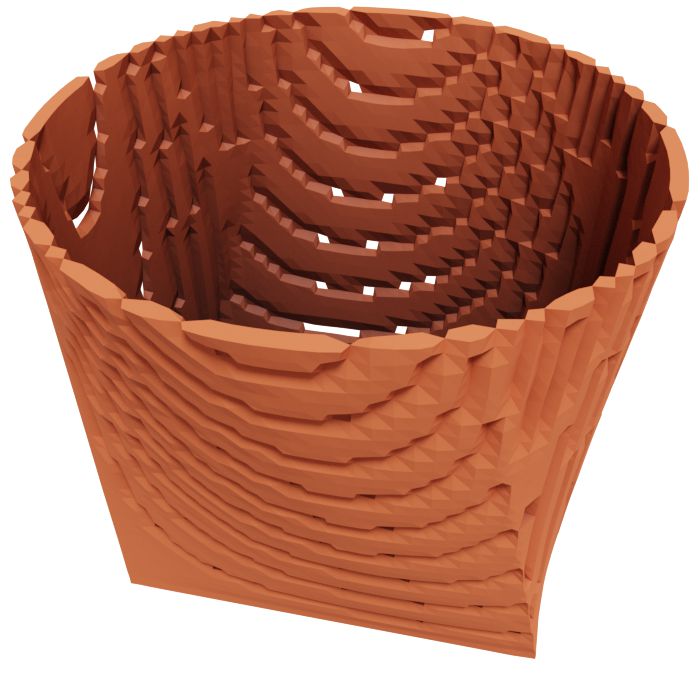} 
\end{subfigure}
 \begin{subfigure}{.18\textwidth}
  \centering
  \includegraphics[width=\linewidth]{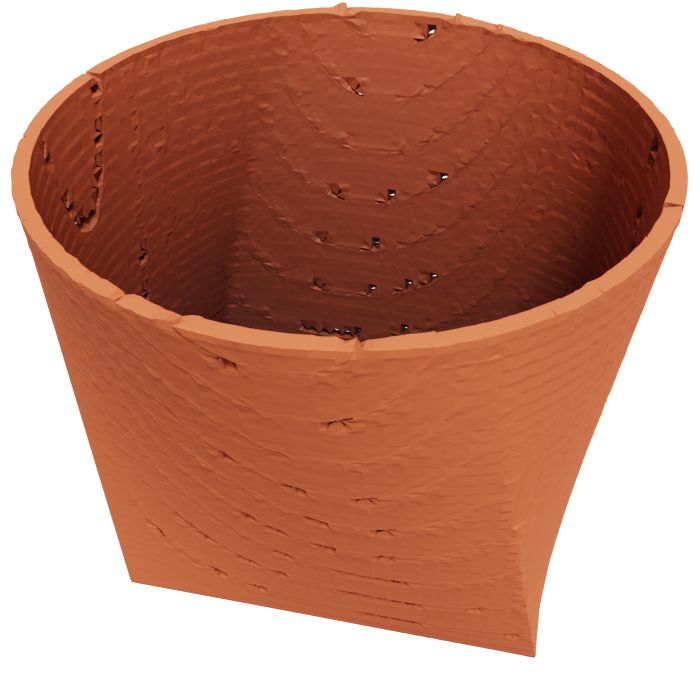} 
\end{subfigure}
 \begin{subfigure}{.18\textwidth}
  \centering
  \includegraphics[width=\linewidth]{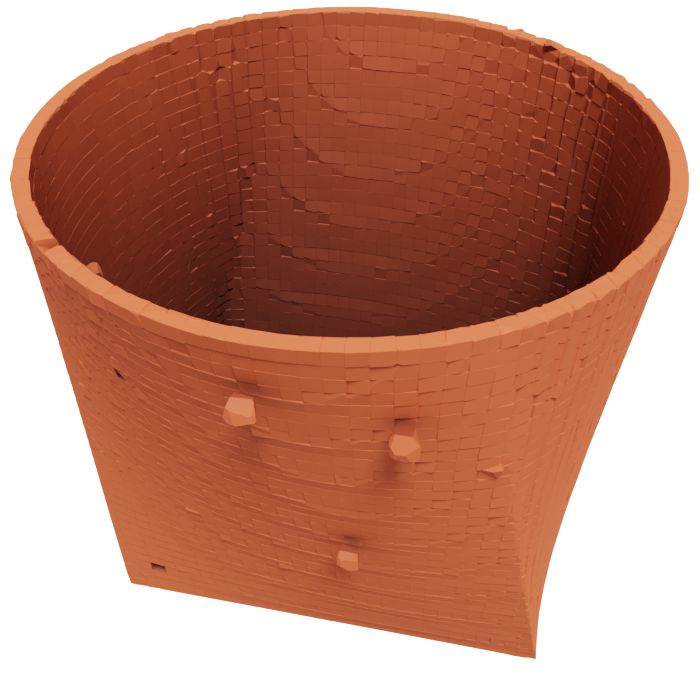} 
\end{subfigure}
 \begin{subfigure}{.18\textwidth}
  \centering
  \includegraphics[width=\linewidth]{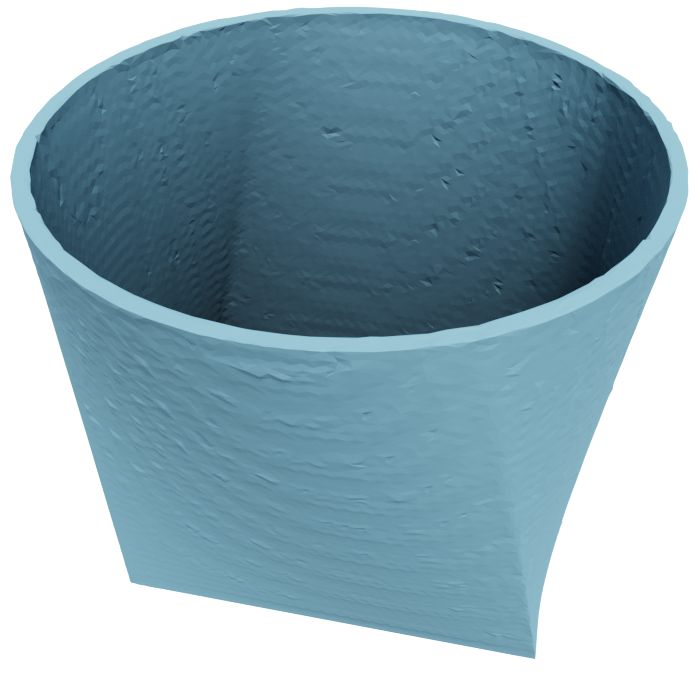} 
\end{subfigure}
 \begin{subfigure}{.18\textwidth}
  \centering
  \includegraphics[width=\linewidth]{figs/learning/gt_test_984_64.jpeg} 
\end{subfigure}
 \begin{subfigure}{.18\textwidth}
  \includegraphics[width=\linewidth]{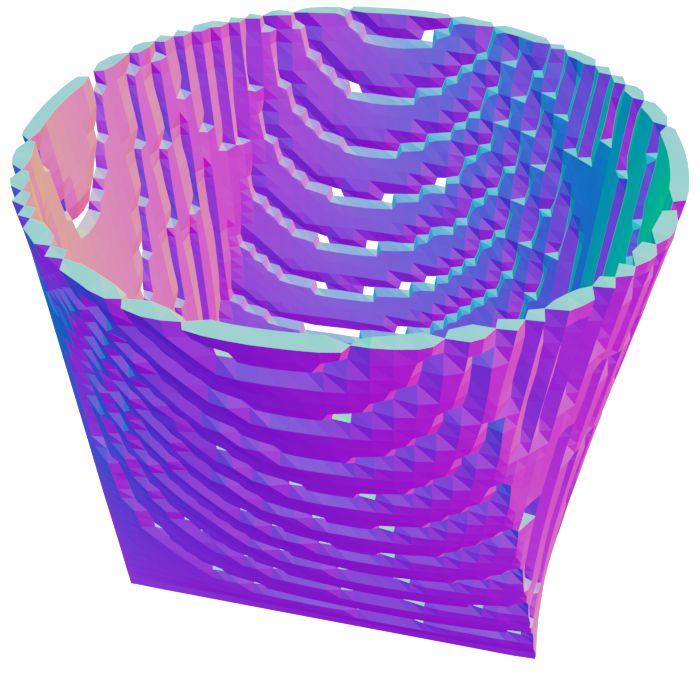}
   \caption{NDC}
\end{subfigure}
 \begin{subfigure}{.18\textwidth}
  \centering
  \includegraphics[width=\linewidth]{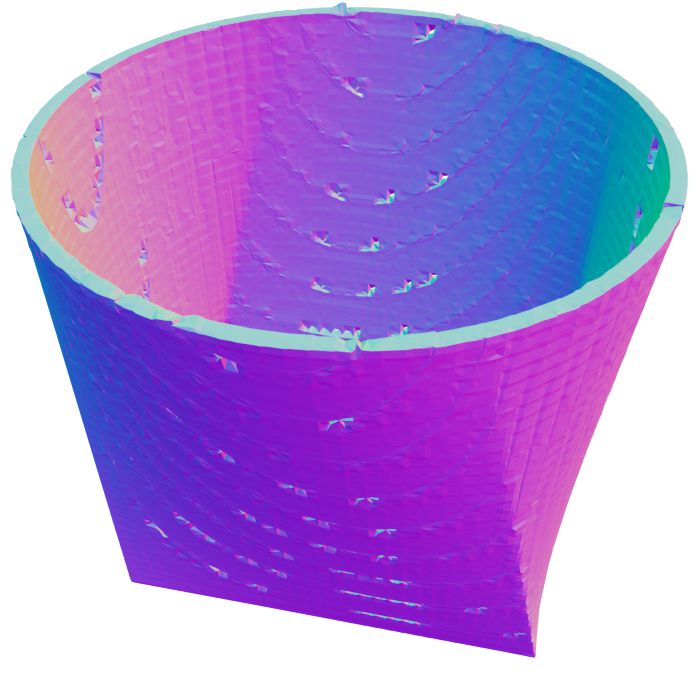}
    \caption{NMC}
\end{subfigure}
 \begin{subfigure}{.18\textwidth}
  \centering
  \includegraphics[width=\linewidth]{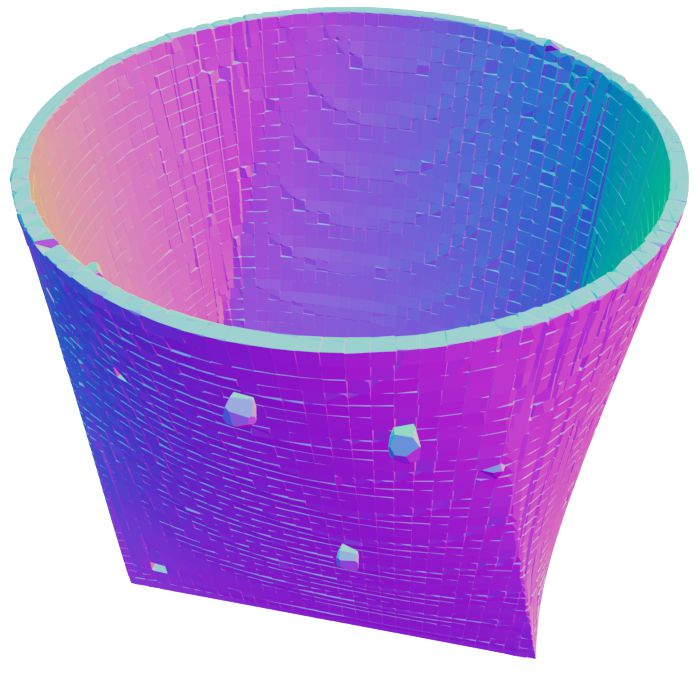}
  \caption{VoroMesh}
\end{subfigure}
 \begin{subfigure}{.18\textwidth}
  \centering
  \includegraphics[width=\linewidth]{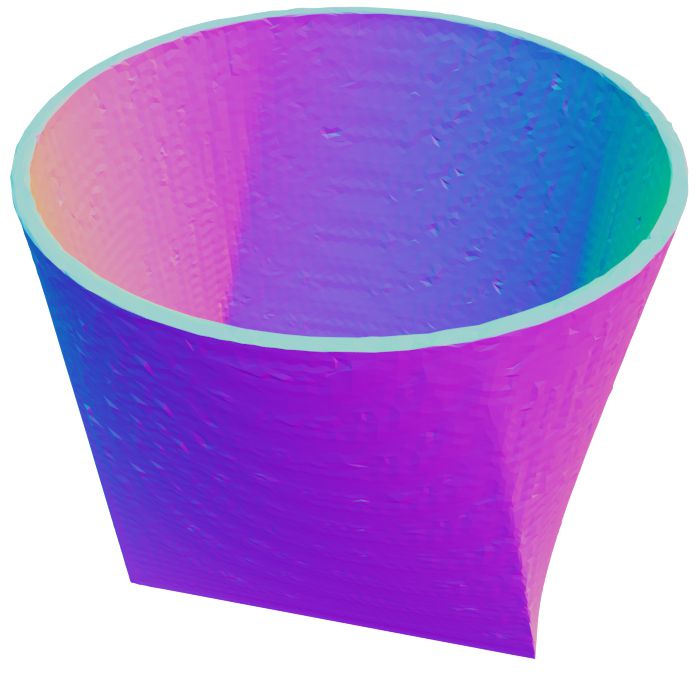}
   \caption{PoNQ}
\end{subfigure}
 \begin{subfigure}{.18\textwidth}
  \centering
  \includegraphics[width=\linewidth]{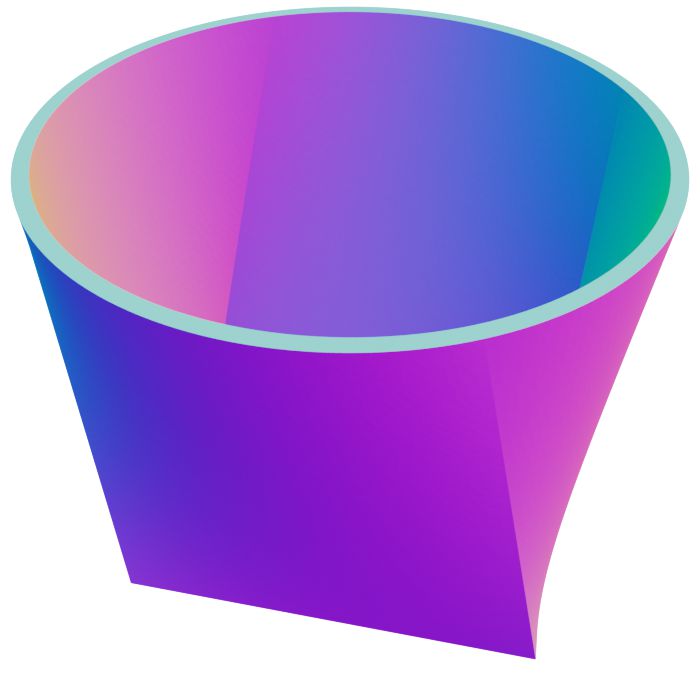} 
   \caption{Gr. Truth}
\end{subfigure} 
\caption{Learning results (top: $32^3$; bottom: $64^3$) on ABC.
}\label{fig:supl_sdf_abc2}
\end{figure*}

\begin{figure*}[t]
\centering
 \begin{subfigure}{.13\textwidth}
  \centering
  \includegraphics[width=\linewidth]{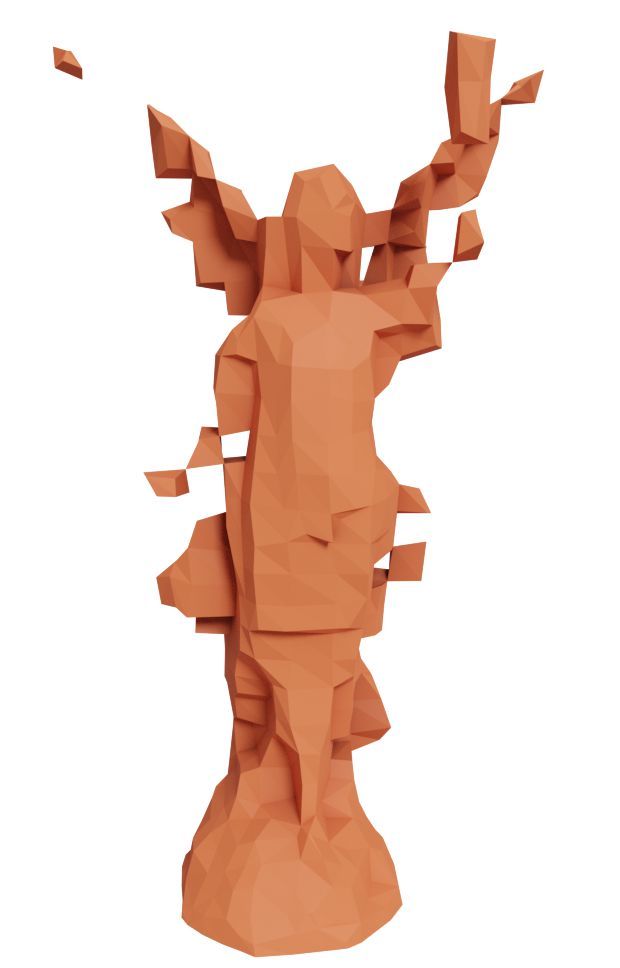} 
\end{subfigure}
 \begin{subfigure}{.13\textwidth}
  \centering
  \includegraphics[width=\linewidth]{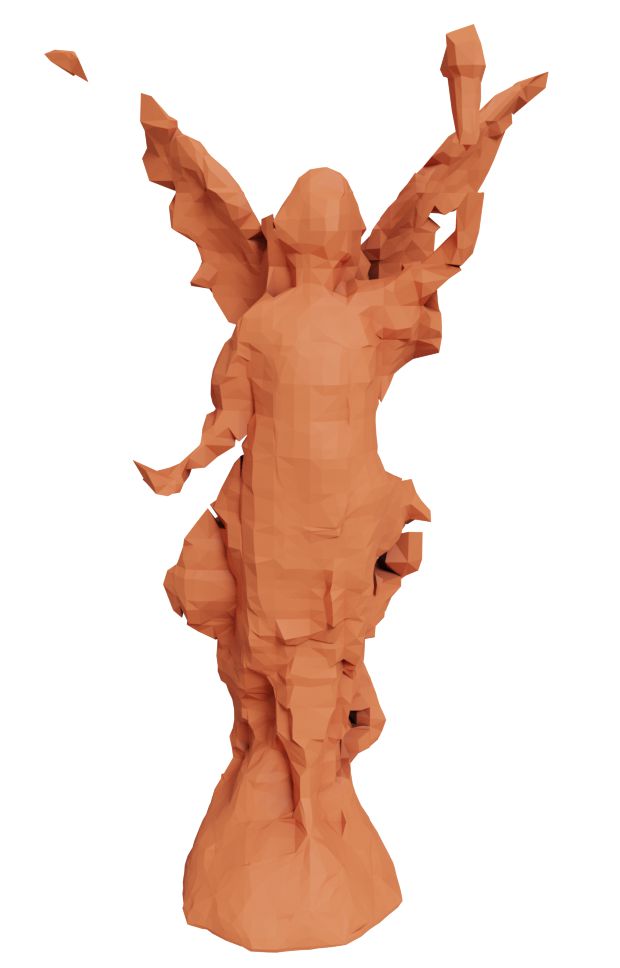} 
\end{subfigure}
 \begin{subfigure}{.13\textwidth}
  \centering
  \includegraphics[width=\linewidth]{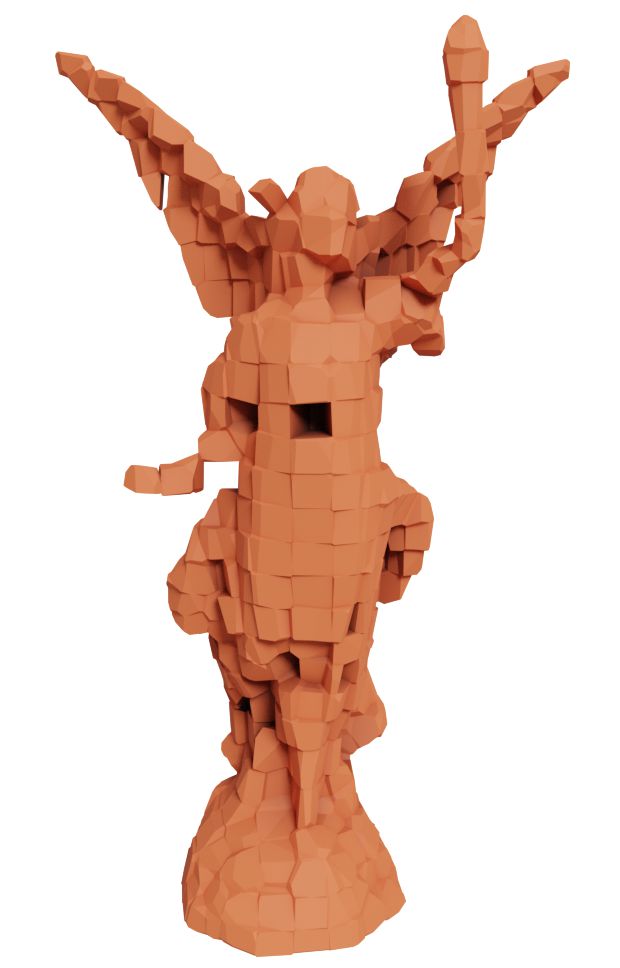} 
\end{subfigure}
 \begin{subfigure}{.13\textwidth}
  \centering
  \includegraphics[width=\linewidth]{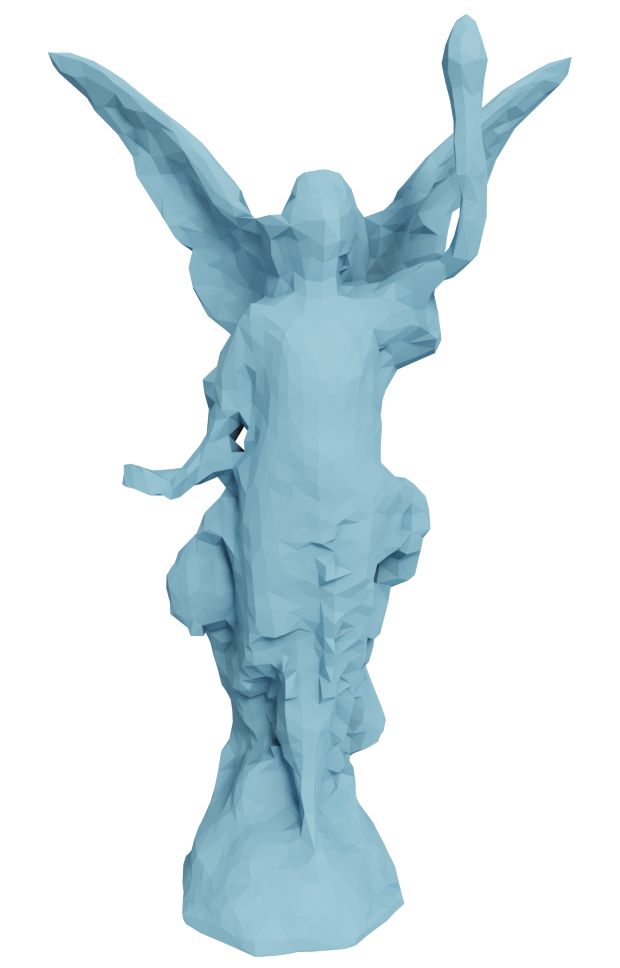} 
\end{subfigure}
 \begin{subfigure}{.13\textwidth}
  \centering
  \includegraphics[width=\linewidth]{figs/groundtruth/252119.jpeg} 
\end{subfigure}

\begin{subfigure}{.13\textwidth}
  \centering
  \includegraphics[width=\linewidth]{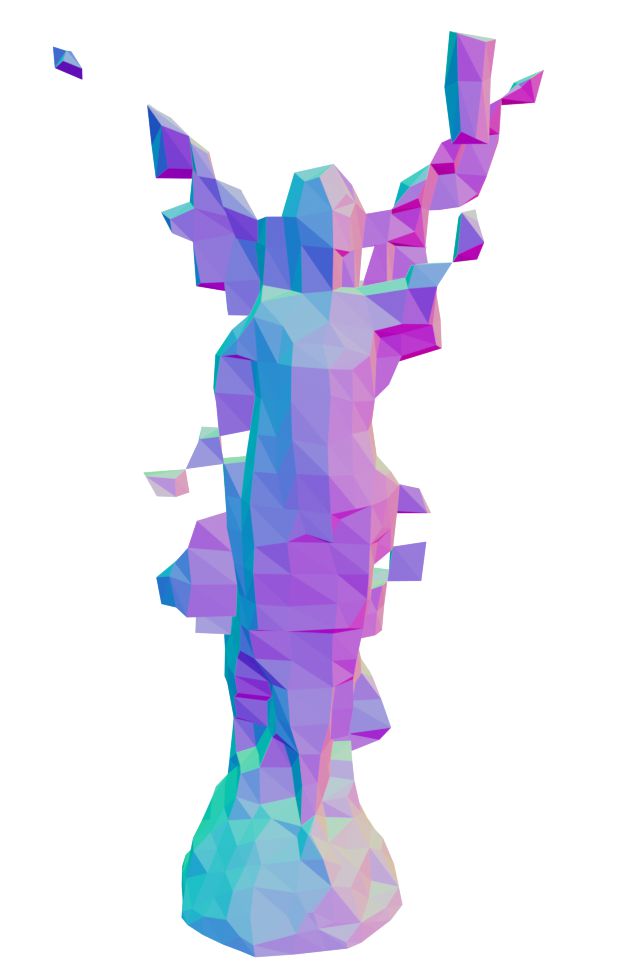} 
\end{subfigure}
 \begin{subfigure}{.13\textwidth}
  \centering
  \includegraphics[width=\linewidth]{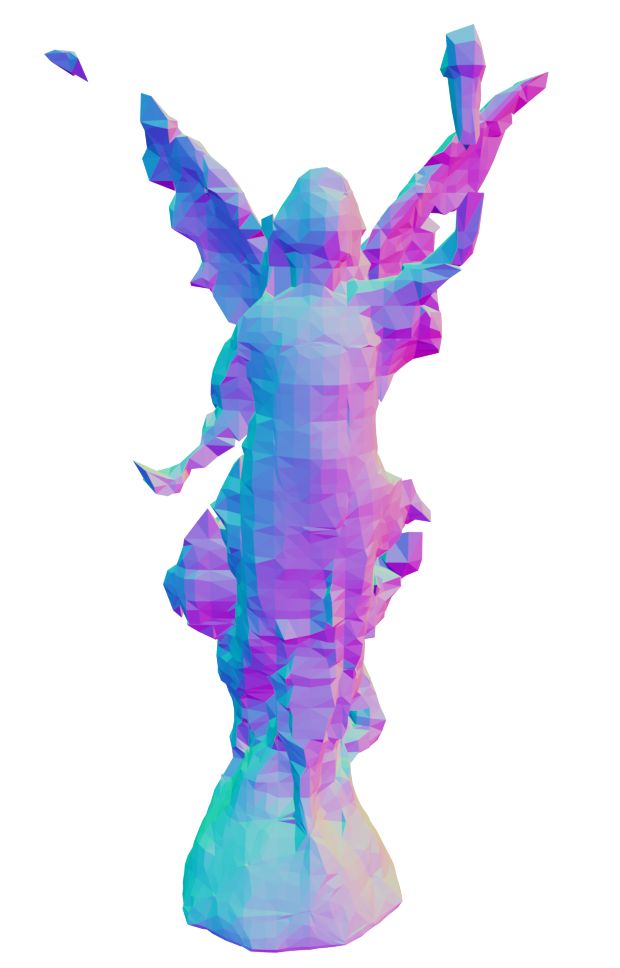} 
\end{subfigure}
 \begin{subfigure}{.13\textwidth}
  \centering
  \includegraphics[width=\linewidth]{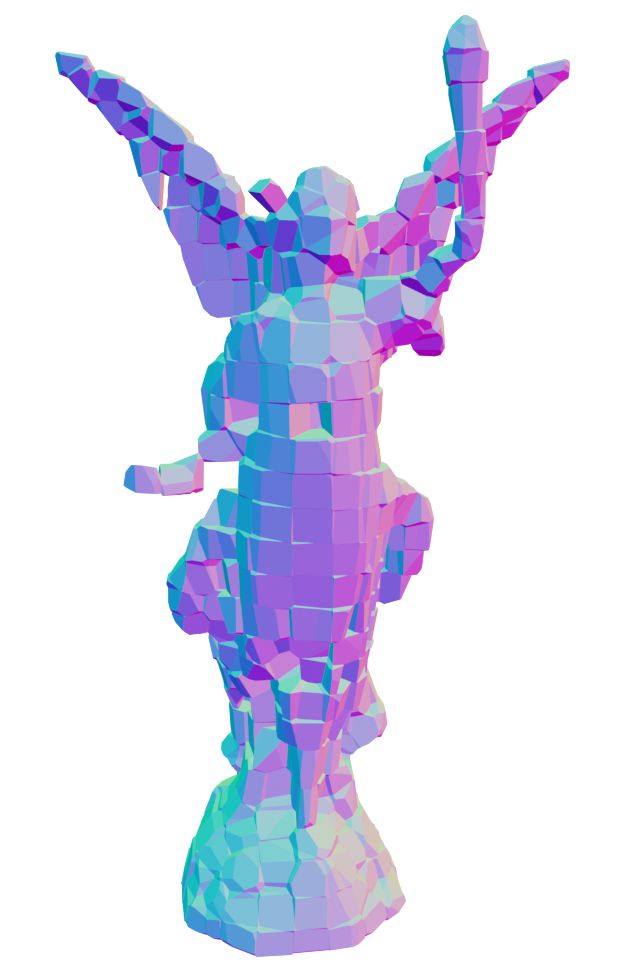} 
\end{subfigure}
 \begin{subfigure}{.13\textwidth}
  \centering
  \includegraphics[width=\linewidth]{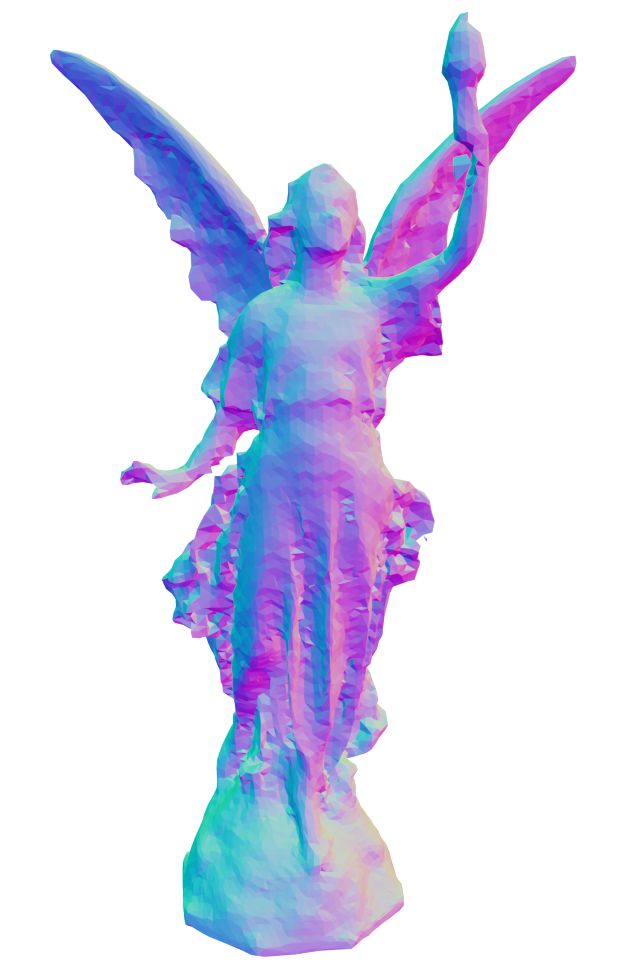} 
\end{subfigure}
 \begin{subfigure}{.13\textwidth}
  \centering
  \includegraphics[width=\linewidth]{figs/groundtruth/252119_normal.jpeg} 
\end{subfigure}

\centering
 \begin{subfigure}{.13\textwidth}
  \centering
  \includegraphics[width=\linewidth]{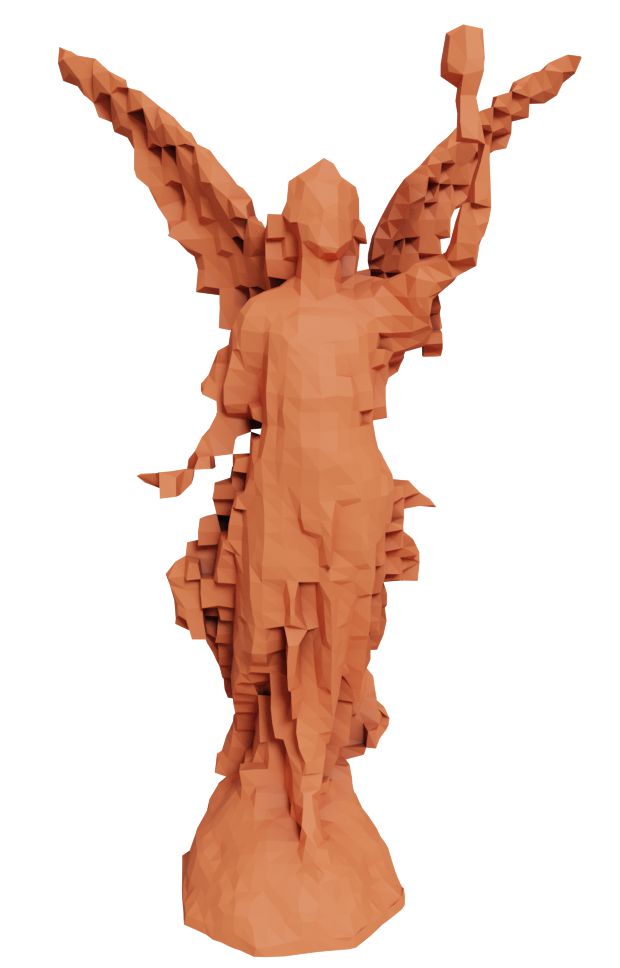} 
\end{subfigure}
 \begin{subfigure}{.13\textwidth}
  \centering
  \includegraphics[width=\linewidth]{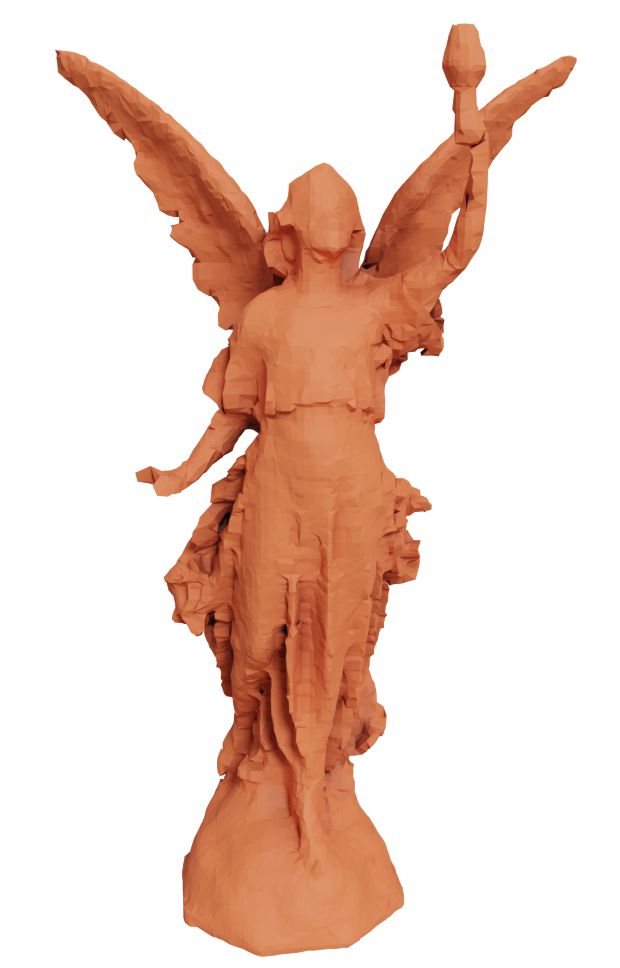} 
\end{subfigure}
 \begin{subfigure}{.13\textwidth}
  \centering
  \includegraphics[width=\linewidth]{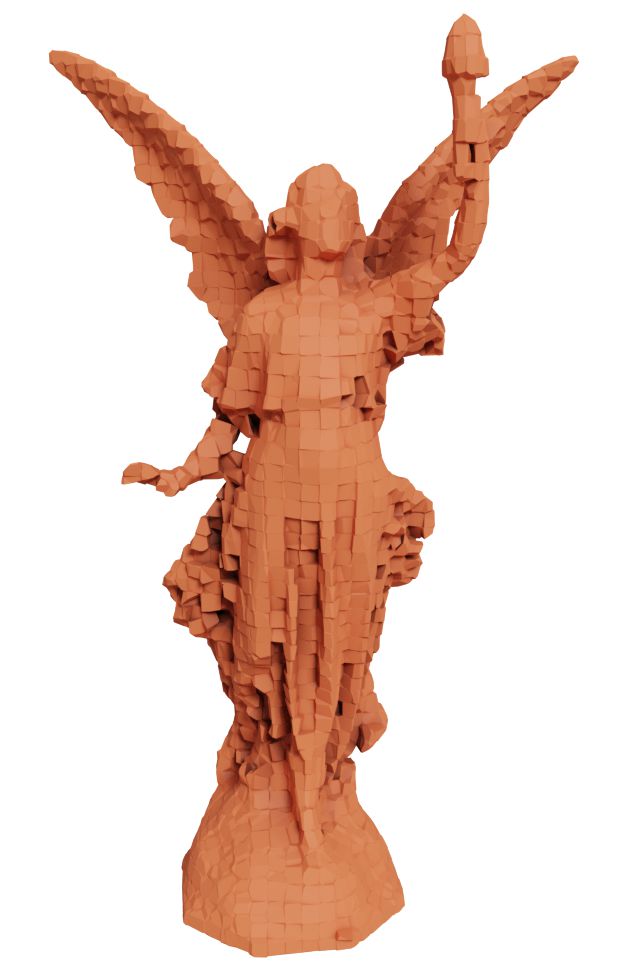} 
\end{subfigure}
 \begin{subfigure}{.13\textwidth}
  \centering
  \includegraphics[width=\linewidth]{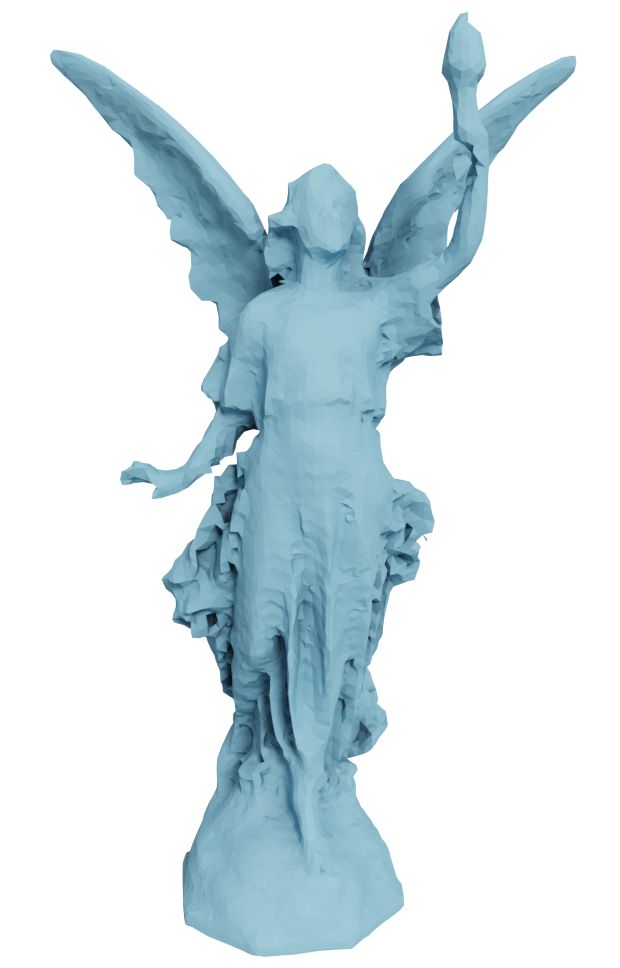} 
\end{subfigure}
 \begin{subfigure}{.13\textwidth}
  \centering
  \includegraphics[width=\linewidth]{figs/groundtruth/252119.jpeg} 
\end{subfigure}

\begin{subfigure}{.13\textwidth}
  \centering
  \includegraphics[width=\linewidth]{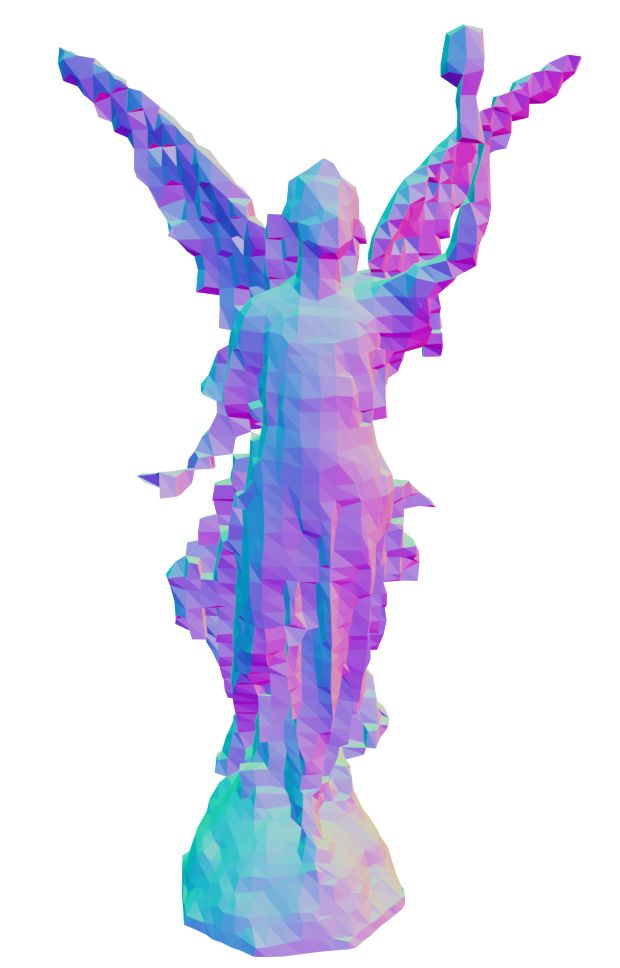} 
\end{subfigure}
 \begin{subfigure}{.13\textwidth}
  \centering
  \includegraphics[width=\linewidth]{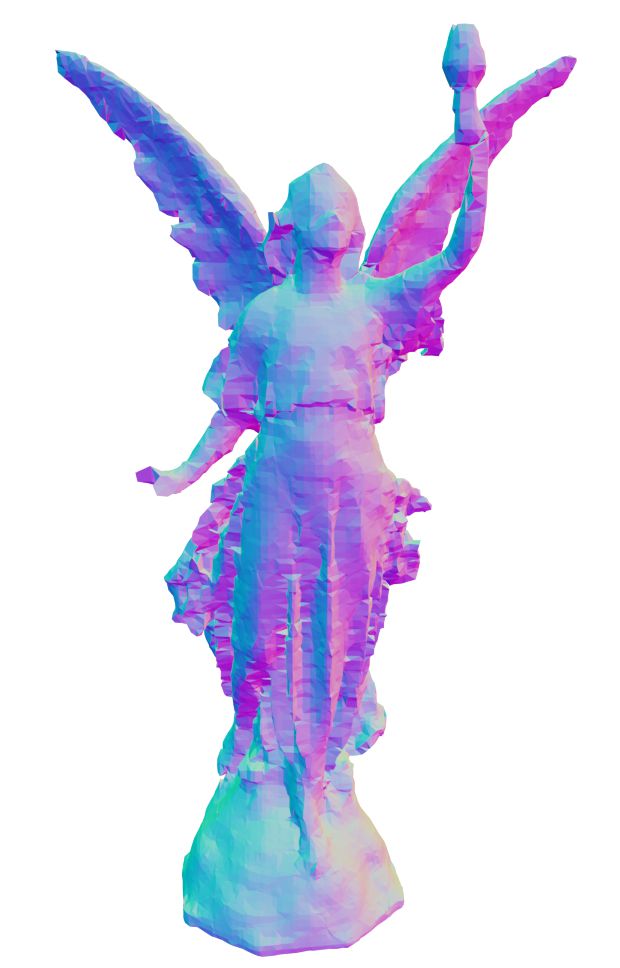} 
\end{subfigure}
 \begin{subfigure}{.13\textwidth}
  \centering
  \includegraphics[width=\linewidth]{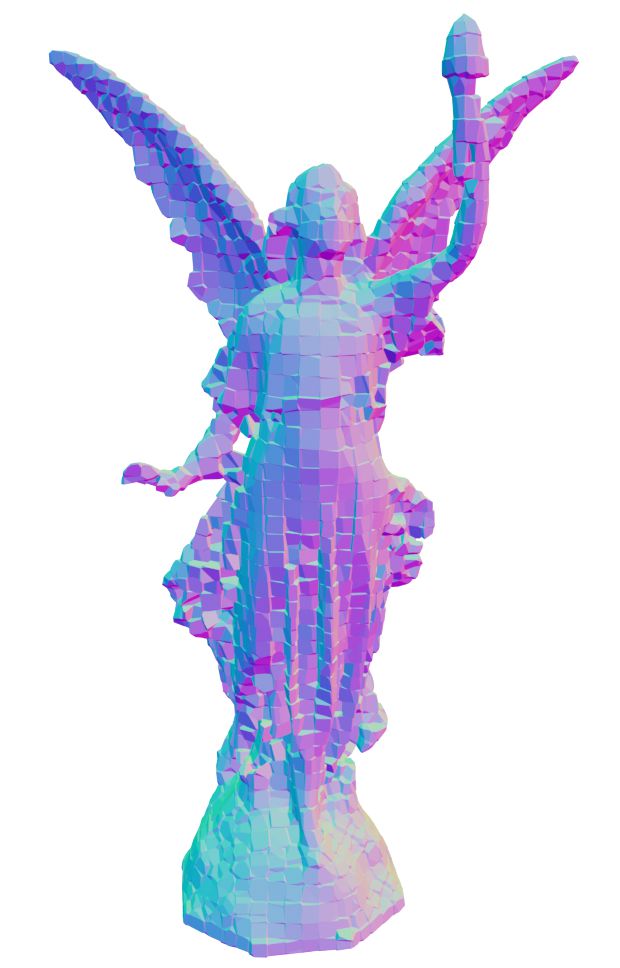} 
\end{subfigure}
 \begin{subfigure}{.13\textwidth}
  \centering
  \includegraphics[width=\linewidth]{figs/learning/quadric_252119_normal_64.jpeg} 
\end{subfigure}
 \begin{subfigure}{.13\textwidth}
  \centering
  \includegraphics[width=\linewidth]{figs/groundtruth/252119_normal.jpeg} 
\end{subfigure}

\centering
 \begin{subfigure}{.13\textwidth}
  \centering
  \includegraphics[width=\linewidth]{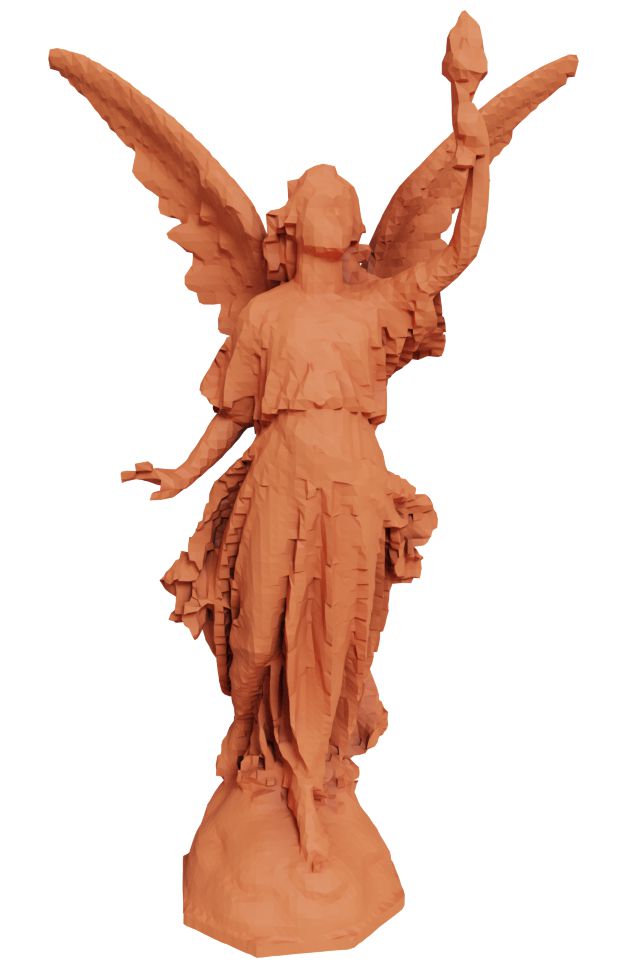} 
\end{subfigure}
 \begin{subfigure}{.13\textwidth}
  \centering
  \includegraphics[width=\linewidth]{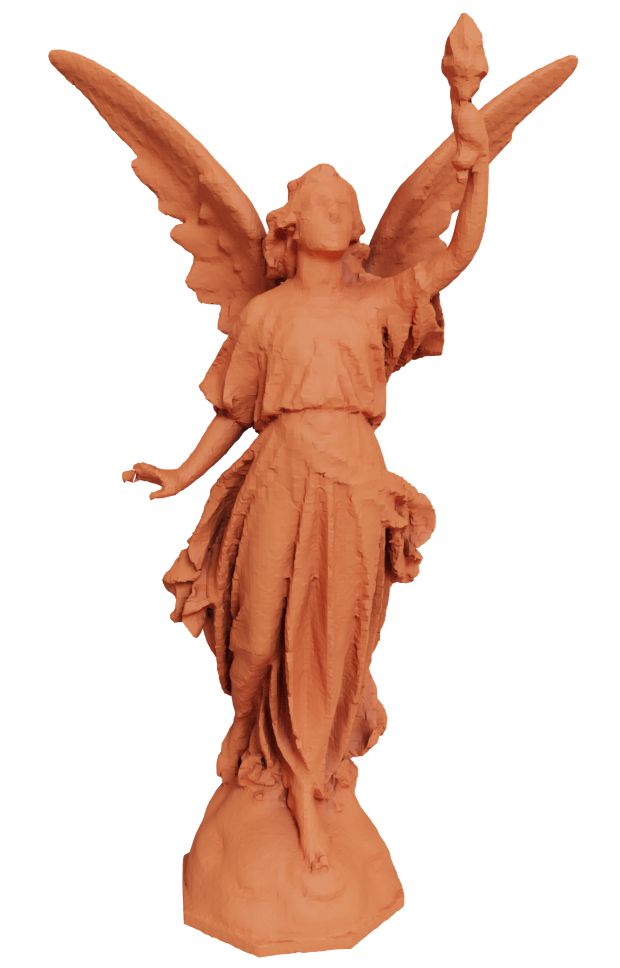} 
\end{subfigure}
 \begin{subfigure}{.13\textwidth}
  \centering
  \includegraphics[width=\linewidth]{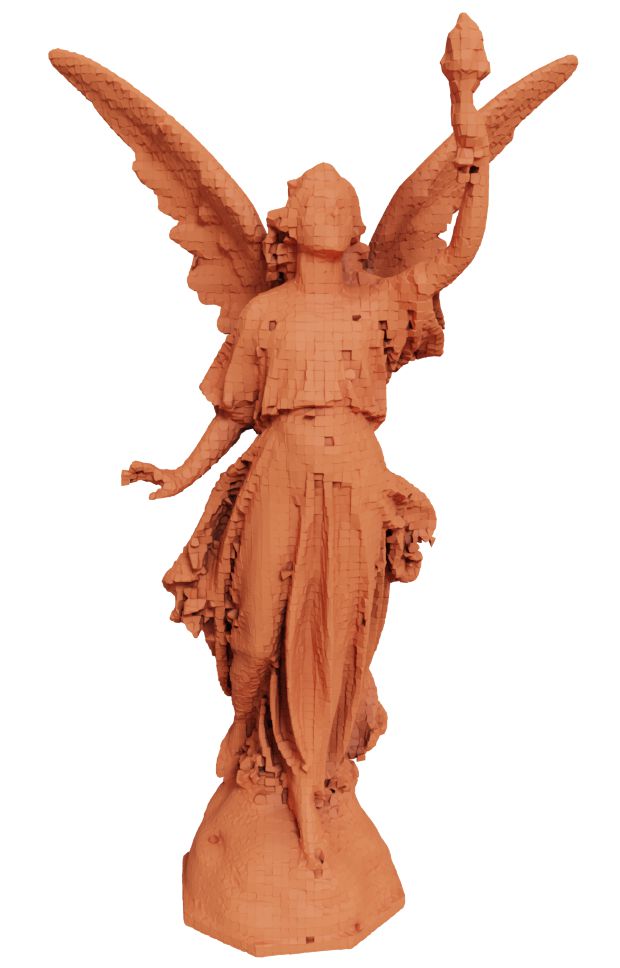} 
\end{subfigure}
 \begin{subfigure}{.13\textwidth}
  \centering
  \includegraphics[width=\linewidth]{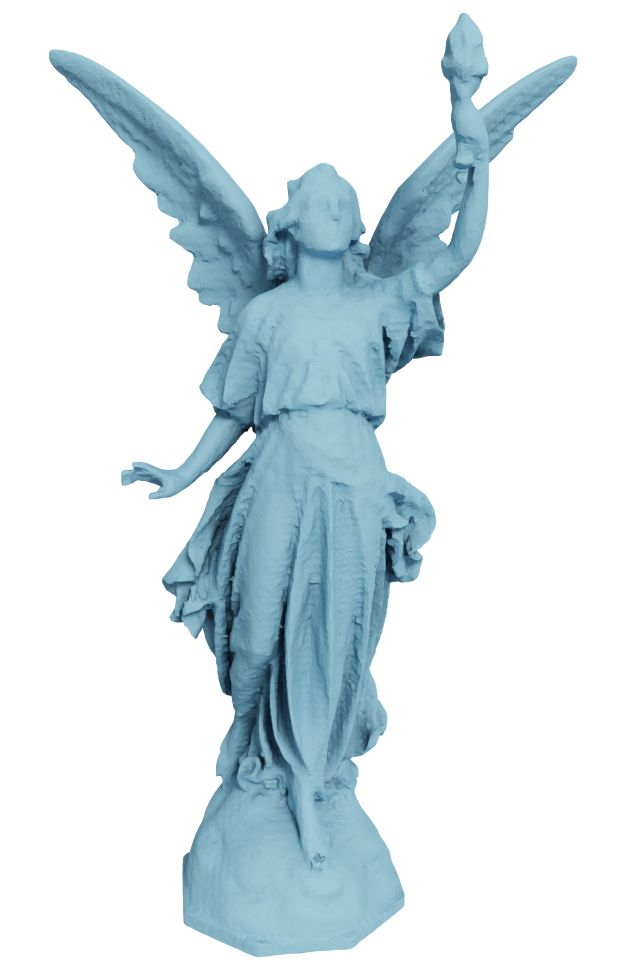} 
\end{subfigure}
 \begin{subfigure}{.13\textwidth}
  \centering
  \includegraphics[width=\linewidth]{figs/groundtruth/252119.jpeg} 
\end{subfigure}

\begin{subfigure}{.13\textwidth}
  \centering
  \includegraphics[width=\linewidth]{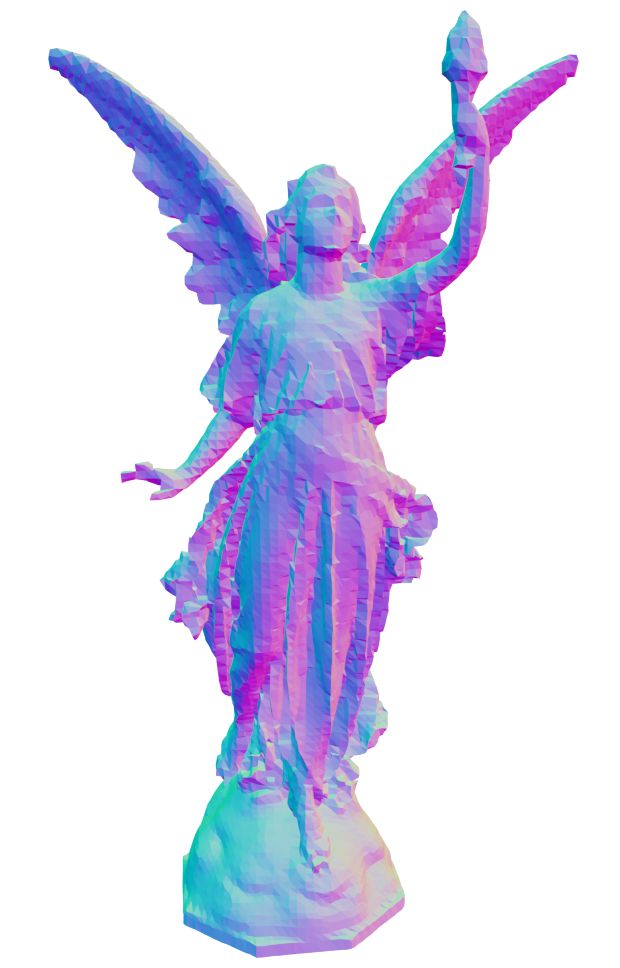} 
     \caption{NDC}
\end{subfigure}
 \begin{subfigure}{.13\textwidth}
  \centering
  \includegraphics[width=\linewidth]{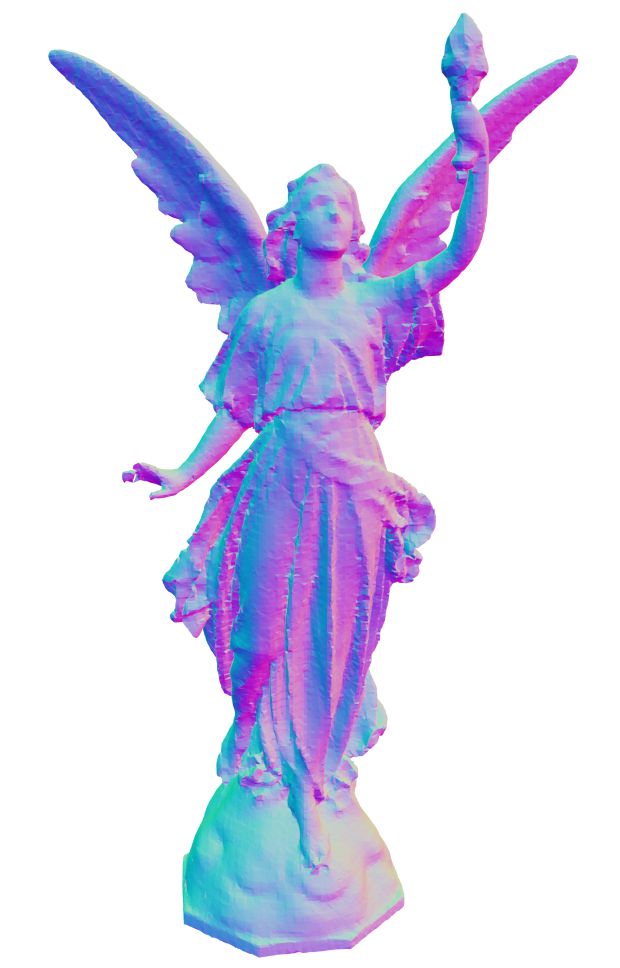} 
     \caption{NMC}
\end{subfigure}
 \begin{subfigure}{.13\textwidth}
  \centering
  \includegraphics[width=\linewidth]{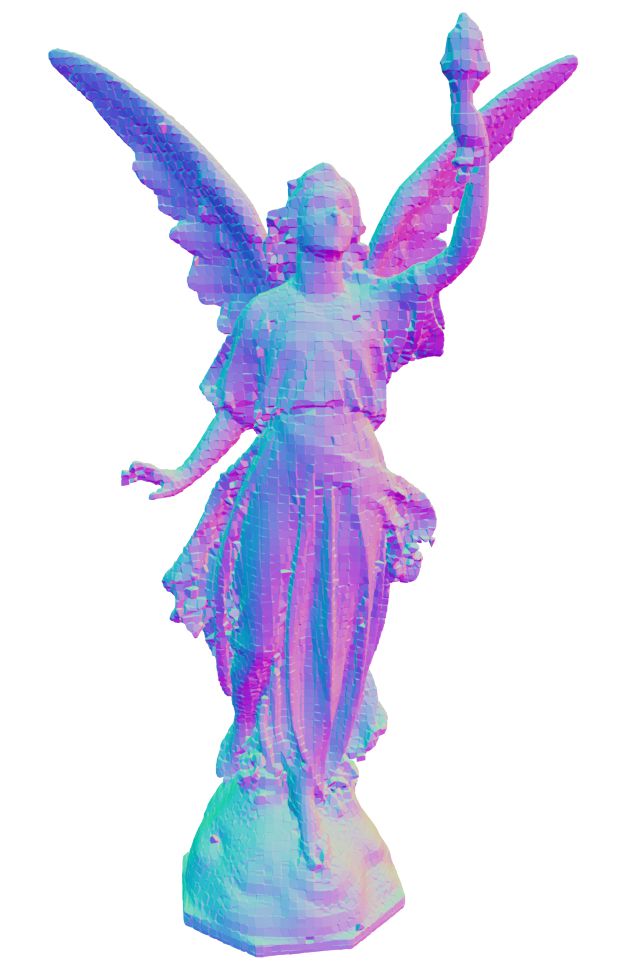} 
     \caption{VoroMesh}
\end{subfigure}
 \begin{subfigure}{.13\textwidth}
  \centering
  \includegraphics[width=\linewidth]{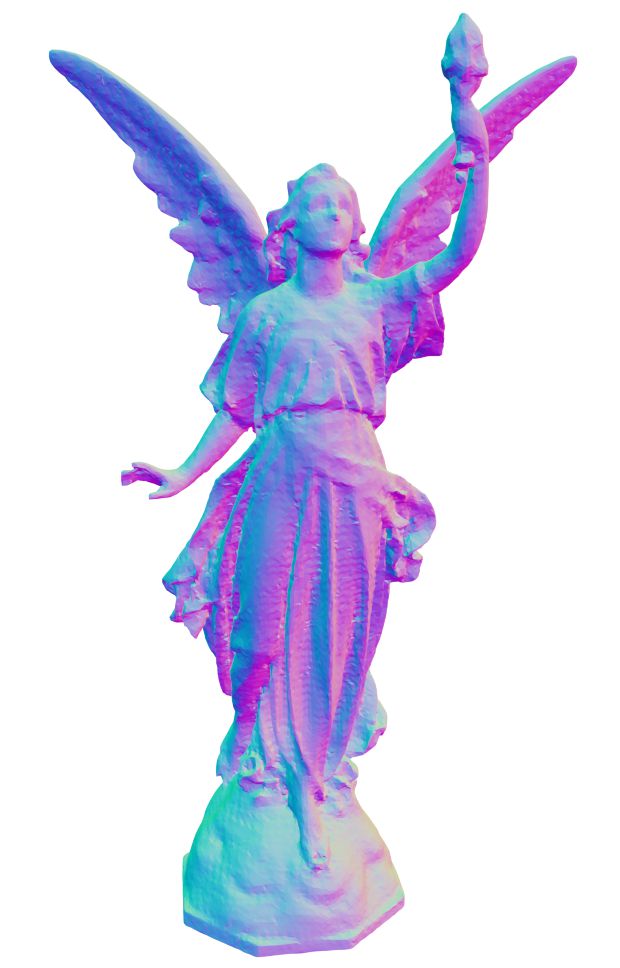} 
     \caption{PoNQ}
\end{subfigure}
 \begin{subfigure}{.13\textwidth}
  \centering
  \includegraphics[width=\linewidth]{figs/groundtruth/252119_normal.jpeg} 
     \caption{Gr. Truth}
\end{subfigure}

\caption{Learning results (top to bottom: $32^3$, $64^3$, $128^3$) on Thingi30. Networks trained on ABC.
}
  \label{fig:sdf_thingi_supl}
\end{figure*}

\begin{figure*}[t]
\centering
 \begin{subfigure}{.13\textwidth}
  \centering
  \includegraphics[width=\linewidth]{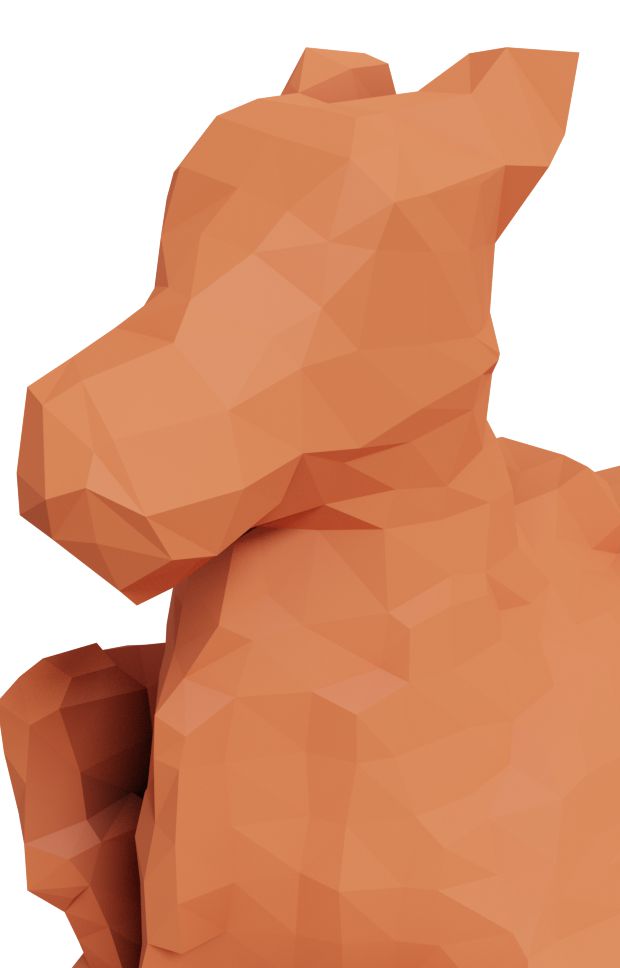} 
\end{subfigure}
 \begin{subfigure}{.13\textwidth}
  \centering
  \includegraphics[width=\linewidth]{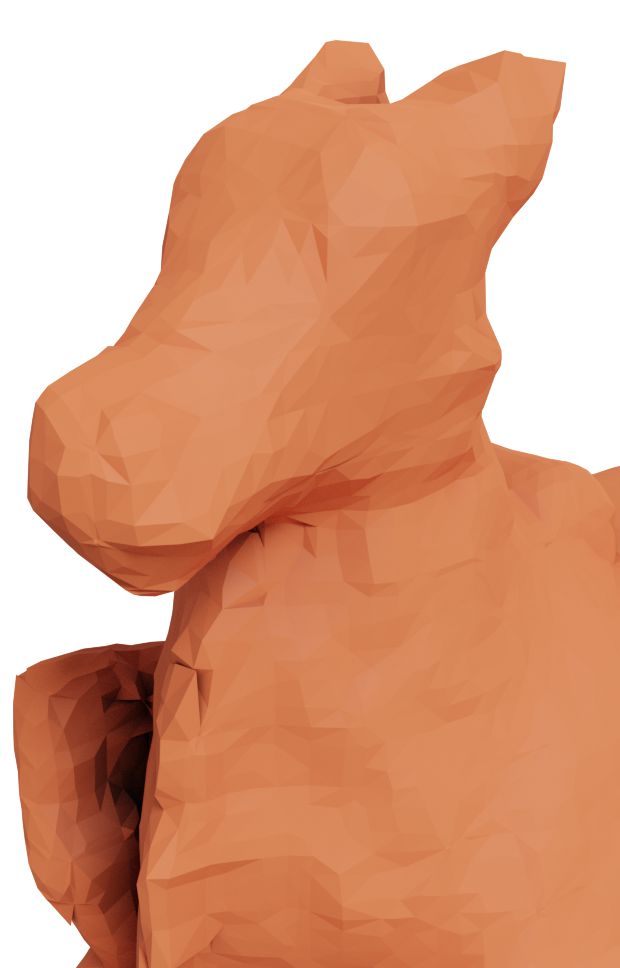} 
\end{subfigure}
 \begin{subfigure}{.13\textwidth}
  \centering
  \includegraphics[width=\linewidth]{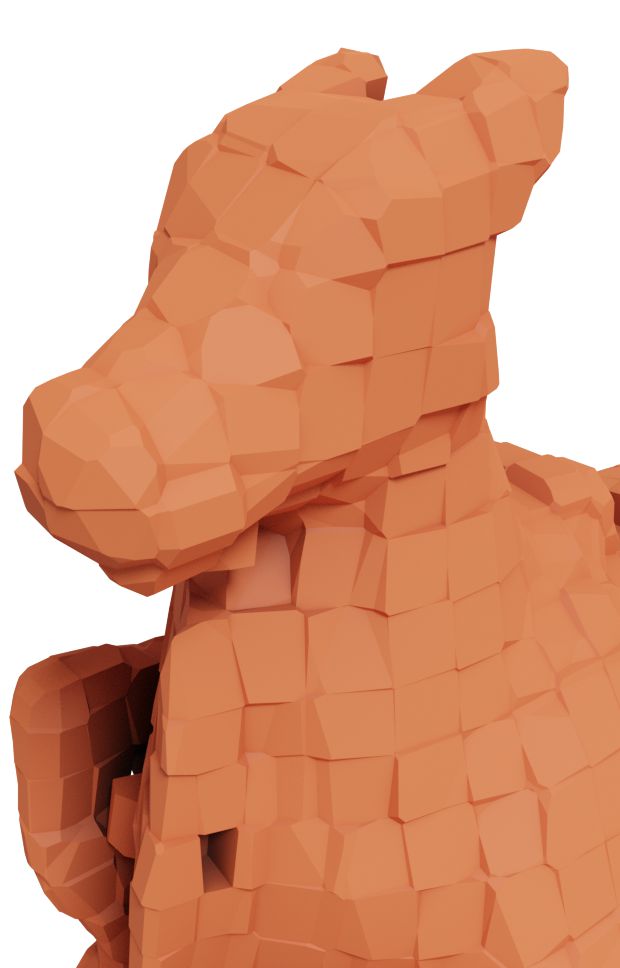} 
\end{subfigure}
 \begin{subfigure}{.13\textwidth}
  \centering
  \includegraphics[width=\linewidth]{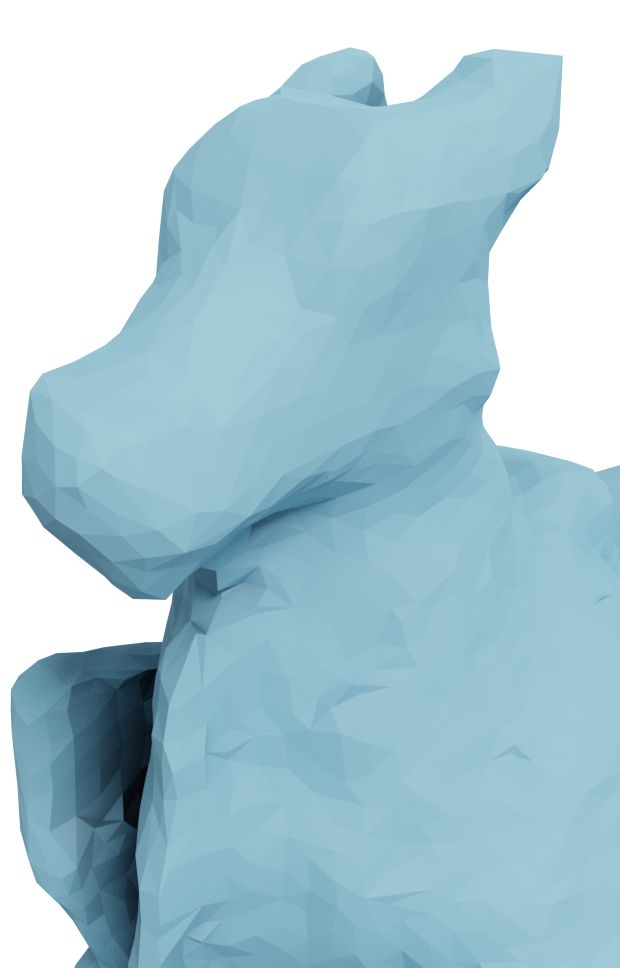} 
\end{subfigure}
 \begin{subfigure}{.13\textwidth}
  \centering
  \includegraphics[width=\linewidth]{figs/groundtruth/90889.jpeg} 
\end{subfigure}

\begin{subfigure}{.13\textwidth}
  \centering
  \includegraphics[width=\linewidth]{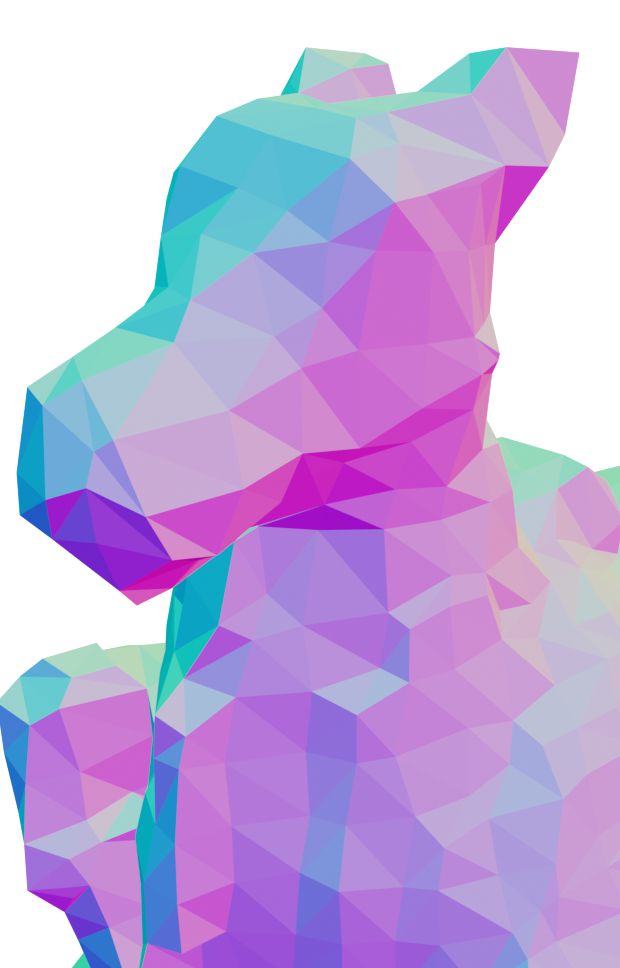} 
\end{subfigure}
 \begin{subfigure}{.13\textwidth}
  \centering
  \includegraphics[width=\linewidth]{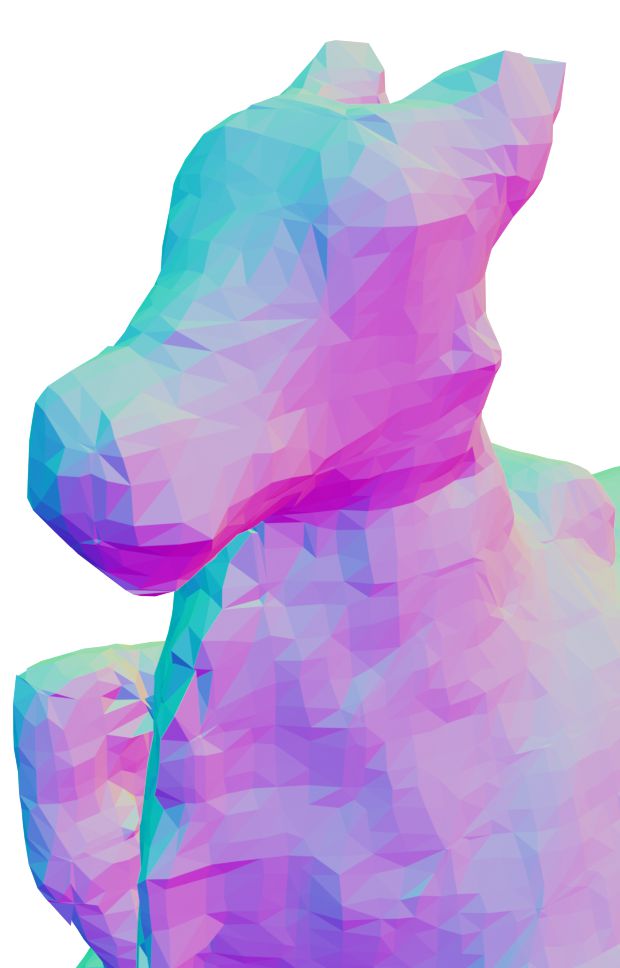} 
\end{subfigure}
 \begin{subfigure}{.13\textwidth}
  \centering
  \includegraphics[width=\linewidth]{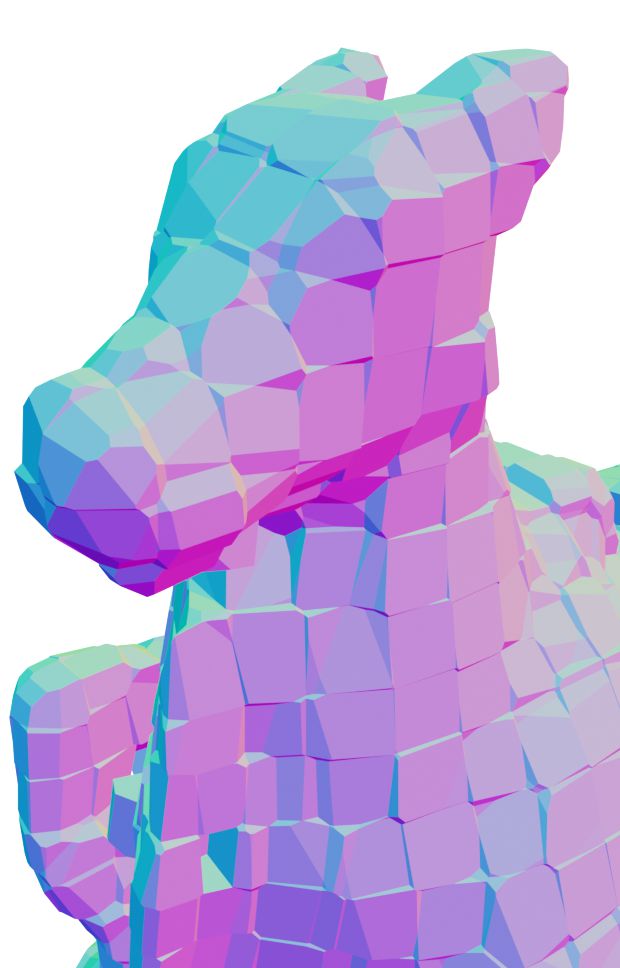} 
\end{subfigure}
 \begin{subfigure}{.13\textwidth}
  \centering
  \includegraphics[width=\linewidth]{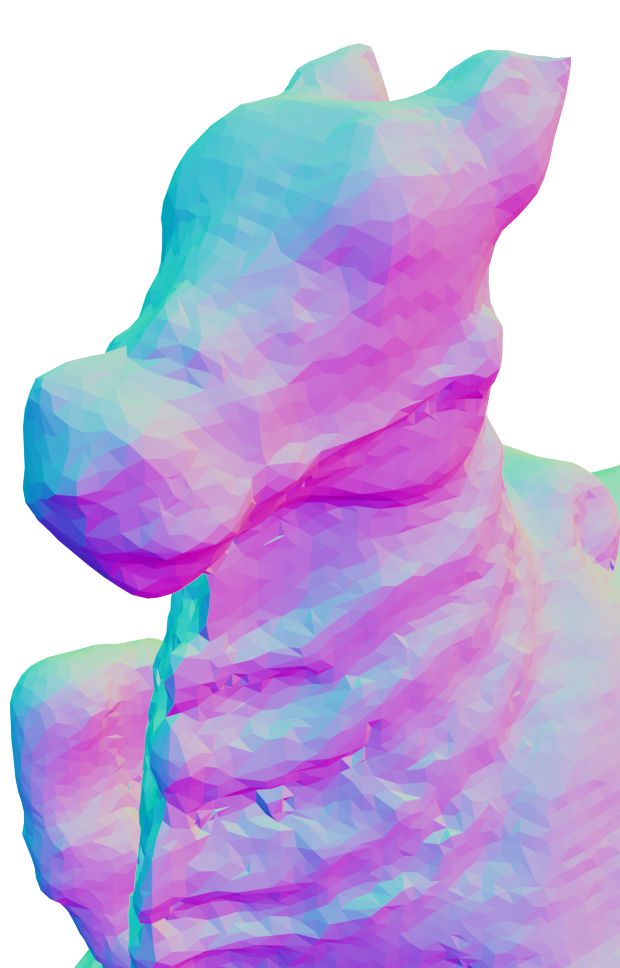} 
\end{subfigure}
 \begin{subfigure}{.13\textwidth}
  \centering
  \includegraphics[width=\linewidth]{figs/groundtruth/90889_normal.jpeg} 
\end{subfigure}

\centering
 \begin{subfigure}{.13\textwidth}
  \centering
  \includegraphics[width=\linewidth]{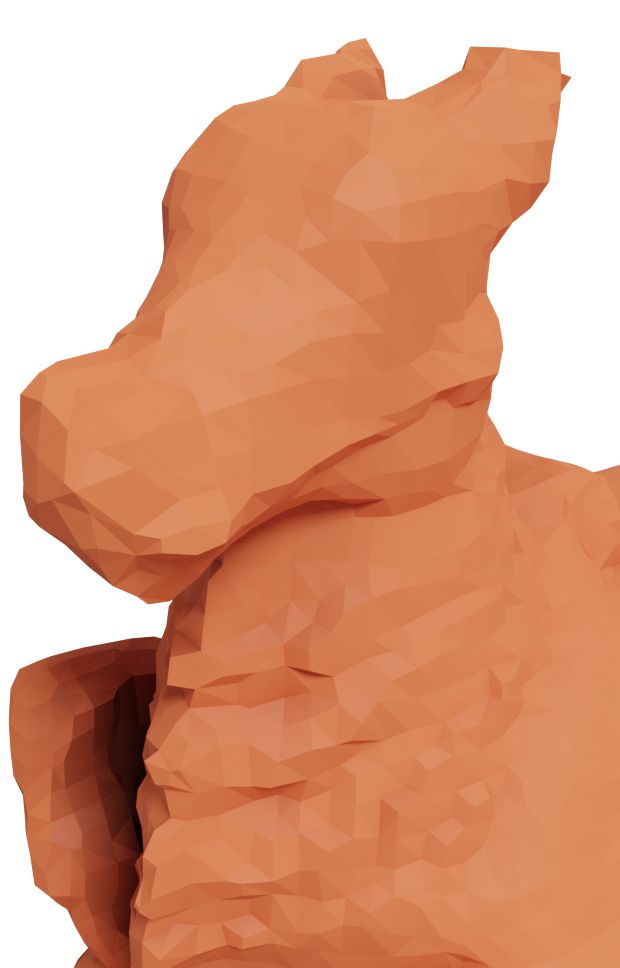} 
\end{subfigure}
 \begin{subfigure}{.13\textwidth}
  \centering
  \includegraphics[width=\linewidth]{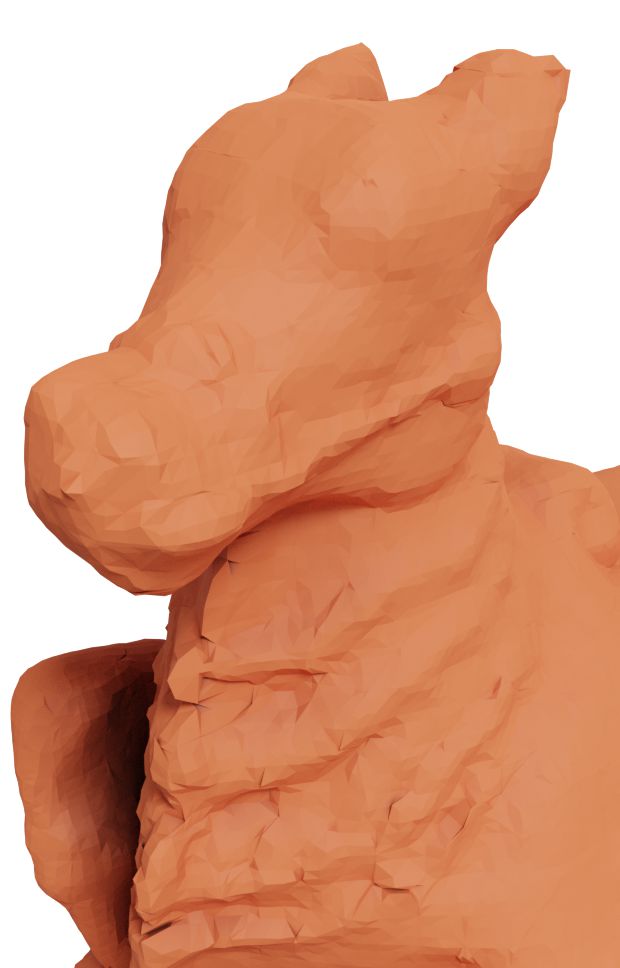} 
\end{subfigure}
 \begin{subfigure}{.13\textwidth}
  \centering
  \includegraphics[width=\linewidth]{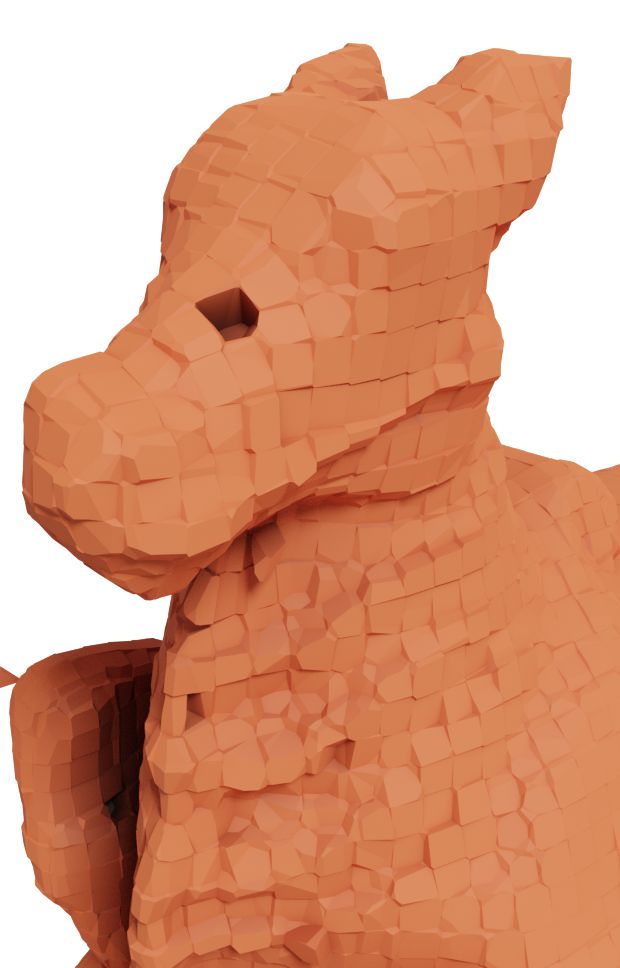} 
\end{subfigure}
 \begin{subfigure}{.13\textwidth}
  \centering
  \includegraphics[width=\linewidth]{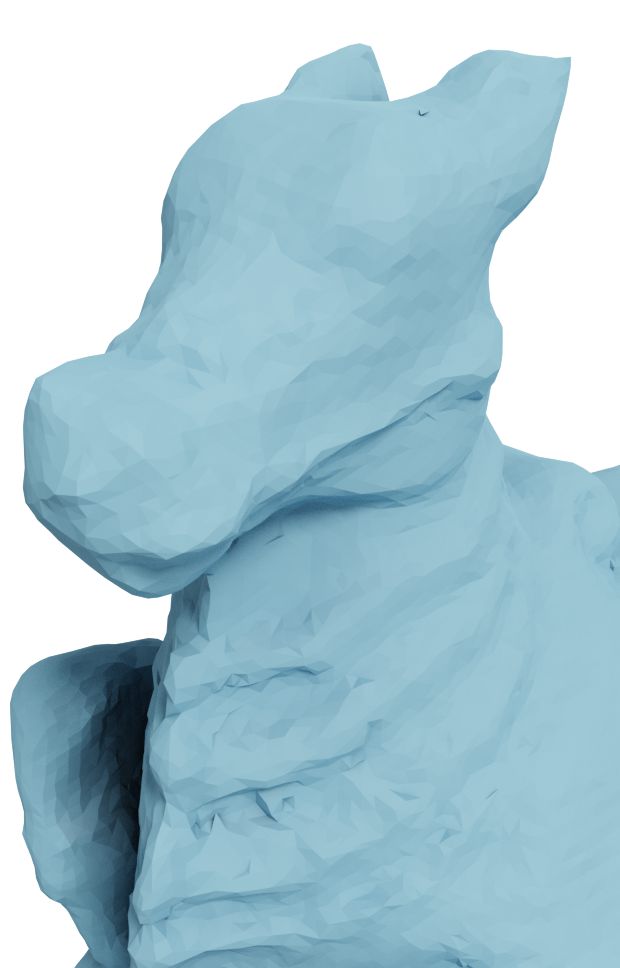} 
\end{subfigure}
 \begin{subfigure}{.13\textwidth}
  \centering
  \includegraphics[width=\linewidth]{figs/groundtruth/90889.jpeg} 
\end{subfigure}

\begin{subfigure}{.13\textwidth}
  \centering
  \includegraphics[width=\linewidth]{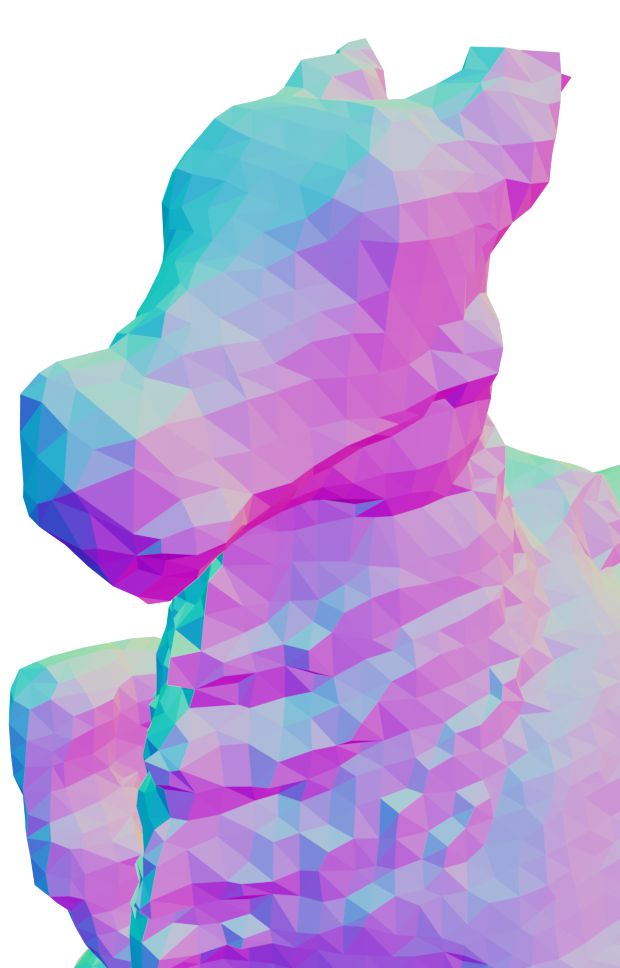} 
\end{subfigure}
 \begin{subfigure}{.13\textwidth}
  \centering
  \includegraphics[width=\linewidth]{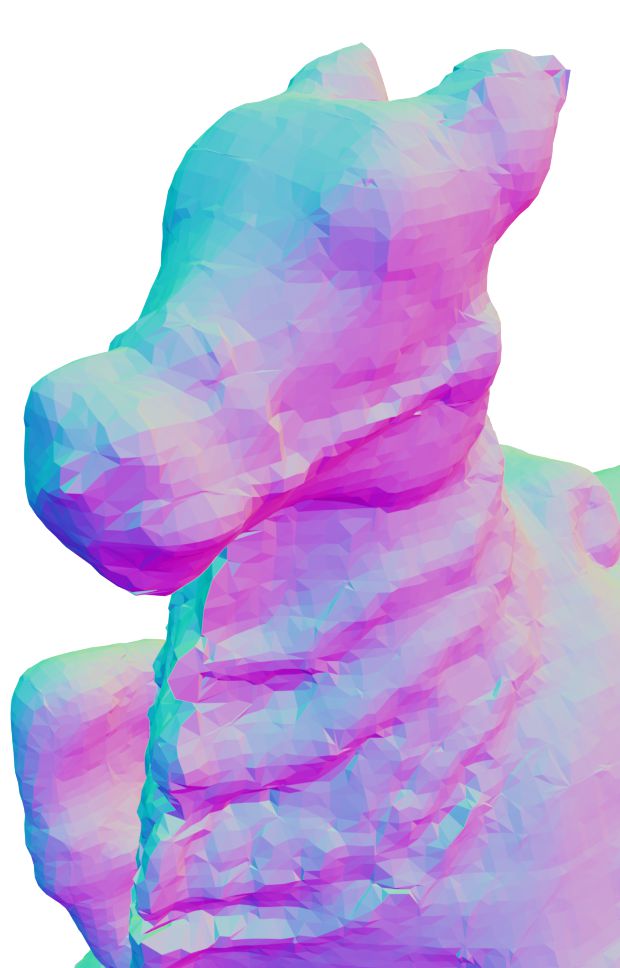} 
\end{subfigure}
 \begin{subfigure}{.13\textwidth}
  \centering
  \includegraphics[width=\linewidth]{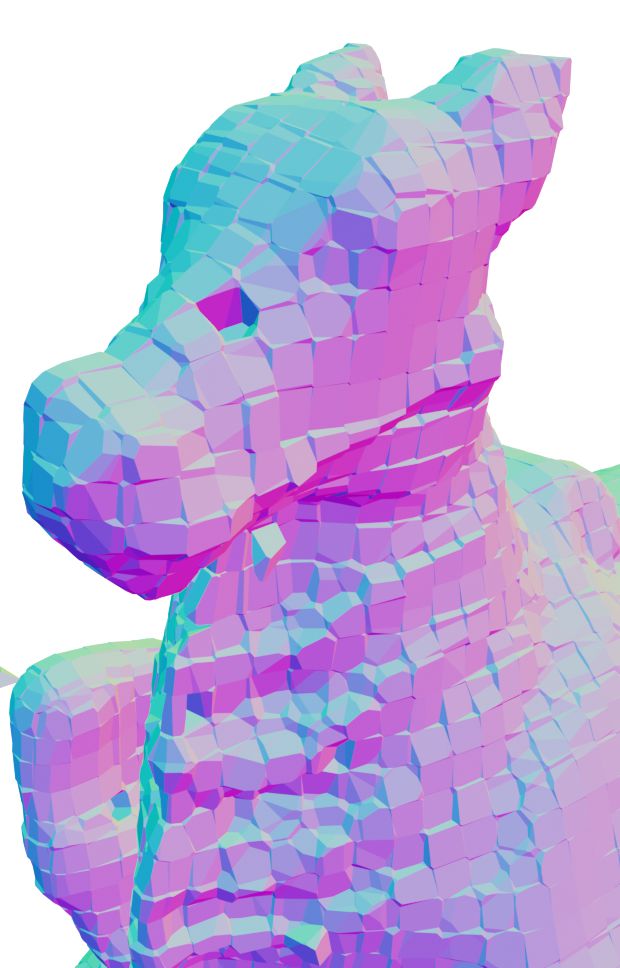} 
\end{subfigure}
 \begin{subfigure}{.13\textwidth}
  \centering
  \includegraphics[width=\linewidth]{figs/learning/quadric_90889_normal_64.jpeg} 
\end{subfigure}
 \begin{subfigure}{.13\textwidth}
  \centering
  \includegraphics[width=\linewidth]{figs/groundtruth/90889_normal.jpeg} 
\end{subfigure}

\centering
 \begin{subfigure}{.13\textwidth}
  \centering
  \includegraphics[width=\linewidth]{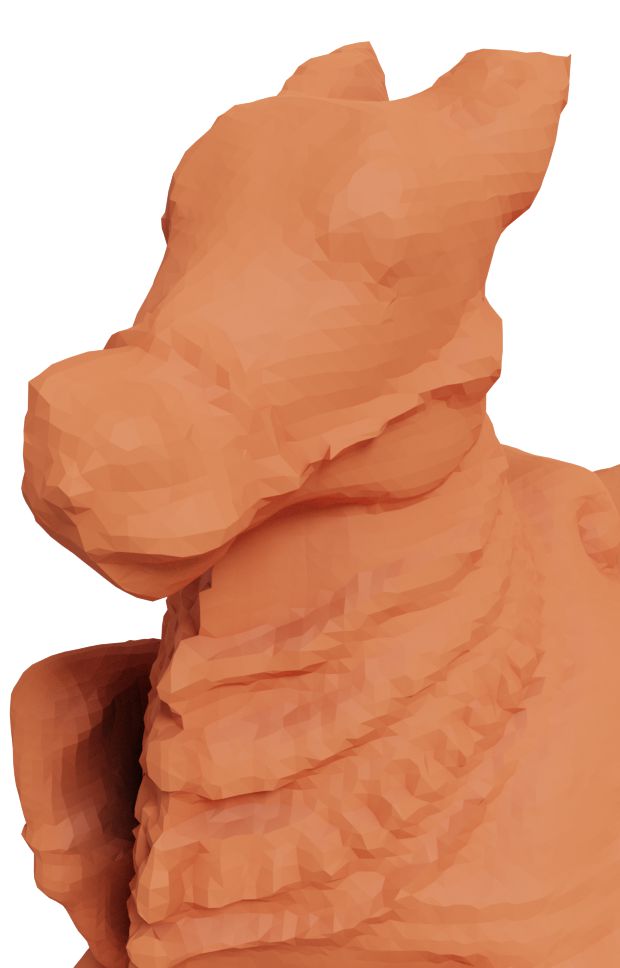} 
\end{subfigure}
 \begin{subfigure}{.13\textwidth}
  \centering
  \includegraphics[width=\linewidth]{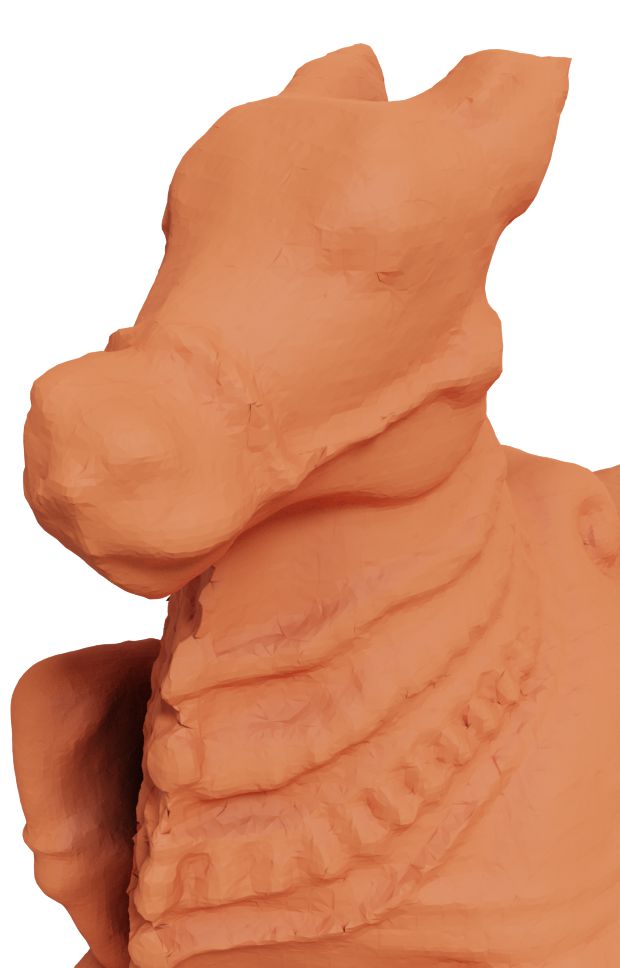} 
\end{subfigure}
 \begin{subfigure}{.13\textwidth}
  \centering
  \includegraphics[width=\linewidth]{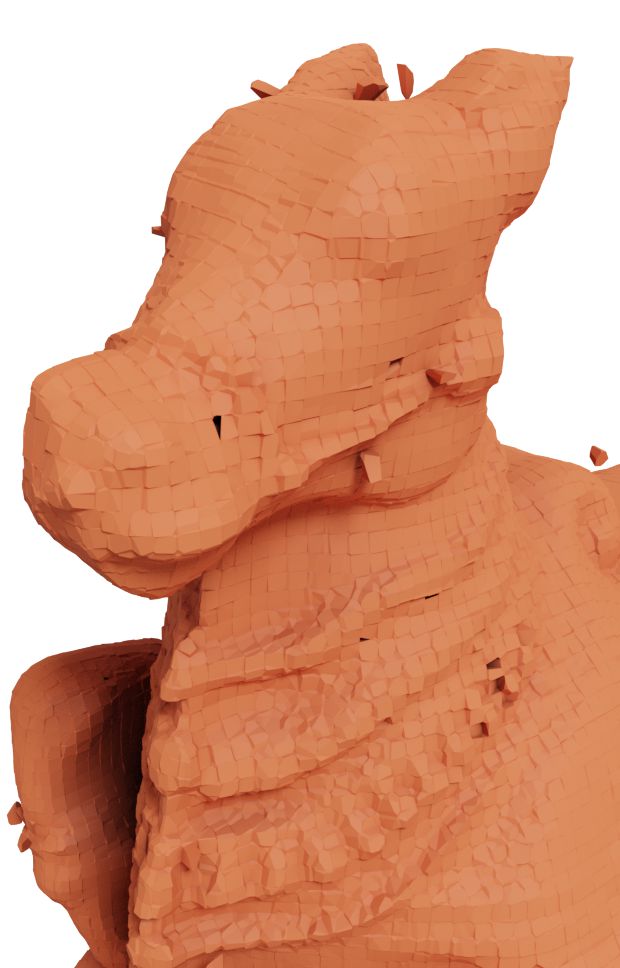} 
\end{subfigure}
 \begin{subfigure}{.13\textwidth}
  \centering
  \includegraphics[width=\linewidth]{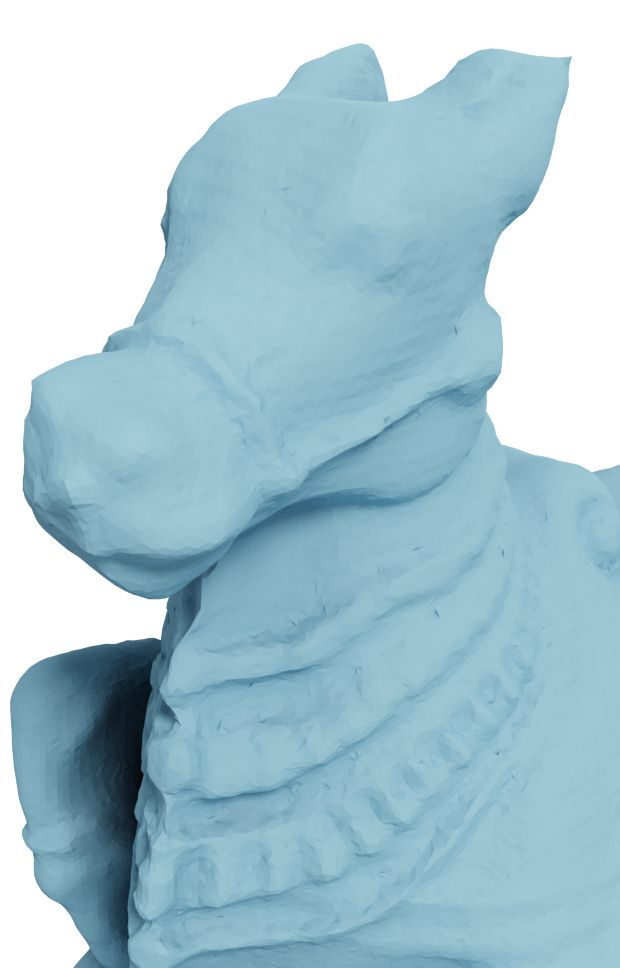} 
\end{subfigure}
 \begin{subfigure}{.13\textwidth}
  \centering
  \includegraphics[width=\linewidth]{figs/groundtruth/90889.jpeg} 
\end{subfigure}

\begin{subfigure}{.13\textwidth}
  \centering
  \includegraphics[width=\linewidth]{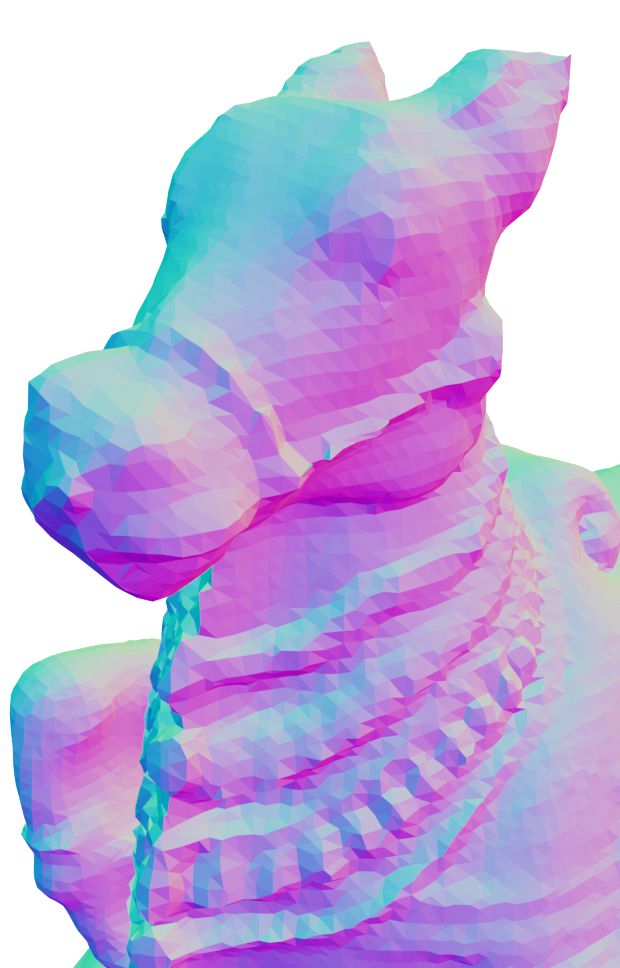} 
     \caption{NDC}
\end{subfigure}
 \begin{subfigure}{.13\textwidth}
  \centering
  \includegraphics[width=\linewidth]{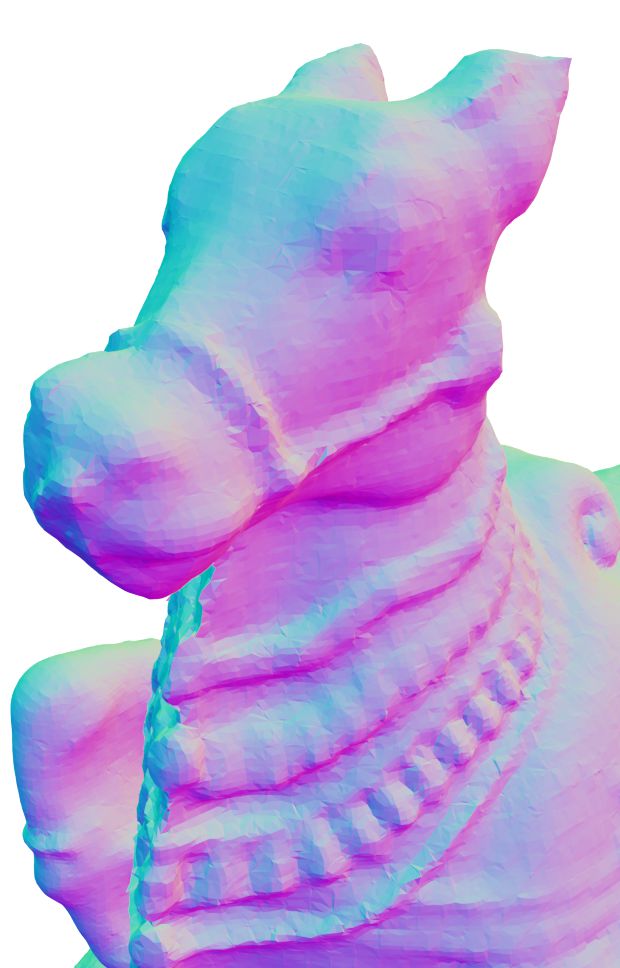} 
     \caption{NMC}
\end{subfigure}
 \begin{subfigure}{.13\textwidth}
  \centering
  \includegraphics[width=\linewidth]{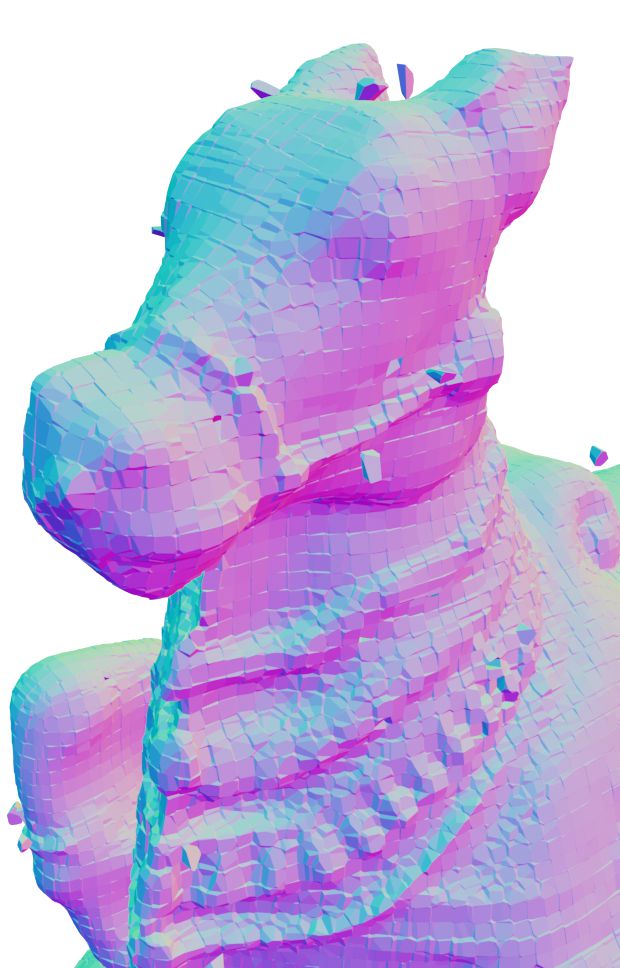} 
     \caption{VoroMesh}
\end{subfigure}
 \begin{subfigure}{.13\textwidth}
  \centering
  \includegraphics[width=\linewidth]{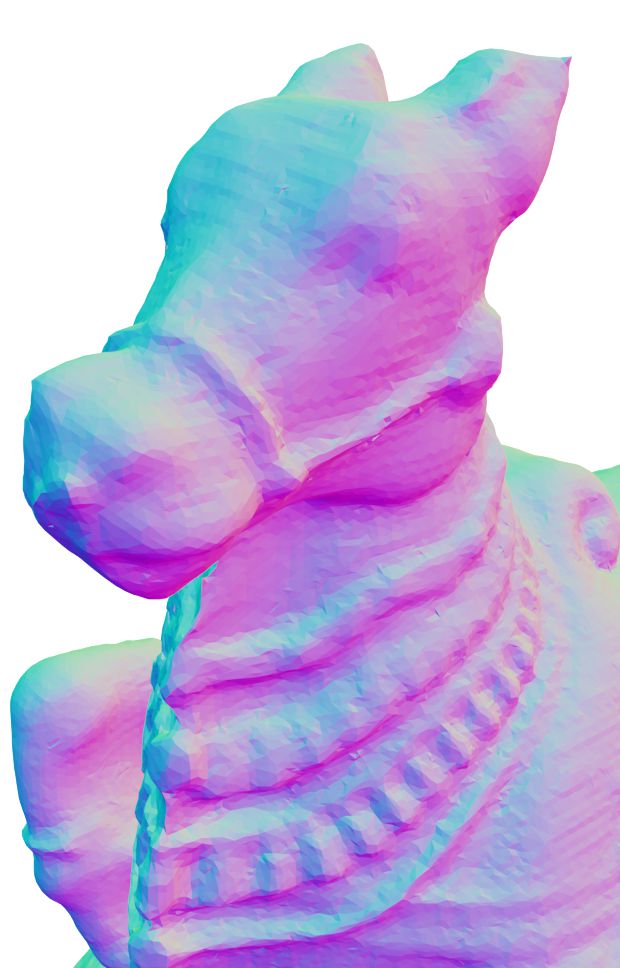} 
     \caption{PoNQ}
\end{subfigure}
 \begin{subfigure}{.13\textwidth}
  \centering
  \includegraphics[width=\linewidth]{figs/groundtruth/90889_normal.jpeg} 
     \caption{Gr. Truth}
\end{subfigure}

\caption{Learning results (top to bottom: $32^3$, $64^3$, $128^3$) on Thingi30. Networks trained on ABC.
}
  \label{fig:sdf_thingi_supl2}
\end{figure*}

Finally, Figs.~\ref{fig:optim_supl}-\ref{fig:sdf_thingi_supl2} exhibit further results comparing PoNQ to previous works.

\end{document}